The Dissertation Committee for Ye Zhao
certifies that this is the approved version of the following dissertation:

# A Planning and Control Framework for Humanoid Systems: Robust, Optimal, and Real-time Performance

Committee:

Luis Sentis, Supervisor

Benito R. Fernandez

Peter Stone

Efstathios Bakolas

Ufuk Topcu

Eric P. Fahrenthold

# A Planning and Control Framework for Humanoid Systems: Robust, Optimal, and Real-time Performance

by

**Ye Zhao, B.E.; M.S.E.**

**DISSERTATION**

Presented to the Faculty of the Graduate School of

The University of Texas at Austin

in Partial Fulfillment

of the Requirements

for the Degree of

**DOCTOR OF PHILOSOPHY**

Dedicated to my loving parents and fiancée who have been consistently supporting many years of my life and education.

# Acknowledgments

I would like to acknowledge many people who make this thesis possible.

First and foremost, I would like to express my extreme gratitude to my advisor and mentor, Prof. Luis Sentis, for his endless support and invaluable guidance over the course of my graduate years at The Human Centered Robotics Laboratory, UT Austin. It is my incredible joy and honor to work with him in such a world-renowned robotics lab. His thought-provoking instructions motivate me to persistently sense interesting problems and pursue convincing solutions. Many ideas are inspired by our deep conversations. Unforgettable memories of numerous walking-home chats deeply engrave on my mind. His foresight and vision profoundly inspire me to foresee how the robotics field will evolve in the next five to ten years, and greatly encourage me to chase academic frontiers. It is beyond doubt that these great minds will be a tremendous fortune for my career.

I would like to thank my thesis committee. I am obliged to Prof. Benito Fernandez for many insightful discussions. His keen advice on mathematical methodologies and practical approaches are fairly precious. I appreciate Prof. Peter Stone's perceptive criticism on hardware implementations, which deeply inspires myself to ruminate about the right way of targeting solid research. I appreciate Prof. Efstathios Bakolas for his instructions on optimal control theory. I gained not only theories but also many penetrating insights into interesting control problems. I am also grateful to Prof. Ufuk Topcu for his inspiring instructions on formal methods for controller verification and synthesis. His timely response and effective pointers make him an affable and approachable professor, and I feel quite comfortable during our discussions. I also express my appreciations to Prof. Eric Fahrenthold for his serving as my committee member and providing valuable suggestions. I would like to thank Prof. Behcet Acikmese for his instructions on convex optimization. My thanks also go to Prof. Dragan Djurdjanovic, Prof. Dongmei Chen, Prof. Ashish Deshpande and Prof. Joseph Beaman, in particular for their help and instructions during my first graduate year. Finally, I would like to thank my undergraduate research advisors Prof. Huijun Gao and Prof. Lixian Zhang for their guidance in opening my research journey.

I am deeply indebted to my lab-mates. Nicholas Paine provides numerous stimulating conversations on series elastic actuation and impedance control. The distributed control work would be impossible without him. I am fortunate to work in an excellent locomotion team with Donghyun Kim and Gray Thomas. I have lost count of the number of times that deep discussions on the blackboard last until far into the night. Donghyun's great efforts on the hardware implementation enabled amazing results of our highly-dynamic walking machine. I learned a lot of system integration techniques from him. Gray is a great creator of many fantastic demo ideas, which deeply inspired me to think about the extreme behavior and physical limit of legged robots. I would like to thank Kwan Suk Kim for his generous help in solving many software issues. He masters numerous coding skills and tricks. I retain a clear memory that there were a few times when he immediately pointed out solutions by merely printing out a message. I appreciate many helps from Steven Jorgensen for paper reviewing and suggestions, significantly improving the quality of published works. I appreciate the Whole-Body Control and software instructions from Chien-Liang Fok. His profoundness of knowledge and willingness to patiently help make him an indispensable resource. It has also been a privilege to obtain guidance from him. Kenan Isik, Gwen Johnson, Jack Hall, Mike Slovich, Kwan-Woong Gwak and Seung Kyu Park, Joshua James, and Cory Crean have not only been a crucial source of research brainstorming but also made my graduate years joyful and memorable. Thank you to Travis Llado, Michael Casterlin and Alan Kwok in building the mechanical facilities for bipedal robot experiments. I would like to thank Ajinkya Jain, Binghan He, Shih-Yun Lo, Mahsa Ghasemi, Orion Campbell for asking me questions about research ideas and topics, which helped me a lot to introspect my works. I appreciate Jonathan Samir Matthis and Sean L. Barton for our collaborations. Their unique insights enable me to inspect my work from the human biomechanics viewpoint. I thank Shiqi Zhang for instructing me of probabilistic motion planning and POMDP. Many thanks to my friends and tennis fellows, who I could not mention one by one.

I would like to express my special thank to the UT Austin team of NASA DARPA Robotics Challenge (DRC) lead by Prof. Luis Sentis and made up of Nicholas Paine, Chien-Liang Fok, and Gwen Johnson. I am deeply influenced by the tremendous efforts from them and the whole NASA-DRC team on building and controlling the Valkyrie robot. It is unimaginable how much work need to be achieved from all levels including mechanical design, embedded electronics, software architecture and planning and control strategies, and how to uniformly integrate them into a full humanoid robot and make it function well. This project significantly changed my perception of achieving the goal of sending a dexterous and safe humanoid robot to the space.

I would like to attribute my accomplishment to the love, patience, and support of my family. My parents, Chunfang Zhao and Shujuan Li, are always by my side to support me. Without them, I would not be able to accomplish this dissertation. Thank forever to my fiancee, Anqi (Angela) Wu, who has offered me great support and understanding. Our mutual trust overcomes all types of difficulties, especially those caused by the far-away distance. I owe much to her. Family is always my power in doing anything.

The work presented in this dissertation was supported by our funding agency Office of Naval Research (ONR) for the HAWK: Hyper-Agile Walking Controller for Bipedal Robots Aboard Navy Vessel project (grant #N0001412 10663), NASA Johnson Space Center for the NSF/NASA NRI project (grant #NNX12AM03G), and The University of Texas at Austin.

**A Planning and Control Framework for Humanoid Systems: Robust, Optimal, and Real-time Performance**

Ye Zhao, Ph.D.
The University of Texas at Austin, 2016

Supervisor: Luis Sentis

Humanoid robots are increasingly demanded to operate in interactive and human-surrounded environments while achieving sophisticated locomotion and manipulation tasks. To accomplish these tasks, roboticists unremittingly seek for advanced methods that generate whole-body coordination behaviors and meanwhile fulfill various planning and control objectives. Undoubtedly, these goals pose fundamental challenges to the robotics and control community. To take an incremental step towards reducing the performance gap between theoretical foundations and real implementations, we present a planning and control framework for the humanoid, especially legged robots, for achieving high performance and generating agile motions. A particular concentration is on the *robust, optimal and real-time* performance. This framework constitutes three hierarchical layers, which are presented from the following perspectives.

First, we present a robust optimal phase-space planning framework for dynamic legged locomotion over rough terrain. This framework is a hybrid motion planner incorporating a series of pivotal components. Via centroidal momentum dynamics, we define a new class of locomotion phase-space manifolds, as a Riemannian distance metric, and propose a robust optimal controller to recover from external disturbances at runtime. The agility and robustness capabilities of our proposed framework are illustrated in (i) simulations of dynamic maneuvers over diverse challenging terrains and under external disturbances; (ii) experimental implementations on our point-feet bipedal robot.

Second, we take a step toward formally synthesizing high-level reactive planners for whole-body locomotion in constrained environments. We formulate a two-player temporal logic game between the contact planner and its possibly-adversarial environment. The resulting discrete planner satisfies the given task specifications expressed as a fragment of temporal logic. The provable correctness of the low-level execution of the synthesized discrete planner is guaranteed through the so-called simulation relations. We conjecture that this theoretical advance has the potential to act as an entry point for the humanoid community to employ formal methods for the planner verification and synthesis.

Third, we propose a distributed control architecture for the latency-prone humanoid robotic systems. A central experimental phenomenon is observed that the stability of high impedance distributed controllers is highly sensitive to damping feedback delay but much less to stiffness feedback delay. We pursue a detailed analysis of the distributed controllers where damping feedback effort is executed in proximity to the control plant, and stiffness feedback effort is im-plemented in a latency-prone centralized control process. Critically-damped gain selection criteria are designed for not only rigid but also series elastic actuators (SEAs). In particular, we devise a novel impedance performance metric, defined as "Z-region", simultaneously quantifying the achievable SEA impedance magnitude and frequency ranges. Finally, this distributed control strategy is generalized to the time-delayed Whole-Body Operational Space Control with SEA dynamics. To ensure passivity, we separate the overall closed-loop system into two subsystems interconnected in a feedback configuration. By designing Lyapunov-Krasovskii functionals, we propose a delay-dependent passivity criterion of the closed-loop system in the form of linear matrix inequalities (LMIs), and solve for the allowable maximum time delays via the passivity criterion. The proposed distributed control strategy is validated through extensive experimental implementations on UT rigid and series elastic actuators, an omnidirectional mobile base Trikey and a SEA-equipped bipedal robot Hume.

# Table of Contents

# List of Tables

# List of Figures

# Chapter 1

# Introduction

## 1.1 Motivation and Goals

Over the past few decades, the complexity of humanoid locomotion and manipulation tasks has grown significantly. A central goal of humanoid robots is to achieve dexterous, versatile and robust behaviors within unstructured environments and along with humans. As a motivating example, the disaster caused by the nuclear reactor melt-down of the Fukushima Daiichi power plant necessitated mobile robots be deployed to survey the damage and assess danger levels (Nagatani et al., 2013). However, simple tracked robots were the only available option, limiting the robotic capability to surveillance of areas accessible to a tracked, tethered vehicle. In response to this disaster, the United States DARPA sponsored a competition – the DARPA Robotics Challenge (DRC)– requiring robots to enter degraded human environments (such as damaged nuclear reactor plants) and actively perform tasks to mitigate damage and threat to human life (DARPA, 2014). The requirements for this competition spawned a number of sophisticated robot control architectures, motion planning frameworks, and decision-making algorithms which were intended to approach human-level dexterity and intelligence. These would help future robotic rescue workers to better surveil a disaster site and act on their own. From a research standpoint, the increase in robot task and dynamics complexity has therefore driven our research towards the following goals (also see Fig. 1.2):

- Seek advanced robot feedback control architectures and high performance control strategies. For instance, Valkyrie, NASA-JSC's entry into the DRC is a humanoid robot with 44 actuated degrees-of-freedom (DOFs) (Paine et al., 2015). Many DOFs of this humanoid have been equipped with series elastic actuators (SEAs) for torque sensing and compliant control. The use of SEAs in torque-controlled robots has gained increasing attention in recent years. In ideal conditions, the SEA control structure should be torque transparent, which means that the torque input sent from the high-level controller should be exactly equal to the torque output sent to the physical joint. But feedback delays, filtering and low-level controllers all prevent SEAs from behaving as an ideal torque source. During our experimental implementations, we realized that system stability and high-performance impedance control, as two foundational performance measures of humanoid control systems, are yet challenging problems to be further explored. This challenge exactly motivates our research on the distributed control architecture and performance evaluations regarding stability, trajectory tracking accuracy and robustness.

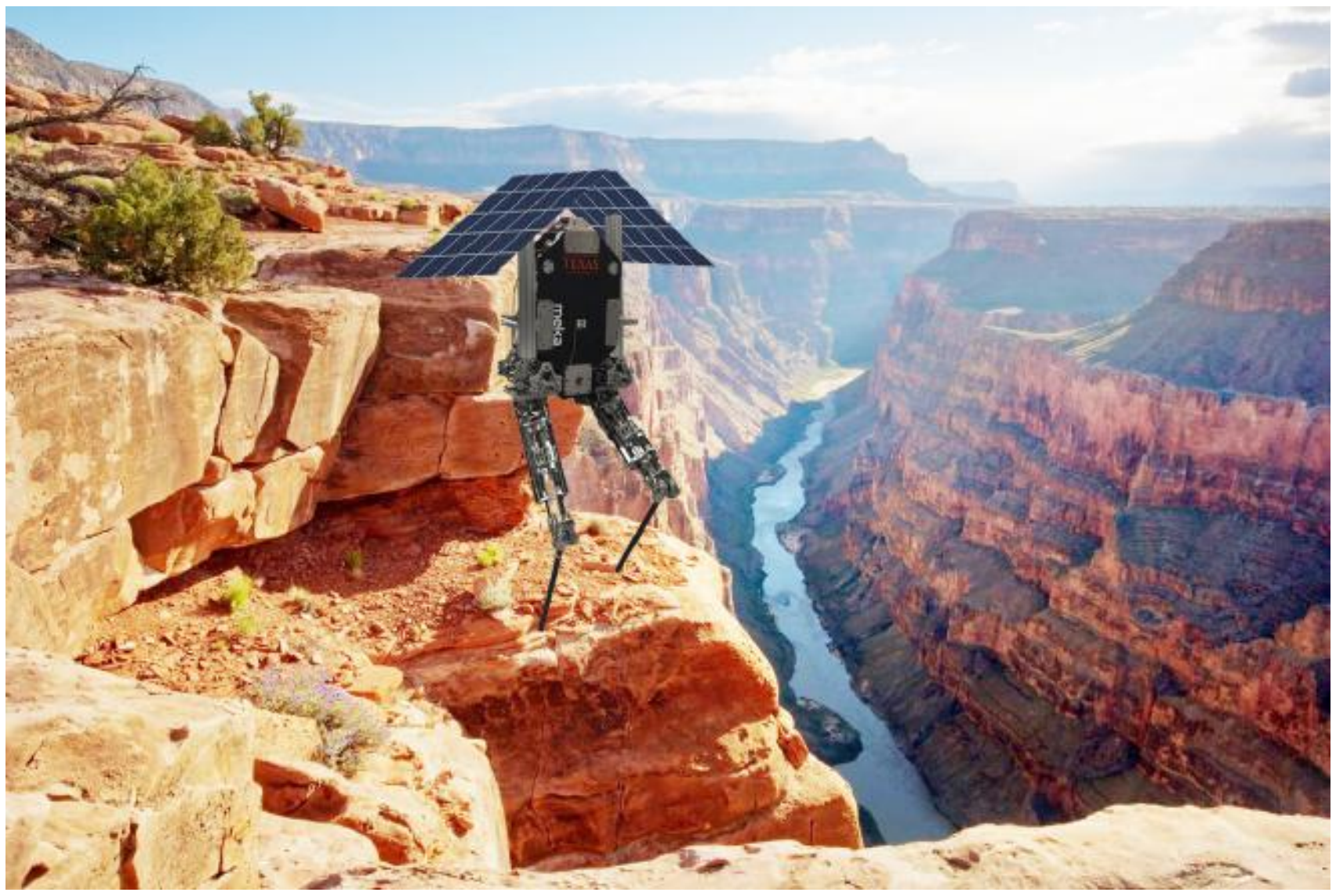

Figure 1.1: A conceptual diagram of a bipedal robot with a solar panel in Grand Canyon.

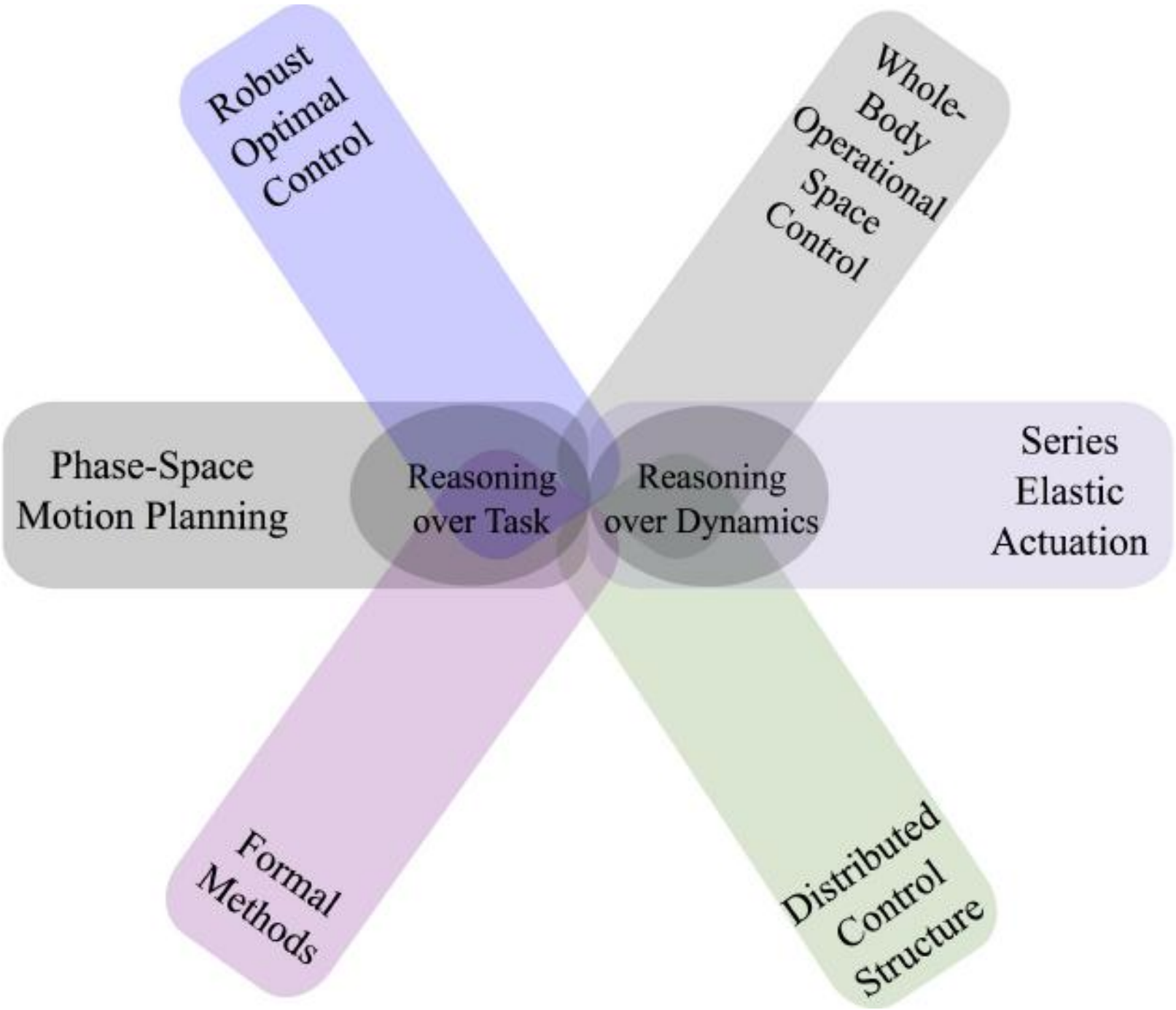

Figure 1.2: Reasoning over tasks and dynamics. This diagram summarizes the core topics of this dissertation.

- Propose a motion planning framework capable of generating agile, optimal, and robust locomotion patterns. Animals and humans have demonstrated excellent locomotion capabilities like walking, running, leaping, and climbing, etc., with superior efficiency and elegance, via full-fledged biomechanics, neural control, and self-learning. However, these behaviors are still unattainable (or at least restricted) for humanoid robots. One reason is the missing generic motion planner that is capable of generating locomotion trajectories and robust control policies to account for external disturbances or model uncertainties. One of the primary objectives of this dissertation is such a planning framework. We propose a phase-space planner for generic terrain maneuvering based on robustly tracking a sequence of keyframe states. To achieve robust recovery, we design a hybrid control strategy via dynamic programming and a foot placement re-planning strategy.

- Devise high-level reactive planners for humanoid task executions. Although widely used for mobile robot motion planning (Wongpiromsarn et al., 2012; Kloetzer and Belta, 2010; Belta et al., 2007; Fu and Topcu, 2016), formal methods from the verification and synthesis communities have not been yet incorporated into the planning sequence for complex mobility behaviors in humanoid robots. A possible reason is that legged robots are high dimensional and possess under-actuated dynamics. Our motion planning strategy focuses on low-dimensional phase-space planning models, circumventing the "curse of dimensionality". Given this advantage, we propose a high-level discrete planner to decide keyframe states and contact configurations at the high level, which have rarely been explored in the locomotion community. The task planner needs to satisfy given task specifications in a provably correct manner. This high-level abstraction opens the potential to generalize existing methods and algorithms to a wider range of robotic systems and applications.

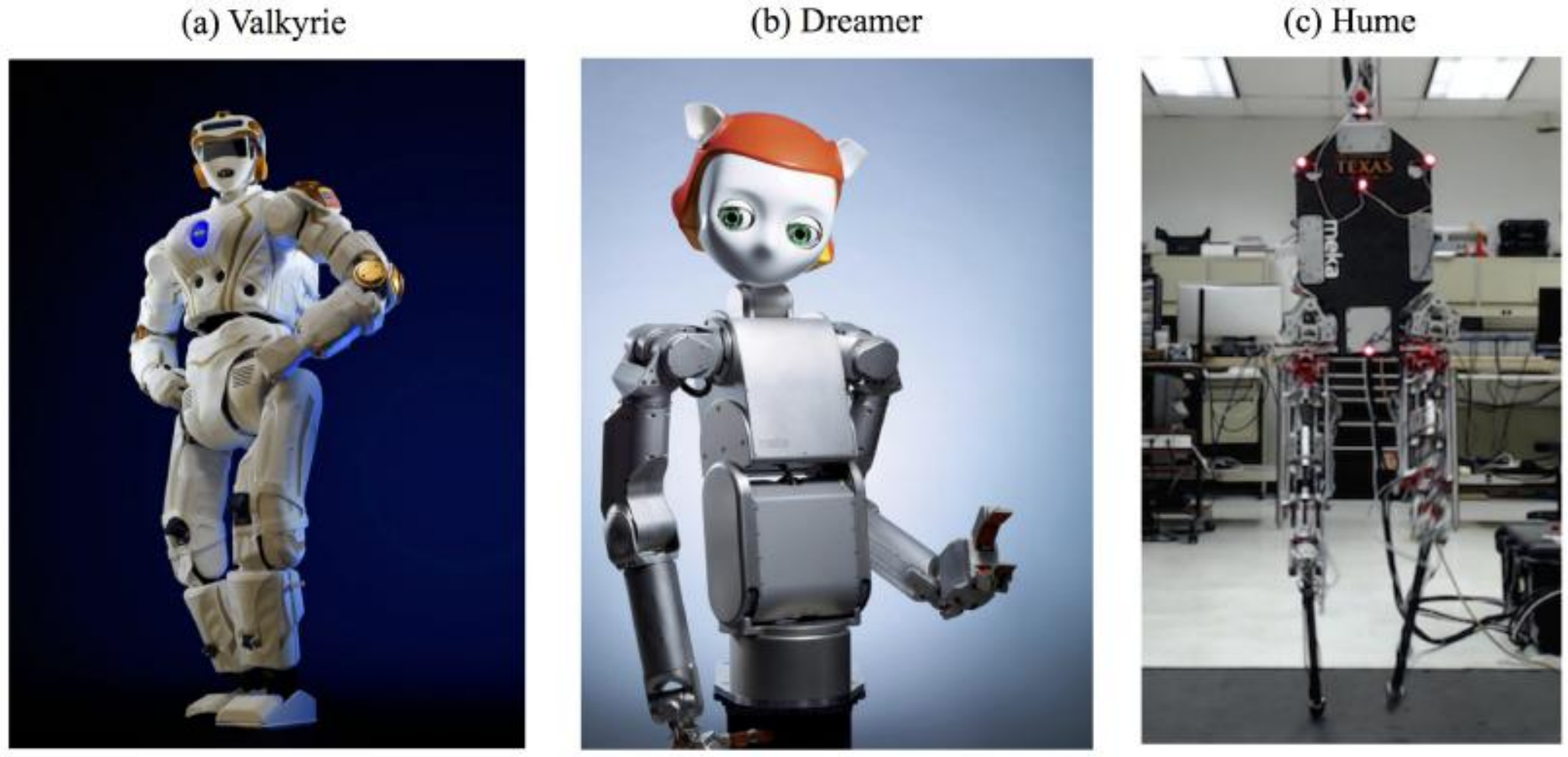

Figure 1.3: HCR lab and NASA Valkyrie robots equipped with series elastic actuators. Phase-space planner (Zhao et al., 2016c) and Whole-Body Operational Space Controller (Kim et al., 2016a; Zhao et al., 2015b; Zhao and Sentis, 2016) are designed to achieve robust and real-time performance while fulfilling prioritized multiple tasks.

## 1.2 Approach and Contribution

The primary objective of this dissertation is to advance the theoretical foundations of humanoid robot planning and control methodologies, and in the meanwhile contribute to the state of the art in experimental evaluations.

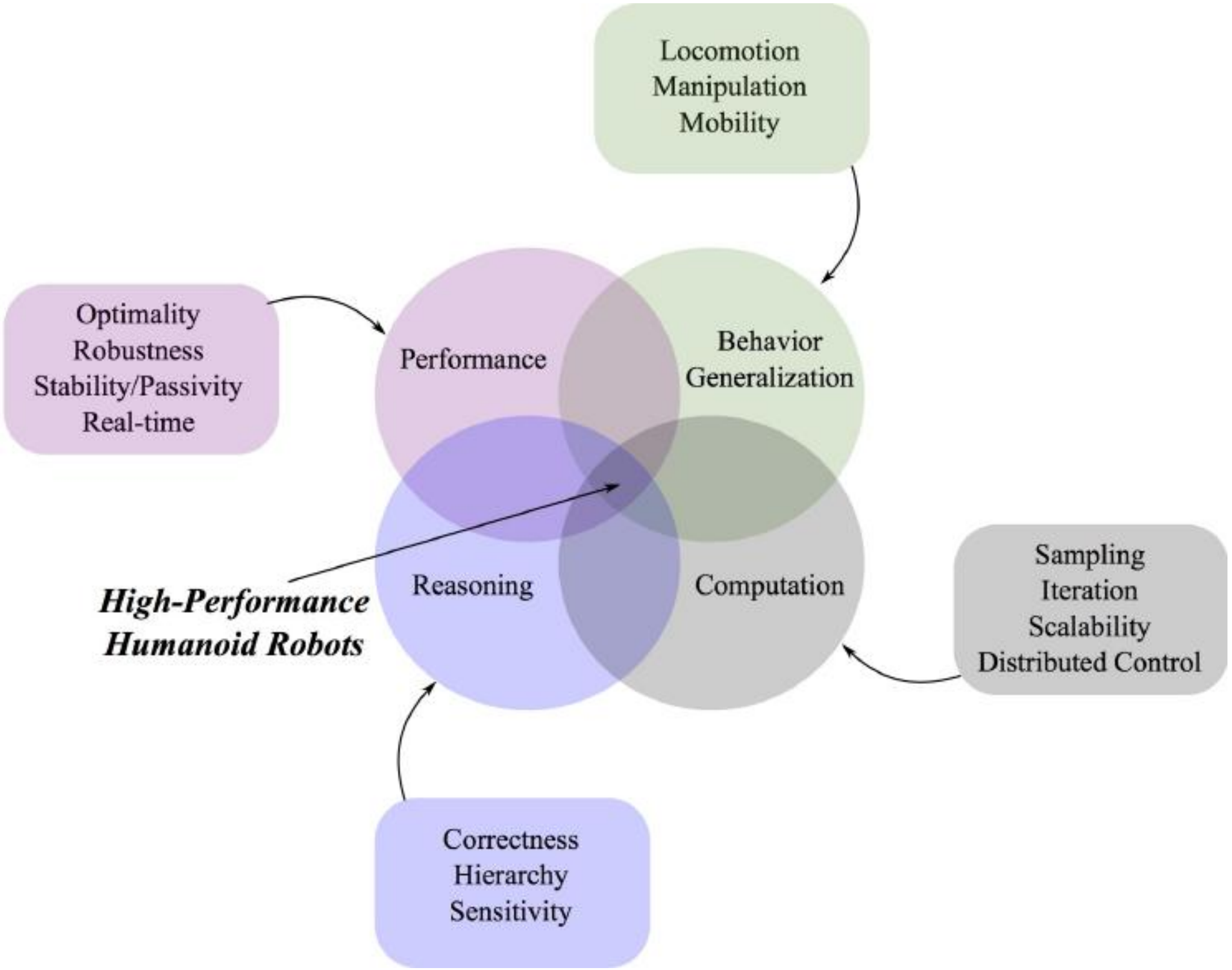

Figure 1.4: Dissertation aim and scheme.

Fig. 1.3 shows the humanoid robots at HCR Lab and NASA for experimental validations. We aim at providing promising theories and implementation methodologies that enable formal analysis and controller design for sophisticated humanoid robot behaviors within unstructured environments or in the presence of external disturbances. Such a goal demands research from four perspectives, that is, *performance, reasoning, computation and behavior generalization* as shown in Fig. 1.4. These perspectives must be unified to advance such an interdisciplinary field. In light of the discussions aforementioned, the approaches and contributions of this dissertation are primarily the following:

First of all, this dissertation formulates a phase-space planning framework for trajectory generations and proposes a robust optimal controller to achieve locomotion over various terrains using phase-space formalism. Based on centroidal momentum dynamics and a desired CoM surface plan, we formulate a hybrid automaton to characterize non-periodic locomotion dynamics. Via this automaton, we synthesize motion plans in the phase-space to maneuver over irregular terrains while tracking a set of desired keyframes. The resulting trajectories are formulated as phase-space manifolds. Borrowing from sliding mode control theory, we use the newly defined manifolds as a Riemannian metric to measure deviations due to external disturbances. A control strategy based on dynamic programming is proposed that steers the locomotion process towards the planned trajectories. Additionally, we devise a foot placement re-planning strategy in the presence of the disturbances making the in-step controller unrecoverable. The proposed framework is validated in various simulations and experimental implementations on a point-feet bipedal robot. Overall, our work focuses on trajectory generation and robust control of non-periodic and hybrid gaits. We are less concentrated on dynamic balance or moving from an initial to a final location, but instead on tracking desired keyframes. As such our method has the great potential to be suitable for designing gaits in complex environments.

Second, we take a step toward formally synthesizing high-level reactive locomotion planners by solving a two-player game between the planner and its possibly-adversarial constrained environment. We employ linear temporal logic (Baier et al., 2008) to specify whole-body locomotion (WBL) behaviors. Locomotion inherently exhibits hybrid dynamics. This property prompts us to focus on discrete planning synthesis instead of adopting abstraction-based methods (Liu et al., 2013; Tabuada, 2009), which are applied to continuous dynamical systems. We rely on a discretization of the phase-space into keyframe states. We concentrate on the communication between the high-level and low-level planners via switching signals, and the correctness of the hierarchical protocol. Various low-level locomotion modes are defined from a centroidal-momentum model that specifies the WBL behaviors. As an extension from rough terrain bipedal locomotion (Sreenath et al., 2013; Englsberger et al., 2015b), we focus not only on whole-body mobility but also on responses to various unforeseen environmental events such as stair cracks and the sudden appearance of a human in the scene (for instance a scenario is shown in Fig. 1.5). Simulations of dynamic locomotion in constrained environments support the effectiveness of the hierarchical planner protocol. To the best of our knowledge, our work is the of the first attempts to employ formal methods for WBL behavior generations and planner synthesis with guarantees of correctness.

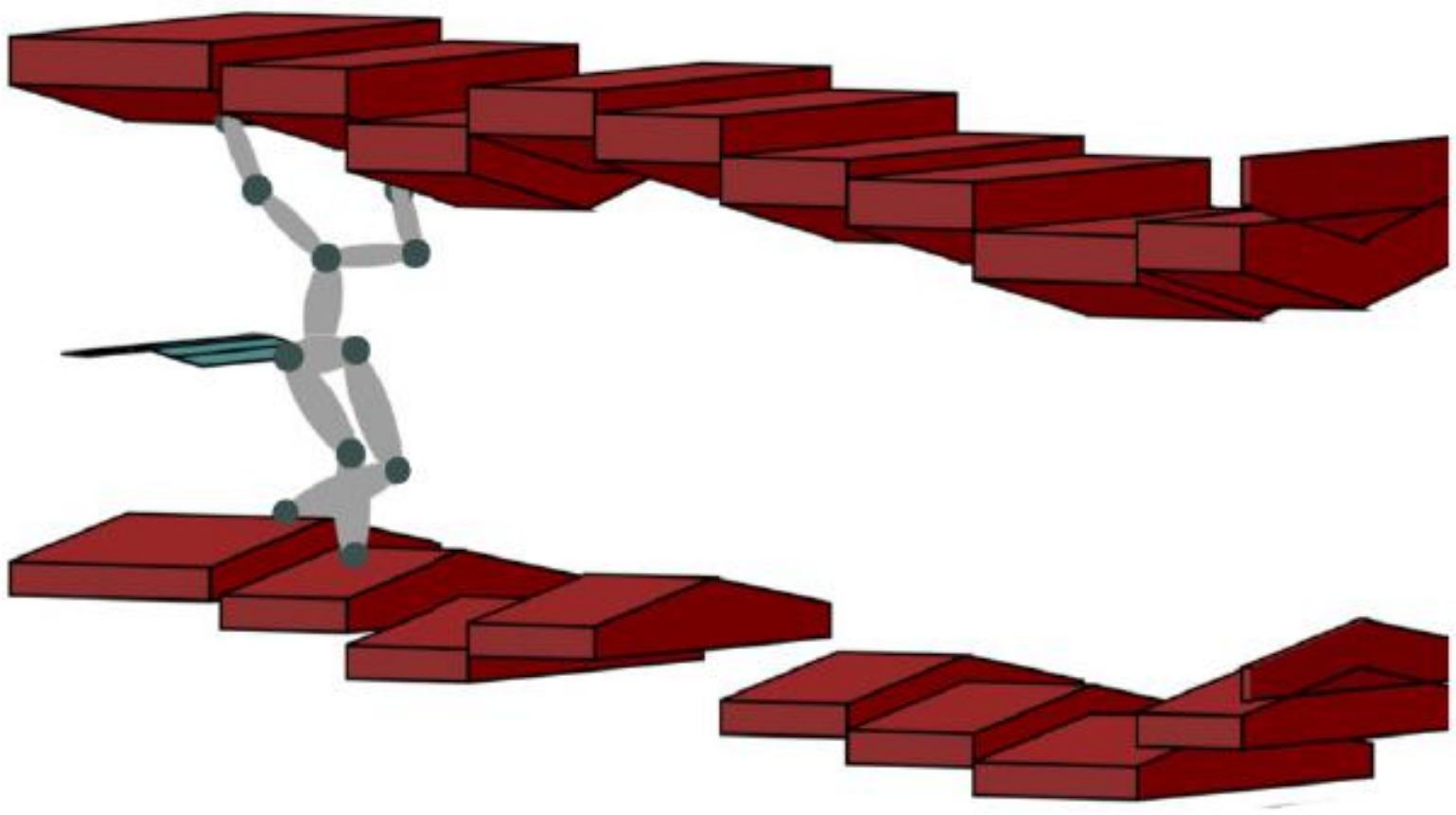

Figure 1.5: A scenario of whole-body locomotion maneuvering in a constrained environment with unforeseen dynamic events.

Third, we propose a class of distributed feedback control architectures which use stiffness servos for centralized WBOSC while realizing embedded-level damping servos as joint space damping processes. Our study reveals that system stability and performance is more sensitive to damping than stiffness servo latencies. We primarily focus on analyzing, controlling, implementing and evaluating actuators and mobile robotic systems with latency-prone distributed architectures to significantly enhance their stability and trajectory tracking capabilities. As will be empirically demonstrated, the benefit of the proposed split control approach over a monolithic controller implemented at the high level is to increase control stability due to the reduced damping feedback delay. As a direct result, closed-loop actuator impedance may be increased beyond the levels possible with a monolithic high-level impedance controller. This technique may be leveraged on many practical systems to improve disturbance rejection by increasing gains without compromising overall controller stability. As such, these findings are expected to be immediately useful on many complex human-centered robotic systems.

Fourth, we leverage the proposed distributed control architecture to series elastic actuators (SEAs) as shown in Fig. 1.6. A critically-damped gain selection criterion is proposed for a cascaded SEA control structure with inner torque and outer impedance feedback loops. Filters and feedback delays are taken into account for stability and impedance performance evaluation. Meanwhile, a trade-off between inner torque gains and outer impedance gains is observed and thoroughly analyzed. A new impedance performance measure named as "Z-region" is proposed to characterize SEA impedance magnitude and frequency range in a unified pattern. The goal is to maximize the SEA impedance range for achieving a wider range of Cartesian impedance tasks. The resulting experimental validations verify our proposed approach. We hope this serves as a stepping stone towards utilizing SEA-equipped humanoid robots for locomotion and manipulation behaviors and interaction with unstructured environments.

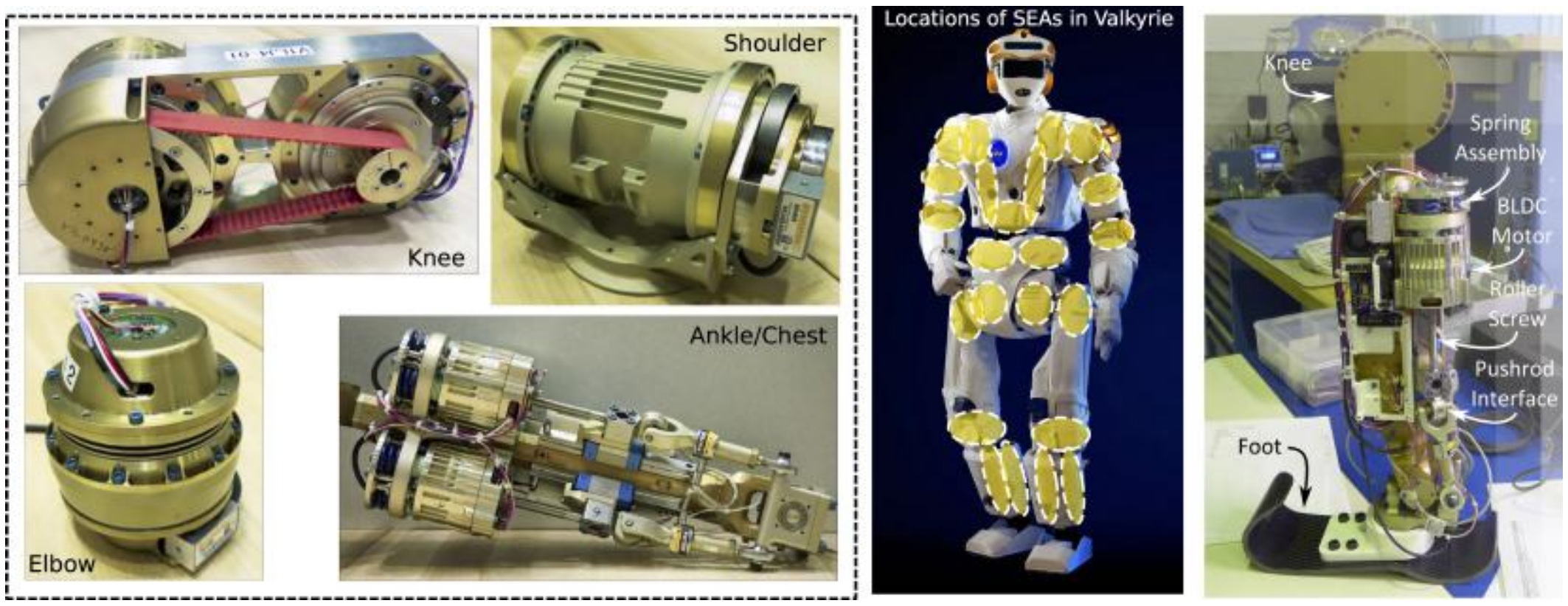

Figure 1.6: Valkyrie robot equipped with series elastic actuators. The left subfigures show a set of high-performance Valkyrie series elastic actuators (SEAs) from NASA; the middle one presents the Valkyrie robot with SEA location annotations; and the right one shows the calf and ankle structure.

Fifth, given the Single-Input Single-Output system discussed above, we generalize the distributed control architecture to Multi-Input Multi-Output (MIMO) humanoid robot systems. We devise a theoretical formalism for time-delayed WBOSC by incorporating the centralized-level Cartesian stiffness controller and the embedded-level SEA motor damping controller. Additionally, embedded-level torque feedback control is formulated and physically interpreted as a shaping of the motor inertia. For the passivity analysis, the whole system is separated into two interconnected subsystems in a feedback configuration. A delay-dependent passivity criterion is proposed by linear matrix inequality (LMI) techniques. To the best of our knowledge, this is the first time that feedback delays and series elastic actuation are simultaneously incorporated into the WBOSC framework.

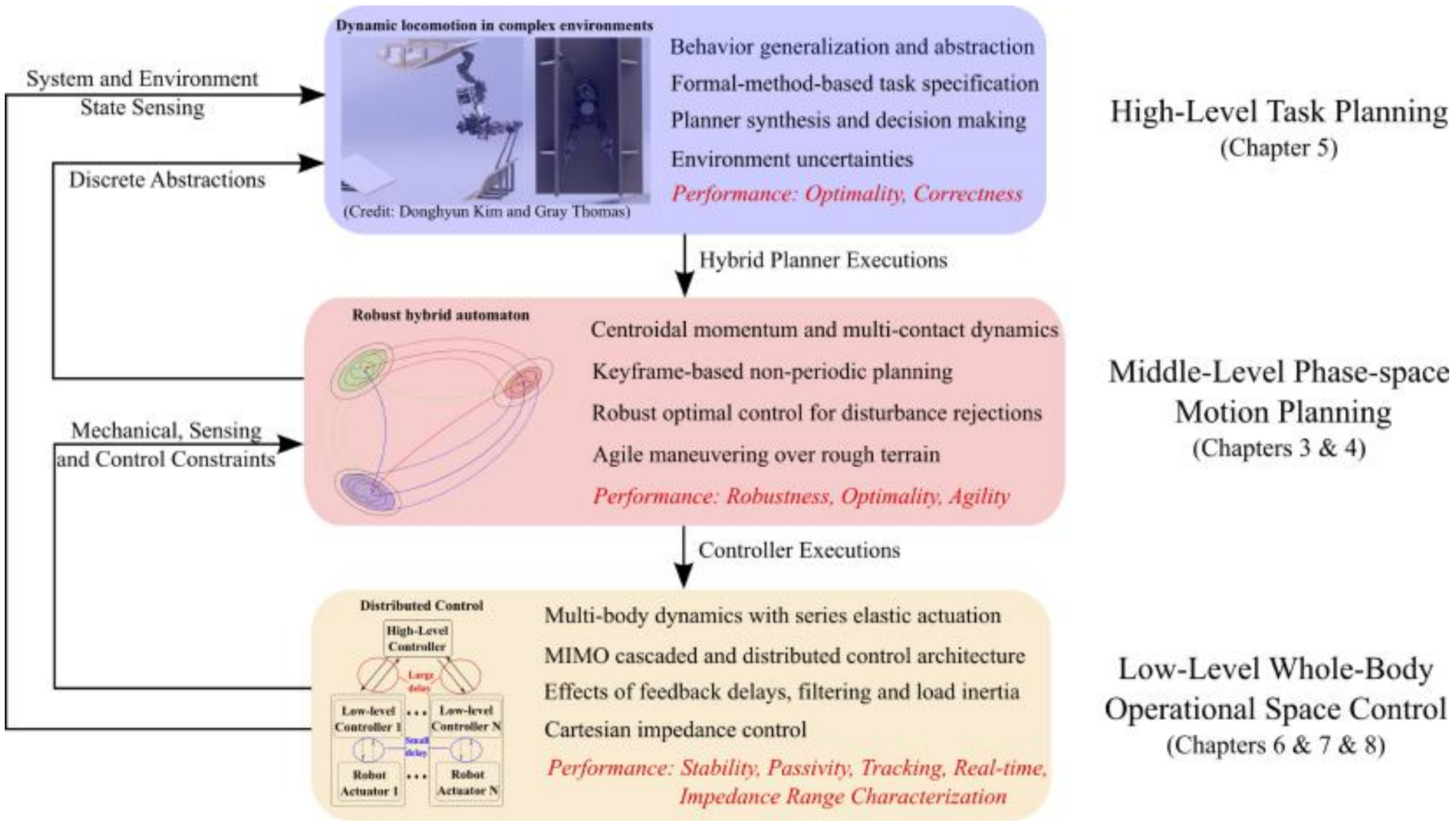

Figure 1.7: A hierarchical structure of the proposed motion planning and control framework.

Although the planning, control and decision-making methods in this dis-sertation are designed and implemented on different types of robots to a certain extent, we intend to highlight the fundamental methodologies and paradigms such as temporal-logic-based planner synthesis, robust optimal controller design and distributed control architecture. A variety of methods ranging from the high-level task planning to the middle-level motion planning to the low-level distributed Whole-Body Operational Space Control (WBOSC) is shown in Fig. 1.7. We hope the proposed methods and tools can be leveraged to other robotic systems.

As a tightly connected component with the task and motion planners, and the low-level actuator controllers, the WBOSC (Sentis, 2007; Kim et al., 2016a) is indispensable. In this dissertation, we use the WBOSC as an underlying full-body controller, such as the ones in simulation results of Section 3.6.3, and experimental results of Sections 4.5 and 6.3.3. In Chapter 8, we formulate and reason about time-delayed WBOSC with SEA dynamics, and analyze its delay-dependent passivity. Given the close communications among planning and control components, the overall system stability and performance are largely determined by the interplay of these components.

## 1.3 Dissertation Outline

This dissertation is outlined as follows. Chapter 2 reviews related literature. In Chapter 3, we present a hybrid phase-space planning framework based on a centroidal momentum model and a series of key planning components. Chapter 4 introduces a robust optimal control strategy to achieve recoveries from external disturbance. In Chapter 5, we focus on synthesizing high-level planners using temporal-logic-based formal method for whole-body locomotion over constrained environments. Chapter 6 proposes a class of distributed feedback controllers where damping feedback effort is executed in proximity to the control plant and stiffness feedback effort is performed in a latency-prone centralized control process. Chapter 7 extends the proposed distributed control architecture to series elastic actuators, designs a critically-damped fourth-order system gain selection criterion, and analyzes its impedance performance. In Chapter 8, we generalize the distributed controller to the time-delayed Whole-Body Operational Space Control and propose a delay-dependent passivity criterion based on linear matrix inequality (LMI) techniques. Chapter 9 summarizes this dissertation and discusses future directions. Table 1.1 shows a summary of the quantified performance in each chapter.

Table 1.1: Key performance classifications of this dissertation.

| Performance index | Phase-space motion planner (Chapters 3 & 4) | High-level task planner (Chapter 5) | Distributed control (Chapters 6 & 7 & 8) |
|---|---|---|---|
| Robustness | External disturbance and model uncertainties | Unforeseen contact environment | Load uncertainties and system passivity |
| Optimality | Dynamic programming (DP) based optimal control | (To be explored) | Optimal phase-margin design via a critically -damped criterion |
| Real-time | Online DP execution and analytical foot placement re-planning strategy | Online environment detection | Distributed control structure with minimized damping feedback delays |

# Chapter 2

# Literature Review

Humanoid and legged robots, to name a few in Fig. 2.1, may soon nimbly maneuver over highly unstructured and cluttered environments. Within the humanoid community, a broad spectrum of control, planning, and decision-making methods have been proposed and evaluated. In this chapter, we present a concise overview on the closely-related literature of legged locomotion, formal methods, series elastic actuation, distributed impedance control, whole-body control and passivity-based control. In particular, the emphasis is placed on recent results directly impacting on our works.

## 2.1 Dynamic Legged Locomotion

### Rough Terrain Locomotion

Dynamic legged locomotion has been a center of attention for the past few decades (Hubicki et al., 2016; Erez and Smart, 2007; Wu and Popović, 2010; Zhao et al., 2014a; Nguyen and Sreenath, 2015; Gan et al., 2016). The work in (Raibert, 1986) pioneered robust hopping locomotion of point-foot monoped and bipedal robots using simple dynamical models but with limited applicability to semi-periodic hopping motions. His focus is on dynamically stabilizing legged robots. Instead, our focus is on precisely tracking keyframe states, i.e. a discrete set of desired robot center of mass (CoM) positions and velocities along the locomotion paths. Such capability is geared towards the design of highly non-periodic gaits in unstructured environments or the characterization of dynamic gait structure in a generic sense. (Pratt et al., 2001) achieved point foot biped walking using a virtual model control method but with limited applicability to mechanically supported robots. Unsupported point foot biped locomotion in moderately rough terrains has been recently achieved by (Grizzle et al., 2014) and (Ramezani et al., 2014) using Poincaré stability methods (Westervelt et al., 2007). However, Poincaré maps cannot be leveraged to achieving non-periodic gaits for highly irregular terrains. The work (Yang et al., 2009) devised switching controllers for aperiodic walking via re-defining the notation of walking stability. In contrast, our work focuses on non-periodic gaits for unsupported robots in highly irregular and disjointed terrains.

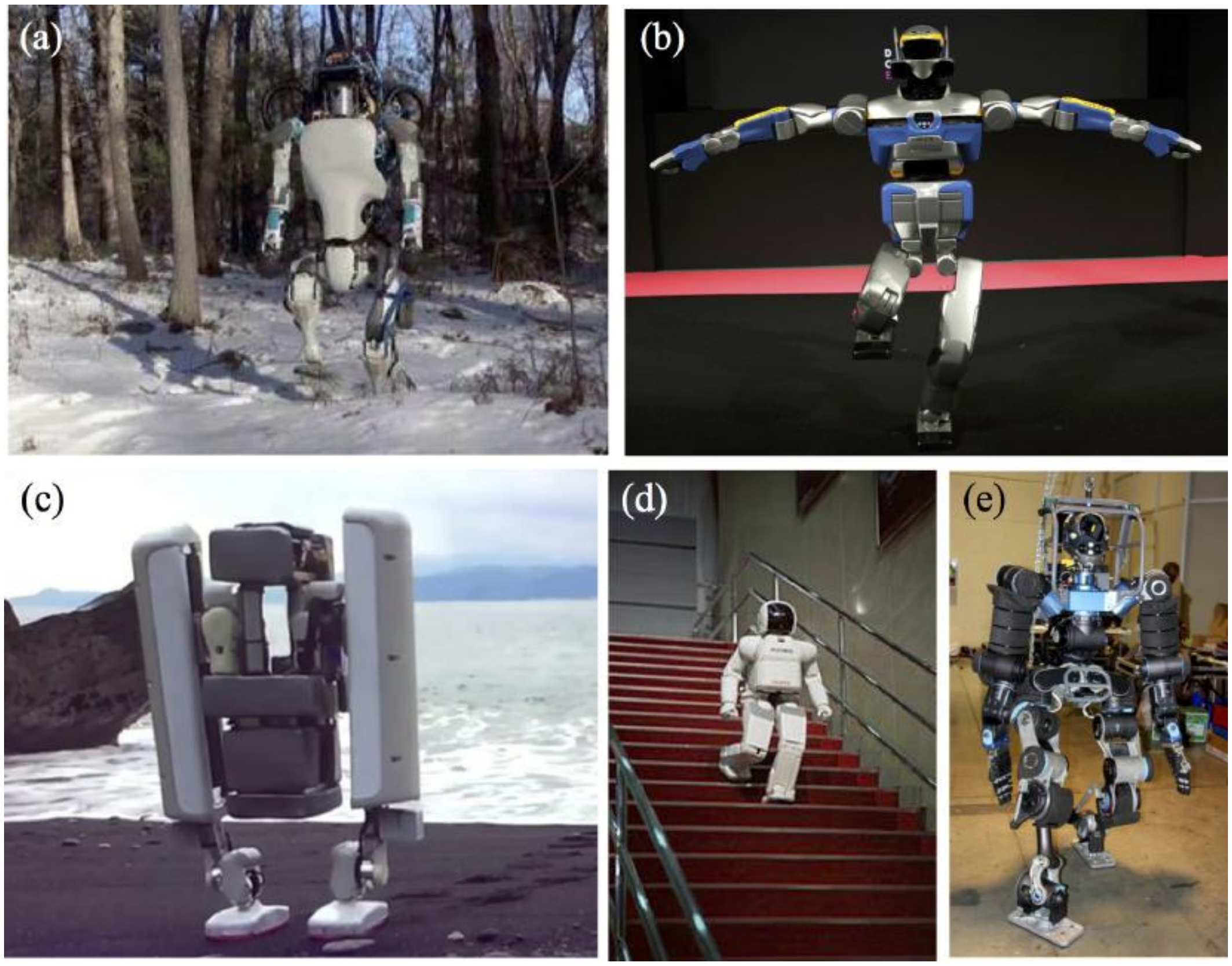

Figure 2.1: A few renowned humanoid robots. (a) Boston Dynamics Atlas robot (Spectrum, 2016); (b) HRP-2 Promet robot (Times, 2016); (c) Google's Alphabet-Schaft robot (Authority, 2016), (d) Honda Asimo robot (Byko, 2003); (e) IIT Walk-man robot (Spectrum, 2015).

The Capture Point method (Pratt et al., 2006) provides one of the most practical frameworks for locomotion. Sharing similar core ideas, divergent component of motion (Takenaka et al., 2009) and extrapolated center of mass (Hof, 2008) were independently proposed. Extensions to the Capture Point method, (Englsberger et al., 2015b; Morisawa et al., 2012), allow locomotion over rough terrains. Recently, the work in (Ramos and Hauser, 2015) generalizes the Capture Point method by proposing a "Nonlinear Inverted Pendulum" model, but it is limited to the two-dimensional case, and angular momentum control is ignored. Motion planning techniques based on interpolation through kinematic configurations have been explored, among other works, by (Hauser, 2014) and (Pham et al., 2013). Those techniques are making great progress towards mobility and locomotion in various kinds of environments. The main difference from these studies is that our controller explicitly accounts for robustness and stability to achieve under-actuated dynamic walking. If these works were to be implemented in unstable robots such as point-foot bipeds, they would lose balance and fail to recover. Our planner allows for continuous recovery without explicitly controlling the robot's center of mass.

Another close work on agile locomotion is (Mordatch et al., 2010) which proposes a physics-based locomotion controller and devises an online motion planner to generate various types of robust gaits over rough terrains. Recently, a general optimal motion control framework for behavior synthesis of human-like avatars is presented in (Mordatch et al., 2012). One key missing aspect is quantifying robustness and analyzing feedback stability. Additionally, this work does not address locomotion of point foot robots.

**Centroidal Momentum Dynamics**

Centroidal-momentum-based control have attracted increasing attentions within the humanoid community. It was first proposed in (Kajita et al., 2003b) for whole body humanoid motions and named resolved momentum control. Later, more detailed properties are systematically formulated in (Orin et al., 2013). The authors achieved that, for a free-floating humanoid robot, its generalized coordinate velocities can be mapped to the centroidal momentum linearly by using a so-called *Centroidal Momentum Matrix* (Orin and Goswami, 2008). Along with the same line of research, centroidal momentum is regarded as a prioritized control task in a humanoid dynamic kicking motion (Wensing and Orin, 2013) and its benefit is demonstrated by flexible arm motions for maintaining balance in challenging environments. Recently, a momentum-based controller (Herzog et al., 2016) is integrated into a complete hierarchical inverse dynamics framework and evaluated on a legged humanoid robot. Oth-ers were designed for balancing over uneven terrains (Lee and Goswami, 2010), or integrated with the concept of capture point (Koolen et al., 2015) and divergent component of motion (Hopkins et al., 2013). A salient advantage of centroidal-momentum-based frameworks is its evasion of complex full-body dynamics. It only comprises six dimensional states of CoM linear and angular momentums while maintaining an equivalent form to the full dynamics of the robot. This simplicity facilitates the trajectory optimization for complex dynamic motions (Dai et al., 2014). In Fig. 2.2, we show a spectrum of different locomotion models ranging from the full-body humanoid model to the simplified linear inverted pendulum model. The centroidal momentum model is a compromise of these two "extreme" models.

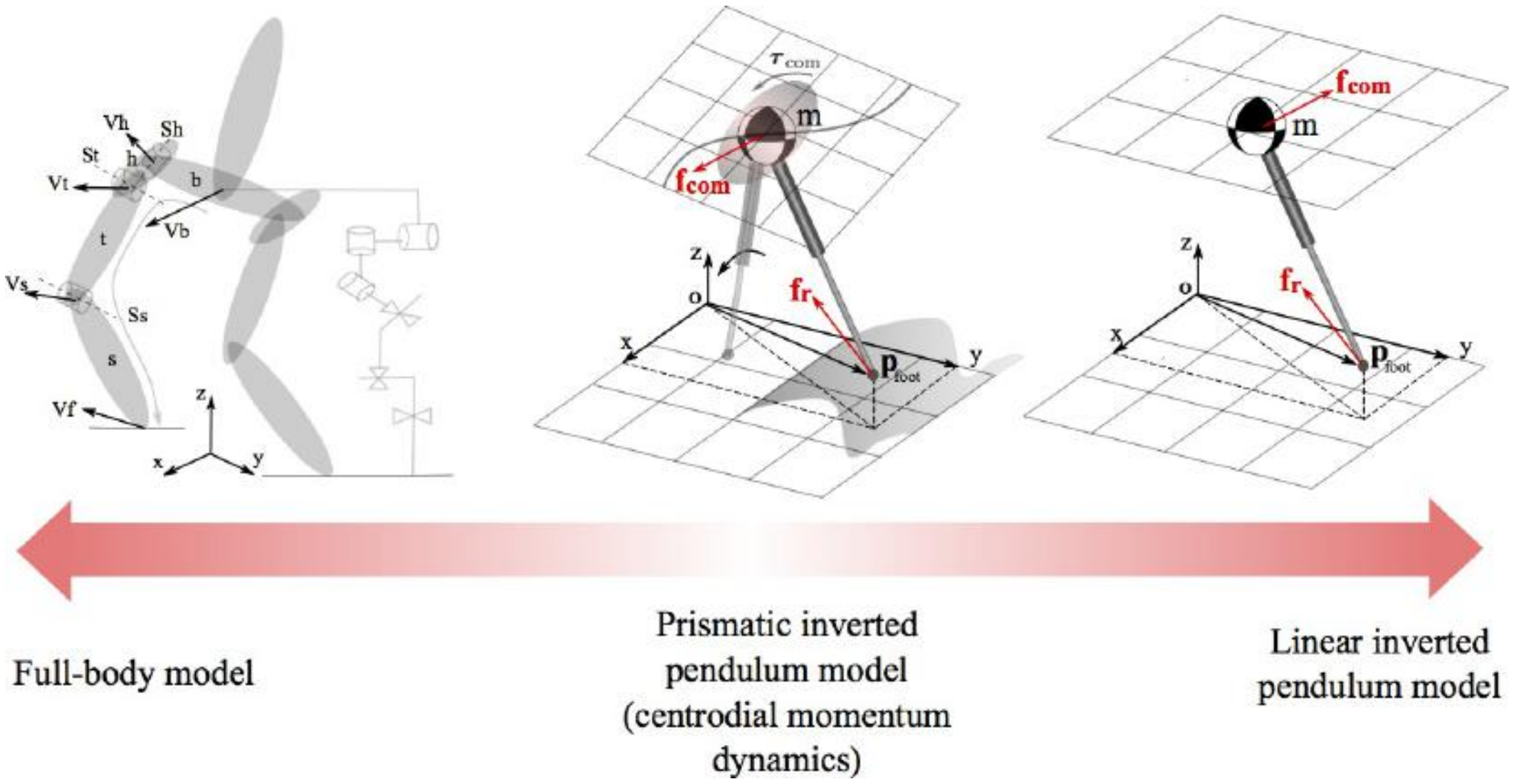

Figure 2.2: A spectrum of bipedal locomotion models associated with different levels of approximation.

### Optimal Control and Planning

Optimal control of legged locomotion over rough terrains are explored in (Feng et al., 2015; Dai and Tedrake, 2012; Byl and Tedrake, 2009). The work in (Manchester et al., 2011) proposed a control technique to stabilize non-periodic motions of under-actuated robots with a focus on walking over uneven terrain. The control is achieved by constructing a lower-dimensional system of coordinates transverse to the target cycle and then computing a receding-horizon feedback controller to exponentially stabilize the linearized dynamics. Recently, progress on this line of work enables the generation of non-periodic locomotion trajectories (Manchester and Umenberger, 2014). Also, in contrast with these works, we propose a metric for robustness to recover from disturbances. In (Saglam and Byl, 2014), a controller switching strategy for walking on irregular terrains is proposed. They optimize policies for switching between a set of known controllers. Their method is further extended to incorporate noise on the terrain and through a value iteration process they achieve a certain degree of robustness through switching. However, they focus on mesh learning by considering a set of known controllers. Instead, our study is focused on creating new controllers from scratch for general types of terrains. Additionally, their work is focused on 2D locomotion whereas we focus on 3D.

Online trajectory optimization is explored in (Tassa et al., 2012; Audren et al., 2014) for complex humanoid behavior generations. (Stephens and Atkeson, 2010) uses model predictive control (MPC) for push recovery by planning future steps. (Wieber, 2006) presents a linear MPC scheme for zero moment point control with perturbations. The problem of MPC is that it cannot find a globally optimal solution due to the finite horizon. In contrast thereto, we use dynamic programming to exhaustively search for the global optimal solution. Since the inverted pendulum model has low-dimensional states, the "curse of dimensionality" is not an issue for our case.

In (Saglam and Byl, 2014), a controller switching strategy for walking on irregular terrains is proposed. They optimize policies for switching between a set of known controllers. Their method is further extended to incorporate noise on the terrain and through a value iteration process they achieve a degree of switching robustness. However, they focus on trajectory mesh learning by considering a set of known controllers. Instead, our work is focused on creating new controllers from scratch for general types of terrains. Additionally, their work is focused on 2D locomotion whereas we focus on 3D.

**Multi-Contact Dynamics**

Humanoid multi-contact planning and control is gaining increasing attention (Sentis et al., 2010; Thomas and Sentis, 2016; Chung and Khatib, 2015; Bouyarmane and Kheddar, 2011; Hauser, 2014; Posa et al., 2016). The authors in (Kudruss et al., 2015) formulated multi-contact centroidal momentum dynamics as an optimal control problem. However, the work is focused on quasi-static mobility behaviors. Instead, our planning strategy is geared towards highly dynamic behavior generations. The work in (Caron et al., 2015) employed contact wrench cones to geometrically construct zero moment point support area in arbitrary virtual planes for multi-contact behaviors. Especially, selecting a ZMP plane on the top of robot's CoM enables the robot to behave as a marginally stable linear pendulum model. Under this model, the support area is guaranteed to be a necessary and sufficient condition of contact stability. However, this work is limited to constant CoM height motions. Our work allows for non-periodic multi-contact dynamic locomotion behaviors over constrained environments.

### Robustness and Recovery Strategies

Numerous studies have focused on recovery strategies upon disturbances (Hofmann, 2006; Zhao et al., 2013a). Various recovery methods have been proposed based on ankle, hip, knee, and stepping strategies (Kuo and Zajac, 1992; Stephens, 2007). In (Hyon and Cheng, 2007), a stepping controller based on ground contact forces is implemented in a humanoid robot. The study in (Komura et al., 2005) controls hip angular momentum to achieve planar bipedal locomotion. In our study, we simultaneously control the rate of change of the torso angular momentum, CoM height and foot placements to achieve unsupported rough terrain walking.

In (Hobbelen and Wisse, 2007), a gait sensitivity norm is presented to measure disturbance rejection during dynamic walking. In (Hamed et al., 2015), sensitivity analysis with respect to ground height variations is performed to model robustness of orbits. These techniques are limited to cyclic walking gaits. The work in (Arslan and Saranli, 2012) unifies planning and control to provide robustness. However, the technique is only applied to planar hopping robots. Recently, robust locomotion over non-cyclic rough terrains is explored in (Saglam and Byl, 2014) by considering noisy terrain slopes and numerical mesh resolution issue. However, robust stability is only quantified by estimating average steps to failure. On the contrary, our study proposes a phase-space manifold to refine the robustness metric and target recovery performance.

Knowledge of the exact terrain profile is partial due to the nature of the sensing processes. Many works in locomotion assume perfect terrain sensing (Feng et al., 2013; Liu et al., 2015). The work of (Byl and Tedrake, 2009) used mean first-passage time to quantify the robustness to unknown terrains, whose height follows a modeled probabilistic distribution. Recently, (Dai and Tedrake, 2013) proposed an $L_2$ gain indicator to characterize the capability of maneuvering over unknown terrains. (Park et al., 2013) devised finite-state-machine-based controllers for unexpected terrain height variations and implemented them in a planar robot. Although our study in this dissertation does not explicitly model terrain uncertainties, our proposed robustness metric and recovery strategies could be applicable to deal with unknown terrains. For instance, it is plausible to analyze the effect of terrain uncertainties in terms of the disturbance categories. As a result, the robust control strategies developed in Chapter 4 could be applied for recovery.

In (Frazzoli, 2001), a robust hybrid automaton is introduced to achieve time-optimal motion planning of a helicopter in an environment with obstacles. The same group studies robustness to model uncertainties (Schouwenaars et al., 2003) but ignores external disturbances. More recently, (Majumdar, 2016) accounts for external disturbances like cross-wind, by computing funnels via sums-of-squares (SOS) programming (Parrilo, 2000) and sequentially switching between these funnels for maneuvering unmanned air vehicles in the presence of obstacles and disturbances. Our work adopts similar ideas of this reference but focuses on point-foot locomotion. Since the locomotion system is inherently hybrid, we propose a hybrid control algorithm that switches states when the physical system changes the number of contacts. The hybrid automaton is used as a tool for planning and control of bipedal locomotion. We in fact extend their use of hybrid automaton to accommodate for hybrid systems. Additionally we re-generate phase-space trajectories on demand while those previous works rely on pre-generated primitives.

## 2.2 Task Planning and Formal Synthesis

### Temporal Logic Motion Planning

Formal methods for motion and task planning have been widely investigated for mobile navigation. The authors in (Kloetzer and Belta, 2010) proposed an automated computational framework for decentralized communications and control of a team of mobile robots from global task specifications. This work suffers from high computational complexity and does not address reactive response to environmental changes. To alleviate the computation burden, the work in (Wongpiromsarn et al., 2012) proposed a receding-horizon-based hierarchical framework that reduced a complex synthesis problem to a set of significantly smaller problems of a shorter horizon. The autonomous vehicle navigation is simulated in the presence of exogenous disturbances. The approach proposed in (Kress-Gazit et al., 2009) allows mobile robots to react to the environment in real time. Deviating from the discretization approaches aforementioned, Signal Temporal Logic (STL) (Donzé and Maler, 2010; Raman et al., 2015) allows to reason about dense-time, real-valued signals. Correspondingly, quantitative semantics are admitted to define the extent to which the specifications are satisfied or violated. This property makes STL especially suitable to quantify robustness measures (Farahani et al., 2015; Deshmukh et al., 2015). Recently, the result of (Sadigh and Kapoor, 2016) extended the STL to a probabilistic framework such that machine learning methods can be used for predictions and safety control under uncertainty. However, all of these works above are applied to mobile robots, which have simple dynamics unlike those of under-actuated humanoid robots.

**Formal Synthesis of Humanoid Robot Control and Planning**

More recently, formal methods such as linear temporal logic (LTL) have been initially used in humanoid robotics, especially in robotic manipulation by (He et al., 2015; Sharan, 2014; Chinchali et al., 2012). However, formal methods are yet to be explored for legged locomotion, or for more complex WBL tasks. Recently, the work in (Maniatopoulos et al., 2016) proposed an end-to-end approach to automatically implement a synthesized LTL-based planner on an Atlas humanoid robot. Reaction to low-level failures is formally incorporated by simply terminating the execution. Nevertheless, the robot behaviors are mainly centralized on manipulation and grasping tasks. The authors in (Antoniotti and Mishra, 1995) determined locomotion goals by using computation tree logic and synthesized controllers for legged motions.

However, this work is restricted to static locomotion tasks and provides little information of the synthesized results. An abstraction based controller was proposed in (Ames et al., 2015) for bipedal hybrid systems, but this work lacks provable correctness guarantees, and the robot behavior is limited to level-ground bipedal walking. The work of (Sreenath et al., 2013) proposed a two-layer hybrid controller hierarchy for locomotion over varying-slope terrains with unprecise sensing. To account for terrain uncertainties, a high-level controller implements a partially observable Markov decision process to make sequential decisions for controller switching. However, this work is limited to mildly rough terrains. In contrast thereto, our environments are composed of very unstructured terrains and our problem requires the synthesis of WBL responsive behaviors.

### Controller Provable Correctness

On controller synthesis, a recent focus has been on provable correctness of controllers (Kloetzer and Belta, 2010; Kress-Gazit et al., 2009). The work of (Liu et al., 2013) extended the controller synthesis with guaranteed-correctness to nonlinear switched systems and reacted to an adversarial environment at runtime. Our approach leverages this work and designs suitable discrete abstractions by approximation techniques. Given a high-level discrete controller encoding reactive task behaviors, the work in (DeCastro and Kress-Gazit, 2015) designed low-level controllers to guarantee the correctness of this high-level controller. However, their robot dynamics are much less challenging than our legged and armed locomotion dynamics, which inherently involve hybrid and under-actuated dynamics. The work in (Bhatia et al., 2010) proposed a multi-layered synergistic framework such that the low-level sampling-based planner communicates with the high-level discrete planner through a middle synergetic layer. This hierarchy facilitates the interaction between high-level and low-level planners. Hybrid robot dynamics are taken into account at the low-level planning. However, their work is not geared towards responding to sudden changes in the environment like we do.

## 2.3 Distributed Feedback Control Architecture

As a result of the increasing complexity of robotic control systems, such as human-centered robots (Paine et al., 2015; Sakagami et al., 2002; Diftler et al., 2011) and industrial surgical machines (Okamura, 2004), new system architectures, especially distributed control architectures (Kim et al., 2005; Santos and Silva, 2006), are often being sought for communicating with and controlling the numerous device subsystems. Often, these distributed control architectures manifest themselves in a hierarchical control fashion where a centralized controller can delegate tasks to subordinate local controllers (see Fig. 2.3). As it is known, communication between actuators and their low-level controllers can occur at high rates while communication between low- and high-level controllers occurs more slowly. The latter is further slowed down by the fact that centralized controllers tend to implement larger computational operations, for instance to compute system models or coordinate transformations online.

**Control Architectures with Time Delays**

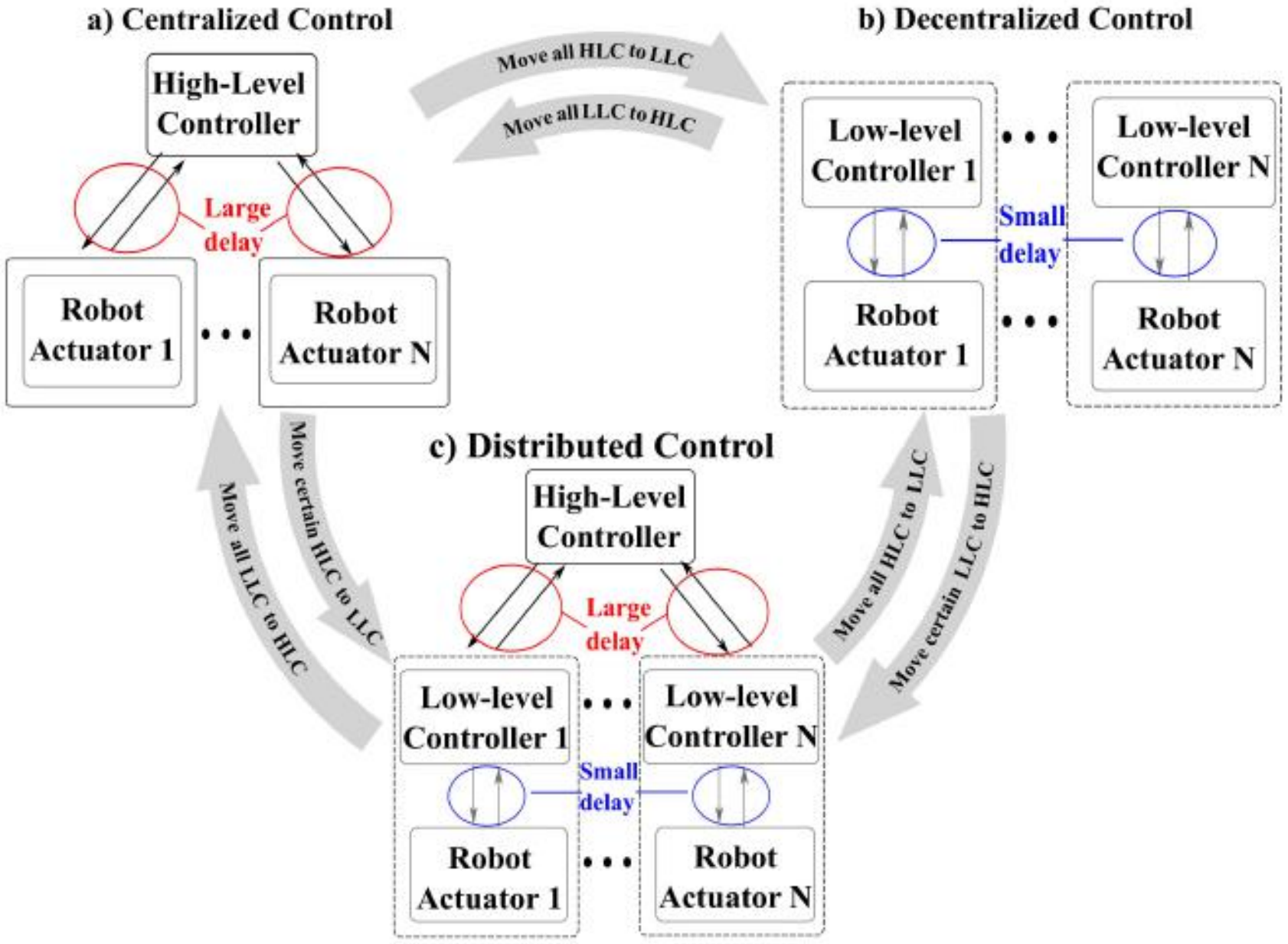

Figure 2.3: Depiction of various control architectures. Many control systems today employ one of the control architectures above: a) Centralized control with only high-level feedback controllers (HLCs); b) Decentralized control with only low-level feedback controllers (LLCs); c) Distributed control with both HLCs and LLCs, which is the focus of our study.

One concern of the control architecture design is that feedback controllers with large delays, such as the centralized controllers mentioned above, are less stable than those with small delays, such as locally embedded controllers. Without the fast servo rates of embedded controllers, the gains in centralized controllers can only be raised to limited values, decreasing their robustness to external disturbances (Lu and Yao, 2014) and unmodelled dynamics (Martin and George, 1981).

As such, why not remove centralized controllers altogether and implement all feedback processes at the low-level? Such operation might not always be possible. For instance, consider controlling the behavior of human-centered robots (e.g. highly articulated robots that interact with humans). Normally this operation is achieved by specifying the goals of some task frames such as the end effector coordinates. One established option is to create impedance controllers on those frames and transform the resulting control references to actuator commands via Operational Space transformations (Khatib, 1987b). Such a strategy requires the implementation of a centralized feedback controller which can utilize global sensing data, access the state of the entire system model, and compute the necessary models and transformations for control. Because of the aforementioned larger delays on high-level controllers, does this imply that high gain control cannot be achieved in human-centered robot controllers due to stability problems? It will be shown that this may not need to be the case. But for now, this delay issue is one of the reasons why various currently existing human-centered robots cannot achieve the same level of control accuracy that it is found in high-performance industrial manipulators. More concretely, this dissertation proposes a distributed impedance controller where only proportional (i.e., stiffness) position feedback is implemented in the high-level control process with slow servo updates. This process will experience the long latencies found in many modern centralized controllers of complex human-centered robots. At the same time, it contains global information of the model and the external sensors that can be used for Operational Space Control. For stability reasons, we proposes to implement the derivative position (i.e., damping) feedback part of the controller in low-level embedded actuator processes which can therefore achieve the desired high update rates.

### Distributed Control Architectures

Advances in distributed control technologies (Zeng and Chow, 2014; Wang and Li, 2011; Santos and Silva, 2006), have enabled the development of decentralized multiple input and multiple output systems such as humanoid systems and highly articulated robots (Sakagami et al., 2002; Diftler et al., 2011; Cheng et al., 2007). Distributed control architectures combine centralized processes with self-contained control units that are in proximity to actuators and sensors. In (Kim et al., 2005), servo motor controllers are used as sub-controllers coordinated via CAN communications by a central controller. As a result, computation burden on the central controller is reduced based on this distributed architecture. Recently, a distributed motion control system (Gu et al., 2010) has been developed to experimentally demonstrate the capabilities of real-time communications and synchronous tracking control. Analogous to human muscle actuation and neural systems, a bio-inspired distributed control infrastructure (Jantsch et al., 2010) reduces controller task complexity by off-loading parts of the controller into the robot's limb processors. Another typical distributed control architecture, related to neuroscience and robotics, is locomotor central pattern generators (CPGs) (Ijspeert, 2008; Ajallooeian et al., 2013). CPGs are modeled as distributed neural networks of multiple coupled oscillators to control articulated robot locomotion. This distributed architecture has the advantage of reducing time delays in the motor control loop (i.e., fast feedback loops through the spinal cord), in order to efficiently coordinate mechanical movements with rhythms.

## 2.4 Impedance Control of Series Elastic Actuators

The use of series elastic actuators (SEAs) in legged robots has become prevailing recent years thanks to SEAs' merits in environment interaction safety, impact absorption and force sensing.

### Cascaded SEA Control Architectures

In the rich control literature, adopting cascaded impedance control architectures for series elastic actuators (SEAs) have attracted increasing investigations over the last few years (Mosadeghzad et al., 2012; Hutter et al., 2013; Vallery et al., 2008; Tagliamonte et al., 2014). Compared with full-state feedback control (Albu-Schäffer et al., 2007; Hutter et al., 2013; De Luca et al., 2005), the cascaded control performs better when the controlled plant comprises relatively slow dynamics and fast dynamics simultaneously. In this case, the inner fast control loop can isolate the outer slow control loop from nonlinear dynamics, such as friction and stiction. The reason why this study focuses on the cascaded control structure is to simulate the distributed control structure for humanoid robots (Sakagami et al., 2002; Feng et al., 2014). To achieve passivity-based stability and robustness, this architecture nests feedback control loops (Mosadeghzad et al., 2012; Vallery et al., 2008), i.e., an inner-torque loop and an outer-impedance loop. The impedance loop is normally designed for the task-level control, such as Cartesian impedance control. Recently, the works in (Vallery et al., 2008; Tagliamonte et al., 2014) proposed embedding a motor velocity loop inside the torque feedback loop. This velocity feedback allows the use of integral gains to counteract static errors such as drivetrain friction while maintaining the system passivity. The authors in (Mosadeghzad et al., 2012) extensively studied the stability, passivity and performance for a variety of cascaded feedback control schemes, incorporating position, velocity and torque feedback loops.

**SEA Impedance Control**

Real world tasks normally necessitate a wide range of impedances. However, existing literature mainly focuses on the low SEA impedance for compliant control (Vallery et al., 2008; Tagliamonte et al., 2014). This imposes a huge restriction on its ability to achieve "stiff" tasks. As an example, legged robots without feedforward compensation require high impedance control to counteract the gravity effect. More importantly, the frequency characteristics of the SEA impedance is commonly ignored, but it does play a vital role in large-variability impedance tasks. To reduce this gap, this study (i) focuses on the maximum achievable SEA impedance range, i.e., Z-width (Colgate and Brown, 1994; Boaventura et al., 2013), with a single control architecture; (ii) quantifies the achievable impedance frequency range, which is defined as "Z-depth". Our analysis reveals that SEA impedance at the low frequency is constrained by the controller-defined active stiffness while its impedance at the high frequency is constrained by the passive spring stiffness.

Optimal controller design methodologies are ceaselessly sought by the robotics and mechatronics community. Our recent works in (Zhao et al., 2015b; Paine and Sentis, 2015) devised a critically-damped controller gain design criterion to accomplish high impedance for rigid actuators. However, inherent fourth-order dynamics make it challenging to design optimal controller gains for the SEA cascaded feedback loops. For the cascade control, a common routine is to tune the inner-loop gains first, followed by an outer-loop gain tuning. Indeed, this routine consumes substantial effort, and the optimality of the resulting gains cannot be quantified. A majority of existing results rely on empirical tuning (Pratt et al., 2004; Mosadeghzad et al., 2012). The work in (Vallery et al., 2008) designed controller gain ranges according to a passivity criterion. However, their gain parameters were coupled in the inequalities, which still leaves the controller gains nondeterministic. In Chapter 7, a fourth-order gain design criterion is proposed by simultaneously solving SEA impedance gains and torque gains. The "optimality" is quantified in the sense of phase-margin-based stability. Given this criterion, the designer only needs to specify one natural frequency parameter, and then all the impedance and torque gains are solved deterministically. This dimensionality reduction and automatic solving process provide considerable convenience for the SEA controller design.

Passivity-based controller gains for series elastic actuators were designed by the authors in (Vallery et al., 2008). However, for impedance feedback loops, that work only considers the stiffness feedback control, but the ignored damping feedback indeed plays a pivotal role, which will be thoroughly analyzed in Chapters 6-8. Damping-type impedance control was investigated in (Tagliamonte et al., 2014). However, it did not analyze the effects of feedback delays and filtering. Although these practical issues were tackled in (Vallery et al., 2008), the feedback delays are significantly smaller than those often found in serial communication buses.

## 2.5 Whole-Body Control

Whole-Body Control (WBC) has spawned a vast number of theoretical and implementation results to enable torque-controlled humanoid robots to perform complex full-body control tasks during the past few decades. The majority of WBC methods fall into two categories: null-space projection-based methods (Khatib, 1987a; Sentis et al., 2010; Ott et al., 2015; Moro et al., 2013; Mansard et al., 2009; Dietrich et al., 2015; Nakanishi et al., 2008), and optimization-based methods (Escande et al., 2014; Feng et al., 2015; Hopkins et al., 2016; Kuindersma et al., 2016; Herzog et al., 2016; Saab et al., 2013). Recently, increasing attention has been placed on hardware implementations such as Whole-Body Operational Space Control (WBOSC) on a point-feet bipedal robot for dynamic balancing (Kim et al., 2016a) and centroidal-momentum-based whole-body controllers (Hopkins et al., 2016; Dai et al., 2014; Herzog et al., 2016). However, real hardware performance frequently falls far short of the expectations, especially due to numerous practical issues arising from delayed communication and sensing processes (Zhao et al., 2015b; Fok et al., 2015), actuator dynamics (Hopkins et al., 2016; Paine et al., 2015; Ott et al., 2008) and unmodelled mechanical compliance (Kim et al., 2016a). These problems motivate our time-delayed WBOSC work which attempts to reduce the performance gap between theoretical foundations and real implementations.

In particular, our focus is to explicitly incorporate time delays and series elastic actuator (SEA) dynamics into the WBOSC formalism and reason about the conditions under which the closed-loop stability and passivity are preserved.

Null-space projection methods are widely used for whole-body controllers with prioritized multiple tasks (Sentis, 2007; Dietrich et al., 2015; Moro et al., 2013). However, these methods did not explore the passivity property, which is usually deteriorated by a non-conservative power transformation during the null-space projection procedure (Dietrich et al., 2016). The work in (Ott et al., 2015) achieved the output passivity conditionally to a positively invariant set containing all system states where certain prioritized tasks are guaranteed to be completed successfully. However, this approach is only applicable to specific scenarios. The authors in (Dietrich et al., 2016) devised an energy-tank-based control method to ensure the passivity of the projection-based null-space controllers at the sacrifice of control performance.

Optimization-based full body control has substantially emerged during the DRC competition. The work in Atlas (Kuindersma et al., 2016) resulted in a family of optimization algorithms for walking and whole system integration. The same team proposed a series of optimization methods at different levels, including active-set based quadratic program (QP) control (Kuindersma et al., 2014), sparse nonlinear optimization based whole-body planning (Dai et al., 2014), robust motion planing via convex optimization (Dai and Tedrake, 2016), and mixed-integer programming based foot placement planning (Deits and Tedrake, 2014). Another related work of (Feng et al., 2015) proposed a two-stage optimization comprising a high-level trajectory optimizer and a low-level QP controller based on free floating whole body inverse dynamics. The hierarchical QP-based inverse dynamics are proposed in (Escande et al., 2014; Saab et al., 2013; Wensing and Orin, 2013). Additionally, the authors of (Herzog et al., 2016) implemented this type of QP controllers on a real torque-controlled humanoid lower leg for the balancing task. Other QP-based controllers are in (Chao et al., 2016; Stephens and Atkeson, 2010; Hopkins et al., 2016; Hutter et al., 2014; Liu et al., 2015; Bouyarmane et al., 2012) and the references therein.

## 2.6 Passivity-based Control

As a robustness property, passivity (Van der Schaft, 2012; Stramigioli, 2001) is an essential requirement to ensure the coupled stability of robotic systems interacting with unknown dynamic environments (Focchi et al., 2016; Albu-Schäffer et al., 2007), networked control systems (Gao et al., 2007) and coordination control (Yin et al., 2016). In the haptic community, the pioneering work in (Colgate and Schenkel, 1997) proposed a necessary and sufficient condition for the passivity of sampled-data systems. However, the proposed passivity criterion is conservative in that the haptic display may remain stable even if violating the passivity condition (i.e., become "active"). The authors in (Buerger and Hogan, 2007) proposed a complementary stability to relax the assumptions in passivity stability. Along the same line of research in (Colgate and Schenkel, 1997), the authors in (Hulin et al., 2014) derived passivity and stability boundary conditions with time delays. Comparisons between passivity and stability are extensively investigated by analyzing the influence of various system parameters including sampling rate, time delay, physical damping and the mass. Additionally, a few other works have invested the effects of quantization, Coulomb friction and amplifier dynamics on system stability (Diolaiti et al., 2006). However, all the results above are restricted to single degree-of-freedom (DOF) systems, which severely limit their applicability to high-DOF humanoid robots.

Passivity-based controllers applicable to multi-DOF robotics and haptic systems were achieved in (Preusche et al., 2003; Hertkorn et al., 2010) along the line of time domain passivity approach (Hannaford and Ryu, 2002), which represents a class of less conservative passivity controller compared to the aforementioned frequency domain approaches. The authors in (Preusche et al., 2003) introduced a geometric condition to distribute the variable damping proposed previously for the single-DOF haptic interaction to six-DOF case. However, this work does not consider the dissipation capability of the kinematically redundant manipulator control, which is further exploited within the result (Ott et al., 2011) by prioritizing the damping in the null-space. Recently, the work in (Hertkorn et al., 2010) incorporated mild and constant time delays and direction-dependent inertia and validated experimental results on the DLR light weight arm robot. Other multi-DOF works include force bounding approach (Kim et al., 2015) and memory-based passivation approach (Ryu and Yoon, 2014).

In Albu-Schäffer and Ott's seminal work on multi-DOF Cartesian impedance control with flexible joint dynamics (Ott et al., 2008; Albu-Schäffer et al., 2007), joint torque feedback was physically interpreted as a scaling of the motor inertia such that the passivity of the closed-loop system is ensured. Nevertheless, these works are limited to manipulations. Applying passivity-based impedance controllers to full-body humanoid control was first proposed in (Hyon et al., 2007). This type of compliant controller designed gravity compensation and adaptation to unknown external forces. The desired ground reaction forces were distributed among a set of predefined contact points and directly mapped to the joint torque. Recently, the authors in (Henze et al., 2016) proposed a compliant multi-contact balancing controller while guaranteeing the overall system passivity. However, the associated tasks are not strictly hierarchical, and prioritized multi-task control based on null-space projection methods are still open to be explored to date. All the above works are limited in that none of them modeled or investigated the effect of time delays, as it is done in Chapters 6-8.

Stability under time delay (Gu et al., 2003) has been extensively studied in the teleoperation community (Lawrence, 1993; Sheridan, 1993). To overcome the instability caused by time delays, a conservative passivity based method (Anderson and Spong, 1989; Lee and Spong, 2006) was proposed to guarantee stable teleoperation performance. The results in (Lawrence, 1993; Hannaford, 1989) revealed a trade-off existing between stability and transparency in practical applications, which will also be explored in the WBOSC of this study. A prior work (Lee and Spong, 2006) employed Lyapunov-Krasovskii functionals to enforce energetic passivity of closed-loop nonlinear teleoperators by passifying the combination of the delayed communication and control blocks altogether. Moreover, this work explicitly used position feedback to prevent position drift during the master-slave coordination, which is a major problem of the conventional scattering-based teleoperation (Niemeyer and Slotine, 2004; Anderson and Spong, 1989; Lawrence, 1993). The delays are allowed to be unknown but have to be finite constant, which is a conservative assumption for the real hardware. In previous work (Hua and Liu, 2010) closely related to our study, asymmetric and time-varying delays were handled by proposing stability criteria of networked teleoperation systems via linear matrix inequality (LMI) techniques. However, their derived LMIs are time-invariant (i.e., no time-varying matrices involved). Recently, a few methods were proposed that dealt with time-varying LMIs by solving them online in (Guo and Zhang, 2014) and by convexifying matrices as a polytope of parametric matrices (Gahinet et al., 1996). In Chapter 8, the proposed LMI-based passivity criterion involves both time-varying delays and time-varying system matrices, and we solve it numerically.

# Chapter 3

# Phase-Space Planning of Dynamic Locomotion

In this chapter, we first present a general mathematical formulation of legged locomotion. Based on centroidal momentum dynamics, we propose a prismatic inverted pendulum model constrained on a parametric surface for single-contact case and a multi-contact model. Given these dynamics, we devise a hybrid phase-space planner, which incorporates a set of key components: (i) a step transition solver that enables dynamically tracking non-periodic keyframe states over various types of terrains, (ii) a robust hybrid automaton to formulate planning and control algorithms effectively, (iii) a lateral foot placement searching algorithm, and (iv) a steering direction model to control the robot's heading. Compared to other locomotion methods, we have a large focus on non-periodic gait generation. Such focus enables the proposed control method to track non-periodic keyframe states over various challenging terrains. Additionally, we explore the planning of a non-periodic leaping maneuver over a disjointed terrain.

---

This chapter incorporates the results from the following publications: (Zhao, Fernandez, and Sentis, 2016a,b; Zhao and Sentis, 2012; Zhao, Kim, Thomas, and Sentis, 2015a). An earlier version of the two-dimensional locomotion planner was proposed by Luis Sentis, and Section 3.1 is mainly contributed by Benito Fernandez.

## 3.1 Problem Definition

To begin with, this section focuses on a general control system formulation of legged robots and system normalization for phase-space design. Our goal is to devise phase-space manifolds as a robust metric to design and control phase-space trajectories for robust locomotion over rough terrain.

### 3.1.1 System Equations

Legged robots can be characterized as Multi-Input Multi-Output (MIMO) systems (Skogestad and Postlethwaite, 2007). Let us assume that the robot can be characterized by $n_j$ DOF for a robot having $n_j$ joints, $\boldsymbol{q} = [q_1, q_2, \ldots, q_{n_j}]^T \in \mathbb{R}^{n_j}$. Letting $\boldsymbol{x}(t) = [\boldsymbol{q}^T(t), \dot{\boldsymbol{q}}^T(t)]^T \in \mathbb{R}^n$, be the state-space vector ($n = 2n_j$), $\boldsymbol{u}(t) \in \mathbb{R}^m$, represents the control input vector (generalized torques and forces), and defining $\boldsymbol{f}(\boldsymbol{x}(t))$, $\boldsymbol{g}(\boldsymbol{x}(t))$, and $\boldsymbol{h}(\boldsymbol{x}(t))$ in the obvious manner, the mechanical model is expressed in state variable form as

$$\dot{\boldsymbol{x}}(t) = \boldsymbol{f}(\boldsymbol{x}(t)) + \boldsymbol{g}(\boldsymbol{x}(t))\boldsymbol{u}(t) + \boldsymbol{J}_d(\boldsymbol{x}(t))\boldsymbol{d}(t), \tag{3.1a}$$

$$\boldsymbol{y}(t) = \boldsymbol{h}(\boldsymbol{x}(t)), \tag{3.1b}$$

where $\boldsymbol{d}(t)$ represent the generalized external disturbance forces, and $\boldsymbol{J}_d(\boldsymbol{q}(t))$ is the disturbance distribution matrix. The output vector, $\boldsymbol{y}(t) = [y_1, y_2, \ldots, y_p]^T \in \mathbb{R}^p$ is generated by $\boldsymbol{h}(\boldsymbol{x}(t))$, that may represent positions and/or velocities in the task space. Without loss of generality, let us consider systems in the normal form in next subsection, where $\boldsymbol{h}(\cdot)$ is at least $\mathcal{C}^r$, where $r$ is the relative order of the output. The disturbances and modeling errors satisfy the matching conditions (Fernández-Rodríguez, 1988).

### 3.1.2 System Normalization for Phase-Space Design

General robotic systems are not in normal form, but we can transform them by finding the relative order of the output derivatives that are explicitly controllable. Each of the outputs $y_i$ in Eq. (3.1) has a relative order $r_i$, defined by the smallest derivative order where the control appears,

$$y_i^{[k]} = \frac{d^k y_i}{dt^k} = \mathcal{L}_{\boldsymbol{f}}^{k}(h_i(\boldsymbol{x})) + \mathcal{L}_{\boldsymbol{g}}(\mathcal{L}_{\boldsymbol{f}}^{k-1}(h_i(\boldsymbol{x})))\boldsymbol{u}, \quad (3.2a)$$

$$\mathcal{L}_{\boldsymbol{g}}(\mathcal{L}_{\boldsymbol{f}}^{k-1}(h_i(\boldsymbol{x}))) = 0 \quad \text{for} \quad 0 \le k < r_i, \quad (3.2b)$$

$$y_i^{[r_i]} = \mathcal{L}_{\boldsymbol{f}}^{r_i}(h_i(\boldsymbol{x})) + \mathcal{L}_{\boldsymbol{g}}(\mathcal{L}_{\boldsymbol{f}}^{r_i-1}(h_i(\boldsymbol{x})))\boldsymbol{u}, \quad (3.2c)$$

$$\mathcal{L}_{\boldsymbol{g}}(\mathcal{L}_{\boldsymbol{f}}^{r_i-1}(h_i(\boldsymbol{x}))) \neq 0 \qquad \forall \boldsymbol{x} \in \mathbb{S}_i \subset \mathbb{R}^n, \quad (3.2d)$$

where $\mathcal{L}_{\boldsymbol{f}}^{0}(h_i(\boldsymbol{x})) = h_i(\boldsymbol{x})$, $\mathcal{L}_{\boldsymbol{f}}(\boldsymbol{h})$ and $\mathcal{L}_{\boldsymbol{g}}(\boldsymbol{h})$ are the directional Lie derivatives of function $\boldsymbol{h}(\boldsymbol{x})$ in the directions of $\boldsymbol{f}(\boldsymbol{x})$ and $\boldsymbol{g}(\boldsymbol{x})$ respectively (Isidori, 1985), and $\mathbb{S}_i$ is the output-controllable subspace, where the Lie derivative in Eq. (3.2) does not vanish,

$$\mathbb{S}_i = \Big\{ \boldsymbol{x} \in \mathbb{R}^n \;\Big|\; \mathcal{L}_{\boldsymbol{g}}(\mathcal{L}_{\boldsymbol{f}}^{r_i-1}(h_i(\boldsymbol{x}))) \neq 0 \Big\}. \quad (3.3)$$

The relative order tells us that the $r_i^{\text{th}}$-derivative of output $y_i$ can be explicitly controlled. The region where $\mathbb{S}_i$ vanishes, entails the system looses relative order and hence the $r_i^{\text{th}}$-derivative is no longer controllable (at least explicitly). For a controllable system, $r_i \le n$. Following the normalization procedure, we get the output controllable subspace defined by $\boldsymbol{\xi}$,

$$\xi_{i,1} = y_i = h_i(\boldsymbol{x}) = \mathcal{L}_{\boldsymbol{f}}^0(h_i(\boldsymbol{x})), \tag{3.4}$$

$$\cdots$$

$$\xi_{i,j} = y_i^{[j-1]} = \dot{\xi}_{i,j-1} = \mathcal{L}_{\boldsymbol{f}}^{j-1}(h_i(\boldsymbol{x})) \quad \text{for} \quad 1 < j < r_i, \tag{3.5}$$

$$\cdots$$

$$y_i^{[r_i]} = \dot{\xi}_{i,r_i} = \mathcal{L}_{\boldsymbol{f}}^{r_i}(h_i(\boldsymbol{x})) + \mathcal{L}_{\boldsymbol{g}}(\mathcal{L}_{\boldsymbol{f}}^{r_i-1}(h_i(\boldsymbol{x})))\boldsymbol{u}(t). \tag{3.6}$$

The output space variables, $\boldsymbol{\xi}_i = [\xi_{i,1}, \xi_{i,2}, \ldots, \xi_{i,r_i-1}]^T \in \mathbb{R}^{r_i}$ represent the phase-space for the $i$-th output. For instance, the output phase-space for locomotion control could be chosen to be the robot's center of mass. We can concatenate all $\boldsymbol{\xi}_i$, $\forall i = 1, 2, \ldots, m$ into a single phase-space vector $\boldsymbol{\xi} = [\boldsymbol{\xi}_1^T, \boldsymbol{\xi}_2^T, \ldots, \boldsymbol{\xi}_m^T]^T \in \mathbb{R}^r$, where, $r = \sum r_i$.

**Definition 3.1 (Phase-Space Manifold).** *For phase-space motion, we define a phase-space manifold $\mathcal{M}_i$ for each task-space output $y_i$ in terms of its phase-space vector $\boldsymbol{\xi}_i$,*

$$\mathcal{M}_i = \Big\{ \boldsymbol{\xi}_i \in \mathbb{R}^{r_i} \subset \mathbb{R}^n \;\Big|\; \sigma_i \triangleq \sigma_i(\boldsymbol{\xi}_i) = 0 \Big\}, \tag{3.7}$$

*where $\sigma_i$ is referred to as the $i^{\text{th}}$ element of deviation vector, which measures the deviation distance from the manifold $\mathcal{M}_i$ using a Riemannian metric.*

More details are shown in Appendix A. In order to be able to control this deviation, the order of the manifold is one less than the relative order of the $i^{\text{th}}$-output, i.e., $r_i - 1$. For most legged robots (not considering actuator dynamics), the relative order is $r = 2$.

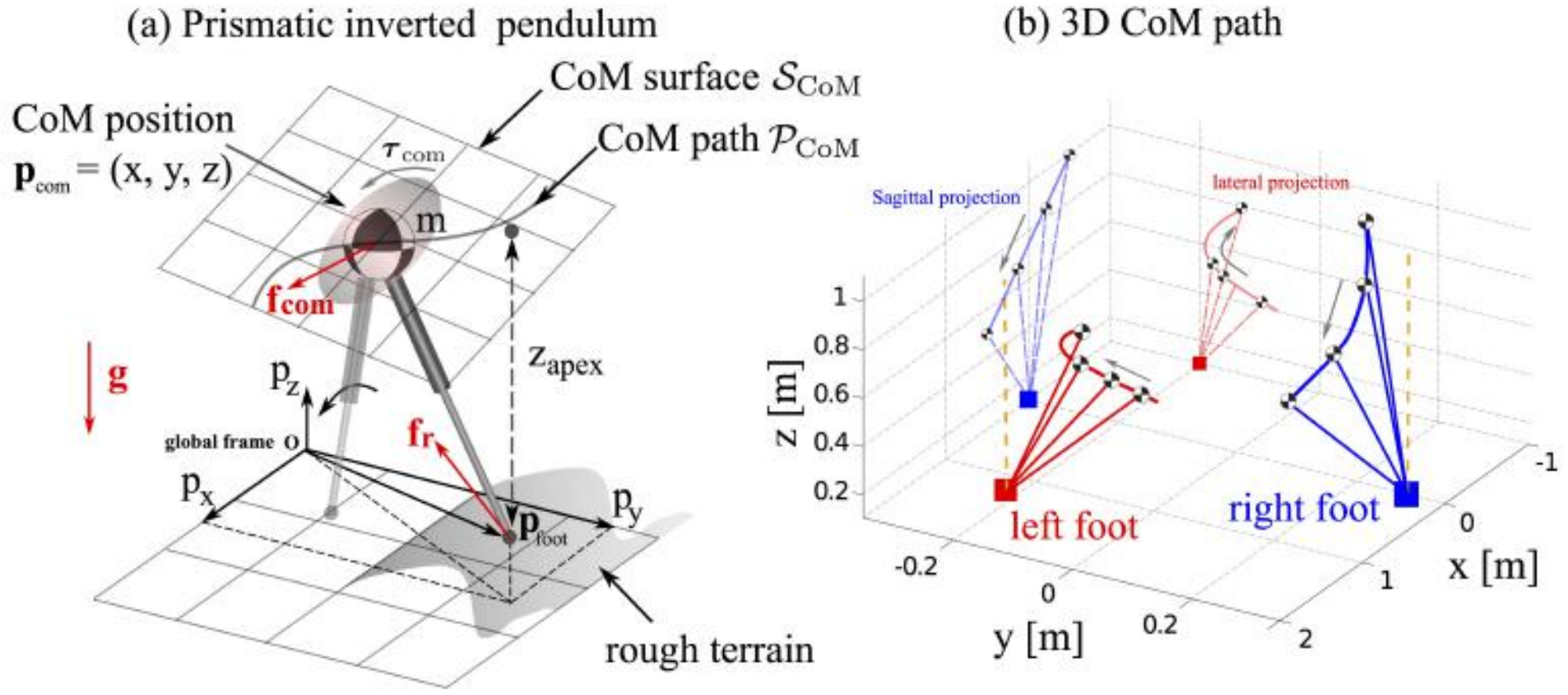

Figure 3.1: 3D prismatic inverted pendulum model. (a) We define a prismatic inverted pendulum model with all of its mass located at its base while equipping it with a flywheel to generate moments. We restrict the movement of the center of mass to 3D planes (surfaces) $\mathcal{S}_{\text{CoM}}$. Three red arrows represent the CoM inertial force $\boldsymbol{f}_{\text{com}}$, the ground reaction force $\boldsymbol{f}_r$ and the gravity force $m\boldsymbol{g}$, respectively. These forces satisfy $\boldsymbol{f}_{\text{com}} = \boldsymbol{f}_r - m\boldsymbol{g}$. (b) shows motions of pendulum dynamics restricted to a 3D plane. Note that, our study assumes a time-varying leg length. The apex height $z_{\text{apex}}$ is 1 m in this simulation.

## 3.2 Center-of-Mass Parametric Surfaces

The rigid body dynamics of point-foot bipedal robots during single contact resemble those of a simple inverted pendulum model (see Fig. 3.1), as observed by studies in dynamic human walking (Cavagna and Margaria, 1966; Kuo, 2002, 2007). In our case, our model consists of a prismatic massless joint with all the mass concentrated on the hip position (Kajita et al., 2002, 2003a; Koolen et al., 2012), defined as the 3D CoM position, $\boldsymbol{p}_{\text{com}} = (x, y, z)^T$, and a flywheel spinning around it, with orientation angles $\boldsymbol{R} = (\phi, \theta, \psi)^T$. Various human walking (Kuo and Zajac, 1992) and balancing (Winter, 1995) studies emphasize that controlling the centroidal angular momentum can improve CoM tracking, balancing and recovering from disturbances. This observation has been recently achieved to dynamic robot locomotion (Pratt et al., 2006; Komura et al., 2005; Yun and Goswami, 2011). The objective of locomotion is to move the robot's CoM along a certain path from point A to B over a terrain. As such, we first specify a 3D surface, $\mathcal{S}_{\text{CoM}}$, where the CoM path will exist, which in general, may have the following implicit form,

$$\mathcal{S}_{\text{CoM}} \;=\; \Big\{ \boldsymbol{p}_{\text{com}} \in \mathbb{R}^3 \;\;|\;\; \psi_{\text{CoM}}(\boldsymbol{p}_{\text{com}}) \;=\; 0 \Big\}. \tag{3.8}$$

This surface can be specified in various ways, such as via piecewise arc geometries (Mordatch et al., 2010; Srinivasan and Ruina, 2006), and spline functions (Morisawa et al., 2005; Englsberger et al., 2015a). Once the controller is designed, the CoM will follow a concrete trajectory $\mathcal{P}_{\text{CoM}}$ (as shown in Fig. 3.1), which we specify via piecewise splines described by a progression variable, $\zeta \in [\zeta_{j-1}, \zeta_j]$, for the $j^{\text{th}}$ path manifold, i.e.

$$\mathcal{P}_{\text{CoM}} = \bigcup_j \mathcal{P}_{\text{CoM}_j} \subseteq \mathcal{S}_{\text{CoM}}, \quad \mathcal{P}_{\text{CoM}_j} = \Bigg\{ \boldsymbol{p}_{\text{com}_j} \in \mathbb{R}^3 \;\Bigg|\; \boldsymbol{p}_{\text{com}_j} = \sum_{k=0}^{n_p} \boldsymbol{a}_{jk} \zeta^k \Bigg\},$$

where $n_p$ is the order of the spline degree. The progression variable $\zeta$ is therefore the arc length along the CoM path acting as the Riemannian metric for distance. Each $\boldsymbol{a}_{jk} \in \mathbb{R}^3$ is the coefficient vector of $k^{\text{th}}$ order. To guarantee the spline smoothness, $\boldsymbol{p}_{\text{com}}$ requires the connection points, i.e. the knots at progression instant $\zeta_j$, to be $\mathcal{C}^{n_p-1}$ continuous,

$$\boldsymbol{p}^{[l]}_{\text{com}_j}(\zeta_j) = \frac{d^l \boldsymbol{p}_{\text{com}_j}}{d\zeta^l}(\zeta_j) = \boldsymbol{p}^{[l]}_{\text{com}_{j+1}}(\zeta_j), \quad \forall\, 0 \le l \le n_p - 1. \tag{3.9}$$

The purpose of introducing the CoM manifold $\mathcal{S}_{\text{CoM}}$ is to constrain CoM motions on surfaces that are designed to conform to generic terrains while allowing free motion within this surface. Following a concrete CoM path is achieved by selecting proper control inputs as we will see further down. The CoM path manifold, $\mathcal{P}_{\text{CoM}}$ (embedded in $\mathcal{S}_{\text{CoM}}$), can be represented in the phase-space, $\boldsymbol{\xi}$. We call this representation as the *phase-space manifold* and define it as,

$$\mathcal{M}_{\text{CoM}} = \bigcup_j \mathcal{M}_{\text{CoM}_j}, \quad \mathcal{M}_{\text{CoM}_j} = \Big\{ \boldsymbol{\xi} \in \mathbb{R}^6 \;\; | \;\; \sigma_j(\boldsymbol{\xi}) \; = \; 0 \Big\}, \tag{3.10}$$

which is the core manifold used in our phase-space planning and control framework. The function $\sigma_j(\boldsymbol{\xi})$ is a measure of the Riemannian distance from the phase-space manifold.

**Remark 3.1.** *Instead of using high-dimensional generalized coordinates, this study chooses the center of mass phase-space of the robot as the output space. Existing promising techniques for dimensionality reduction include, for instance, differential flatness (Liu et al., 2012; Sreenath and Kumar, 2013), which reduces high dimensional nonlinear systems to a controller design problem involving a chain of integrators, and virtual constraint and partial zero dynamics (Ames et al., 2015) to constrain the high-dimensional robot dynamics to a hybrid-invariant and low-dimensional manifold.*

## 3.3 Centroidal Momentum Model

The centroidal momentum dynamics can be characterized via formulating the dynamic balance of moments around the system's centroidal point.

$$\dot{\boldsymbol{l}} = m\ddot{\boldsymbol{p}}_{\text{com}} = \sum_i^{N_c} \boldsymbol{f}_{r_i} - m\boldsymbol{g}, \quad \dot{\boldsymbol{k}} = \boldsymbol{\tau}_{\text{com}} = \sum_i^{N_c} (\boldsymbol{p}_{\text{foot}_i} - \boldsymbol{p}_{\text{com}}) \times \boldsymbol{f}_{r_i} + \boldsymbol{\tau}_i, \tag{3.11}$$

where $\boldsymbol{l} \in \mathbb{R}^3$ and $\boldsymbol{k} \in \mathbb{R}^3$ represent the centroidal linear and angular momenta, respectively. $\boldsymbol{f}_{r_i} \in \mathbb{R}^3$ is the $i^{\text{th}}$ ground reaction force, $m$ is the total mass of the robot, $\boldsymbol{g} = (0, 0, g)^T$ corresponds to the gravity field, $\boldsymbol{f}_{\text{com}} = m\ddot{\boldsymbol{p}}_{\text{com}} = m(\ddot{x}, \ddot{y}, \ddot{z})^T$ is the vector of center-of-mass inertial forces. The first equation above represents the rate of change of linear momentum being equal to the total action of linear contact forces minus gravitational forces. $\boldsymbol{\tau}_{\text{com}} = (\tau_x, \tau_y, \tau_z)^T$ is the vector of angular moments of the modeled flywheel attached to the inverted pendulum. $\boldsymbol{p}_{\text{foot}_i} = (p_{i,x}, p_{i,y}, p_{i,z})^T$ is the position of the $i^{\text{th}}$ foot contact contact. $\boldsymbol{\tau}_{r_i} \in \mathbb{R}^3$ is the $i^{\text{th}}$ contact torque vector. The second equation above represents the rate of change of angular momentum being equal to the sum of the torques generated by total action of contact wrenches projected to the CoM. In our case, $\boldsymbol{\tau}_{r_i} = \mathbf{0}$ due to having point-foot contacts.

Given this centroidal momentum model, let us focus on designing dynamic locomotion modes which will be composed sequentially to generate the overall hybrid phase-space trajectories.

### 3.3.1 Single-Contact Dynamics

Aside from the CoM path surface previously described, pendulum dynamics can be characterize via formulating the dynamic balance of moments of the pendulum system. For our single contact scenario, the sum of moments, $\mathbf{m}_i$, with respect to the global reference frame (see Fig. 3.1) is

$$\sum_i \mathbf{m}_i = -\boldsymbol{p}_{\text{foot}} \times \boldsymbol{f}_r + \boldsymbol{p}_{\text{com}} \times \Big(\boldsymbol{f}_{\text{com}} + m\,\boldsymbol{g}\Big) + \boldsymbol{\tau}_{\text{com}} = 0, \tag{3.12}$$

The system's linear force equilibrium can be formulated as $\boldsymbol{f}_r = \boldsymbol{f}_{\text{com}} + m\,\boldsymbol{g}$, allowing us to simplified Eq. (3.12) to:

$$\Big(\boldsymbol{p}_{\text{com}} - \boldsymbol{p}_{\text{foot}}\Big) \times (\boldsymbol{f}_{\text{com}} + m\,\boldsymbol{g}) = -\boldsymbol{\tau}_{\text{com}}. \tag{3.13}$$

For our purposes, we consider only the class of prismatic inverted pendulums whose center of mass is restricted to a path surface $\mathcal{S}_{\text{CoM}}$ as indicated in Eq. (3.8). A detailed definition of this 3D plane will be presented in Eq. (3.16). This type of model with varying height is called the Prismatic Inverted Pendulum Model (PIPM) (Zhao et al., 2016c; Zhao and Sentis, 2012).

It is observed that the CoM behavior during human walking approximately follows the slope of terrains (Zhao and Sentis, 2012; Zhao et al., 2016d). Based on this observation, we design piecewise CoM planes approximating terrain slopes and adjust the CoM planes according to the acceleration or deceleration phases. A variety of CoM trajectory design methods have been proposed over the years. The Capture Point method in (Koolen et al., 2012) assumes a constant CoM height. Closely related to us, (Kajita et al., 2003a) constrains the CoM motion to a 3D plane. However, our focus is on robust hybrid control. Designing CoM trajectories with a varying CoM height is described in (Englsberger et al., 2015b). The work described in (Ramos and Hauser, 2015) proposed a Nonlinear Inverted Pendulum model and the CoM path is extended to a parabola, but it focuses on planar locomotion.

Considering as our output state the CoM positions, $\boldsymbol{p}_{\text{com}}$, the state space $\boldsymbol{\xi} = (\boldsymbol{p}_{\text{com}}^T, \dot{\boldsymbol{p}}_{\text{com}}^T)^T = (x, y, z, \dot{x}, \dot{y}, \dot{z})^T \in \mathcal{X} \subseteq \mathbb{R}^6$ is the phase-space vector, where $\Xi$ is the set of admissible CoM positions and velocities. Then from Eq. (3.13), the prismatic inverted pendulum model can be defined below

**Definition 3.2** (**Prismatic inverted pendulum model**). *The prismatic inverted pendulum model for a walking step, indexed by a discrete variable $q$, is represented by the following control system*

$$\dot{\boldsymbol{\xi}} = \boldsymbol{\mathcal{F}}(q, \boldsymbol{\xi}, \boldsymbol{u}) = \begin{pmatrix} \dot{x} \\ \dot{y} \\ \dot{z} \\ \underbrace{\omega_q^2 (x - x_{\text{foot}_q}) - \frac{\omega_q^2}{mg}(\tau_y + b_q \tau_z)}_{A} \\ \underbrace{\omega_q^2 (y - y_{\text{foot}_q}) - \frac{\omega_q^2}{mg}(\tau_x + a_q \tau_z)}_{B} \\ a_q A + b_q B \end{pmatrix}, \tag{3.14}$$

*where the phase-space asymptotic slope is defined as*

$$\omega_q = \sqrt{\frac{g}{z_{\text{apex}_q}}}, \text{ with } z_{\text{apex}_q} = (a_q \cdot x_{\text{foot}_q} + b_q \cdot y_{\text{foot}_q} + c_q - z_{\text{foot}_q}), \tag{3.15}$$

*where $g$ is the gravity constant, $a_q$ and $b_q$ are the slope coefficients while $c_q$ is the constant bias for the linear CoM path surfaces that we consider (see Fig. 3.2), i.e.*

$$\mathcal{S}_{\text{CoM}_q} : \Big\{ (x, y, z) \in \mathbb{R}^3 \;\Big|\; \psi_{\text{CoM}_q}(x, y, z) = z - a_q x - b_q y - c_q = 0 \Big\}. \tag{3.16}$$

Detailed derivations of Eq. (3.14) are provided in Appendix B.1. $z_{\text{apex}_q}$ is the height of the CoM at the apex of its sagittal path "$x$ direction" as shown in Fig. 3.1 such that it corresponds to the vertical distance between the CoM and the location of the foot contact at the instant when the CoM is on the top of the foot location. $\boldsymbol{\mathcal{F}}$ represents a vector field of inverted pendulum dynamics, which is assumed to be infinitely continuously and differentiable (i.e., $\mathcal{C}^\infty$) in the domain $\mathcal{D}(\boldsymbol{\xi})$ and globally Lipschitz in $\Xi$, given fixed control inputs. In general, there will be a hybrid control policy $\boldsymbol{u} = \boldsymbol{\pi}(q, \boldsymbol{\xi})$ defined by the control variables $\boldsymbol{u} = (\omega_q, \boldsymbol{\tau}_{\text{com}_q}, \boldsymbol{p}_{\text{foot}_q})^T \in \mathcal{U}$, where $\mathcal{U}$ is an open set of admissible control values. The sets $\Xi$ and $\mathcal{U}$ are assumed to be compact. Our design of foot placement algorithms further into the study will guarantee the tracking of keyframe states within a specified tolerance.

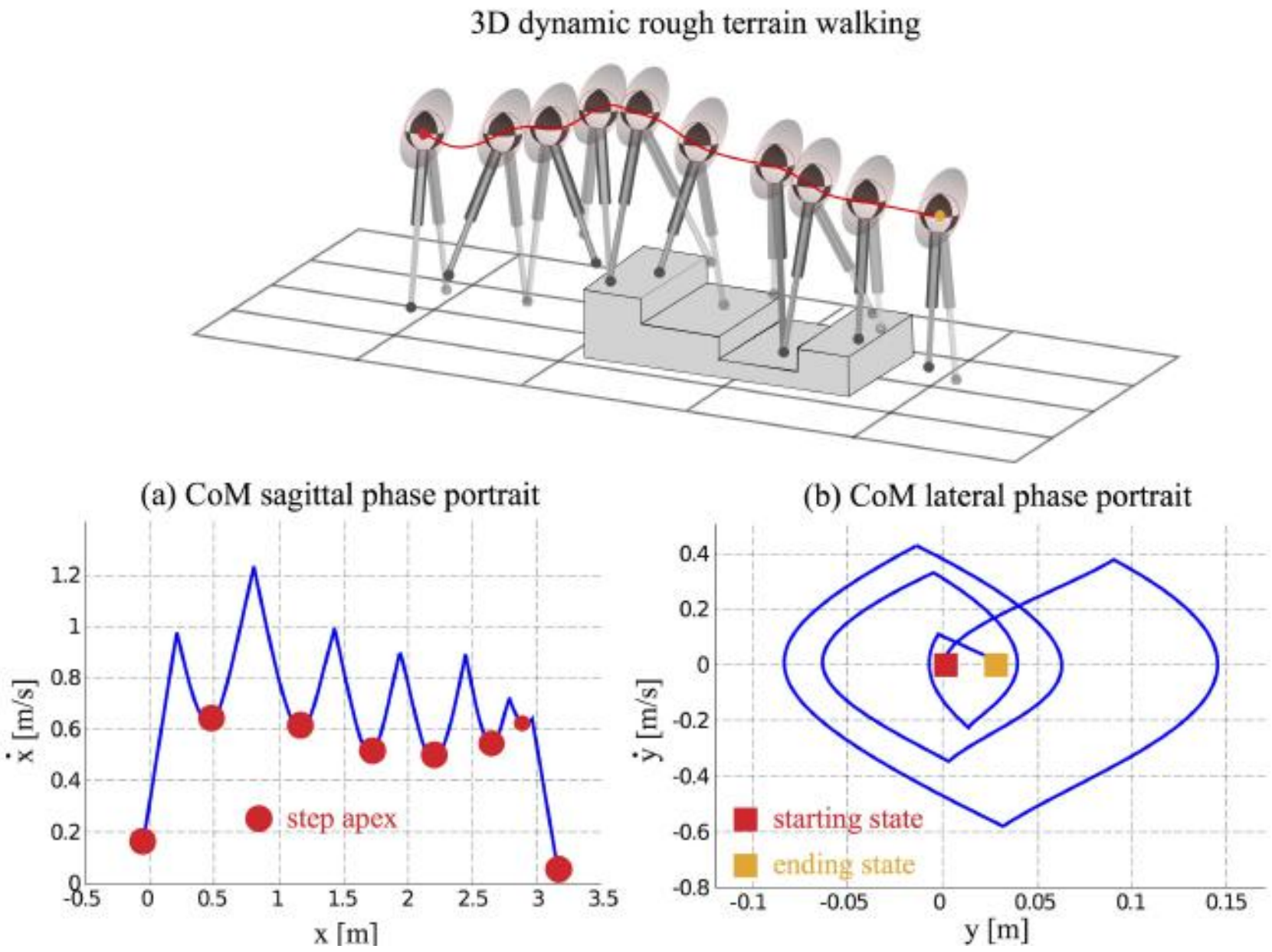

Figure 3.2: 3D phase-space planning. Given step apex conditions, single contact dynamics generate the valley profiles shown in (a). (b) depicts a similar strategy in the lateral plane. However, since foot transitions have already been determined, what is left is to determine foot lateral positions. This is done so the lateral CoM behavior shown in (b) follows a semi-periodic trajectory that is bounded within a closed region.

**Definition 3.3** (**Sagittal and lateral apex**). *The sagittal apex occurs when the projection of the CoM is equal to the location of the foot contact in the system's sagittal axis. The lateral apex is defined as the CoM lateral position when the sagittal apex occurs.*

From a physical perspective, the continuous control input $\omega_q$ in Eq. (3.14) is equivalent to modulating the leg force magnitude, since it can directly change the CoM accelerations by modulating the apex height $z_{\text{apex}}$ as shown in Eq. (3.15). Using piecewise-linear CoM planes can cause sharp changes on the phase trajectories when the center-of-mass switches among multiple steps. To mitigate this problem, we will employ multi-contact strategies to smooth CoM trajectories. Multi-contact consists on controlling the CoM behavior when two feet are in the ground.

### 3.3.2 Multi-Contact Dynamics

We introduce a multi-contact model and briefly present how to modulate the internal tension force such that the friction cone constraints are satisfied. Different from the 2D planar scenario in (Sentis and Slovich, 2011), this study focuses on the 3D case. Based on the virtual linkage model in (Sentis et al., 2010), the dual contact dynamics can be described by a multi-contact/grasp matrix as,

$$\begin{pmatrix} \boldsymbol{f}_{\text{com}} + m\boldsymbol{g} \\ \boldsymbol{\tau}_{\text{com}} \\ f_{\text{int}} \end{pmatrix} = [\boldsymbol{G}]_{7\times 6} \begin{pmatrix} \boldsymbol{f}_{r_{\text{left}}} \\ \boldsymbol{f}_{r_{\text{right}}} \end{pmatrix} \tag{3.17}$$

where $f_{\text{int}}$ represents the internal force along the line of dual feet contact points. $[\boldsymbol{G}]_{7\times 6}$ is the multi-contact/grasp matrix defined as

$$[\boldsymbol{G}]_{7\times 6} = \begin{pmatrix} [\boldsymbol{W}_{\text{com}}]_{6\times 6} \\ [\boldsymbol{W}_{\text{int}}]_{1\times 6} \end{pmatrix} \tag{3.18}$$

By inverting Eq. (3.17), we can solve the ground reaction forces for given center of mass inertial forces and moments, and the internal force

$$\begin{pmatrix} \boldsymbol{f}_{r_{\text{left}}} \\ \boldsymbol{f}_{r_{\text{right}}} \end{pmatrix} = [\boldsymbol{G}]^{+}_{7\times 6} \begin{pmatrix} \boldsymbol{f}_{\text{com}} + m\boldsymbol{g} \\ \boldsymbol{\tau}_{\text{com}} \\ f_{\text{int}} \end{pmatrix} = \boldsymbol{G}_f(\boldsymbol{f}_{\text{com}} + m\boldsymbol{g}) + \boldsymbol{G}_\tau \boldsymbol{\tau}_{\text{com}} + \boldsymbol{G}_{\text{int}} f_{\text{int}} \tag{3.19}$$

Matrices $\boldsymbol{W}_{\text{com}}, \boldsymbol{W}_{\text{int}}, \boldsymbol{G}_f, \boldsymbol{G}_\tau$ and $\boldsymbol{G}_{\text{int}}$ are outlined in (Sentis and Slovich, 2011). Different from the method of simultaneously controlling CoM and internal force behaviors described in (Sentis et al., 2010), this study implements the following procedure: (i) we first design a multi-contact phase trajectory between single contact phases that satisfies CoM position, velocity, and acceleration boundary conditions. The duration of the multi-contact phase and boundary velocities can be chosen by the designer. A similar multi-contact transition strategy, named "Continuous Double Support" (CDS) trajectory generator, was proposed in (Englsberger et al., 2014) to achieve smooth "Enhanced Centroidal Moment Pivot" (eCMP) and leg force profiles. We have ourselves used this strategy as shown in Appendix B.3. (ii) Using Eq. (3.19) and the CoM inertial wrench trajectory, we solve for the internal forces such that they satisfy friction constraints.

## 3.4 Nominal Trajectory Generation

Contact-based locomotion planning involves two components: a sequence of discrete contact configurations and a resulting continuous trajectory concatenating this contact sequence. Every time when contact switches, the robot switches its continuous dynamics. We will first focus on the generation of trajectories in the sagittal plane of the robot's walking reference. Sagittal dynamics are represented - ignoring for simplicity, the discrete variable, $q$, - in the first and fourth row of the system of Eq. (3.14), i.e.

$$\dot{\boldsymbol{x}} = \boldsymbol{\mathcal{F}}_{\boldsymbol{x}}(\boldsymbol{x}, \boldsymbol{u}_{\boldsymbol{x}}) = \begin{pmatrix} \dot{x} \\ \omega^2(x - x_{\text{foot}}) - \frac{\omega^2}{mg}(\tau_y + b_q \tau_z) \end{pmatrix}. \tag{3.20}$$

This system would be fully controllable if its continuous control inputs, $\boldsymbol{u}_{\boldsymbol{x}} = (\omega, \tau_y, \tau_z, x_{\text{foot}})^T$ were unconstrained. However, their limited range urges us to first consider the motion trajectories under nominal (i.e. open loop) values. As we previously motivated in Eq. (3.16), the path manifold, $\mathcal{S}_{\text{CoM}}$ is defined a priori to conform to the terrains via simple heuristic methods previously described in (Zhao and Sentis, 2012; Sentis and Slovich, 2011). From Eq. (3.15), once the path manifold is defined and for known contact locations, the set of phase-space asymptotic slopes, $\omega$, is also known as shown in Eq. (3.15). For simplicity, the nominal flywheel moments are designed to be null, i.e. $\tau_y = 0$. Under these considerations, the following algorithm produces nominal phase-space trajectories of the system's center of mass in the sagittal direction of reference:

**Algorithm 1.** Nominal Phase-Space Trajectory Generation.

**Input:**

**(i)**: $\mathcal{S}_{\text{CoM}} \leftarrow \{\mathcal{S}_{\text{CoM}_q} : [\zeta_{q-1}, \zeta_q] \rightarrow \mathbb{R}^3,\ \forall q = 1, \ldots, N\}$

**(ii)**: $x_{\text{foot}} \leftarrow \{x_{\text{foot}_1}, x_{\text{foot}_2}, \ldots, x_{\text{foot}_N}\}$

**(iii)**: $\dot{x}_{\text{apex}} \leftarrow \{\dot{x}_{\text{apex}_1}, \dot{x}_{\text{apex}_2}, \ldots, \dot{x}_{\text{apex}_N}\}$

**(iv)**: $(\tau_y(t), \tau_z(t)) \leftarrow \mathbf{0}$

**Operation:**

**(i)**: $\omega \coloneqq \{\omega_1, \omega_2, \ldots, \omega_N\}$ is assigned via Eqs. (3.15) and (3.16)

**(ii)**: $(x(t), \dot{x}(t), \ddot{x}(t)) \leftarrow \text{PIPM}(\omega, \tau_y(t), \tau_z(t), x_{\text{foot}})$ via Eq. (3.20) and the analytical solution proposed in Eq. (4.2)

**Output:**

Phase-space trajectories $\mathcal{M}_{\text{CoM}} := \bigcup_q \mathcal{M}_{\text{CoM}_q}$

The reader should refer to Fig. 4.7 to see the end-to-end planning and control process of the proposed locomotion methodology. It is specially important to understand that the desired CoM surfaces, nominal foot positions, keyframe states, and zero flywheel torques are provided a priori by the designer. A similar algorithm could be written to generate motion in the lateral CoM direction via Eq. (3.14). Here, $\dot{x}_{\text{apex}}$, represents desired apex velocities, PIPM represents the prismatic inverted pendulum model on the parametric surfaces defined in Eq. (3.14), used to derive CoM accelerations. Trajectories for multiple steps of a locomotion sequence on rough terrain are simulated using this process in Fig. 3.2.

## 3.5 Hybrid Phase-Space Planning

In this section, we propose a hybrid robust automaton (Branicky et al., 1998; Frazzoli, 2001; Lygeros et al., 1999) with the following key features: I) an invariant set and a recoverability set to characterize control robustness, i.e., the bundle of attractiveness, and II) a non-periodic step transition strategy based on the previously described phase-space trajectories. The hybrid automaton governs the planner's behavior across multiple walking steps and as such constitutes the theoretical core of our proposed phase-space locomotion planning framework. We continue our focus on sagittal plane dynamics first, then extend the planner to all directions. For practical purposes we will use the symbol $\boldsymbol{x} = \{x, \dot{x}\}$ to describe the sagittal state space associated with CoM dynamics. Note that this symbol represents now the output dynamics outlined in Eq. (3.4) instead of the robot plant of Eq. (3.1). Eq. (3.10) can thus be re-considered in the output space as $\mathcal{M}_{\mathrm{CoM}_q} = \{\boldsymbol{x} \in \mathcal{X} \mid \sigma_q(\boldsymbol{x}) = 0\}$ where $\sigma_q$ (the normal distance) representing the deviation from the manifold $\mathcal{M}_{\mathrm{CoM}_q}$.

**Definition 3.4 (Invariant bundle).** *A set $\mathcal{B}_q(\epsilon)$ is an invariant bundle if, given $\boldsymbol{x}_{\zeta_0} \in \mathcal{B}_q(\epsilon)$, with $\zeta_0 \in \mathbb{R}_{\geq 0}$, and increment $\epsilon > 0$, $\boldsymbol{x}_\zeta$ stays within an $\epsilon$-bounded region of $\mathcal{M}_{\mathrm{CoM}_q}$,*

$$\mathcal{B}_q(\epsilon) = \left\{ \boldsymbol{x} \in \mathcal{X} \;\middle|\; |\sigma_q(\boldsymbol{x})| \leq \epsilon \right\}, \tag{3.21}$$

*where, $\zeta_0$ and $\zeta$ are initial and current phase progression variables, respectively. $\boldsymbol{x}_{\zeta_0}$ is an initial condition.*

This type of bundle characterizes "robust subspaces" (analogous to "tubes" and "funnels" (Mason, 1985)) around nominal phase-space trajectories which guarantee that, if the state initializes within this space, it will remain on it.

**Definition 3.5 (Finite-phase recoverability bundle).** *The invariant bundle $\mathcal{B}_q(\epsilon)$ around a phase-space manifold $\mathcal{M}_{\mathrm{CoM_q}}$ has a finite phase recoverability bundle, $\mathcal{R}_q(\epsilon, \zeta_f) \subseteq \mathcal{X}$ defined as,*

$$\mathcal{R}_q(\epsilon, \zeta_f) = \left\{ \boldsymbol{x}_\zeta \in \mathcal{X}, \quad \zeta_0 \leq \zeta \leq \zeta_f \quad \middle| \quad \boldsymbol{x}_{\zeta_f} \in \mathcal{B}_q(\epsilon) \right\}. \tag{3.22}$$

Note that this bundle assumes the existence of a control policy for recoverability. We will later use these metrics to characterize robustness of our controllers. Visualization of the invariant and recoverability bundles are shown in Fig. 3.3. In particular, the subfigure on the right only shows positive bundles of $\sigma$ while the negative ones are symmetric about the $\zeta$ axis. Since this is a Euclidean space, the manifold for a constant $\sigma$ is a horizontal line and constant values of $\zeta$ are vertical lines. If the condition when we expect the transition to occur is at $\zeta = \zeta_f$, the recoverability bundle shows the range of perturbations that can be tolerated at different $\zeta$ – the system recovers to the invariant bundle before $\zeta_f$. A benefit of using this new $\zeta - \sigma$ coordinate lies in that the original nonlinear phase-space, with respect to the nominal PSM, is transformed into a Euclidean space.

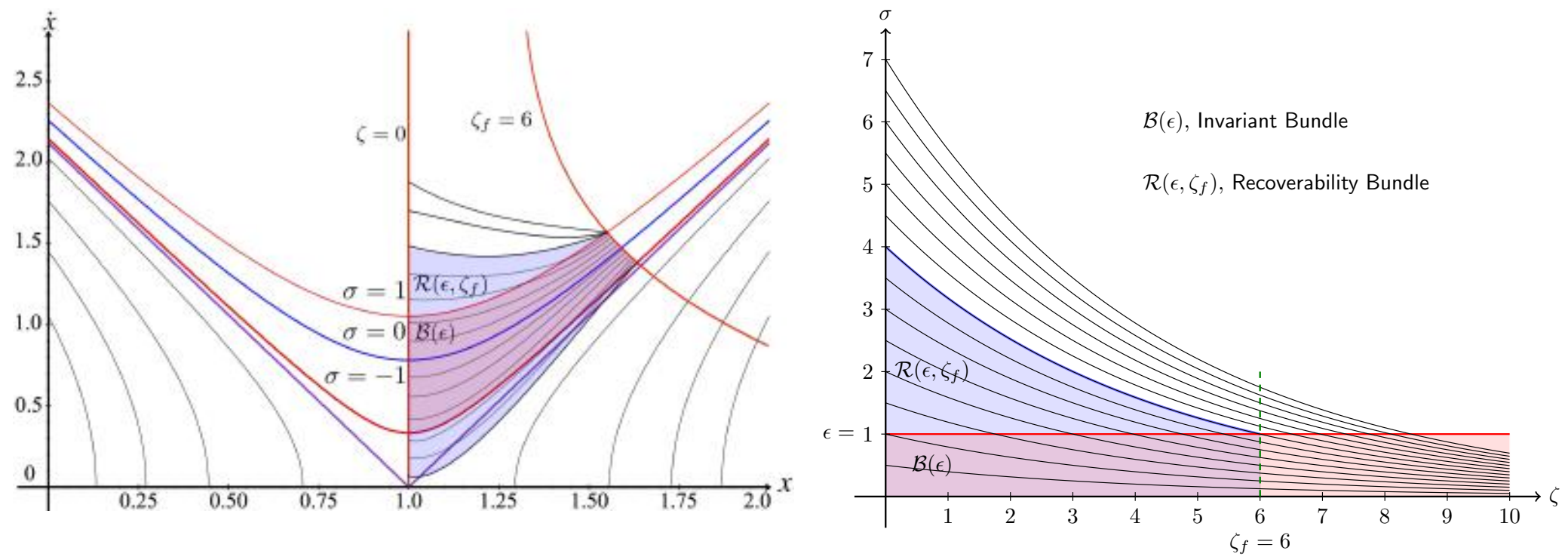

Figure 3.3: Mapping between Cartesian phase-space and $\zeta - \sigma$ space. The two subfigures show the invariant bundle, $\mathcal{B}(\epsilon)$ (shown in red) and the recoverability bundle, $\mathcal{R}(\epsilon, \zeta_f)$ (shown in blue) in different spaces. The left subfigure shows Cartesian phase-space while the right one shows $\zeta - \sigma$ space ($\sigma$ denotes the phase-space manifold as defined in Eq. (4.2) of Section 4.1).

### 3.5.1 Hybrid Locomotion Automaton

Legged locomotion is a naturally hybrid control system, with both continuous and discrete dynamics. We define discrete states $\mathcal{Q} = \{q_l, q_r, q_s\}$ representing the support of the left foot $q_l$ or the right foot $q_r$ or both $q_s$ (stance) feet as shown in Figs. 3.2 and 3.4. On each phase, the continuous dynamics are represented as shown in Eq. (3.20) and over a domain $\mathcal{D}(q)$. We characterize the hybrid system as a directed graph $(\mathcal{Q}, \mathcal{E})$ (see Fig. 3.4), with nodes represented by $q \in \mathcal{Q}$ and edges represented by $\mathcal{E}(q, q + 1)$, that characterize the transitions between nodes. The transitions between states can be grouped into eight classes depending on whether a vector field or variable changes discontinuously and what the trigger mechanism is. Table 3.1 shows the transition classifications.

Table 3.1: Transition classifications. System vector field is $\boldsymbol{\mathcal{F}_x}$ as shown in Eq. (3.20).

| Type | Transition | Switching | Jump |
|---|---|---|---|
| Autonomous | $\Delta_a^{[\tau]}$ | $\boldsymbol{\mathcal{F}_x}^+(\cdot\,,\,\cdot\,,\boldsymbol{x}^+,\,\cdot\,,\,\cdot) \leftarrow \Delta_a^{[\delta_s]}(\boldsymbol{x}^-)$ | $\boldsymbol{x}^+ \leftarrow \Delta_a^{[\delta_j]}(\boldsymbol{x}^-)$ |
| Controlled | $\Delta_c^{[\tau]}$ | $\boldsymbol{\mathcal{F}_x}^+(\cdot\,,\,\cdot\,,\,\cdot\,,\boldsymbol{u_x}^+,\,\cdot) \leftarrow \Delta_c^{[\delta_s]}(\boldsymbol{u_x}^-)$ | $\boldsymbol{u_x}^+ \leftarrow \Delta_c^{[\delta_j]}(\boldsymbol{u_x}^-)$ |
| "Timed" | $\Delta_t^{[\tau]}$ | $\boldsymbol{\mathcal{F}_x}^+(\zeta,\,\cdot\,,\,\cdot\,,\,\cdot\,,\,\cdot\,,\,\cdot) \leftarrow \Delta_t^{[\delta_s]}(\zeta)$ | $\boldsymbol{x}^+ \leftarrow \Delta_t^{[\delta_j]}(\zeta)$ |
| "Disturbed" | $\Delta_d^{[\tau]}$ | $\boldsymbol{\mathcal{F}_x}^+(\cdot\,,\,\cdot\,,\,\cdot\,,\,\cdot\,,w_d) \leftarrow \Delta_d^{[\delta_s]}(w_d)$ | $\boldsymbol{x}^+ \leftarrow \Delta_d^{[\delta_j]}(w_d)$ |

The hybrid automaton state is given by: $\boldsymbol{s} = (\zeta, q, \boldsymbol{x}^T)^T$. $\tau \in \{\delta_s, \delta_j\}$ represents the "switching" or "jump" transition types respectively. $\mu \in \{a, c, t, d\}$ represents the "autonomous", "controlled", "timed" and "disturbed" transitions, respectively. The transition map $\Delta_\mu^{[\tau]}(\cdot)$ is described in further details in Appnedix C. Other details for these types of transitions can be found in (Branicky et al., 1998). The condition that triggers the type of event (switching or jump) is determined by a guard $\mathcal{G}(q, q+1)$ for the particular edge $\mathcal{E}_{q,q+1}$. With this information, let us formulate a robust hybrid automaton for our locomotion planning.

**Definition 3.6 (Robust hybrid automaton).** *A phase-space robust hybrid automaton is a dynamical system, described by a n-tuple*

$$\text{PSRHA} := (\zeta, \mathcal{Q}, \mathcal{X}, \mathcal{U}, \mathcal{W}, \mathcal{F}, \mathcal{I}, \mathcal{D}, \mathcal{R}, \mathcal{B}, \mathcal{E}, \mathcal{G}, \mathcal{T}, \Delta), \tag{3.23}$$

*where, $\zeta$ is the phase-space progression variable, $\mathcal{Q}$ is the set of discrete states, $\mathcal{X}$ is the set of continuous states, $\mathcal{U}$ is the set of control inputs, $\mathcal{W}$ is the set of disturbances, $\mathcal{F}$ is the vector field, $\mathcal{I}$ is the initial condition, $\mathcal{D}$ is the domain, $\mathcal{R}$ is the collection of recoverability sets, $\mathcal{B}$ is the collection of invariant bundles, $\mathcal{E} := \mathcal{Q} \times \mathcal{Q}$ is the edge, $\mathcal{G} : \mathcal{Q} \times \mathcal{Q} \to 2^{\mathcal{X}}$ is the guard, $\mathcal{T}$ is the transition termination set, $\Delta$ is the transition map.*

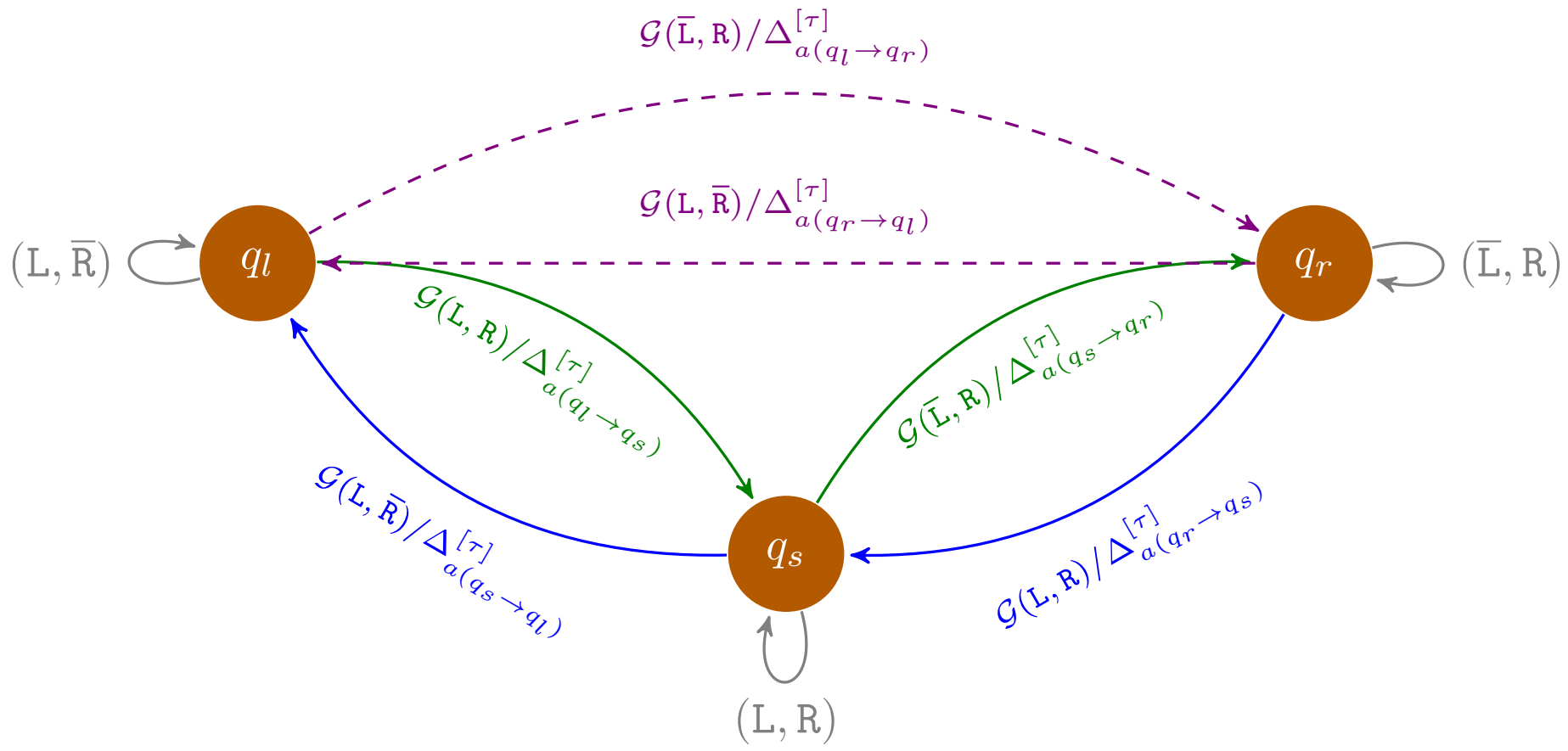

Figure 3.4: Hybrid locomotion automaton for the bipedal robot walking. This automaton has three generic continuous modes, $\mathcal{Q} = \{q_l, q_s, q_r\}$, that represent when the robot is standing in the left leg only ($q_l$), standing in the right leg only ($q_r$), and when the robot is stance (standing in both legs) ($q_s$). Shown in the edges are the guard $\mathcal{G}(q, q+1)$ and the transition map $\Delta^{[\tau]}_{a(q \to q+1)}$. This locomotion automaton has non-periodic mode transitions.

More detailed definitions of these symbols are provided in Appendix C, including arguments and subscripts. This automaton will be used to represent non-periodic trajectories since our planning process focuses on walking over irregular and disjointed terrain. A directed diagram of this non-periodic automaton is shown in Fig. 3.4.

To the best of the authors' knowledge, this is the first formulation of a robust hybrid automaton used for dynamic locomotion. In Section 4.2, more details will be provided for how this automaton governs the hierarchical optimization sequence. To demonstrate the usefulness of this hybrid automaton, we provide an example of a planning sequence as follows.

**Example 3.1.** *Consider a phase-space trajectory that contains two consecu-*

*tive walking steps $\mathcal{Q} = \{q, q+1\}$ (e.g., left and right feet). Given an initial condition $(\zeta_0, q, \boldsymbol{x}_q(\zeta_0)) \in \mathcal{I}$, the hybrid system will evolve following the dynamics of Eq. (3.14) as long as the continuous state $\boldsymbol{x}_q$ remains in $\mathcal{D}(q)$ (e.g., one foot in the ground the other one is swinging). If at some point $\boldsymbol{x}_q$ reaches the guard $\mathcal{G}(q, q+1)$ (e.g., the right foot touches the ground) of some edge $\mathcal{E}(q, q+1)$, the discrete state will switch from $q$ to $q+1$. At the same time the continuous dynamics will reset to some value via $\Delta^{[\tau]}_{a(q\to q+1)}$. After this transition, the whole procedure repeats.*

### 3.5.2 Step Transition Strategy

Step transitions could be idealized as an instataneous contact or have a short dual-contact phase (Fig. 3.6(a)). We first create a strategy for instantaneous contact switch then extend it to multi-contact in Appendix B.3.

**Definition 3.7** (**A phase-space walking step**). *A $q^{\text{th}}$ walking step is defined as a phase-space trajectory in domain $\mathcal{D}(q)$, which has two boundary guards $\mathcal{G}(q-1, q)$ and $\mathcal{G}(q, q+1)$.*

To characterize the non-periodic stability associated with walking in rough terrains, we define a keyframe map between apex states.

**Definition 3.8** (**Keyframe map of non-periodic gaits**). *We define the keyframe map of non-periodic gaits as the progression map, $\Phi$, that takes the robot's center of mass from one desired apex state, $(\dot{x}_{\text{apex}_q}, x_{\text{foot}_q}, \theta_q)$, to the*

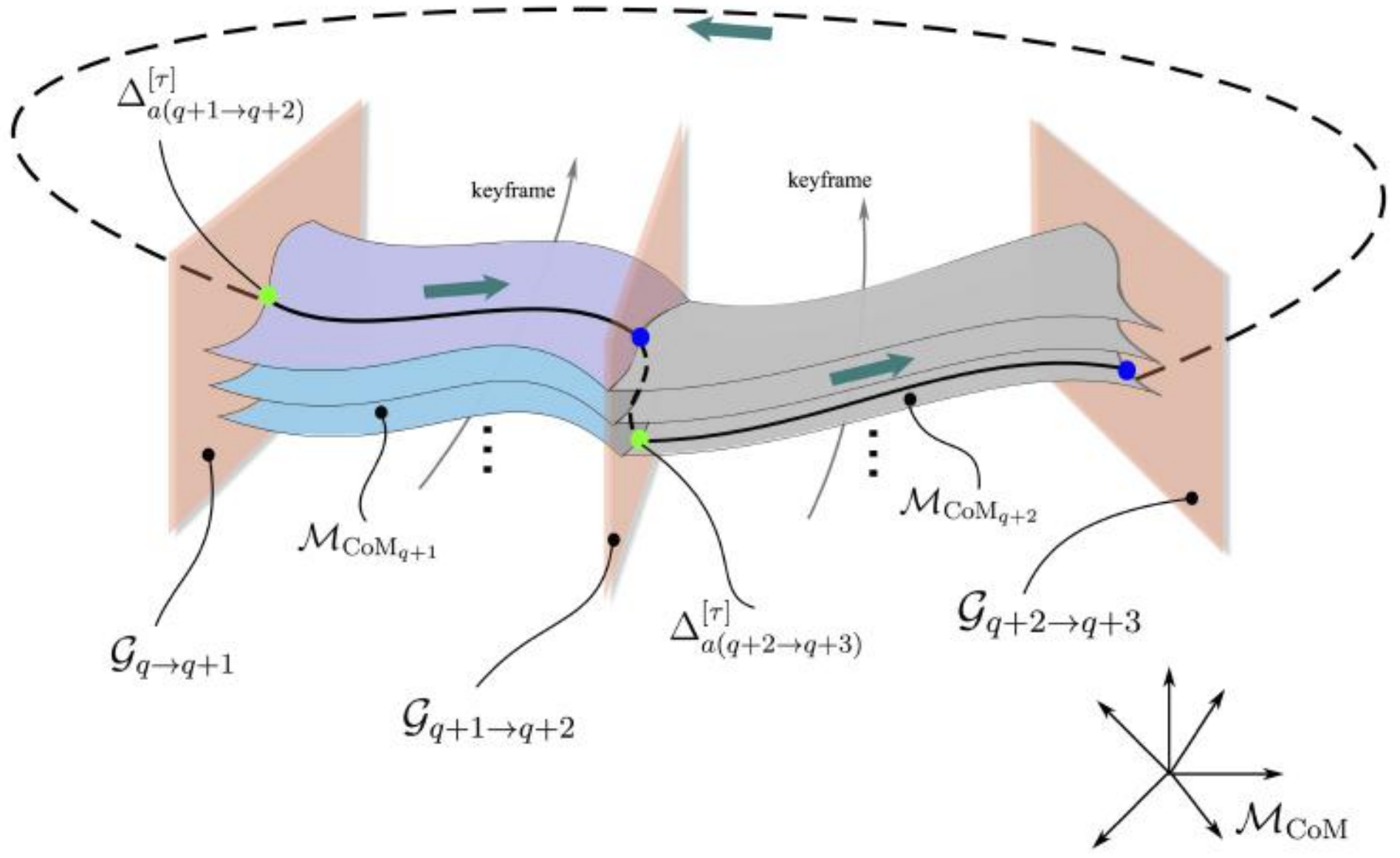

Figure 3.5: Non-periodic keyframe map. This figure illustrates a multi-domain keyframe map in the high-dimensional phase space $\mathcal{M}_{\mathrm{CoM}}$. Within each walking step, there is a stack of modes (i.e., manifold) characterized by the keyframe state. The continuous locomotion dynamics progress on one of them. When the state hits the guard $\mathcal{G}_{(\cdot\to\cdot)}$ at the blue dot, the keyframe is mapped to a new one and the continuous dynamics evolve from a new initial condition at green dot (i.e., reset by a transition map $\Delta^{[\tau]}_{a(\cdot)}$). This procedure is repeated in a non-periodic pattern.

*next one on the phase-space manifold, and via the control input,* $\boldsymbol{u}$*, i.e.*

$$(\dot{x}_{\mathrm{apex}_{q+1}}, x_{\mathrm{foot}_{q+1}}, \theta_{q+1}) = \Phi(\dot{x}_{\mathrm{apex}_q}, x_{\mathrm{foot}_q}, \theta_q, \boldsymbol{u}_{\boldsymbol{x}}). \tag{3.24}$$

*where* $\theta_q$ *represents the heading of the* $q^{\mathrm{th}}$ *walking step.*

Fig. 3.5 shows an illustration of this keyframe map. We will use this map for the steering walking model of Section 3.5.4. The above map addresses the nature of "non-periodic" gaits by enabling arbitrary keyframe specifications. Users can design "non-periodic" keyframes that change the speed and steer the robot through its walk. For this study, we use heuristics to design keyframes. More recently, we have proposed to use a keyframe decision maker based on linear temporal logic (Zhao et al., 2016f,g), which will be introduced in Chapter 5.

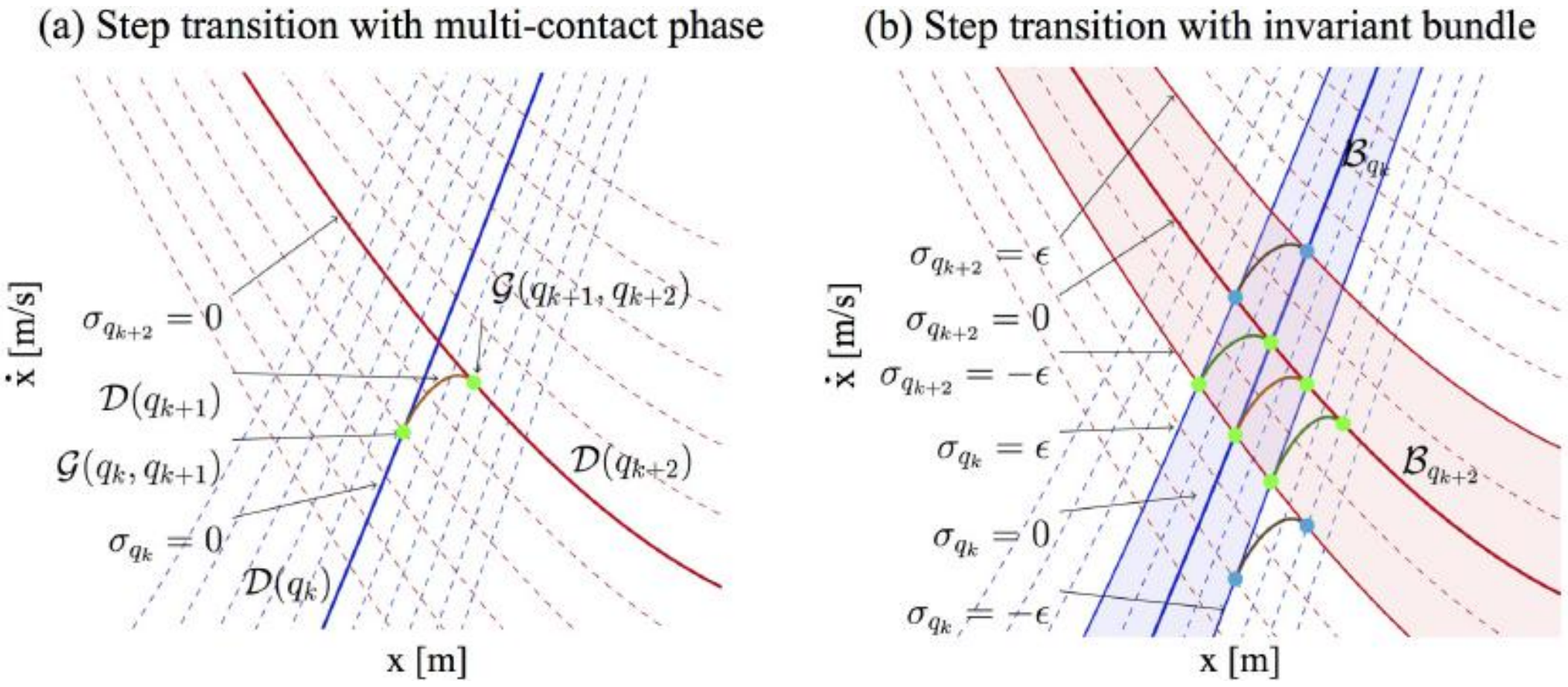

Figure 3.6: Step transitions. This figure illustrates two types of step transitions in the sagittal phase-space, associated with $\sigma$-isolines. (a) switches between two single contacts with a multi-contact phase. (b) shows several guard alternatives for multi-contact transitions, from the current single-contact manifold value $\sigma_{q_k}$ to the next single-contact manifold $\sigma_{q_{k+2}}$. In particular, the invariant bundle bounds, $\sigma_{q_k} = \pm\epsilon$ are shown. The transition phase in green reattaches to the nominal manifold, $\sigma_{q_{k+2}} = 0$, while the transition phase in brown maintains its $\sigma$ value, i.e., $\sigma_{q_{k+2}} = \sigma_{q_k}$.

Our motion planner employs CoM apex states instead of touchdown states as keyframes due to our focus on non-periodic CoM dynamics. CoM apex states represent practical salient states for agile walking and help to design walking directions and velocities in a versatile fashion.

**Definition 3.9** (**Phase progression transition value**). *A phase progression transition value* $\zeta_{\text{trans}} : \mathcal{Q} \times \mathcal{X} \to \mathbb{R}_{\geq 0}$ *is the value of the phase progression* *variable when the state* $\boldsymbol{x}_q$ *intersects a guard* $\mathcal{G}$*, i.e.,*

$$\zeta_{\text{trans}} \coloneqq \inf\{\zeta > 0 \mid \boldsymbol{x}_q \in \mathcal{G}\}. \tag{3.25}$$

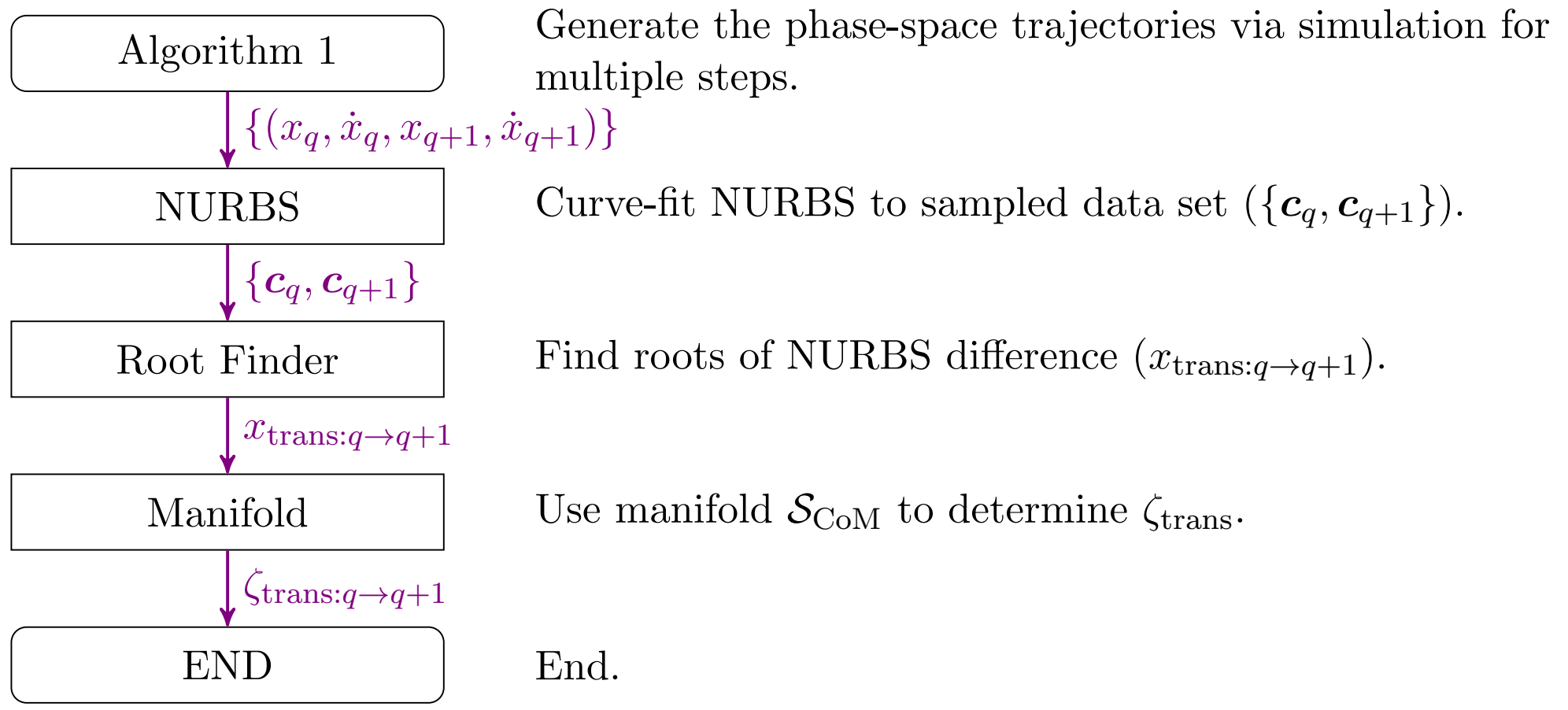

We propose an algorithm to find transitions between adjacent steps, which occur at $\zeta_{\text{trans}}$. We first consider the case of *walking steps* as defined in Def. 3.7. Given known step locations and apex conditions, phase-space curves can be numerically obtained using Algorithm 1. Phase-space curves in the general case have infinite slopes when crossing the zero-velocity axis. One of the characteristics of pendulum dynamics portrayed in the phase-space is displaying infinite slopes when crossing the zero-velocity axis (Zhao et al., 2013b). To deal effectively with this difficulty, we use NURBS (non-uniform rational B-splines)[2] to the manifolds of the generated data (see Fig. 4.1 for an illustration of adjacent step manifolds). Subsequently, finding step transitions just consists on finding the root difference between adjacent NURBS, which reduces to a simple polynomial root-finding problem. The pipeline to find step intersections is shown below. For clarity, sagittal apices are the states relating the robot's CoM velocities to their positions when crossing the sagittal contact positions. On the other hand, the instants at which contact transitions occur, derived from the above algorithm, are halfway between apices.

[2]Different from polynomials, non-rational splines or Bézier curves, NURBS can be used to precisely represent conics and circular arcs by adding weights to control points.

In many cases, the hybrid dynamics of a robot impacting the ground are significant (Yang et al., 2009). This has prompted the consideration of a discrete map to represent the sudden joint velocity changes that occur during impact (Grizzle et al., 2014). However, since our planning and control algorithms focus exclusively on the CoM behavior, we use a model that predicts negligible change of the CoM velocity during impact. Here we assume idealized point-foot robots (Kim et al., 2016a) with 1) most of the robot's mass concentrated on the upper body and having lightweight legs, 2) having some elastic properties such as foot bumpers or series elastic elements on the leg actuators, 3) having frictionless actuators, and 4) having the robot's upper body fairly decoupled from the impact point due to the leg's kinematic chain.

### 3.5.3 Lateral Foot Placement Searching

To complete the 3D walking planner, we formulate a searching strategy for lateral foot placement that complies with the timing of sagittal step transitions. The main objective of the lateral dynamics is to return the robot's center

**Algorithm 2** Newton-Raphson Search for Lateral Foot Placement

Initialize iteration index $n \leftarrow 1$, maximum iterations $n_{\max}$, tolerance $\dot{y}_{\text{tol}}$ and initial state $y_{\text{init}}, \hat{y}_{\text{foot}}(1), \ddot{y}_{\text{apex}}(1)$
$\dot{y}_{\text{apex}}(1) \leftarrow$ integration of inverted pendulum model given in Eq. (3.26) with $\hat{y}_{\text{foot}}(1)$
**while** $n < n_{\max}$ and $|\dot{y}_{\text{apex}}(n)| > \dot{y}_{\text{tol}}$ **do**
  $\hat{y}_{\text{foot}}(n+1) = \hat{y}_{\text{foot}}(n) - \dot{y}_{\text{apex}}(n)/\ddot{y}_{\text{apex}}(n)$ by Newton-Raphson method
  $\dot{y}_{\text{apex}}(n+1) \leftarrow$ integration of inverted pendulum in Eq. (3.26) with $\hat{y}_{\text{foot}}(n+1)$
  $\ddot{y}_{\text{apex}}(n+1) = (\dot{y}_{\text{apex}}(n+1) - \dot{y}_{\text{apex}}(n))/(\hat{y}_{\text{foot}}(n+1) - \hat{y}_{\text{foot}}(n))$
  $n \leftarrow n+1$
**end while**

of mass to a walking center through a semi-periodic cycle. If lateral foot placements are not adequately picked, the lateral behavior will drift away or become unstable. According to Eq. (3.14), lateral center of mass dynamics are equal to

$$\dot{\boldsymbol{y}} = \boldsymbol{\mathcal{F}}_{\boldsymbol{y}}(\boldsymbol{y}, \boldsymbol{u}_{\boldsymbol{y}}) = \begin{pmatrix} \dot{y} \\ \omega^2(y - y_{\text{foot}}) - \frac{\omega^2}{mg}(\tau_y + a_q \tau_z) \end{pmatrix}, \tag{3.26}$$

which can be numerically simulated adapting Algorithm 1 to the lateral dynamics (see Fig. B.1 for simulations of lateral dynamics).

To generate bounded lateral trajectories, we choose the simple criterion of achieving zero apex lateral velocity, $\dot{y}_{\text{apex}} = 0$ at the instant when the CoM lateral apex position, $y_{\text{apex}}$ is located between the two feet. Here $y_{\text{apex}}$ and $\dot{y}_{\text{apex}}$ are the lateral center of mass position and veloicty when the center of mass crosses the sagittal apex as defined in Def. 3.3. Algorithm 2 achieves this objective by using the Newton-Raphson method, which approximates the roots of real-valued functions. In our case, this function is chosen to be the apex lateral velocity $\dot{y}_{\text{apex}}$ with the lateral foot placement $\hat{y}_{\text{foot}}$ as its independent variable (as shown in line 4 of Algorithm 2). In this algorithm, $\hat{y}_{\text{foot}}(n)$ represents the estimated lateral foot placements in the $n^{\text{th}}$ search iteration. A foot placement range constraint $\hat{y}_{\text{foot,min}} \leq \hat{y}_{\text{foot}} \leq \hat{y}_{\text{foot,max}}$ and the maximum iteration step constraint $n < n_{\text{max}}$ are also provided. Examples of usage are given in Fig. B.1.

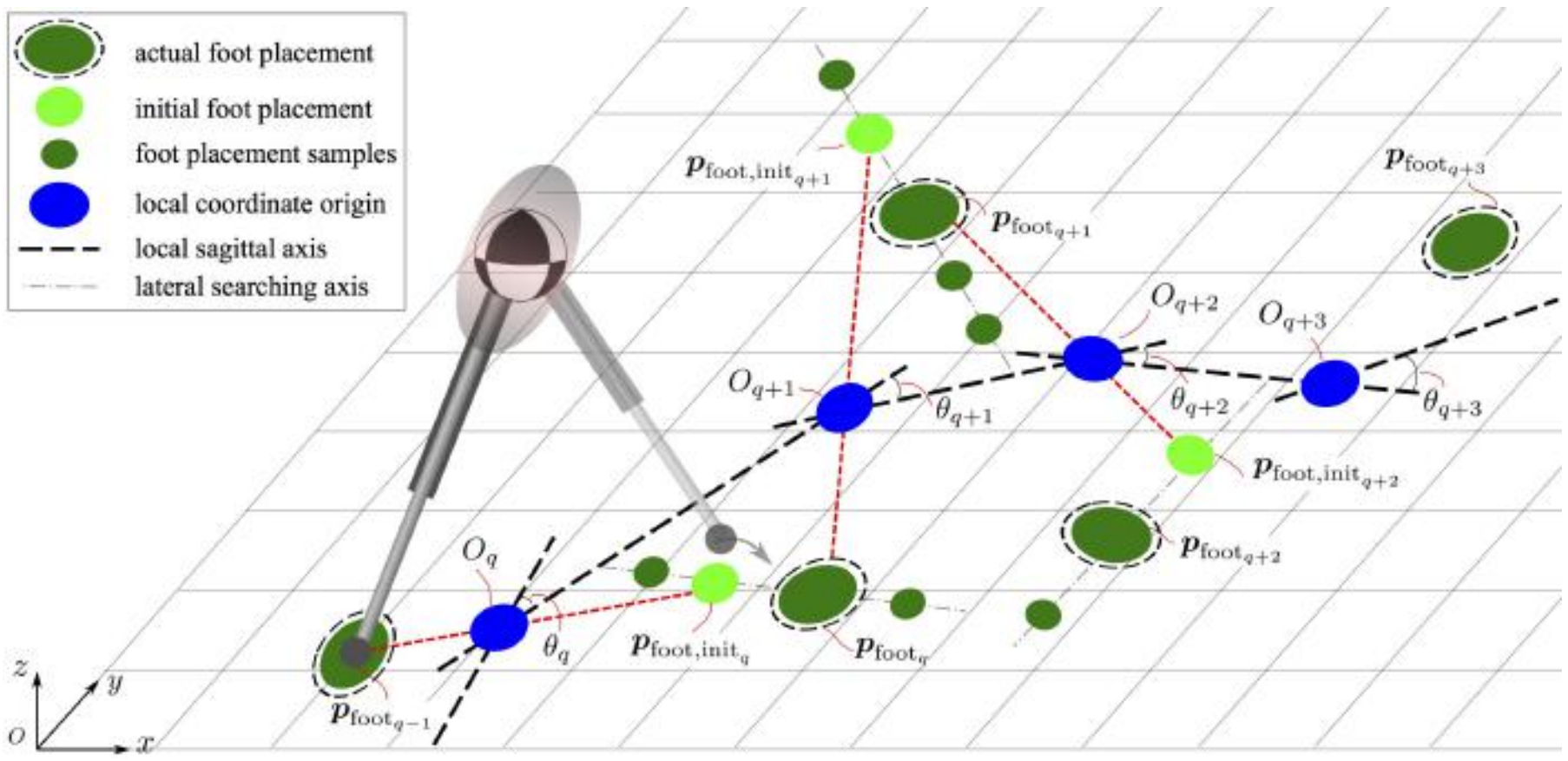

Figure 3.7: Strategy for steerable walking. We define a local coordinate frame with origin represented by  and local sagittal axes represented by the black dash lines. The lateral foot placement searching algorithm described in Algorithm 2 is applied using the newly defined local frames.  represents the final foot locations found via this procedure.

### 3.5.4 Steerable Walking Model

To plan practical walking behaviors, we introduce a steerable walking model based on local coordinates as shown in Fig. 3.7. The pipeline for the steerable walking process is as follows: (i) define a local sagittal axis (black dash line) projected to level ground which specifies the heading angle $\theta_q$ for the $q^{\text{th}}$ step; (ii) define the local origin $O_q$ (represented by ) as the intersection of the local sagittal axis and a dash line (red dash line) connecting the previous foot placement $\boldsymbol{p}_{\text{foot}_{q-1}}$ (represented by 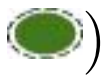) and an initial guess of the foot placement $\boldsymbol{p}_{\text{foot,init}_q}$ (represented by 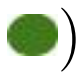); (iii) search the lateral foot placement with respect to the local frame (note that the lateral search line, shown as a gray color dash-dot line, is orthogonal to the local sagittal axis); (iv) once determined the foot placement $\boldsymbol{p}_{\text{foot}_q}$, we generate CoM and foot trajectories for the walking step; (v) after this step, we provide the new desired heading angle $\theta_{q+1}$ and re-start the planning process to step (i). Once all trajectories in local frame are obtained, we transform them to global frame via the recorded heading angles. A circular walking example is provided in Section 3.6.2 using the above planning strategy.

## 3.6 Example: 3D Dynamic Maneuvering

We evaluate various locomotion scenarios for the purpose of testing our planner's ability to walk on rough terrains. Our robot model uses six-dimensional free floating states, three degrees-of-freedom (DOFs) per leg and one DOF for torso pitch flexion/extension. Each leg has three actuated joints: hip abduction/adduction, hip flexion/extension and knee flexion/extension. We assume that each actuated joint has enough torque capability to achieve the planned motion. This model has a 0.55 m torso length, a 0.56 m hip width, a 0.6 m thigh length and a 0.55 m calf length, respectively. Given CoM trajectories and foot locations generated from the planner, we use inverse kinematics to obtain corresponding joint angles. On the other hand, because the feet contact transitions are discrete, we create smooth swinging trajectories to land the feet at the desired locations with the given time stamps. An accompanying video of the dynamic walking over various terrain topologies is available in the video (Rou, 2016).

### 3.6.1 Walking over Rough Terrain

In this section, we validate the versatility of phase-space planning by performing locomotion over stochastic terrains. Three challenging terrains are implemented as shown in Fig. 3.8: (a) concave terrain, (c) convex terrain and (d) inclined terrain. The height discrepancy, $\Delta h_k$, of two consecutive stairs is randomly sampled following a uniform distribution,

$$\Delta h_k = h_{k+1} - h_k \sim \mathrm{U}\left\{(-\Delta h_{\max}, -\Delta h_{\min}) \cup (\Delta h_{\min}, \Delta h_{\max})\right\}, \qquad (3.27)$$

where, $h_k$ represents the height of the $k^{\text{th}}$ stair, $\Delta h_{\min} = 0.1$ m, $\Delta h_{\max} = 0.3$ m (this represents 33% of the total length of a stretched-out leg). A 10 degree tilt angle is assigned to the slope of the surface. Foot placements are chosen a priori using simple kinematic rules and considering the length of the terrain steps. We design apex velocities according to a heuristic that accounts for terrain heights, and we use an average apex velocity of 0.6 m/s. We choose CoM surfaces that conform to the terrains. We then apply the trajectory generation and controller pipeline procedures outlined in previous sections including the generation of trajectories based on Algorithm 1 and the search for step transitions based on the procedures of Section 3.5.2.

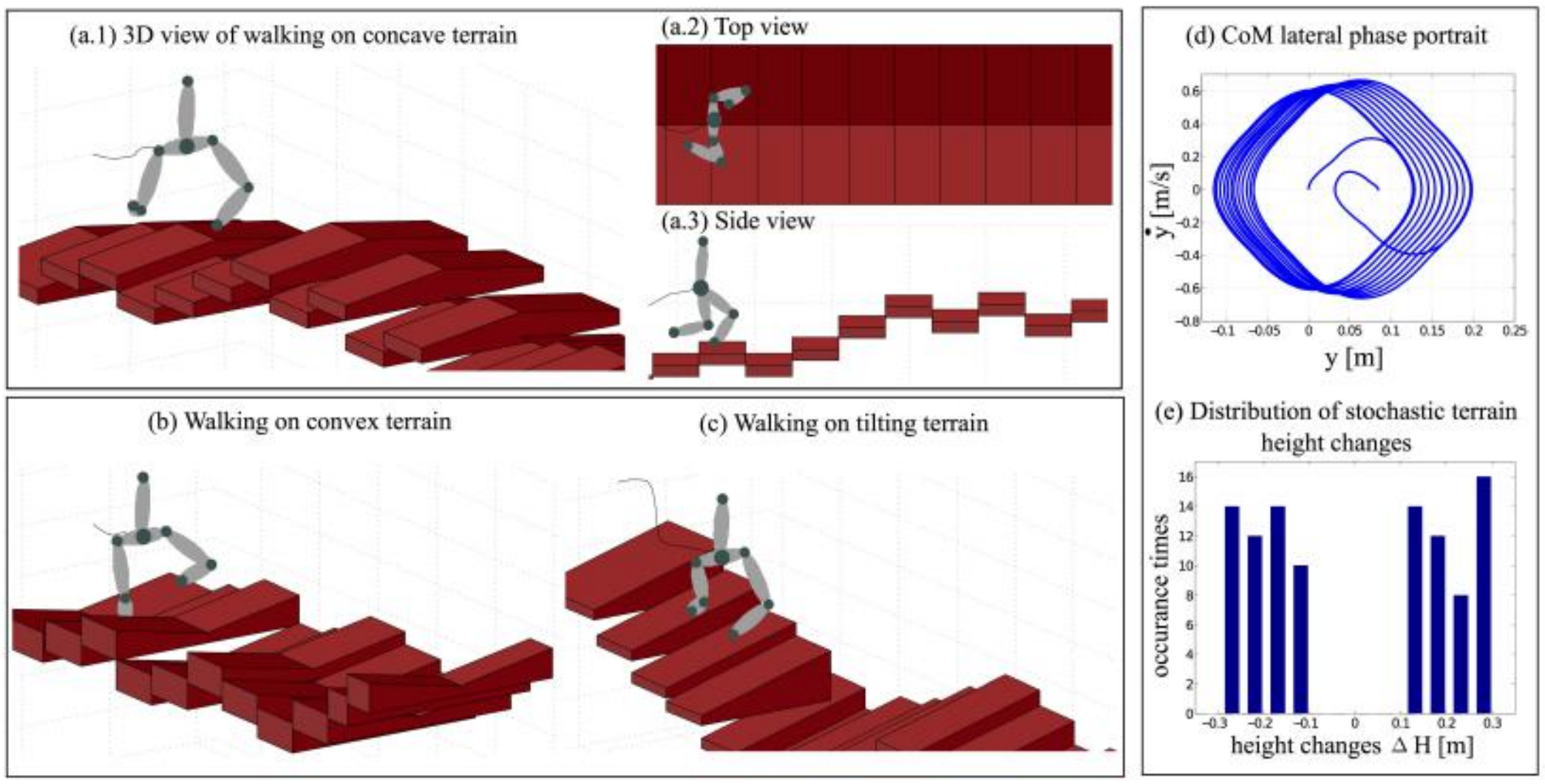

Figure 3.8: Traversing different rough terrains and distribution of stochastic terrains. The subfigures on the left block visualize dynamic locomotion over stochastic rough terrains. The block on the lower right shows the height variation distribution of 100-step walking over stochastic terrains.

Fig. 3.8 (a.1) shows a snapshot of the bipedal walking on a terrain with rough concave surfaces. The side and top visualizations are shown in Fig. 3.8 (a.2)-(a.3). The ability to navigate through such diverse terrain illustrates the agile walking capabilities. The lateral CoM phase portrait in Fig. 3.8 (b) shows a walking sequence with 25 steps. Since this is forward walking, the lateral CoM phase portrait shows semi-cyclic motions. Fig. 3.8 (c) and (d) test the applicability of our gait generator to other types of rough terrains by simulating 100 steps of dynamic walking on the stochastically generated terrains. Fig. 3.8 (e) shows the sagittal-vertical Cartesian position in local contact foot coordinate. The bar graph in Fig. 3.8 (f) shows the distribution of sampled stochastic terrain height changes. This set of simulations demonstrate the effectiveness of our planning framework for challenging yet stochastic terrain maneuvering. Walking over these three types of terrains are attached in the supplementary video.

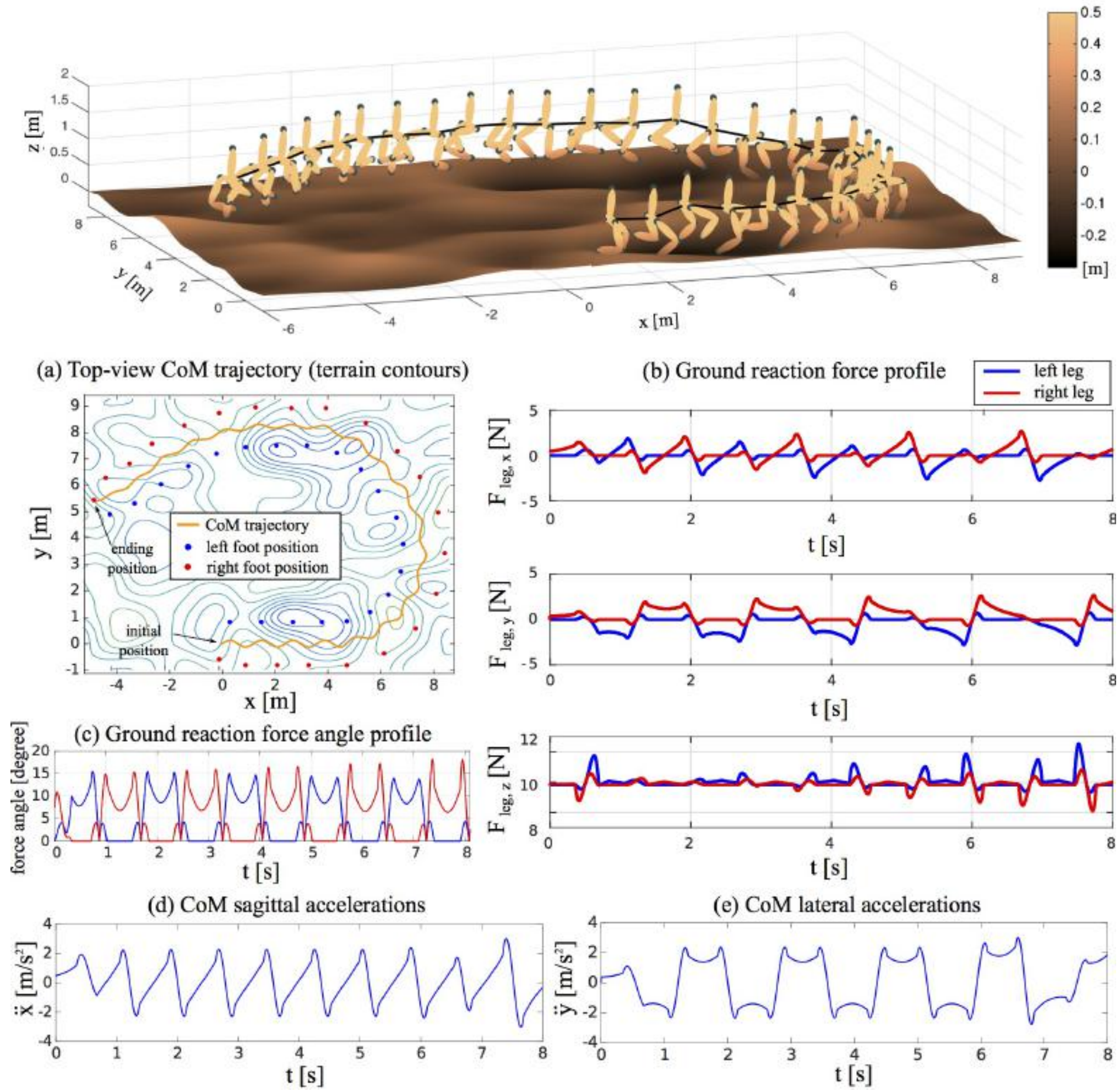

Figure 3.9: Circular walking over a random rough terrain. The figure above shows dynamic walking while steering in 3D. Subfigure (a) shows the top view of the CoM trajectory and the foot locations given the terrain contour. (b) shows each leg's ground reaction forces in local coordinate. The reaction forces at step transitions are smooth thanks to the dual-contact control policy. (c) shows that the angle of reaction forces is constrained within the 45° friction cone. (d) and (e) show smooth CoM sagittal and lateral accelerations.

### 3.6.2 Circular Rough Terrain Walking

Circular walking over a random rough terrain is shown in Fig. 3.9. The terrain height randomly varies within [-0.24, 0.3] m. We use this example to validate the steering capability of our planner. The walking direction is defined by the heading angle $\theta$ shown in Def. 3.8. The planning process is performed in the robot's local coordinate with respect to the heading angle. We then apply a local-to-global transformation. Also, this simulation validates the steering direction model introduced in Section 3.5.4 and smoothness of the leg force profile by using multi-contact dynamics in Section 3.3.2.

### 3.6.3 Bouncing over A Disjointed Terrain

A more challenging locomotion scenario is explored using a disjointed terrain. The slope of the surfaces is 70°. The goal is to step up over the surfaces by bouncing over the terrain. A physics based dynamic simulation called SrLib is used for validation and a Whole-Body Operational Space Controller (Kim et al., 2016a) is implemented to follow the locomotion plans. The robot in the dynamic simulator has masses and inertias distributed across its body compared to the previous simulations. It possesses the same degrees of freedom, actuation joints and kinematic parameters as those in previous simulations. Another difference is that this simulation is planarized, meaning that the robot is not allowed to move laterally. Snapshots of a one-step bouncing behavior are shown in Fig. 3.10 (a). To successfully bounce over the terrain, we design a CoM path manifold, shown in Fig. 3.10 (b.1), that mimics that of a pre-recorded human jumping motion (Sentis and Slovich, 2011). During the multi-contact phase, we apply a 250 N internal tension force, shown in Fig. 3.10 (c), between the two surfaces to avoid sliding down due to the weight of the robot. The torso angular moment is also controlled immediately before and after the stepping-up motion. Our planner for this scenario operates in the $x$–$\dot{z}$ phase-space as shown in Fig. 3.10 (b.2). This is more convenient as $\dot{z}$ captures the moment at which the center-of-mass starts falling down. More details on this strategy are discussed in (Sentis and Slovich, 2011). Note that the keyframe in this case also becomes defined as a state $(x, \dot{z})$, shown as a red circle in Fig. 3.10 (b.2). Even though the bouncing behavior on the disjointed terrain is intrinsically different from the previously-studied rough terrain walking, we still use the proposed single contact inverted pendulum model of Section 3.3.1 and the multi-contact dynamics of Section 3.3.2. The overall behavior is essentially different than the walking cases. The main reasons are that: (i) internal force control is needed to overcome gravity forces on the highly inclined surfaces and that (ii) the multi-contact phase is more dominant than the single contact phase.

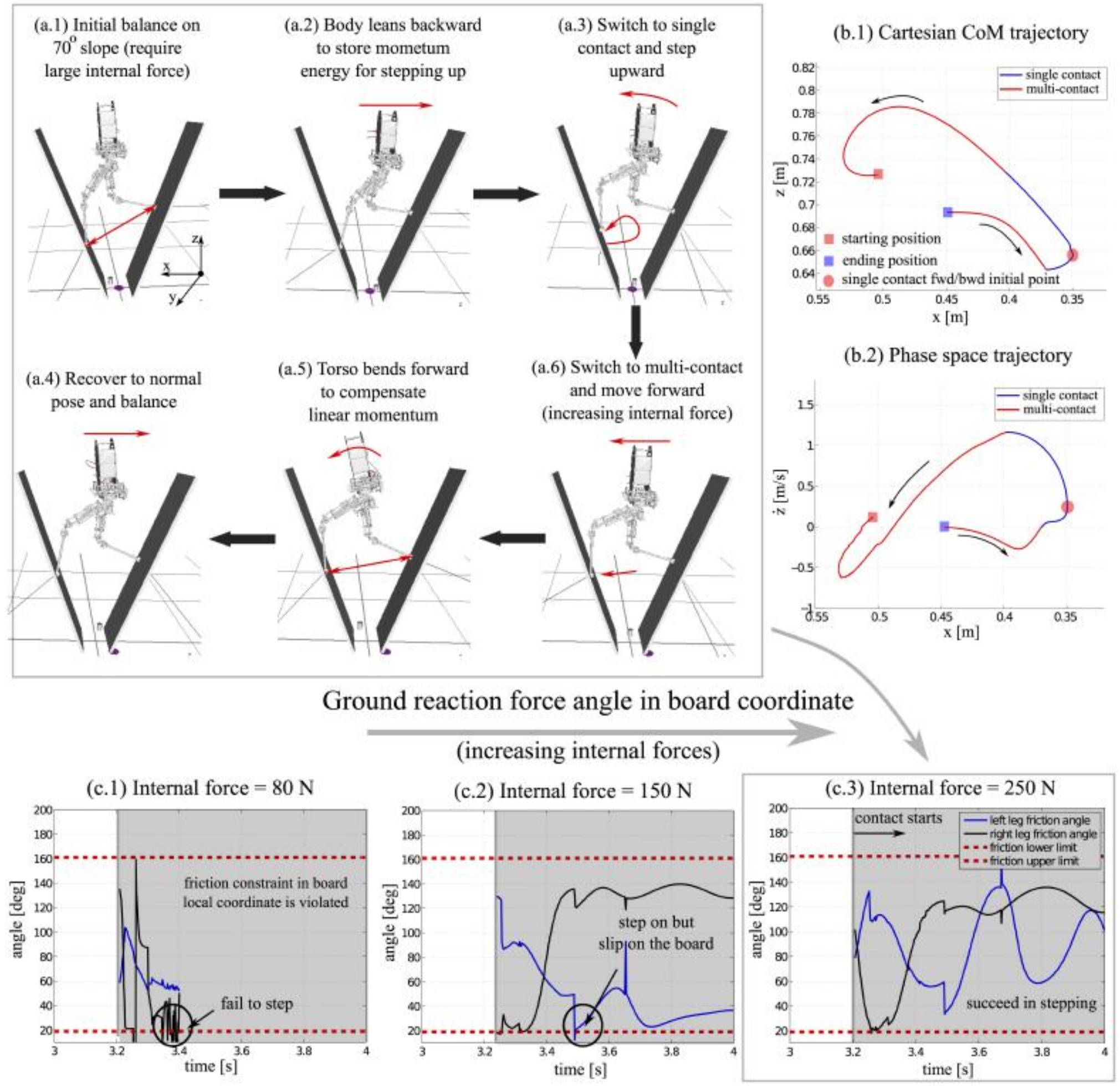

Figure 3.10: Bouncing over a disjointed terrain. In this dynamic simulation, a biped balances on a steep disjointed terrain and dynamically bounces upwards. Internal force and torso pitch torque are controlled appropriately to achieve this motion. Subfigure (b) shows the 2D Cartesian CoM trajectory and the $x - \dot{z}$ phase portrait. (c) shows contact reactions for three different desired internal force scenarios.

## 3.7 Discussions and Conclusions

The primary focus of this chapter has been on addressing the needs for planning non-periodic bipedal locomotion behaviors. These types of behaviors arise in situations where terrains are non-flat, disjointed, or extremely rough. The majority of bipedal locomotion methodologies have been historically focused on flat terrain or mildly rough terrain locomotion behaviors. Some of them are making their way into dynamically climbing stairs or inclined terrains. Additionally, Raibert experimented with planar hopping locomotion over rough terrains in the middle-80s (Raibert, 1986). In contrast, our effort is centered around the goals of (i) generalizing gaits to any types of surfaces including disjointed terrains, and (ii) providing formal tools to study planning and reachability of the non-periodic gaits.

The planning methods and tools described throughout this chapter have been initially implemented on our bipedal robot for proof-of-concept. We will talk about the experimental evaluation in next chapter.

In the nominal trajectory generation process of Algorithm 1, we assume a sequence of foot placements given a priori. There exist optimization methods determining discrete foot placements, and therefore it has not been a focus of ours to explore this issue. As to the apex state design, the results in this chapter use a heuristic related to the terrain heights. In Chapter 5, we will propose an advanced keyframe decision maker based on temporal-logic-based formal methods (Zhao et al., 2016g,f). Choosing apex states in our planner strategy is not only a mathematical convenience but also enables designers to plan non-periodic apex velocities which is related to the walking speed. Apex states are a type of salient points more natural to regulate the walking speed.

Our choice of providing a priori CoM surfaces can be traced back to our initial design methodology for this line of work. Initially, we extracted the CoM trajectory from capturing human walking over rough terrains (Zhao and Sentis, 2012; Zhao et al., 2016d). We observed that the CoM trajectory approximately conforms to the terrain height and slope. This observation prompted us to use the following three-step procedure: 1) design the piecewise-linear CoM plane approximately in parallel with the terrain slope; 2) design heuristics to adjust the CoM plane sagittal and lateral slopes (i.e., tilting angles) according to the walking phases (step acceleration or deceleration phases).

Our method could use generic CoM surfaces like in (Morisawa et al., 2005), but the dynamics of Eq. (3.14) would become more complicated. In that case, deriving an analytical phase-space distance metric should be done based on numerical approximation and curve fitting (i.e. NURBS). To avoid this added complexity, we chose to rely on the piecewise-linear CoM surface model and smooth it out using multi-contact dynamics. We believe that our current method presented in this chapter is sufficient to achieve smooth locomotion without using a more complicated metric.

Zero lateral velocity at the sagittal apex is a simple heuristic that prevents the center of mass from drifting away from the local frame. It is important to remark that this heuristic is specified in the local frame, and therefore it accounts for the steering angle. As such, when considering the global frame, the lateral velocity at the end of each step is effectively non-zero.

The lateral foot placement is an output of the planner. Each time a new sagittal foot placement is re-planned in an online fashion, the lateral foot placement has also to be re-planned. We view this online re-planning stage, described in Sections 4.2.2 and 3.5.3, as a controller which is a part of the runtime methodology that should be implemented in real experiments.

# Chapter 4

# Robust Optimal Planning under Disturbance

The main objective of this chapter is to propose a phase-space manifold as a Riemannian distance metric to measure nominal trajectory deviations, and further use it to design a robust hierarchical controller to achieve runtime disturbance rejection: the continuous control is based on dynamic programming while the discrete controller is designed for foot placement re-planning. Additionally, we propose a theorem to prove the attractiveness of the continuous controller.

## 4.1 Phase-Space Manifold Analysis

In this Section, we focus on formulating a metric to measure deviations from the phase-space manifolds (PSM) of planned trajectories derived in the previous chapter. A phase-space sensitivity norm is also formulated to study the effect of disturbances. Various disturbance patterns and suggested recovery strategies are considered.

---

This chapter incorporates the results from the following publications: (Zhao, Fernandez, and Sentis, 2016a), (Zhao, Fernandez, and Sentis, 2016b). Benito Fernandez contributes Fig. 4.2. Section 4.5 is mainly lead by Donghyun Kim and the results are shown in (Kim, Zhao, Thomas, Fernandez, and Sentis, 2016a).

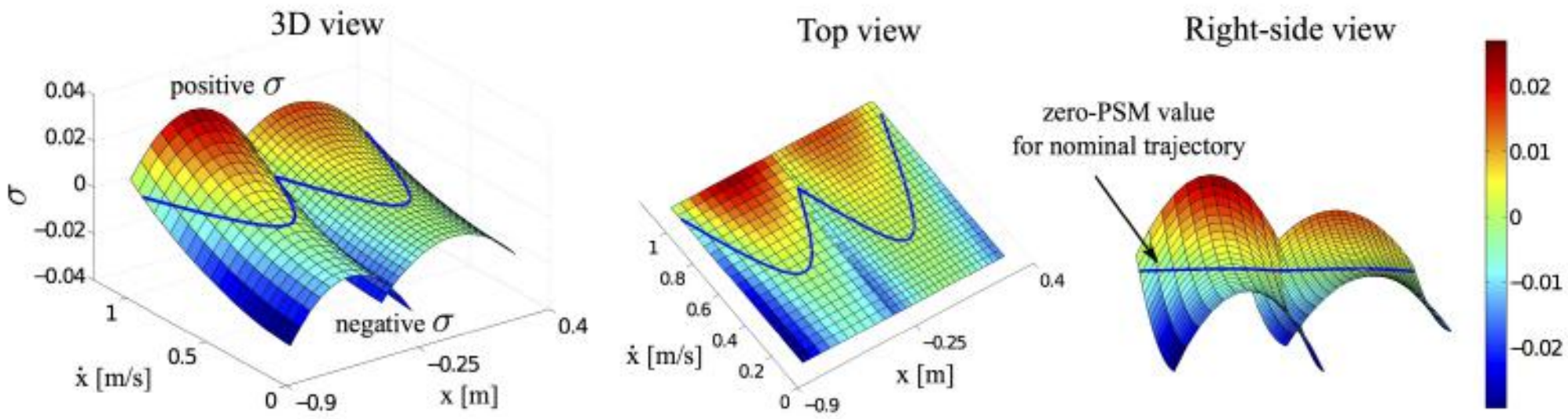

Figure 4.1: Phase-space manifold isoline plot. This three-dimensional space demonstrates our phase-space manifold isolines defined in Eq. (4.1) by the color map.

**Proposition 4.1 (Phase-space tangent manifold).** *Given the prismatic inverted pendulum dynamics of Eq. (3.20) with an initial condition* $(x_0, \dot{x}_0)$ *and foot placement* $x_{\text{foot}}$*, the phase-space tangent manifold is*

$$\sigma := (x_0 - x_{\text{foot}})^2 \Big( 2\dot{x}_0^2 - \dot{x}^2 + \omega^2 (x - x_0)(x + x_0 - 2x_{\text{foot}}) \Big) - \dot{x}_0^2 (x - x_{\text{foot}})^2 + \dot{x}_0^2 (\dot{x}^2 - \dot{x}_0^2)/\omega^2, \tag{4.1}$$

*where the condition* $\sigma = 0$ *is equivalent to the trajectory manifolds. Therefore,* $\sigma$ *represents the distance to the estimated locomotion trajectories.*

*Proof.* Given in the Appendix A. □

If we use the apex conditions as initial values, i.e. $(x_0, \dot{x}_0) = (x_{\text{foot}}, \dot{x}_{\text{apex}})$, the tangent manifold becomes

$$\sigma(x, \dot{x}, \dot{x}_{\text{apex}}, x_{\text{foot}}) = \frac{\dot{x}_{\text{apex}}^2}{\omega^2} \Big( \dot{x}^2 - \dot{x}_{\text{apex}}^2 - \omega^2 (x - x_{\text{foot}})^2 \Big). \tag{4.2}$$

Since the tangent manifold is considered as a trajectory deviation metric in the phase-space, we will use it in the next chapter as a feedback control parameter to ensure robustness. Fig. 4.1 provides an illustration of the value of $\sigma$ as a function of the state. The horizontal plane represents the sagittal phase-space while the vertical third dimension represents the non-zero $\sigma$ value in Eq. (4.1). As we can see, the blue nominal trajectory has a zero $\sigma$ value. Phase-space region above the nominal trajectory has positive $\sigma$ values while the lower region have negative $\sigma$ values. The same type of analysis can be done for lateral trajectory deviations using the lateral pendulum dynamics of Eq. (3.26).

**Proposition 4.2** (**Phase-space cotangent manifold**). *Given the pendulum system of Eq. (3.20), the cotangent manifold is equal to*

$$\zeta = \zeta_0 \left(\frac{\dot{x}}{\dot{x}_0}\right)^{\omega^2} \frac{x - x_{\text{foot}}}{x_0 - x_{\text{foot}}}, \tag{4.3}$$

*and represents the arc length along the tangent manifold of Eq. (4.2).* $\zeta_0$ *is a nonnegative scaling factor, which represents the initial condition of a cotangent manifold. We choose it as the phase progression value when a contact switch occurs.*

*Proof.* Given in the Appendix A. □

**Remark 4.1.** *These low-dimensional phase-space manifolds are embedded in high-dimensional full-state CoM phase space. This feature is analogous to the low-dimensional manifold control studies in (Erez and Smart, 2007; Ramamoorthy and Kuipers, 2008). However, neither of these two works provide metrics of disturbance deviations or strategies to handle these disturbances.*

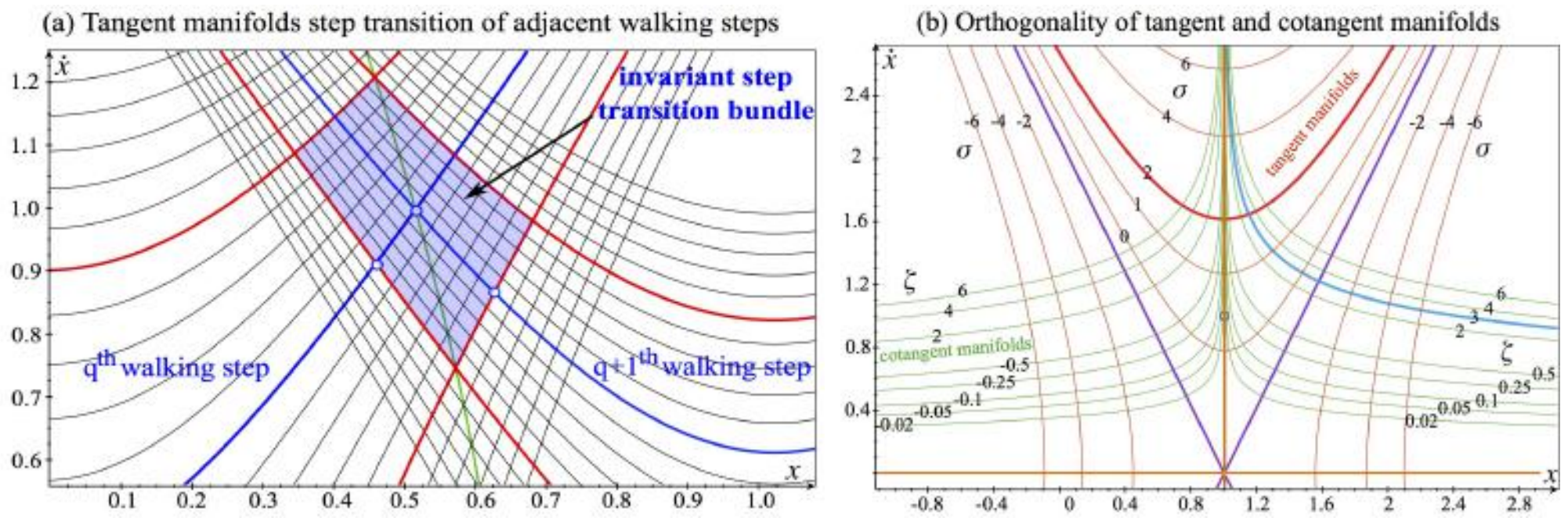

Figure 4.2: Orthogonal tangent and cotangent manifolds. The left subfigure shows the tangent manifolds for two consecutive walking steps. Nominal manifolds are shown as thick blue trajectories. The subfigure on the right shows orthogonal tangent and cotangent manifolds (i.e., iso-parametric curves) in phase-space.

Illustration of the tangent and cotangent manifolds are given in Fig. 4.2. In subfigure 4.2(a), their intersection corresponds to the phase progression transition value $\zeta_{\text{trans}}$ and the guard $\mathcal{G}_{q\to q+1}$. Shown in red are the boundary manifolds $\sigma = \pm\epsilon$ of the invariant bundle $\mathcal{B}_q(\epsilon)$. For the current $q^{\text{th}}$ step, we can use the $\sigma = -\epsilon$ manifold of the next-step invariant bundle $\mathcal{B}_{q+1}(\epsilon)$ as the guard, namely, $\mathcal{G}_{q\to q+1} = \{(x, \dot{x}) \mid \sigma_{q+1} = -\epsilon\}$. In subfigure (b), the numbers represent the manifold deviations from the nominal one. For instance, the cotangent manifold $\zeta = 4$ has a tangent Riemannian distance 4 to the nominal cotangent manifold $\zeta = 0$. The red tangent manifolds are shown as curves of constant $\sigma$ as defined in Eq. (4.2). Thick lines in purple are the asymptotes of tangent manifolds[2] where the thick red line illustrates a specific manifold $\sigma = 2$. The asymptotes intercept the saddle point $(x_{\text{foot}}, 0)$, where $x_{\text{foot}}$ is the sagittal foot position. The green cotangent manifold are curves of constant $\zeta$ that are orthogonal to the tangent manifolds. Horizontal and vertical lines in orange are the asymptotes of cotangent manifolds, and the thick cyan line represents a specific manifold $\zeta = 3$. The vertical asymptote represents a manifold $\zeta = 0$.

[2]These two asymptotes are equivalent to the eigenvector lines in Capture Point (Pratt et al., 2006). The one in the second quadrant is stable while the one in the first quadrant is not.

In robust control theory (Zhou et al., 1996), input-output system response can be evaluated through the use of system norms. In this spirit, we define a new norm that characterizes sensitivity to disturbances of our non-periodic gaits, as

**Definition 4.1 (Phase-space sensitivity norm).** *Given a disturbance d, the phase-space sensitivity norm is defined as*

$$\kappa\Big(\sigma(x_{\zeta_d}, \dot{x}_{\zeta_d})\Big) = \Big(\frac{1}{\zeta_{\text{trans}} - \zeta_d} \int_{\zeta_d}^{\zeta_{\text{trans}}} \sigma(x_\zeta, \dot{x}_\zeta)^2 d\zeta\Big)^{1/2}, \tag{4.4}$$

*where, $\zeta_d$ corresponds to the phase value when a disturbance occurs and $\zeta_{\text{trans}}$ is the phase transition defined in Def. 3.9 for a given step.*

Contrary to other sensitivity norms (Hobbelen and Wisse, 2007; Hamed et al., 2015), our gait norm evaluates disturbance sensitivity for non-periodic gaits. It does so by explicitly accounting for disturbance magnitude and for the instant where disturbances occur. And it does not rely on approximate linearization nor Taylor series expansion as the previous periodic gait norms require. We will use the proposed norm in the control chapter for dynamic programming. Let us consider various types of disturbances and corresponding recovery strategies.

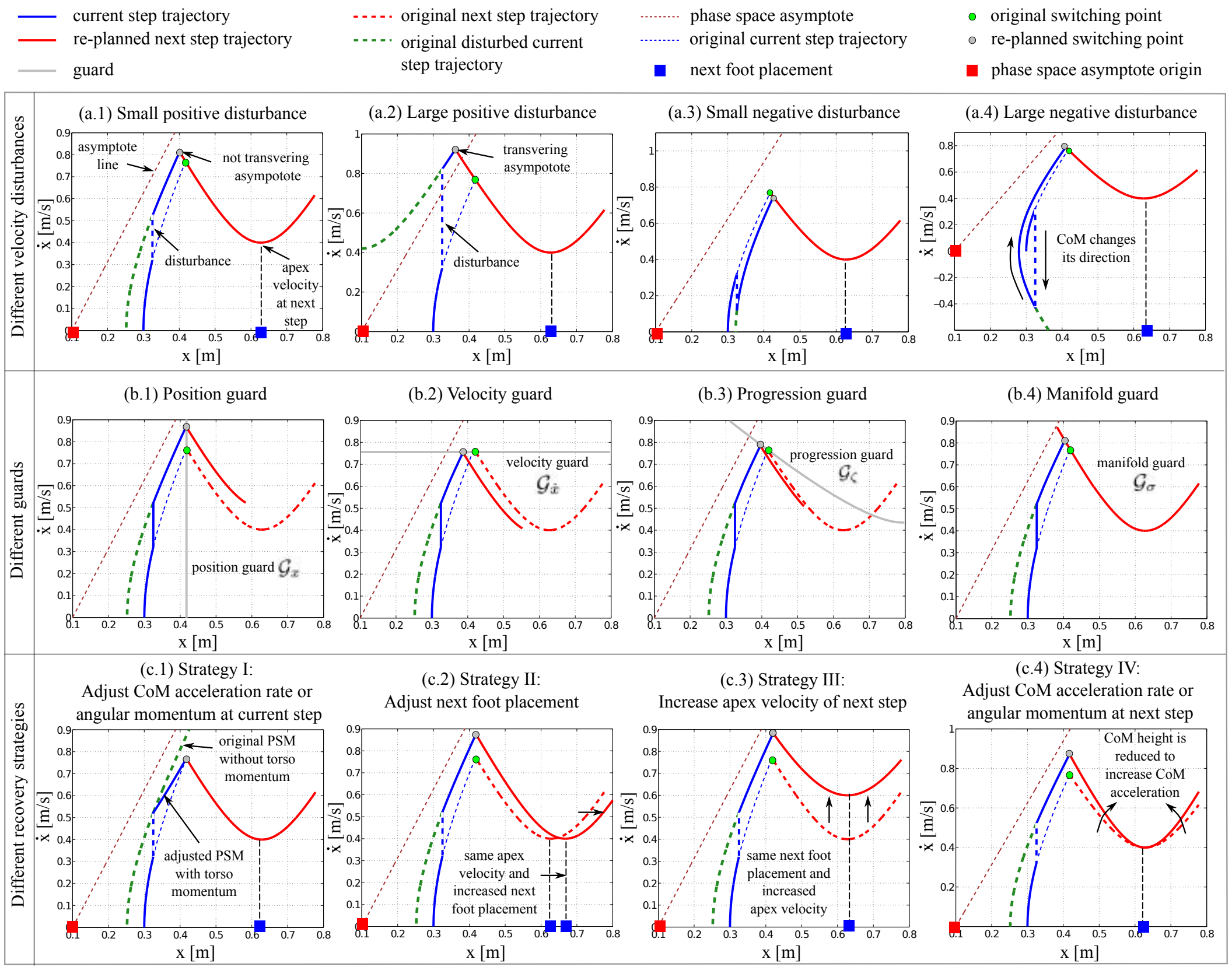

Figure 4.3: Disturbance pattern, guard and recovery strategy classification. Four different velocity disturbance cases are shown in subfigures (a.1)-(a.4). The second row shows four proposed guards for next step transition while the last row provides four recovery strategies.

**Definition 4.2** (**Phase-space disturbances**). *Disturbances in the phase-space can be categorized based on four characteristics: (1) the disturbance direction, (2) the disturbance magnitude, (3) the terminal asymptote-region, and (4) the change of the motion direction.*

Fig. 4.3 (a.1)-(a.4) illustrates those four scenarios, respectively. Here, we assume that the disturbances are impulses that change the velocity instantly[3]. (a.2) has a larger positive disturbance than (a.1) such that velocity after disturbance crosses the asymptote of the inverted pendulum model. On the other hand, (a.3) has a smaller negative disturbance such that velocity after disturbance keeps the same direction while (a.4) does not. In general, a disturbance can be characterized by its direction and magnitude. However, our study provides support for designing recovery strategies using the proposed phase-space control strategies. In such case, we need to understand whether the disturbances cross terminal regions. This is the reason why we incorporate additional disturbance categories.

More disturbance scenarios could be defined, depending on specific phase-space state locations and disturbance characteristics. Given these disturbed phase-space trajectories, new step transition strategies need to be considered. Here we propose four types of guard strategies for next step transition in Fig. 4.3 (b.1)-(b.4). The guards shown are: position guard $\mathcal{G}_x$ (vertical line), velocity guard $\mathcal{G}_{\dot{x}}$ (horizontal line), progression guard $\mathcal{G}_\zeta$ ($\zeta$-isoline), and manifold guard $\mathcal{G}_\sigma$ ($\sigma$-isoline). We find each guard such that they have the same transition point for the nominal PSM. Although this guard recovering strategy forces the motion to be adjusted, it might not be sufficient. If that is the case, we consider designing more recovery strategies by properly using control inputs. In the last four subfigures of Fig. 4.3, four recovery strate-gies are illustrated. These strategies are inspired by observations of human walking behaviors (Hofmann, 2006), (Kuo and Zajac, 1992), (Abdallah and Goswami, 2005) and by the experiences gained during extensive simulations. In order to fulfill those recovery strategies, a proper control policy will need to be designed.

[3]The disturbance could also be impulses changing the position or a combination of both position and velocity. In any case, the proposed disturbance characteristics and recovery strategies are still valid.

## 4.2 Robust Optimal Control Strategy

This section formulates a two-stage control procedure to recover from disturbances. When a disturbance occurs, the robot's CoM deviates from the planned phase-space manifolds obtained via Algorithm 1. Various control policies could be used for recovery from disturbances. We used dynamic programming to find an optimal policy of the continuous control variables for recovery, and, when necessary, feet placements are re-planned from their initial locations based on a policy that involves the guards defined in Fig. 4.3. Our proposed controller, relies on the distance metric of Eq. (4.2) to steer the robot current's trajectory to the planned manifolds.

### 4.2.1 Continuous Control: Dynamic Programming

We propose a dynamic programming-based controller for the continuous control of the sagittal locomotion behavior. To robustly track the planned CoM manifolds, we minimize a finite phase quadratic cost function and solve for the continuous control parameters, i.e.

$$\begin{aligned} &\min_{\boldsymbol{u}_{\boldsymbol{x}}^c} \ \mathcal{V}_N(q,\ \boldsymbol{x}_N) + \sum_{n=0}^{N-1} \eta^n \mathcal{L}_n(q, \boldsymbol{x}_n, \boldsymbol{u}_{\boldsymbol{x}}^c) \\ &\text{subject to}: \ \dot{\boldsymbol{x}} = \boldsymbol{\mathcal{F}}_{\boldsymbol{x}}(\boldsymbol{x}, \boldsymbol{u}_{\boldsymbol{x}}^c) \text{ in Eq. (3.20)} \\ &\qquad\qquad \omega^{\min} \leq \omega \leq \omega^{\max}, \\ &\qquad\qquad \tau_y^{\min} \leq \tau_y \leq \tau_y^{\max}, \end{aligned} \tag{4.5}$$

where $\boldsymbol{u}_{\boldsymbol{x}}^c = \{\omega, \tau_y\}$ corresponds to the continuous variables of the hybrid control input $\boldsymbol{u}_{\boldsymbol{x}}$ of Eq. (3.20)[4], $\omega$ and $\tau_y$ are scalars in this case, $0 \leq \eta \leq 1$ is a discount factor, $N$ is the number of discretized stages until the next step transition, $\zeta_{\text{trans}}$, the terminal cost is, $\mathcal{V}_N(q, \boldsymbol{x}_N) = \alpha(\dot{x}(\zeta_{\text{trans}}) - \dot{x}(\zeta_{\text{trans}})^{\text{pred}})^2$. Here, $\dot{x}(\zeta_{\text{trans}})$ is the actual velocity at the instant of the next step transition and $\dot{x}(\zeta_{\text{trans}})^{\text{pred}}$ is the predicted transition velocity. Additionally, $\mathcal{L}_n$ is the one step cost-to-go function at the $n^{\text{th}}$ stage defined as a weighted square sum of the tracking errors and control variables:

$$\mathcal{L}_n(q, \boldsymbol{x}_n, \boldsymbol{u}_{\boldsymbol{x}}^c) = \int_{\zeta_{q,n}}^{\zeta_{q,n+1}} \big[\beta\sigma^2 + \Gamma_1 \tau_y^2 + \Gamma_2(\omega - \omega^{\text{ref}})^2\big] d\zeta, \tag{4.6}$$

where $\sigma$ is the tangent bundle of Eq. (4.1) used as a feedback control parameter, $\zeta_{q,n}$ and $\zeta_{q,n+1}$ are the starting and ending phase progressions for the $n^{\text{th}}$ stage of the $q^{\text{th}}$ walking step, $\alpha$, $\beta$, $\Gamma_1$ and $\Gamma_2$ are weights, and $\omega^{\text{ref}}$ is the reference phase-space asymptote slope given in Algorithm 1. Eq. (4.5) is solved in a backward propagation pattern. More details of dynamic programming are provided in Appendix D. This optimal control process is applied only when a disturbance occurs. In disturbance-free scenario, no control adjustments are required, as the system naturally follows its CoM dynamics.

[4]For simplicity, the yaw torque $\tau_z$ is assumed to be zero in the disturbance case and thus it is not included in $\boldsymbol{u}_x^c$.

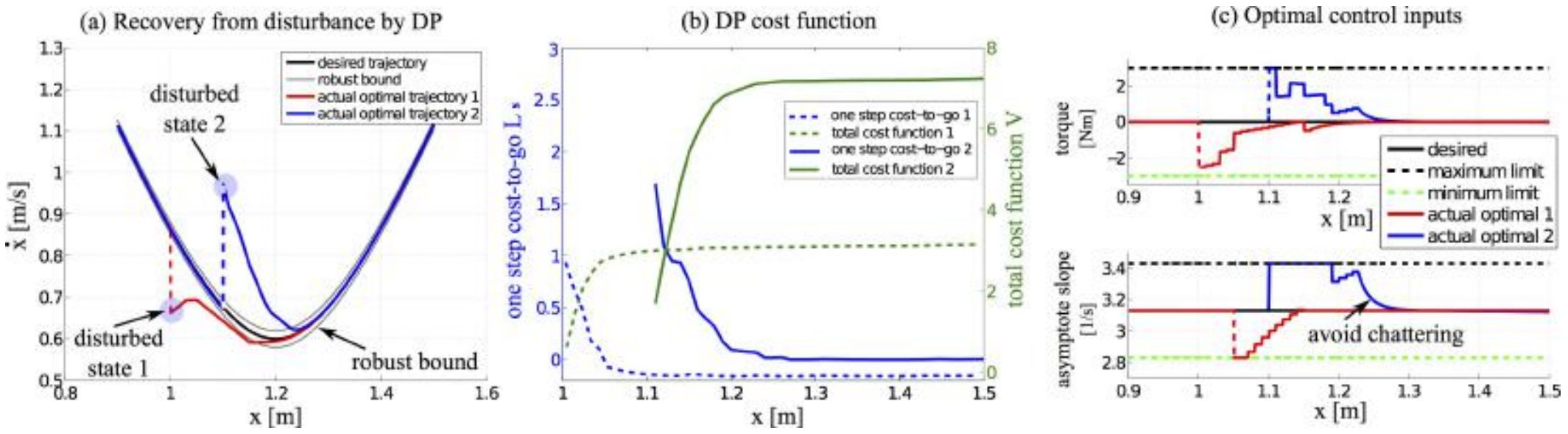

Figure 4.4: Chattering-free recoveries from disturbance by the proposed optimal recovery control law. Subfigure (a) shows two random disturbances, where disturbed state 1 has a negative impulse while the disturbed state 2 has a positive impulse. Control variables are piece-wise constant within one stage as shown in subfigure (c).

To avoid chattering effects[5] in the neighborhood of the planned manifold, a boundary layer is defined and used to saturate the controls, i.e.

$$\boldsymbol{u}_{\boldsymbol{x}}^{c'} = \begin{cases} \boldsymbol{u}_{\boldsymbol{x}}^{c} & |\sigma| > \epsilon \quad (4.7a) \\ \dfrac{|\sigma|}{\epsilon}\boldsymbol{u}_{\boldsymbol{x}}^{c,\epsilon} + \dfrac{\epsilon - |\sigma|}{\epsilon}\boldsymbol{u}_{\boldsymbol{x}}{}^{c,\text{ref}} & |\sigma| \leq \epsilon \quad (4.7b) \end{cases}$$

where $\epsilon$ corresponds to the boundary value of an invariant bundle $\mathcal{B}(\epsilon)$ as defined in Def. 3.4, $\boldsymbol{u}_{\boldsymbol{x}}^{c,\epsilon} = \{\omega^{\epsilon}, \tau_y^{\epsilon}\}$ are control inputs at the instant when the trajectory enters the invariant bundle $\mathcal{B}(\epsilon)$, $\boldsymbol{u}_{\boldsymbol{x}}^{c,\text{ref}}$ are nominal control inputs defined in Algorithm 1. The indication of smoothness of the above control follows from (Utkin, 2013). As Eq. (4.7) shows, when $|\sigma| \leq \epsilon$, the control effort, $\boldsymbol{u}_{\boldsymbol{x}}^{c'}$ is scaled between $\boldsymbol{u}_{\boldsymbol{x}}^{c,\epsilon}$ and $\boldsymbol{u}_{\boldsymbol{x}}^{c,\text{ref}}$. This control law is composed of an "inner" and an "outer" controller. The "outer" controller steers states into $\mathcal{B}(\epsilon)$ while the "inner" controller maintains states within $\mathcal{B}(\epsilon)$. Note that, this controller performs better than asymptotic stability since the invariant bundle $\mathcal{B}(\epsilon)$ is reached in finite time. Recovery trajectories are shown in Fig. 4.4 for two scenarios in the presence of random disturbance (opposite directions). In these simulations, angular momentum control range is $[-3, 3]$ Nm and acceleration rate range is $[2.83, 3.43]$ 1/s. Other parameters are shown in Table 4.1.

[5]This chattering is caused by the digital controllers with finite sampling rate. In theory, an infinite switching frequency will be required. However, the control input in practice is constant within a sampling interval, and thus the real switching frequency cannot exceed the sampling frequency. This limitation leads to the chattering.

Table 4.1: Dynamic programming parameters

| Parameter | Range | Parameter | Range |
|---|---|---|---|
| nominal pitch torque $\tau_y^{\text{ref}}$ | 0 Nm | nominal asymptote slope $\omega^{\text{ref}}$ | 3.13 1/s |
| pitch torque range $\tau_y^{\text{range}}$ | [-3, 3] Nm | asymptote slope range $\omega^{\text{range}}$ | [2.83, 3.43] 1/s |
| apex height $z_{\text{apex}}$ | 1 m | mass $m$ | 1 kg |
| stage range | [0.9, 1.5] m | state range | [0.03, 1.5] m/s |
| stage resolution | 0.01 m | state resolution | 0.01 m/s |
| disturbed initial state $s_{\text{initial}}$ | (1.1 m, 0.7 m/s) | nominal apex velocity $\dot{x}_{\text{apex}}$ | 0.6 m/s |
| weighting scalar $\Gamma_1$ | 5 | weighting scalar $\Gamma_2$ | 5 |
| weighting scalar $\beta$ | $4 \times 10^4$ | weighting scalar $\alpha$ | 100 |

Since the control is constrained, we need to define the finite-phase (control-dependent) recoverability bundle for the Sagittal phase-space. Given an acceptable deviation $\epsilon_0$ from the manifold, the practical invariant bundle is $\mathcal{B}(\epsilon_0)$. The control policy in Eq. (42) generates a control-dependent practical recoverability bundle defined as follows.

**Definition 4.3** (**Control-dependent recoverability bundle**). *A control-dependent recoverability bundle, a.k.a., region of attraction to the "boundary-*

*layer", is defined as*

$$\mathcal{R}(\epsilon, \zeta_{\text{trans}}) = \Big\{(x, \dot{x})_\zeta,\ \zeta_0 \le \zeta \le \zeta_{\text{trans}} \;\Big|\; (x, \dot{x})_{\zeta_{\text{trans}}} \in \mathcal{B}(\epsilon),\ \boldsymbol{u}^c_{\boldsymbol{x}} \in \boldsymbol{u}^{c,\text{range}}_{\boldsymbol{x}}\Big\}.$$

Unlike the one given in Def. 3.5, the recoverability bundle $\mathcal{R}$ above has two specific features: i) dependence on accessible control ranges, and ii) a finite convergence phase.

**Theorem 4.1 (Existence of recoverability bundle).** *Given a Lyapunov function $V = \sigma^2/2$, a phase progression transition value $\zeta_{\text{trans}}$, and the control policy in Eq. (4.7), a recoverability bundle $\mathcal{R}(\epsilon, \zeta_{\text{trans}})$ exists and can be estimated by a practical recoverability bundle $\mathcal{R}^o(\sigma_{\text{init}}, \zeta_{\text{trans}})$, where $\sigma_{\text{init}}$ is the tube initial radius.*

*Proof.* We use a Lyapunov function to prove the existence of $\mathcal{R}(\epsilon, \zeta_{\text{trans}})$. First, let us define $V = \sigma^2/2$ based on Eq. (4.7a). If there exists a control policy such that $\exists\ \sigma_0 > \epsilon, \sigma_{\text{trans}} \le \epsilon$, then, $\mathcal{R}(\epsilon, \zeta_{\text{trans}})$ is composed of the range of values $(x, \dot{x})_\zeta,\ \zeta_0 \le \zeta \le \zeta_{\text{trans}}$, such that $V_{\text{trans}} = \sigma^2_{\text{trans}}/2 \le \epsilon^2/2$. Taking the derivative of $V$ along the pendulum dynamics in Eq. (3.14), we have

$$\begin{aligned}
\dot{V} = \sigma\dot{\sigma} &= \sigma\dot{x}^2_{\text{apex}}\Big(-2\dot{x}(x - x_{\text{foot}}) + 2\dot{x}\ddot{x}/\omega^2\Big)\\
&= \sigma\dot{x}^2_{\text{apex}}\Big(-2\dot{x}(x - x_{\text{foot}}) + 2\dot{x}\big((x - x_{\text{foot}}) - \frac{\tau_y}{mg}\big)\Big)\\
&= -\frac{2\dot{x}^2_{\text{apex}}\sigma\dot{x}\tau_y}{mg} = -\frac{2\sqrt{2}\dot{x}^2_{\text{apex}}\dot{x}\tau_y \cdot \text{sign}(\sigma)}{mg}\sqrt{V} \le 0. \qquad (4.8)
\end{aligned}$$

which proves the stability (i.e., attractiveness) of $\sigma = 0$. For example, consider the case of walking forward, $\dot{x} > 0$. Then, as long as $\sigma \cdot \tau_y > 0$, i.e., the pitch torque has the same sign as $\sigma$, attractiveness is guaranteed. That is, if $\sigma > 0$ (the robot moves forward faster than expected), then we need $\tau_y > 0$ to slow down, and vice-versa. If $\tau_y = 0$, then $\dot{V} = 0$, which implies a zero convergence rate. This means that the CoM state will follow its natural inverted pendulum dynamics without converging. As such, in order to converge to the desired invariant bundle, $\tau_y$ control action is required.

To estimate $\mathcal{R}(\epsilon, \zeta_{\text{trans}})$, we can use (i) the optimal control policy proposed in Eq. (4.7), defining an "optimal" recoverability bundle; or the maximum control inputs (without any regards to optimality) by selecting $\boldsymbol{u}_{\boldsymbol{x}}^{c}$ as the bounds of $\boldsymbol{u}_{\boldsymbol{x}}^{c,\text{range}}$, defining the "maximum" recoverability bundle. These two cases are derived below.

**Case I: DP based Control.** If $\tau_y$ is solved by the optimization in Eq. (4.5), we have $|\tau_y| > |\tau_y^{\epsilon}|$. Then Eq. (4.8) becomes

$$\dot{V} < -\frac{2\sqrt{2}\dot{x}_{\text{apex}}^2 \dot{x} |\tau_y^{\epsilon}|}{mg}\sqrt{V} < 0. \tag{4.9}$$

**Case II: Supremum (Bang-Bang) Control.** If we design $\tau_y = \tau_y^{\max}\text{sign}(\dot{x})$ and $\dot{x} > 0$, then,

$$\dot{V} = -\frac{2\sqrt{2}\dot{x}_{\text{apex}}^2 \dot{x} \tau_y^{\max}\text{sign}(\dot{x})}{mg}\sqrt{V} = -\frac{2\sqrt{2}\dot{x}_{\text{apex}}^2 \dot{x} \tau_y^{\max}}{mg}\sqrt{V} < 0. \tag{4.10}$$

Note that, the bounded $\dot{V}$ in Eqs. (4.9) and (4.10) have similar forms and can be integrated to

$$\int_{V_0}^{V_{\text{trans}}} \frac{dV}{\sqrt{V}} = -\int_{t_0}^{t_{\text{trans}}} \mu \dot{x} \tau_y dt = -\mu \tau_y (x_{\text{trans}} - x_0), \tag{4.11}$$

where $\mu = (2\sqrt{2}\dot{x}_{\text{apex}}^2)/(mg)$, $\tau_y = \tau_y^{\epsilon}$ for Case I while $\tau_y = \tau_y^{\max}$ for Case II. Eq. (4.11) can be simplified to

$$\sqrt{V_0} = \sqrt{V_{\text{trans}}} + \frac{1}{2}\mu \cdot (x_{\text{trans}} - x_0) \cdot \tau_y \quad \Rightarrow \quad \sigma_0 = \epsilon + \frac{\sqrt{2}}{2}\mu \cdot (x_{\text{trans}} - x_0) \cdot \tau_y. \tag{4.12}$$

Thus, the practical recoverability bundle, $\mathcal{R}^o(\sigma_{\text{init}}, \zeta_{\text{trans}})$, can be expressed in terms of the initial recoverability bundle radius $\sigma_{\text{init}}$.

$$\begin{aligned} \mathcal{R}(\epsilon, \zeta_{\text{trans}}) &\leq \mathcal{R}^o(\sigma_{\text{init}}, \zeta_{\text{trans}}) \\ &= \Big\{(x, \dot{x})_{\zeta},\ \zeta_0 \leq \zeta \leq \zeta_{\text{trans}} \ \Big|\ \sigma_{\text{init}} \leq \sigma_0(\epsilon, x_{\text{trans}}, \dot{x}_{\text{apex}}, \tau_y)\Big\}. \end{aligned} \tag{4.13}$$

Since Eq. (4.9) has an inequality bound while Eq. (4.10) has an equality bound, DP based control is an optimal but conservative estimation of the true recoverability bundle while supremum control is an accurate but non-optimal estimation for the recoverability bundle. There are trade-offs between these two control strategies. □

Since our study aims at optimal performance, we will use the control policy generated from dynamic programming to estimate the recoverability bundle. To estimate $\mathcal{R}(\epsilon, \zeta_{\text{trans}})$, we perform a grid sampling for the initial condition $\boldsymbol{x}_{\zeta_0}$, based on the ranges of stage $x$ and state $\dot{x}$ in Table 4.1. That is, we execute the optimization in Eq. (4.5) for each sampled $\boldsymbol{x}_{\zeta_0}$ (treated as a realization) and repeat this procedure for all the $\boldsymbol{x}_{\zeta_0}$. Then the feasible realizations of recovery trajectories (here, "recovery" means the convergence into $\mathcal{B}(\epsilon)$ before $\zeta_{\text{trans}}$) constitute the recoverability bundle[6]. An example of an estimated recoverability bundle is shown in Fig. 4.5 (a). The control constraint is: $\omega$ range [-0.3, 0.3] 1/s and $\tau_y$ range [-3, 3] N/m. For better visualization, Fig. 4.5 (b) uses the range value to represent the control inputs in the horizontal axes. For instance, if the torque range is $r$, then it implies $\tau_y \in [-r/2, r/2]$.

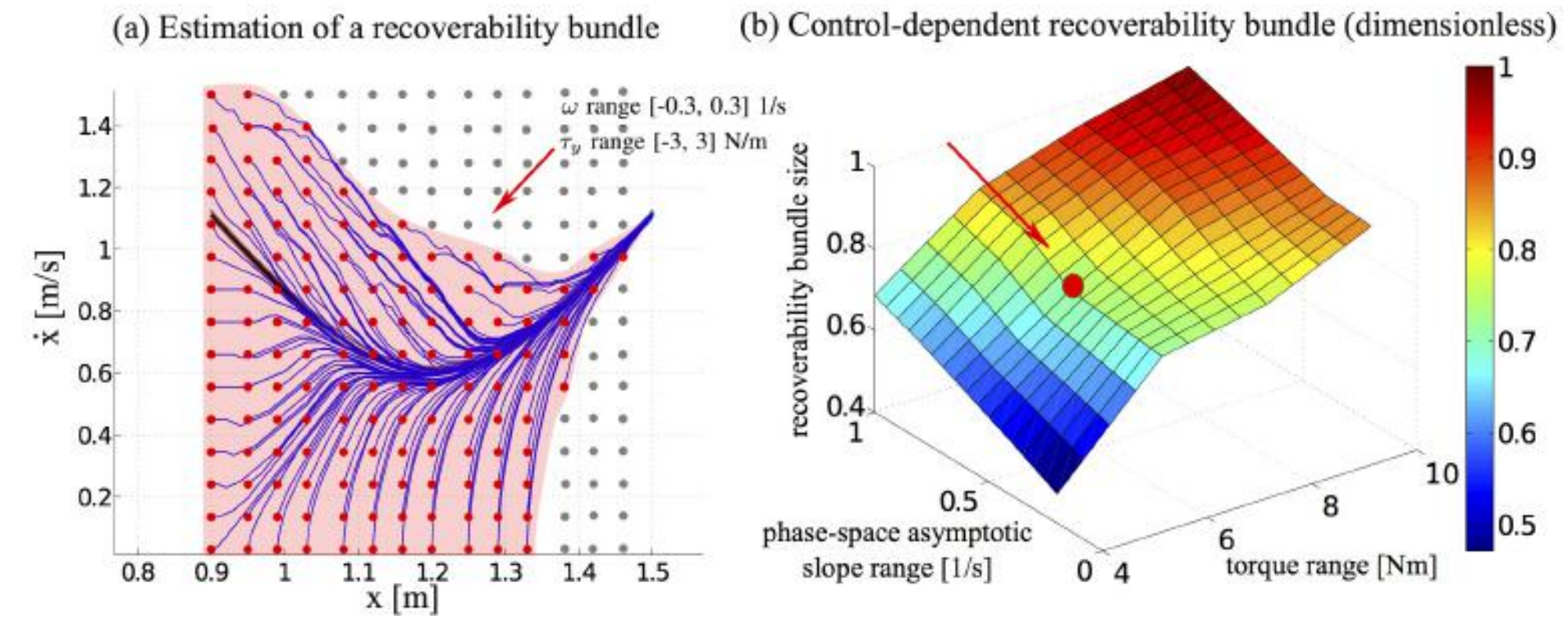

Figure 4.5: Estimation of control-dependent recoverable region. In the left figure, disturbed states are sampled in a discretized grid and the shaded region represents the recoverability bundle. As it shows, a larger recoverable region is achieved in the beginning phase (i.e., before apex state). In the ending phase, the recoverable region shrinks to the invariant bundle. The right figure illustrates the dependence of the recoverable region size on allowed control ranges.

**Remark 4.2.** *Other related methods for computing recoverability or reachable sets are (i) adopting Lyapunov function methods to estimate the recoverability bundle (Frazzoli, 2001), LQR-Trees (Tedrake et al., 2010), sums-of-squares*

[6]Only forward walking is considered in this study. Recovery from disturbances during backward or forward-to-backward walking could be achieved in a similar manner. If we take the backward walking for instance, all that is needed is to plan a proper sequence of apex states and integrate phase-space trajectories in a backward pattern, detect the PSM deviation via Eq. (4.2) and look up an offline precomputed DP policy table designed for backward walking.

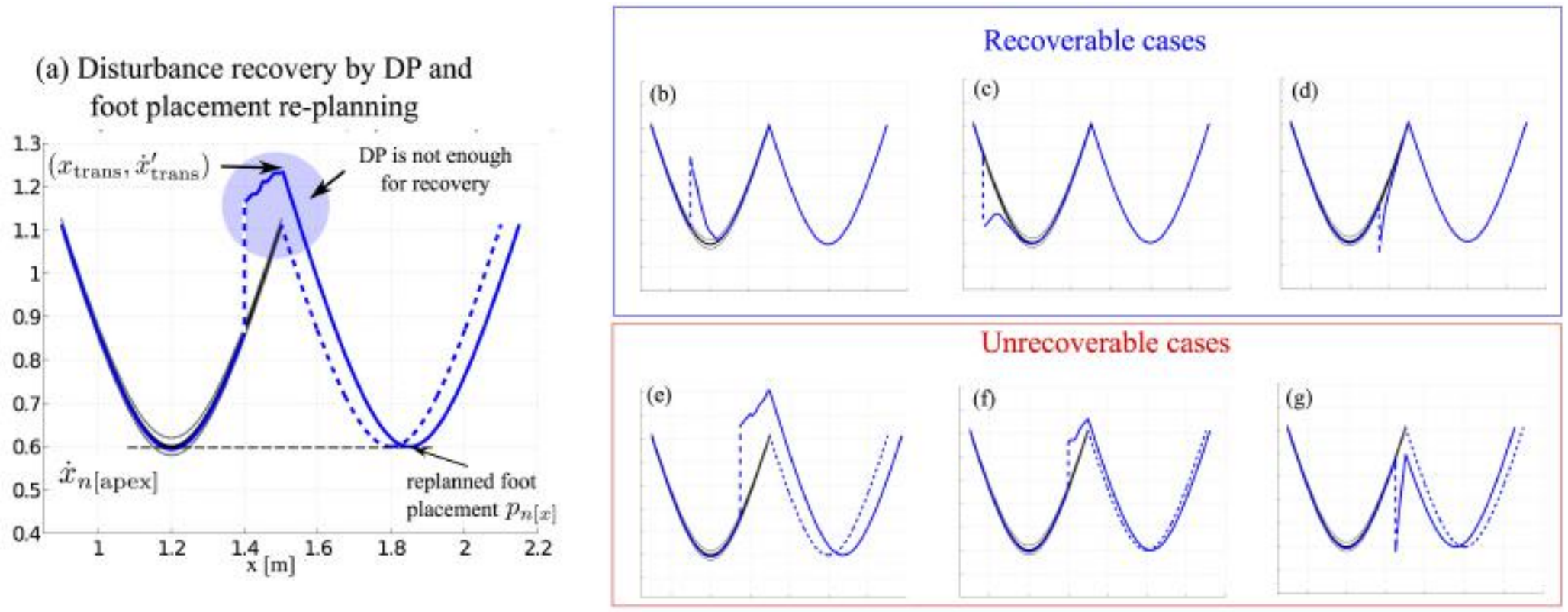

Figure 4.6: Recovery from disturbance by re-planning sagittal foot placement. Here, next apex velocity is given *a priori*. In our case, we assume apex velocity is maintained to be the same. In subfigures (b)-(d), first-stage DP optimization is sufficient to achieve the recovery. However, cases in subfigures (e)-(g) can not. These happened when the combination of disturbance occurred to close to transition and the size of the disturbance resulted in not enough time and power to recover. Instead, it automatically re-plans its new next foot placement based on Eq. (4.14).

*(SOS) (Majumdar, 2016) or its scaled-up version Scaled-Diagonally-Dominant SOS (Ahmadi and Majumdar, 2014) (suitable for high-dimensional systems) and (ii) using dynamic game formulation based Hamilton-Jacobi methods (Mitchell et al., 2005; Gillula et al., 2011).*

### 4.2.2 Discrete Control: Foot Placement Re-planning

As mentioned before, when the disturbance is large enough to take the system outside the recoverability manifold, $\mathcal{R}(\epsilon, \zeta_f)$, the controller will not be able to recover to the invariant manifold within a single stepping cycle. Any of the guard strategies proposed in Section 4.1 can be used. Here, we will use the position guard $\mathcal{G}_x$ as an example, and re-plan the foot placement of the next step as proposed in Strategy (ii) of Fig. 4.3. We try to maintain the next apex velocity $\dot{x}_{\text{apex}_{q+1}}$ as planned. Hence, the re-planned foot placement is analytically solved from PSM in Eq. (4.1). Let us define the re-planned phase-space transition state as $(x_{\text{trans}}, \dot{x}^{\text{rep}}_{\text{trans}})$, where only velocity $\dot{x}^{\text{rep}}_{\text{trans}}$ varies. Since $\dot{x}_{\text{apex}_{q+1}}$ is unchanged, the adjusted sagittal foot placement $x_{\text{foot}_{q+1}}$ is solved by

$$x_{\text{foot}_{q+1}} := x_{\text{trans}} + \frac{1}{\omega}(\dot{x}^{\text{rep2}}_{\text{trans}} - \dot{x}^2_{\text{apex}_{q+1}})^{1/2}. \tag{4.14}$$

In forward walking, the condition $x^{\text{rep}}_{\text{foot}_{q+1}} > x_{\text{trans}}$ holds, prompting us to ignore the solution with the negative square root. Note that if $\dot{x}_{\text{apex}_{q+1}} = 0$, i.e., the robot is coming to a stop, Eq. (4.14) becomes $x^{\text{rep}}_{\text{foot}_{q+1}} = x_{\text{trans}} + \dot{x}^{\text{rep}}_{\text{trans}}/\omega$, which is equivalent to the Capture Point dynamics described in (Englsberger et al., 2011).

**Proposition 4.3** (**Foot placement re-planning**). *Given a disturbance that occurred at $\zeta_d$, which moved the state $\boldsymbol{x}_{\zeta_d}$ outside the practical recoverability bundle, $\mathcal{R}^o(\sigma_{\text{init}}, \zeta_{\text{trans}})$, i.e., $\sigma_{\text{init}} \leq \sigma_0(\epsilon, x_{\text{trans}}, \dot{x}_{\text{apex}}, \tau_y)$, then the next foot placement will be re-planned by Eq. (4.14).*

To evaluate the performance of this step re-planning method, we consider the six disturbances scenarios of Fig. 4.6. The top three scenarios are recoverable using the DP-based continuous controller that we presented earlier. In the bottom three scenarios the disturbance occurs too close to the transition or is too large and therefore requires the foot placement re-planner described above to be executed. Once foot placements have been re-planned in the sagittal direction, lateral foot placements are re-planned using Algorithm 2.

To conclude, the two-stage procedure discussed in this section constitutes the core process of our robust-optimal phase-space planning strategy. The combined locomotion planning procedure is shown in Algorithm 3 in the Appendix. We use the continuous control strategy first to better track the desired locomotion trajectories. As such, the continuous controller represents a servo process that is always on. When deviations are too large for recovery, we apply the foot placement re-planner.

**Remark 4.3.** *The computational burden of our control process is low. The reason lies in that the DP-based continuous controller is designed offline to precompute a table storing all possible policies for any admissible disturbance. Therefore, once disturbances are detected, the offline preprocessing table is quickly looked at in real-time. A similar procedure is proposed in (Majumdar, 2016) for funnel-based robust feedback motion planning. In the case of the discrete foot placement re-planner, it is fast to compute due to its closed-form solution given in Eq. (4.14).*

## 4.3 End-to-End Phase-Space Planning Procedure

Robust optimal phase-space planning is proposed for agile locomotion planning and targets hierarchical control for disturbance rejections. This framework uses phase-space techniques to design a robust hybrid automaton in Section 3.5.4 to achieve 3D agile and versatile locomotion behaviors. In particular, a low-dimensional manifold is proposed in Section 4.1 as a robustness metric to quantify phase-space deviations and a sensitivity norm is used as a penalty metric in dynamic programming for disturbance rejections in Section 4.2. An overall planning and control procedure is shown in Fig. 4.7. The locomotion designer first provides the following pieces of information: (i) desired CoM surfaces, (ii) nominal foot positions, (iii) desired keyframe states, and (iv) flywheel torque limits. Algorithm 1 produces the locomotion trajectories and the distance metric aided by the step transition solver, the lateral foot placement planner, and the steering model. A disturbance checker verifies if the current trajectories are within the invariant bundle. If they are not, the results stored on a table from the DP-based controller are utilized as a control policy and if this is not sufficient, new steps are re-planned in an online fashion.

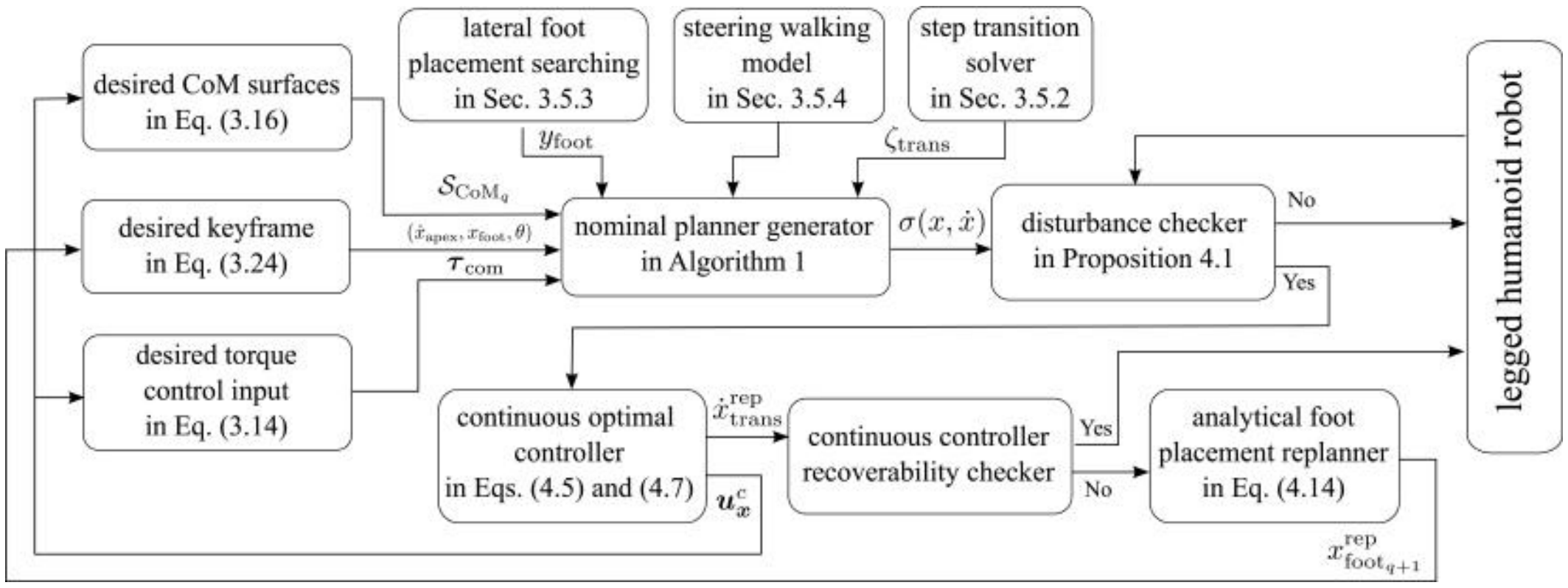

Figure 4.7: End-to-end planning and control process. This diagram describes the pipeline for generating locomotion plans in the phase-space and robust control strategies.

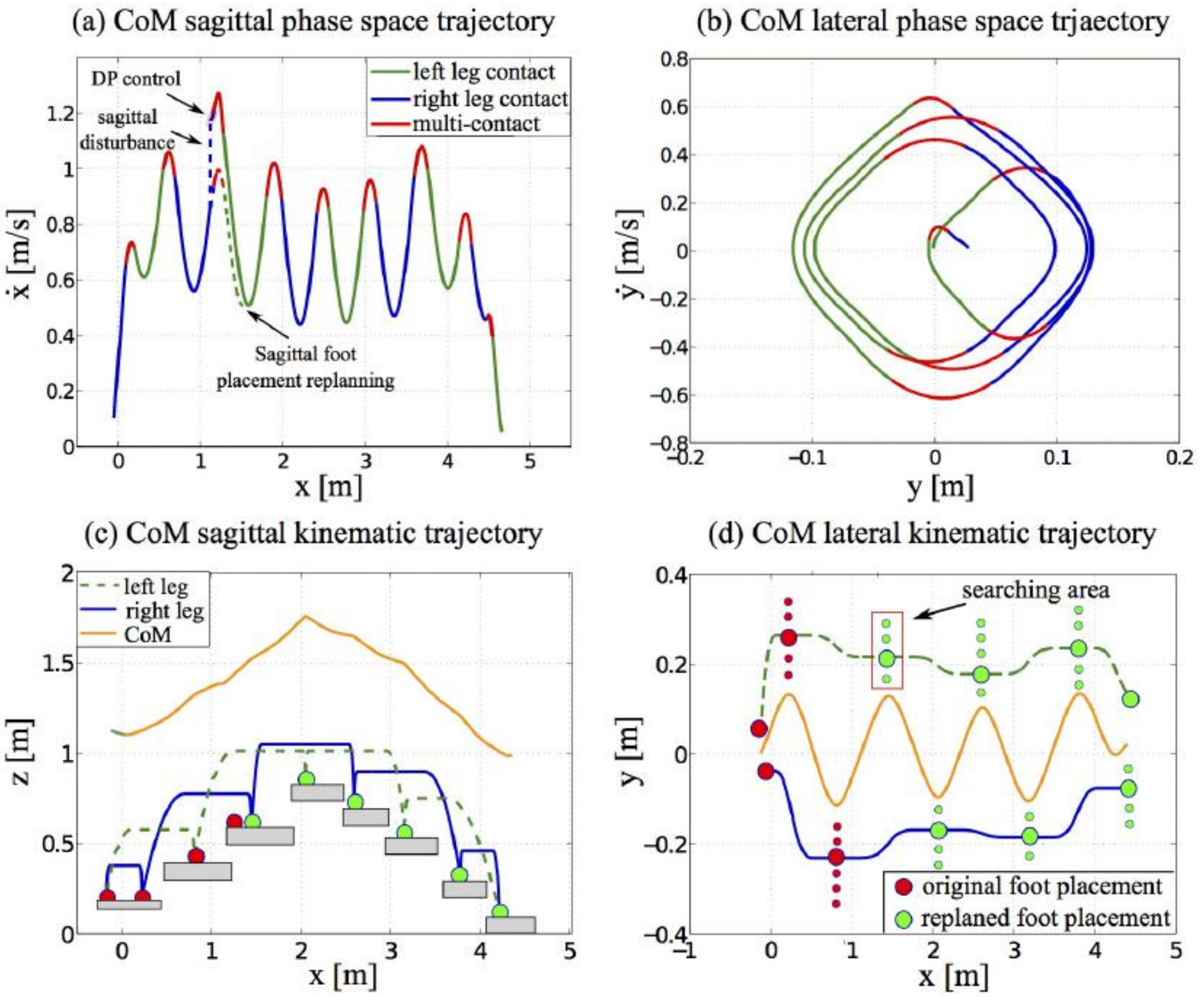

Figure 4.8: Rough terrain recovery from sagittal disturbance. To recover from a sagittal push with a 0.4 m/s CoM velocity jump, the planner uses both DP continuous control and discrete foot placement re-planning in a sequential manner. (red circle) denotes the pre-defined foot placement before the disturbance while (green circle) denotes the re-planned foot placement after the disturbance.

## 4.4 Example: Dynamic Walk under Disturbance

In this section, we focus on two recovery scenarios from external disturbance. The first is to validate the effectiveness of dynamic programming based optimization for continuous control when sagittal disturbance occurs, while the second is to verify the lateral foot placement re-planning strategy.

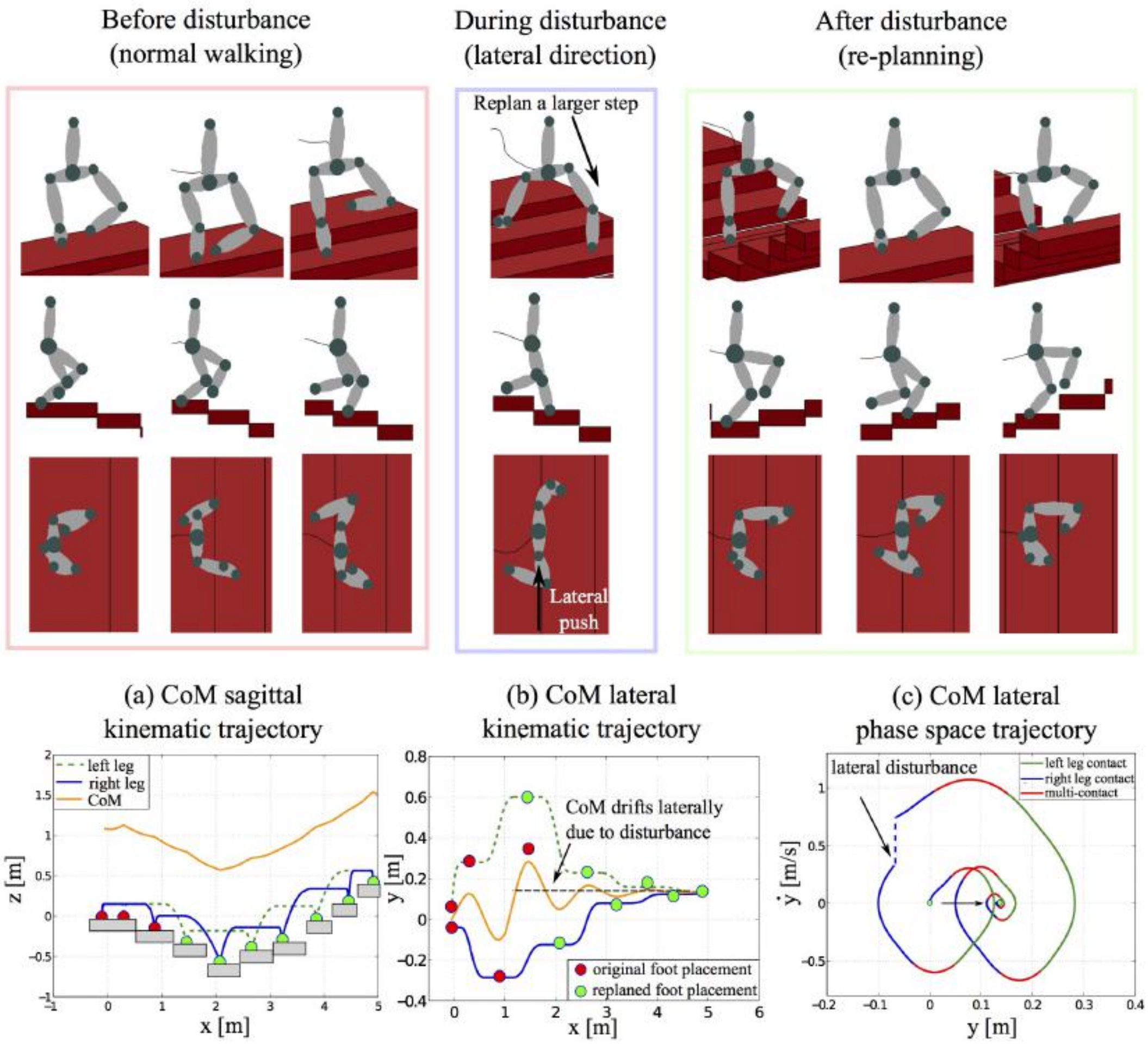

Figure 4.9: Rough terrain dynamic walking under lateral disturbance. During the lateral disturbance phase, the robot re-plans its foot placement to avoid falling down. In (d), the CoM trajectory drifts away to the left side. Also in the phase portrait (f), a lateral CoM velocity jump emerges and the lateral phase portrait is moved aside. After the disturbance, the foot planner re-plans the desired lateral foot placement and regulates the CoM trajectory to achieve balanced walking.

### 4.4.1 Optimal Control for Recovery from Sagittal Disturbance

A sagittal push is exerted on the CoM, which causes an instantaneous velocity jump as shown in Fig. 4.8 (b). This disturbance is so large that the phase-space state can not recover to its nominal PSM in one single step. Thus, a sagittal foot re-planning strategy is executed to re-compute a sagittal foot step immediately while maintaining the same next apex velocity. The dash line in Fig. 4.8 (b) represents the original phase-space primitive while the blue solid line represents the re-planned one. Specifically, to make the animation more consistent with real robot kinematics, the robot's torso and hip are colored in red. Also, instead of an instantaneous step transition, a multi-contact transition is used as shown in Fig. 4.8 (a) and (d).

### 4.4.2 Re-planning Strategy for Lateral Disturbance

In this case, we specifically validate the lateral foot placement re-planning strategy in our second-stage optimization by exerting a lateral disturbance. In the third step, the robot bears a lateral CoM disturbance, which similarly causes a lateral velocity jump in Fig. 4.9 (f) and a lateral CoM kinematic trajectory drift in Fig. 4.9 (d). To deal with this, a new lateral foot placement is re-planned and a semi-limit cycle property is maintained for the lateral phase-space primitive. Note that, since angular momentum is not considered in lateral inverted pendulum dynamics, the unique strategy to recover from lateral disturbance is to adjust its foot placement.

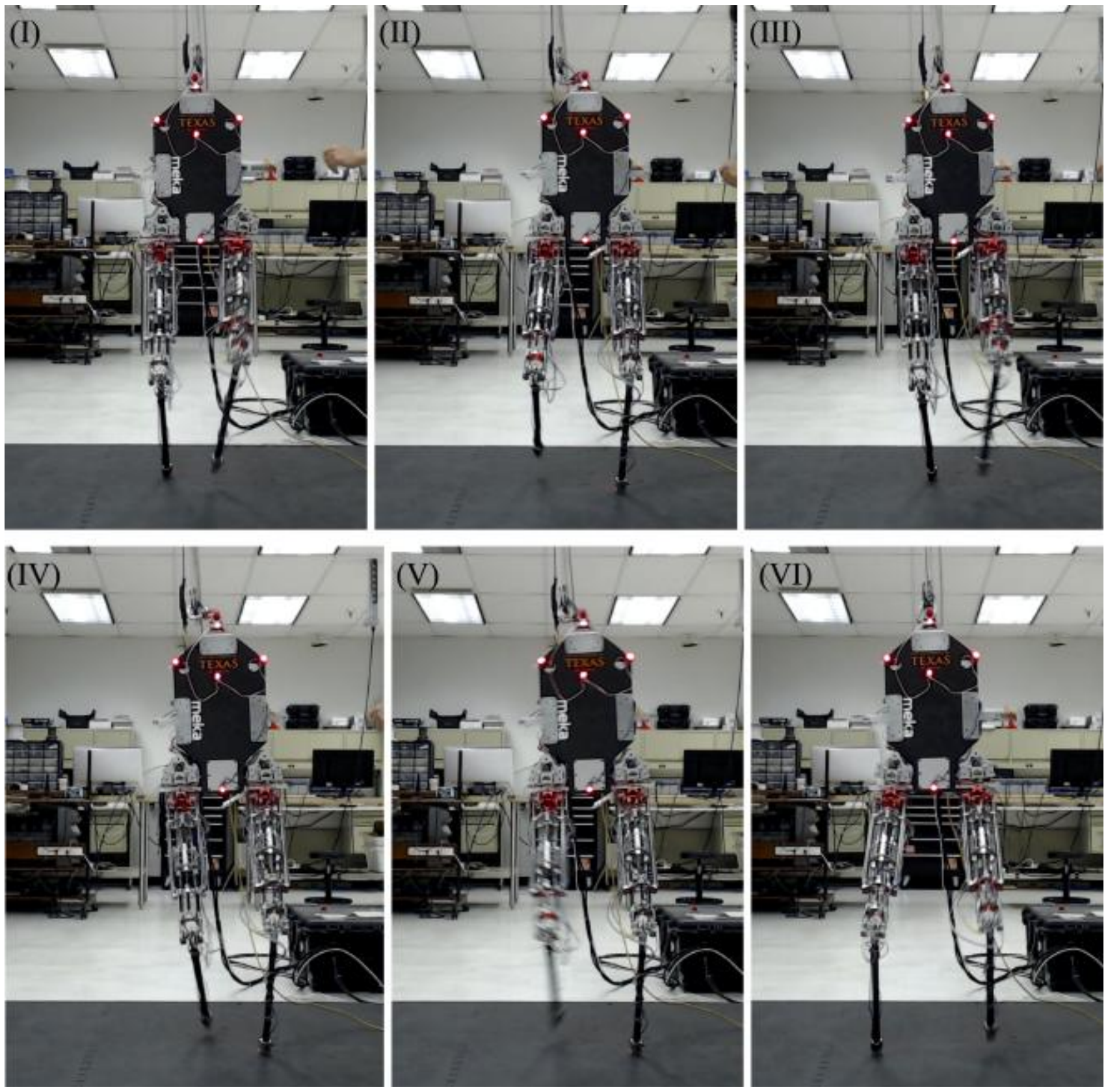

Figure 4.10: Snapshots of unsupported dynamic balancing experiment. This figure shows a portion of an experiment in which our point foot biped robot, Hume, accomplishes 18 steps of unsupported dynamic balancing. The snapshots here show the $3^{\text{th}} - 6^{\text{th}}$ walking steps.

## 4.5 Experimental Evaluations

In this section[7], we briefly demonstrate a proof of concept of the phase-space planner by experimental implementations on our bipedal point-feet robot Hume. For more details about robot specifications and the Whole-Body Operational Space Control (WBOSC), please refer to the recent work in (Kim et al.,

[7] Experimental results in this subsection are primarily lead by Donghyun Kim.

2016a). Regarding non-trivial mechanical uncertainties and sensing noise, it is difficult to follow the planned multi-step phase-space trajectory accurately. To avoid this problem, the robot has to re-plan its CoM trajectory and foot placements after each step in an online pattern. The practical walking performance reveals a great level of robustness. That is the reason why we present this experimental evaluation in this chapter focusing on the robust control.

Achieving dynamic walking of a point-feet bipedal robot is an undoubted challenge. The robot in this experiment aims at achieving dynamic balancing without any external support. Footstep locations are computed at runtime such that the proposed prismatic inverted pendulum model is stabilized. This strategy should result in a stable balance behavior assuming the trajectory generators and WBOSC, i) successfully place the feet at the desired locations, ii) achieve the desired height of the center of mass, and iii) fix the orientation of the robot's torso.

The foot swing motion constitutes two phases: lifting and landing phases. The landing location is designed before the landing phase starts. During each walking step, when the lifting phase reaches 70% completion, the planner computes the next footstep location. The value of 70% was empirically determined to ensure the planner completes before the landing phase starts. It is a processor intensive task that must be run outside of the real-time thread. The operational space set-point trajectory for the swing foot is then defined based on a polynomial function and the desired landing position, with the trajectory ending once ground contact is sensed. If the ground is at the expected height and the position tracking is ideal, the footstep will land after the nominal swing time. If the planned step is outside the mechanical limits of the robot, the planner chooses the closest reachable step. For more details of this footstep placement algorithm, please refer to the results in (Kim et al., 2014).

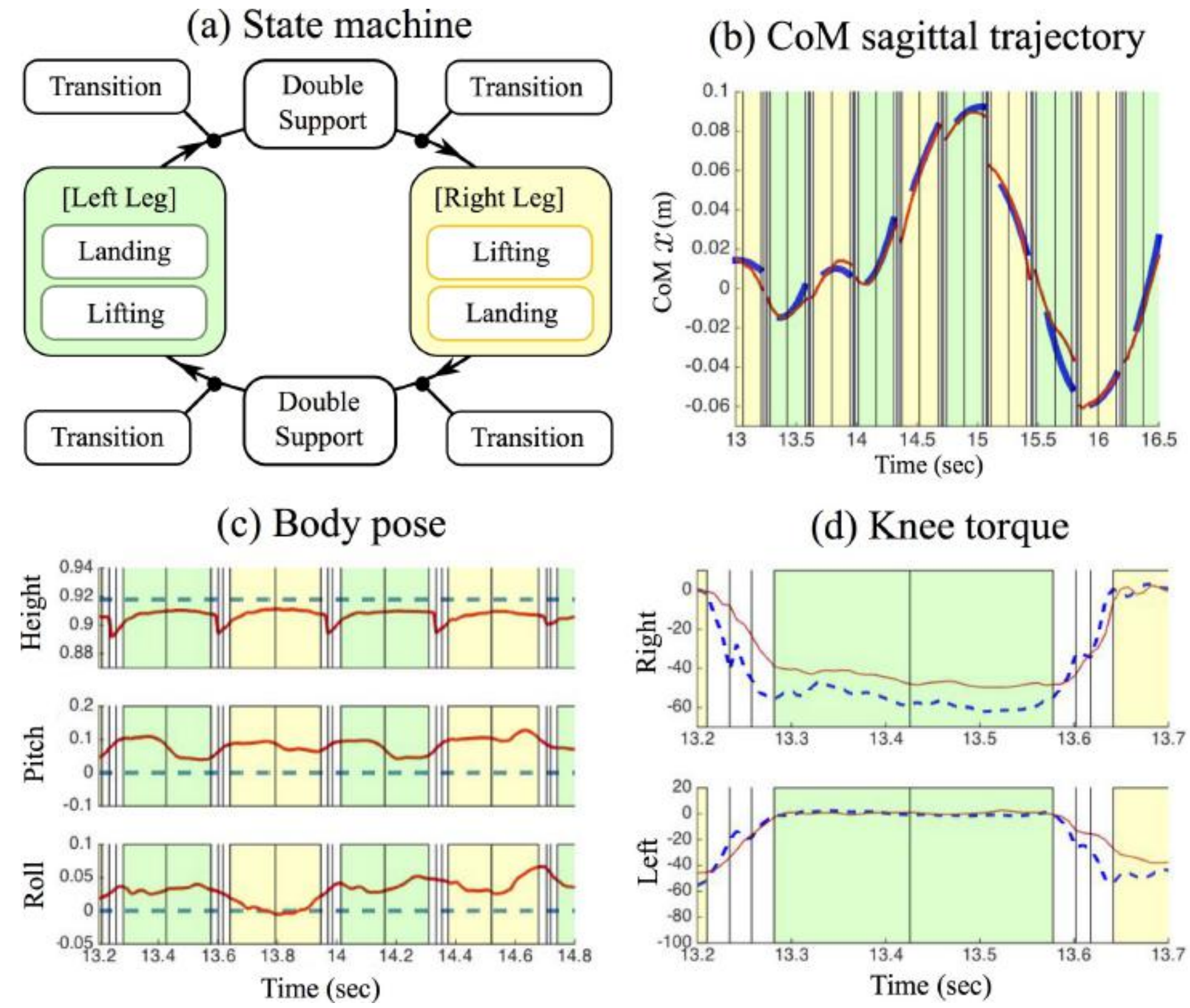

Figure 4.11: State machine and data of dynamic balancing experiment. (a) Motion within the balancing experiment is divided into three states: dual support, swing leg lift, and swing leg landing. Additionally, a transition state exists at the beginning of the lifting phase and the end of the landing phase to avoid the jerk caused by a sudden change in controller constraints. (b) and (c) show the $x$ and $y$-directional trajectory of the CoM. The blue line represents actual data while the red lines indicate the trajectories estimated by the planner. (d) Position and orientation task tracking are plotted over a representative portion of the stepping experiment, with sensor data in red and desired values, dotted, in blue. Height refers to CoM height.

To initiate the experiment, Hume is slightly supported while lifting its body to the desired height for balancing. Once this desired height is reached, the experimenter balances the robot carefully and lets it go as it takes its first step. Once free, the robot balances itself by continuously taking steps. There is a harness rope, slack when the robot is at its starting height, which catches it if it falls to prevent major damage. The power and Ethernet tether hang slack from another rope.

The motion sequence follows a state machine as shown in Fig. 4.11(a). Since the states are symmetric with respect to the supporting leg being either right or left, states are categorized in two different compound tasks with left and right single support having symmetric structures. The WBOSC compound task, $\boldsymbol{x}_{\text{task}}$, differs between dual support and single support phases of the stepping state machine. In dual support, the compound task coordinates are $[z, \phi, \theta, \psi]^T$, where $z$ represents the height of the center of mass. $\phi$, $\theta$ and $\psi$ are body yaw, pitch and roll angles, respectively. Those coordinates are controlled via the acceleration input of WBOSC, $\boldsymbol{u}_{\text{task}}$, shown in (Eq. (44), Kim et al. (2016a)) and via PD or PID control laws. In single support, the compound task is $[z, \theta, \psi, x_{\text{foot}}, y_{\text{foot}}, z_{\text{foot}}]^T$. The desired height is set to the initial height when Hume begins to step and the body orientation is set to be straight up. The desired foot trajectory for the lifting phase consists on first reaching a predefined height while keeping its sagittal and lateral position constant. Then, a $3^{\text{rd}}$ order polynomial is used to generate the desired landing trajectory. Since Hume has point-feet, it is not possible to control the yaw motion $\phi$ in single support phase. To compensate for this deficiency, we use the short duration that the robot spends in dual support to correct for it. In the balancing experiment, we do not use internal force feedback control to reduce the complexity of the sequencing process. Since the standing surfaces are flat, internal force regulation is not needed to keep the contacts stable. All control parameters for the single and dual support phases, and for the stance and swing legs are shown in (Table I, Kim et al. (2016a)). The parameters of the planner used in this experiment are shown in (Table II, Kim et al. (2016a)).

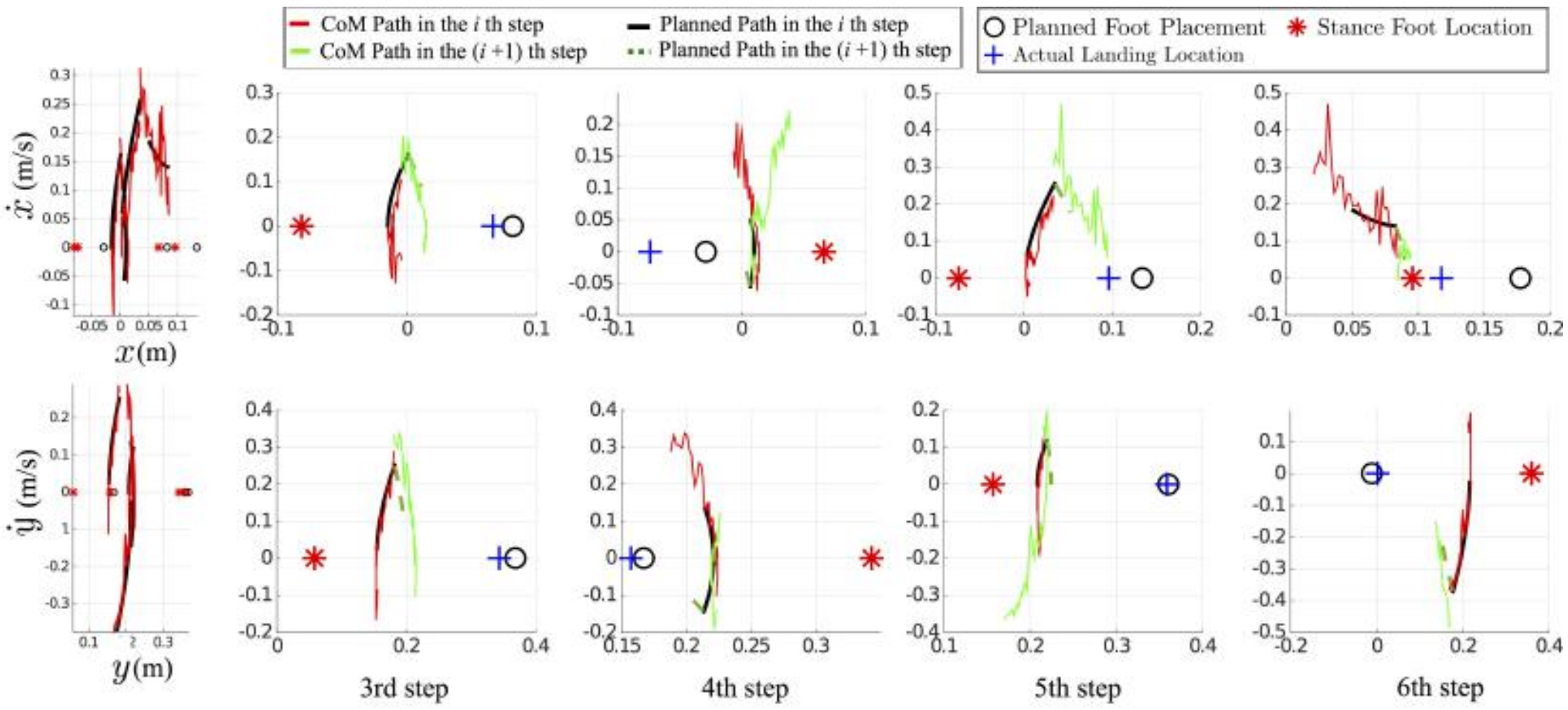

Figure 4.12: Phase-space data of the unsupported dynamic balancing experiment. The phase paths of steps 3-6 are expanded into individual step planning plots.

Phase-space trajectories are shown in Fig. 4.11. In (b) and (c), the sagittal and lateral CoM trajectories (red color) are superimposed on the one-step predicted path by our planner (blue color). Although predicting CoM path for multiple steps is difficult, computing the CoM path for a single step using the PIPM dynamics closely approximates the actual CoM motion. The orientation error is bounded by 0.05 rad (Fig. 4.11(d)). Given the model disturbances and the impacts, this error is reasonable small to validate the controller's performance. Fig. 4.12 shows snapshots and phase paths for this experiment. For each step, a red line marks the actual CoM path up to the switching state, and a green line continues the trajectory after the switch. The robot's initial stance foot in the step planning plot is denoted with a $\ast$, the planned second footstep with a black circle, and the achieved second stance foot location with a blue cross. Therefore, a green line in the $i^{\text{th}}$ step plot is the same path as the red line in the $(i+1)^{\text{th}}$ step plot. A black line and a dark green dotted line are, respectively, the PIPM's predicted paths before and after the switching state. The phase-space data is corrupted by high-frequency noise from the IMU sensor signal and the joint encoders that are combined to compute the CoM velocity.

Transition smoothness and gain scheduling techniques are indispensable to achieve stable step transitions. In Fig. 4.11(d), the commanded torque (blue color) varies from 0 to -60 N/m without significant jerk due to our contact transition technique. When the right leg switches to a stance leg (green background), around 20 N/m of torque tracking error appears. This is expected because we detune the low-level torque controller to achieve stiffer position control by the WBOSC controller. The controller corrects the yaw error during the dual support phase. Additionally, Hume incurs a significant bending of the stance leg which results in the uncertainty of the CoM position with respect to the stance foot.

In spite of all these sources of error, the robot was able to dynamically balance unsupported for 18 steps using its point contacts.

## 4.6 Discussions and Conclusions

This chapter is centered around the goals of (i) providing metrics of robustness to quantify the phase-space deviations, (ii) classifying various disturbance pattern, guard and recovery strategies in phase-space, (iii) design robust hybrid controllers for the locomotion behaviors in rough terrain, which provides fast computation guarantee at runtime, and (iv) demonstrating the ability of our method to deal with large external disturbances.

The characterized disturbances are assumed to be impulses that change the CoM velocity instantaneously. In fact, the disturbance can be of various types: (i) instantaneous changes to the CoM behavior; (ii) continuous perturbations (Englsberger et al., 2015b); and (iii) friction-like drag forces. As for continuous force disturbances, the method proposed in (Hyon et al., 2007) can be used to estimate the effect of undesired external forces. In any case, the proposed disturbance characteristics and recovery strategies are still valid.

Our planner is based on a simplified locomotion model, which ignores swing leg dynamics. However, this type of dynamics can significantly affect the actual motion tracking performance. In the future, we will explore more sophisticated models that include this type of dynamics. In the dynamic programming approach of Eq. (4.5), we only constrain the pitch torque while the pitch angle does not have limits. The focus of the manuscript so far has been on the generation of the trajectories and on outlining a robust control approach. However, for real implementation users need to incorporate the dynamics of the flywheel to constraint the torso pitch's range of motion.

Although the robust optimal controller is designed for sagittal dynamics, similar controllers can be formulated for the lateral and vertical CoM behaviors, given the PIPM dynamics of Eq. (3.14). Correspondingly, the yaw and roll angular torques can be incorporated into the optimal control inputs.

# Chapter 5

# High-level Reactive Task Planner Synthesis

In this chapter, we focus on synthesizing a high-level reactive planner for the whole-body locomotion (WBL) over cluttered dynamic environments. To accomplish WBL behaviors, we will compose a sequence of locomotion modes with planned keyframes. This mode composition can be achieved by synthesizing a high-level planner protocol which makes proper contact decisions like the ones shown in Fig. 5.2, and determines the switching strategy of the low-level planner. This hierarchy motivates us to use temporal-logic-based formal methods for task specifications and synthesizing the switching strategy such that the switched system satisfies, by construction, all the specifications.

## 5.1 Whole-Body Locomotion Dynamics

For a general locomotion process, the evolution of a family of its dynamics modes is represented as

$$\dot{\boldsymbol{\xi}} = \boldsymbol{\mathcal{F}}_p(\boldsymbol{\xi}, \boldsymbol{u}), \quad p \in \mathcal{P} \tag{5.1}$$

---

This chapter incorporates the results from the following publications: (Zhao, Topcu, and Sentis, 2016f,g).

of which Eq. (3.14) is a specific case. $\boldsymbol{\xi}(\zeta) \in \mathbb{R}^n$ denotes the system state vector. $\boldsymbol{u} \in \mathbb{R}^m$ denotes the control input; The switching signal $p$ indexes a specific planner mode. $\boldsymbol{\mathcal{F}}_p$ denotes a vector field associated with mode $p$. The overall system can be expressed as a switched system (Liberzon, 2012), $\dot{\boldsymbol{\xi}} = \boldsymbol{\mathcal{F}}_\sigma(\boldsymbol{\xi}, \boldsymbol{u})$, where $\sigma$ is the continuous switching signal taking values in a finite mode set $\mathcal{P}$.

Given this general model, certain assumptions are commonly imposed to make its dynamics more tractable (Audren et al., 2014), (Kajita et al., 2003b). In our case, four specific planner modes, under different mild assumptions, are proposed according to specific locomotion contact configurations.

**Mode (a): Prismatic Inverted Pendulum Model.** For this single foot contact, we have derived the PIPM dynamics of Eq. (3.14) in Section 3.3.1. For the results in this Chapter, the unique simplification we make is to assume a lateral-invariant piece-wise linear CoM path surface, i.e., $\psi_{\text{CoM}_q}(x, y, z) = z - ax - c = 0$. The sagittal and lateral dynamics of Eq. (3.14) can be easily derived.

**Mode (b): Pendulum Model**. When the robot grasps the loop on the ceiling wall to swing over an unsafe region, the system dynamics behave as a pendulum model (PM). For this single arm contact, we have

$$\begin{pmatrix} \ddot{x} \\ \ddot{y} \end{pmatrix} = -\omega_{\text{PM}}^2 \begin{pmatrix} (x - x_{\text{arm}}) - \frac{\tau_y}{mg} \\ (y - y_{\text{arm}}) - \frac{\tau_x}{mg} \end{pmatrix}, \tag{5.2}$$

where PM phase-space asymptotic slope is defined as

$$\omega_{\mathrm{PM}} = \sqrt{\frac{g}{z_{\mathrm{PM}}^{\mathrm{apex}}}},\ z_{\mathrm{PM}}^{\mathrm{apex}} = (z_{\mathrm{arm}} - a \cdot x_{\mathrm{arm}} - b),$$

with the same parameters and CoM path surface in Mode (a). An essential difference between Modes (a) and (b) lies in that PM dynamics are stable since the CoM is always enforced to move towards the apex position while the PIPM dynamics are unstable.

**Mode (c): Stop-Launch Model.** When a human appears, the robot has to come to a stop and wait until human disappears. The task in this mode is solely to decelerate the CoM motion to zero and accelerate from zero again, built upon its multi-contact configuration. We name this model as a stop-launch model (SLM).

$$\dot{l}_x = m\ddot{x},\ \dot{l}_y = m\ddot{y},\ \dot{l}_z = m\ddot{z} \tag{5.3}$$

where the control input $\boldsymbol{u} = (\ddot{x}, \ddot{y}, \ddot{z})^T$ are CoM accelerations. For simplicity, constant accelerations are used and the phase-space CoM motion is a parabolic manifold.

**Mode (d): Modified Multi-Contact Model.** When the robot maneuvers through constrained rough terrains, arms and legs in contact can accelerate and decelerate the CoM according to terrain height variations. In this case, a new multi-contact model is proposed built on centroidal momentum under mild assumptions, named as modified multi-contact model (MMCM). To make the dynamics tractable, we assume a known constant vertical acceleration $\ddot{z}$ in one step and neglection of the angular momentum $h_z$ around $z$-axis (Audren et al., 2014). Via these assumptions, we have a constant resultant vertical external force, i.e., $\sum_i^{N_c} f_{iz} = m(\ddot{z} - g)$, where $N_c$ is the number of limb contacts. More details can be referred to (Zhao et al., 2016f).

In the light of these four modes, the switching mode set is defined as

**Definition 5.1** (**Switching modes**). *Phase-space planner switching mode $\mathcal{P}$ is composed of*

$$\mathcal{P} \coloneqq \{\mathrm{PIPM}, \mathrm{PM}, \mathrm{SLM}, \mathrm{MMCM}\} \tag{5.4}$$

*where each component corresponds to one aforementioned mode.*

Given the planner mode commanded from the discrete contact planner, the hybrid phase-space planner generates the continuous CoM trajectory based on this designated mode and compute the mode switching instant. A logic-based planner structure is shown in Fig. 5.1. In next section, we will discuss more details about how to design the mode switching by formal methods.

This study especially concentrates on the multi-contact mobility of a humanoid legged robot in a constrained environment as shown in Fig. 1.5. In particular, besides commonly-used legs during dynamic walking, an additional pair of arms are incorporated to achieve a considerably large range of accelerated or decelerated CoM motions. More than that, certain extreme cases, such as when stair cracks, require the arm to maintain contacts so as to smoothly traverse unsafe areas.

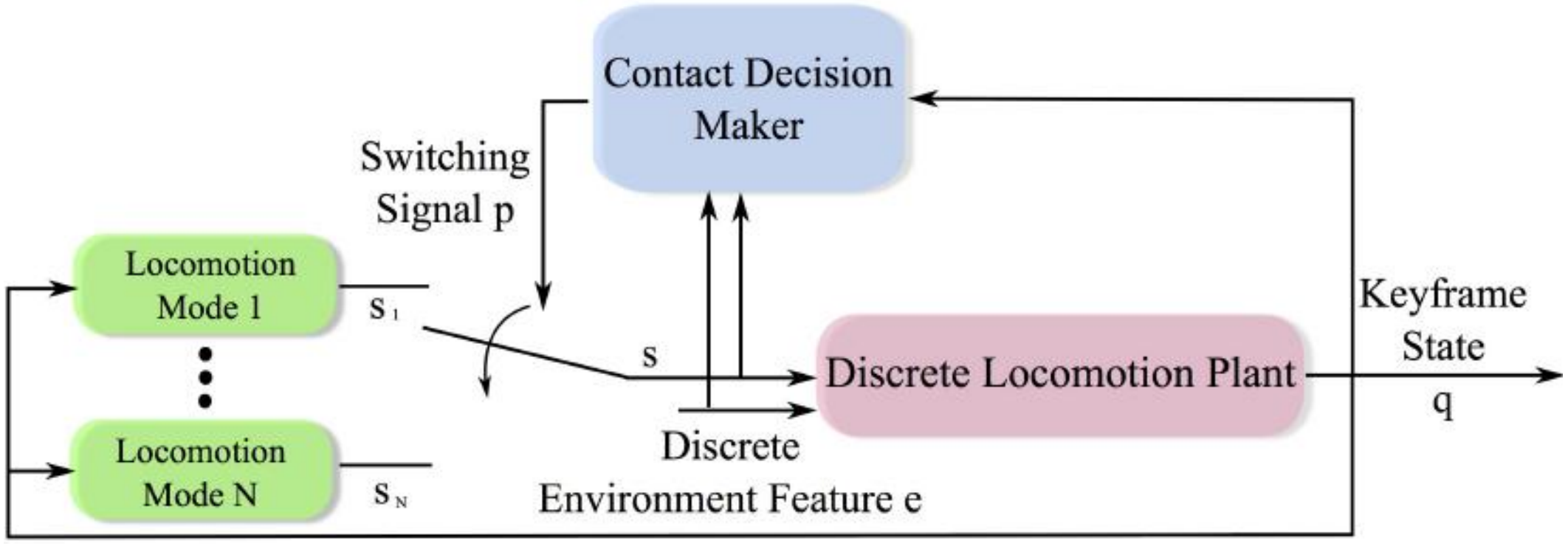

Figure 5.1: Logic-based planner structure. Each mode, indexed by a switching signal $p$, corresponds to a planner model. Four modes in total are modeled for our constrained environment maneuver. Uncontrollable environment state $e$ is denoted by next stair height while controllable system action $u$ is represented by the limb contact configuration. Discretized phase-space state $q = (p_{\text{contact}}, \dot{x}_{\text{apex}})$.

## 5.2 Linear Temporal Logic Background

We now define an open finite transition system and its execution, describe temporal logic preliminaries.

**Definition 5.2** (**Open finite transition system**). *An open finite transition system (OFTS) $\mathcal{TS}$ is a tuple,*

$$\mathcal{TS} := (\mathcal{Q}, \mathcal{P}, \mathcal{E}, \mathcal{S}, \mathcal{T}, \mathcal{I}, AP, \mathcal{L}), \tag{5.5}$$

*where $\mathcal{Q}$ is a finite set of states, $\mathcal{P}$ is a set of system modes defined in Eq. (5.1), $\mathcal{E}$ is a finite set of uncontrollable environmental actions, $\mathcal{S}$ is a finite set of controllable robot contact actions, $\mathcal{T} \subseteq (\mathcal{Q} \times \mathcal{P}) \times \mathcal{E} \times \mathcal{S} \times (\mathcal{Q} \times \mathcal{P})$ is a transition, $\mathcal{I} = \mathcal{Q}_0 \times \mathcal{P}_0 \subseteq \mathcal{Q} \times \mathcal{P}$ is a set of initial states, $AP$ is a set of atomic propositions, $\mathcal{L} : (\mathcal{Q} \times \mathcal{P}) \to 2^{AP}$ is a labeling function mapping the state to an atomic proposition. $\mathcal{TS}$ is finite if $\mathcal{Q}, \mathcal{P}, \mathcal{E}, \mathcal{S}$ and $AP$ are finite.*

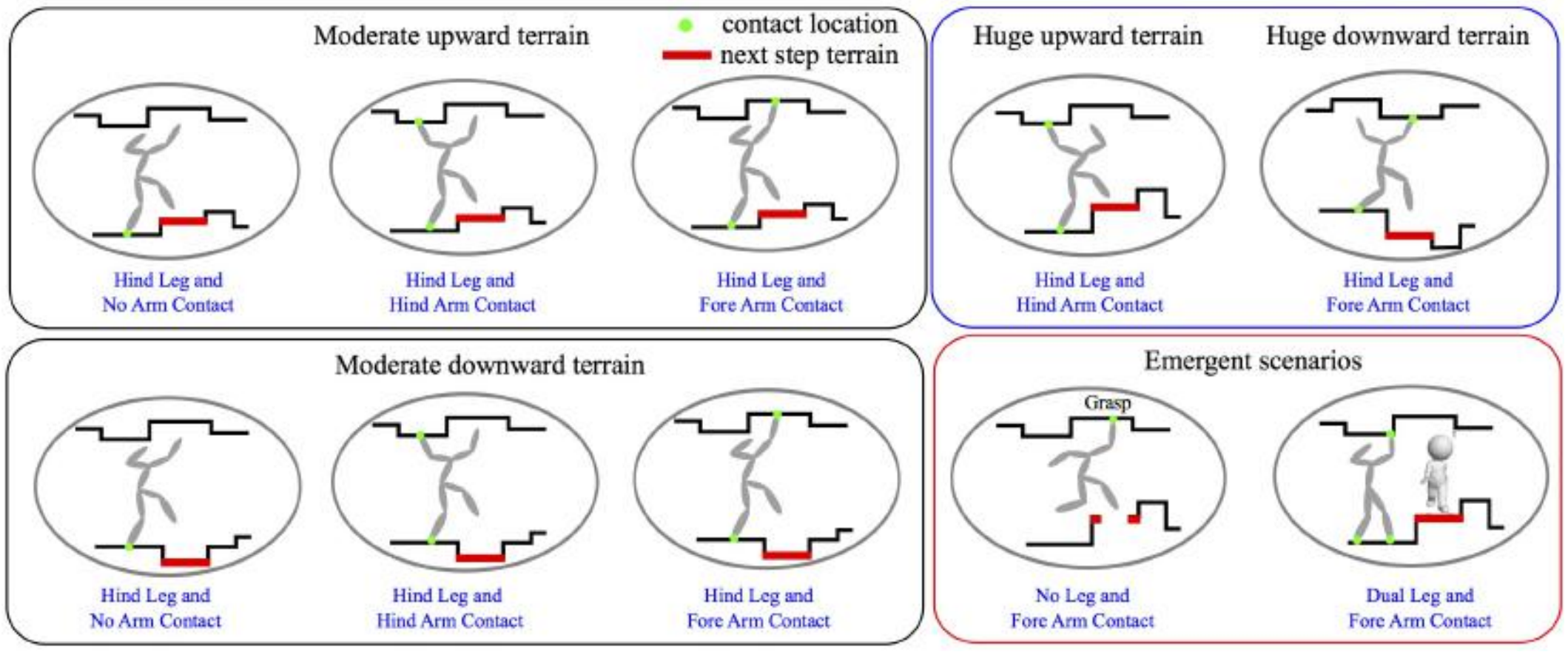

Figure 5.2: Various contact planning strategies according to rough terrains with different discretized heights. Various contact configurations, i.e., system actions, are specifically designed. Two emergent scenarios are incorporated.

The $\mathcal{TS}$ defined above is "open" because it has uncontrollable actions $\mathcal{E}$. Without loss of generality, it is assumed that for every pair $(q, p) \in \mathcal{Q} \times \mathcal{P}$, there exists at least one pair $(q', p')$ such that $(q, p) \xrightarrow{\mathcal{T}} (q', p')$. The OFTS considered in this study has non-deterministic transitions.

**Definition 5.3** (**Execution and word of an OFTS**). *An execution $\boldsymbol{\gamma}$ of an OFTS $\mathcal{TS}$ is an infinite path sequence $\boldsymbol{\gamma} = (q_0, p_0, e_0, s_0)(q_1, p_1, e_1, s_1)(q_2, p_2, e_2, s_2)\ldots$, with $\gamma_i = (q_i, p_i, e_i, s_i) \in \mathcal{Q} \times \mathcal{P} \times \mathcal{E} \times \mathcal{S}$ and $\gamma_i \xrightarrow{\mathcal{T}} \gamma_{i+1}$. The word generated from $\boldsymbol{\gamma}$ is $w_\gamma = w_\gamma(0)w_\gamma(1)w_\gamma(2)\ldots$, with $w_\gamma(i) = \mathcal{L}(\gamma_i)$, $\forall i \geq 0$.*

The word $w_\gamma$ is said to satisfy a LTL formula $\varphi$, if and only if the execution $\boldsymbol{\gamma}$ satisfies $\varphi$. If all executions of $\mathcal{TS}$ satisfy $\varphi$, we say that $\mathcal{TS}$ satisfies $\varphi$, i.e., $\mathcal{TS} \models \varphi$.

For LTL syntax and semantics, please refer to the definitions in Appendix E and the references therein.

## 5.3 Planner Synthesis Formulation

Given the preliminaries above, we formulate a discrete planner synthesis problem and we introduce a specific fragment of the temporal logic which will be used in the task specifications.

**Definition 5.4** (**Discrete contact planner switching synthesis**). *Given a transition system $\mathcal{TS}$ and a LTL specification $\varphi$ following the assume-guarantee form,*

$$\varphi := \big((\varphi_q \wedge \varphi_e) \Rightarrow \varphi_s\big), \tag{5.6}$$

*where $\varphi_q$ represents a liveness assumption incorporating transience property*[2]*, $\varphi_e$ and $\varphi_s$ are the propositions for admissible environment and system actions, respectively, we synthesize a contact planner switching strategy $\gamma$ that generates only correct executions $(q, p, e, s)$, i.e., $(q, p, e, s) \models \varphi$.*

**Remark 5.1.** *For the discrete synthesis problem, we use a particular type of abstraction named as over-approximation (Liu et al., 2013) (or similarly $\epsilon$-inflation (Fainekos et al., 2009)) to approximate the underlying continuous dynamics. A detailed definition is provided in (Liu et al., 2013). Intuitively, a finite transition system $\mathcal{TS}$ is named as an over-approximation of the continuous system if each transition $q \xrightarrow{\mathcal{T}} q'$ is included in $\mathcal{TS}$, as long as the*

[2]Roughly speaking, a set is transient if and only if all the trajectories starting from a certain state within this set, will leave this set in a finite time. Namely, a transient set does not comprise any positively invariant sets. For more details, please refer to (Batt et al., 2008), (Liu et al., 2012).

*strategy is feasible to be continuously implemented, regardless of either a coarse partition or external disturbances. The discrepancy between the conservative over-approximation and the underlying continuous system is accounted as non-determinism and treated as adversary.*

To make the computation tractable, we employ a class of LTL formulae with favorable polynomial complexity, named as the Generalized Reactivity (1) (GR (1)) formulae (Bloem et al., 2012). This class of formulae is expressed as, for $v \in \{e, s\}$,

$$\varphi_v = \varphi^v_{\text{init}} \bigwedge_{i \in I_{\text{safety}}} \Box \varphi^v_{\text{trans},i} \bigwedge_{i \in I_{\text{goal}}} \Box \Diamond \varphi^v_{\text{goal},i}, \tag{5.7}$$

where $\varphi^v_{\text{init}}$ are the propositional formulae defining initial conditions. $\varphi^v_{\text{trans},i}$ refer to the transitional propositional formulae (i.e., safety conditions) incorporating the state at next step. $\varphi^v_{\text{goal},i}$ are the propositional formulae describing the goals to be reached infinitely often (i.e., liveness conditions).

**Remark 5.2.** *As to computational complexity, restricting to the favorable GR (1) formulae provides an advantage of translating LTL formulae to the corresponding automaton in polynomial $O(N^3)$ time, where $N$ is the number of state space.*

### 5.3.1 Specifications for Environment and System Actions

This subsection defines an environment action set $\mathcal{E}$ and a system action set $\mathcal{S}$. Then we propose the corresponding specifications for these actions.

**Definition 5.5** (**Environment actions**). *The environment action is denoted by the set of the variation levels in the stair height*

$$\mathcal{E} := \{e_{md}, e_{hd}, e_{mu}, e_{hu}, e_{sc}, e_{ha}\}, \tag{5.8}$$

*where the first four actions denote different compositions of* downward *and* upward *stairs with* moderate *and* huge *height variations, respectively.*

For instance, $e_{md}$ denotes moderateDownward. The last two actions $e_{sc}$ and $e_{ha}$ represent unexpected events: stairCrack and humanAppear.

**Definition 5.6** (**System actions**). *Given these environment actions, the system actions of the robot are denoted by a contact configuration set*

$$\mathcal{S} := \{a_{li\text{-}aj},\ \forall (i,j) \in \mathcal{A}_{\text{index}}\}, \tag{5.9}$$

*where the indices 'l' and 'a' are short for* leg *and* arm*, respectively.* $(i,j)$ *corresponds to the contact limb.* $\mathcal{A}_{\text{index}} = \{(h,n),(h,h),(h,f),(n,f),(d,h),(d,f)\}$*, where the letters 'h', 'f', 'd' and 'n' represent* hind, fore, dual *and* no *contacts, respectively.*

For instance, $a_{lh\text{-}af}$ denotes the legHindArmFore contact configuration.

We now describe environment and system action specifications. The environment actions are assumed not to occur at the initial time, and therefore $\varphi^e_{\text{init}} = \neg e_{sc} \wedge \neg e_{ha}$. Accordingly, the robot can initially not take the actions designated for these unexpected events, i.e., $\varphi^s_{\text{init}} = \neg a_{ln\text{-}af} \wedge \neg (a_{ld\text{-}ah} \vee a_{ld\text{-}af})$. Transitional and goal specifications are defined next

- (S0) None of the huge terrain height variations (i.e., hugeDownward or hugeUpward), humanAppear and stairCrack occur in two consecutive steps,

$$\square(e_{hd} \Rightarrow \bigcirc\neg e_{hd}) \bigwedge \square(e_{hu} \Rightarrow \bigcirc\neg e_{hu})$$
$$\bigwedge \square(e_{ha} \Rightarrow \bigcirc\neg e_{ha}) \bigwedge \square(e_{sc} \Rightarrow \bigcirc\neg e_{sc}).$$

- (S1) In normal scenarios, the mappings between environment and system actions, as illustrated in Fig. 5.2, are specified by

$$\square(e_{md} \Rightarrow (a_{lh\text{-}an} \vee a_{lh\text{-}ah} \vee a_{lh\text{-}af})) \bigwedge \square\big(e_{mu} \Rightarrow (a_{lh\text{-}an} \vee a_{lh\text{-}ah} \vee a_{lh\text{-}af})\big)$$
$$\bigwedge \square(e_{hu} \Rightarrow a_{lh\text{-}ah}) \bigwedge \square(e_{hd} \Rightarrow a_{lh\text{-}af}),$$

  where moderate terrain variations allow more non-deterministic contact actions than those allowed by huge terrain variations. For instance, if $e = e_{hu}$, i.e., hugeUpward, the robot has only one action option, which makes its hind arm in contact such that its CoM gains a larger acceleration to step up as shown in Fig. 5.2.

- (S2) If a stair crack occurs, the robot will grab a handle on the ceiling wall (i.e., $a_{ln\text{-}af}$). On the contrary, when the stair is not cracked, we obviously have $a \neq a_{ln\text{-}af}$:

$$\square(e_{sc} \Rightarrow a_{ln\text{-}af}) \bigwedge \square(\neg e_{sc} \Rightarrow \neg a_{ln\text{-}af}).$$

- (S3) If a human appears in front of the robot, the robot comes to a stop with the action $a_{ld\text{-}af}$. Contrarily, when the human disappears, the robot starts to walk from where it stops:

$$\square\big(e_{ha} \Rightarrow (a_{ld\text{-}ah} \vee a_{ld\text{-}af})\big) \bigwedge \square\big(\neg e_{ha} \Rightarrow \neg(a_{ld\text{-}ah} \vee a_{ld\text{-}af})\big).$$

- (S4) Neither unexpected events (i.e., humanAppear and stairCrack) occur infinitely often:

$$\square\lozenge\neg e_{ha} \bigwedge \square\lozenge\neg e_{sc}.$$

Among these specifications, we have (S0) representing $\varphi^e_{\rm trans}$, (S1)-(S3) representing $\varphi^s_{\rm trans}$ and (S4) representing $\varphi^e_{\rm goal}$, where $\varphi^e_{\rm trans}, \varphi^s_{\rm trans}$ and $\varphi^e_{\rm goal}$ are defined in Eq. (5.7). Note that, the contact planner requires at least one limb in contact. This requirement is automatically encoded into the system contact actions thus we do not need to explicitly assign specifications for this.

### 5.3.2 Specifications for Keyframe States

Our discrete phase-space planner has a keyframe state vector $q = \{p_{\rm contact}, \dot{x}_{\rm apex}\}$. Since we always focus on the keyframe state for next walking step, the discretization of a particularly selected phase-space region is shown in Fig. 5.3.

**Definition 5.7 (Phase-space keyframe).** *The keyframe is represented by the following discrete state,*

$$\mathcal{Q} := \{q_{i\text{-}j},\ \forall(i,j) \in \mathcal{Q}_{\rm index}\} \bigcup \{q_{\rm swing}, q_{\rm stop}\}, \tag{5.10}$$

*where the apex velocity index $i$ and the step length index $j$ are assigned to three different "levels of degree" (LOD), i.e., $s$ (Small), $n$ (Normal) and $l$ (Large), respectively.*

For instance, $q_{s\text{-}l}$ represents velSmallstepLarge, a keyframe state with a small apex velocity and a large step length. The index set $\mathcal{Q}_{\rm index}$ comprises 9 ele-ments in total. Additionally, two more keyframe states $q_{\text{swing}}$ (velSwingstepSwing) and $q_{\text{stop}}$ (velStopstepStop) are designed for two unexpected events, respectively.

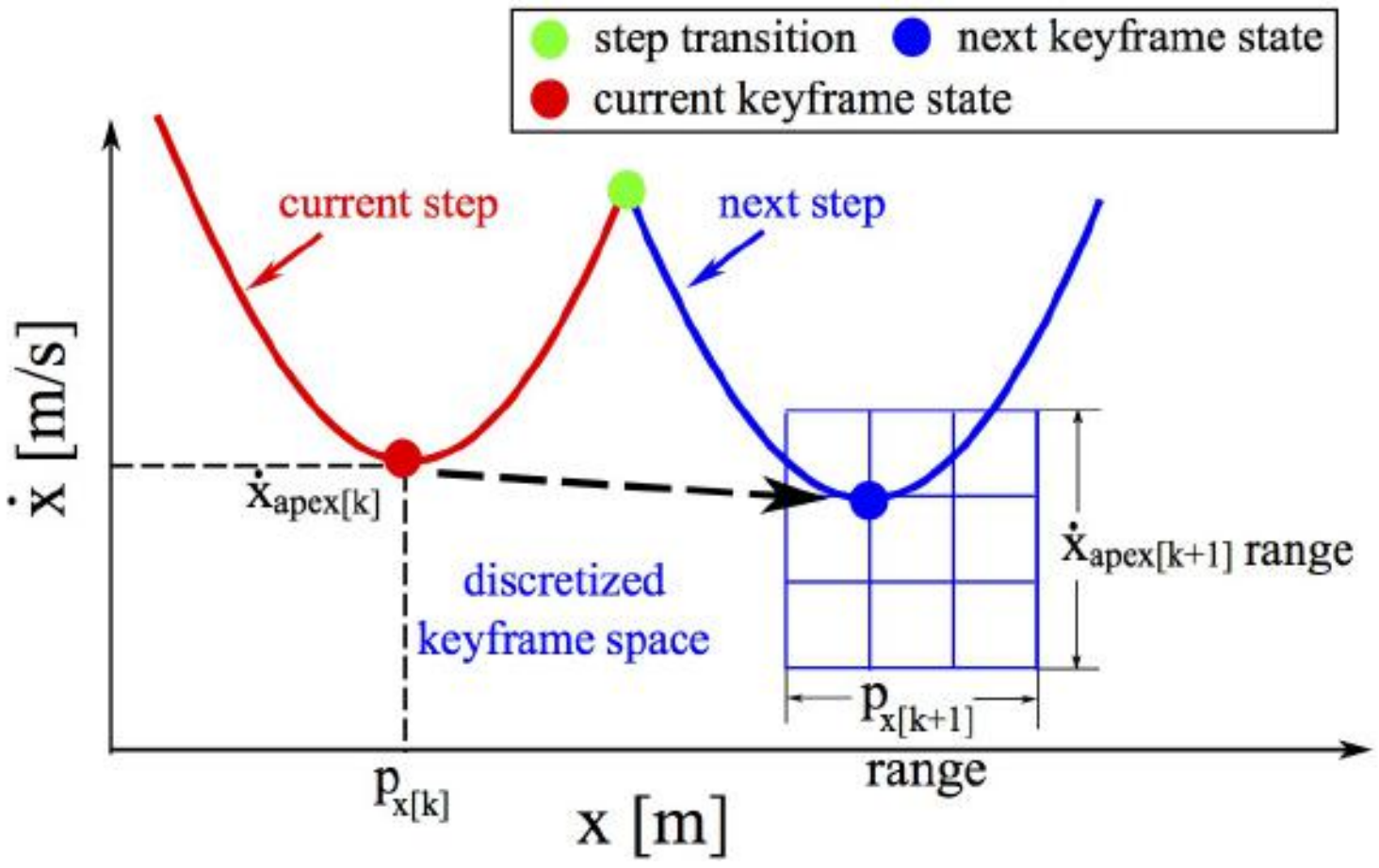

Figure 5.3: Keyframe discretization for next walking step. We choose a phase-space region based on robot kinematics and discrete this region into a grid. The keyframe state is determined according to environment actions.

**Remark 5.3.** *While we use a fixed discretization over the range of CoM apex velocity and step length, finer or varying discretization may enhance the flexibility of the method at the expense of an increase of the computational cost.*

Given environment actions, the propositions for the keyframe states are designed as follows.

- (S5) If $\bigcirc e = \bigcirc e_{md}$ (i.e., moderateDownward), the LOD for the next keyframe state $\bigcirc q$ remains constant or increases by one either from step

length or apex velocity:

$$\Box\big((q_{s\text{-}s} \wedge \bigcirc e_{md}) \Rightarrow \bigcirc(q_{s\text{-}s} \vee q_{s\text{-}n} \vee q_{n\text{-}s})\big) \bigwedge \Box\big((q_{s\text{-}n} \wedge \bigcirc e_{md}) \Rightarrow \bigcirc(q_{s\text{-}n} \vee q_{s\text{-}l} \vee q_{n\text{-}n})\big)$$

$$\ldots$$

$$\bigwedge \Box\big((q_{l\text{-}n} \wedge \bigcirc e_{md}) \Rightarrow \bigcirc(q_{l\text{-}n} \vee q_{l\text{-}l})\big) \bigwedge \Box\big((q_{n\text{-}l} \wedge \bigcirc e_{md})$$

$$\Rightarrow \bigcirc(q_{n\text{-}l} \vee q_{l\text{-}l})\big) \bigwedge \Box\big((q_{l\text{-}l} \wedge \bigcirc e_{md}) \Rightarrow \bigcirc q_{l\text{-}l}\big)$$

$$\bigwedge \Box\Big(\big((q_{\text{swing}} \vee q_{\text{stop}}) \wedge \bigcirc e_{md}\big) \Rightarrow \bigcirc(q_{s\text{-}n} \vee q_{n\text{-}n} \vee q_{l\text{-}n})\Big),$$

where, if $q = q_{s\text{-}s}$, $\bigcirc q$ can choose $q_{s\text{-}s}$ (remaining constant), $q_{s\text{-}n}$ (step length increases one degree) or $q_{n\text{-}s}$ (apex velocity increases one degree). All the other keyframes in normal scenarios follow the same pattern. There are three special cases: (i) when $q = q_{l\text{-}n}$, there are only two choices for $\bigcirc q$, i.e., $q_{l\text{-}n}$ and $q_{l\text{-}l}$; (ii) the same situation applies to $q_{n\text{-}l}$; (iii) when $q = q_{l\text{-}l}$, the only choice is $\bigcirc q = \bigcirc q_{l\text{-}l}$. In unexpected cases, we assign $\bigcirc q$ as one of $q_{s\text{-}n}$, $q_{n\text{-}n}$ and $q_{l\text{-}n}$.

- (S6) If $\bigcirc e = e_{hd}$ (i.e., hugeDownward), the level of degree for next keyframe state increases by one or two, either from step length or apex velocity. The only exception is when $q = q_{l\text{-}l}$, next step state $q$ can only maintain $q_{l\text{-}l}$.

$$\square\big((q_{s\text{-}s} \wedge \bigcirc e_{hd}) \Rightarrow \bigcirc(q_{n\text{-}s} \vee q_{s\text{-}n} \vee q_{l\text{-}s} \vee q_{s\text{-}l} \vee$$

$$q_{n\text{-}n})\big) \bigwedge \square\big((q_{s\text{-}n} \wedge \bigcirc e_{hd}) \Rightarrow \bigcirc(q_{s\text{-}l} \vee q_{n\text{-}n} \vee q_{n\text{-}l})\big)$$

$$\vdots$$

$$\bigwedge \square\Big(\big((q_{l\text{-}n} \vee q_{n\text{-}l} \vee q_{l\text{-}l}) \wedge \bigcirc e_{hd}\big) \Rightarrow \bigcirc q_{l\text{-}l}\Big) \bigwedge \square$$

$$\Big(\big((q_{\text{swing}} \vee q_{\text{stop}}) \wedge \bigcirc e_{hd}\big) \Rightarrow \bigcirc(q_{s\text{-}n} \vee q_{n\text{-}n} \vee q_{l\text{-}n})\Big)$$

where, if $q = q_{s\text{-}s}$, $\bigcirc q$ increases the level of degree by one, i.e., $q_{s\text{-}n}$ and $q_{n\text{-}s}$, or by two, i.e., $q_{n\text{-}n}, q_{l\text{-}s}$ and $q_{s\text{-}l}$. Special cases are $q_{l\text{-}n}, q_{n\text{-}l}$ and $q_{l\text{-}l}$ where $q = q_{l\text{-}l}$ is the only choice. Emergent cases follow the same principle as those in (S5).

- (S7) If $\bigcirc e = \bigcirc e_{sc}$ (i.e., stairCrack), then $\bigcirc q = \bigcirc q_{\text{swing}}$ regardless of the current $q$:

$$\square(\bigcirc e_{sc} \Rightarrow \bigcirc q_{\text{swing}}).$$

  Note that $q_{\text{swing}}$ is excluded from the current state $q$ since we specify that a stair crack can not appear two steps consecutively as shown in (S0).

- (S8) If $\bigcirc e = \bigcirc e_{ha}$ (i.e., humanAppear), then $\bigcirc q = \bigcirc q_{\text{stop}}$ regardless of the current $q$:

$$\square(\bigcirc e_{sc} \Rightarrow \bigcirc q_{\text{stop}}).$$

The remaining eight normal scenarios involving different environment and system action compositions are defined in a similar pattern and they are omitted here for brevity. The supplementary material illustrates one scenario when $\bigcirc e = \bigcirc e_{hd}$ (i.e., $\mathsf{hugeDownward}$). All the specifications in (S5)-(S7) represent $\varphi^{e}_{\text{trans}}$. The goal specification $\varphi^{s}_{\text{goal}}$ in our scenario is trivial, that is, repeatedly selecting contact configurations among $\mathcal{S}$. Now all the specifications have been proposed, and $\varphi = \big((\varphi_q \wedge \varphi_e) \Rightarrow \varphi_s\big)$ holds.

## 5.4 Game Based Reactive Planner Synthesis

In this section, we focus on reactive planner synthesis by formulating the high-level locomotion planning problem as a game between the robot and its adversarial constrained environment. Given task specifications introduced next, a reactive control protocol is synthesized such that the controlled legged robot behaviors satisfy all the designed specifications whatever admissible uncontrollable environment behaviors are.

**Definition 5.8** (**Game of the whole-body locomotion planner**). *A game structure for the legged and armed locomotion planning is a tuple*

$$\mathcal{G} := \langle \mathcal{V}, \mathcal{X}, \mathcal{Y}, \theta_e, \theta_s, \rho_e, \rho_s, \phi_{\text{win}} \rangle \tag{5.11}$$

*where*

- $\mathcal{V} = \mathcal{X} \times \mathcal{Y}$ *is a finite set of proposition state variables over finite domains in the game.*
- $\mathcal{X} := \mathcal{E} \times \mathcal{Q}$ *is a set of input variables, defined by the keyframe state and environment (i.e., player 1).*

- $\mathcal{Y} := \mathcal{S} \times \mathcal{P}$ *is a set of output variables, defined by the system contact configuration and switching signal (i.e., player 2).*
- $\theta_e$ *is an atomic proposition over* $\mathcal{X}$ *characterizing initial states of the environment.*
- $\theta_s$ *is an atomic proposition over* $\mathcal{Y}$ *characterizing initial states of the system.*
- $\rho_e(\mathcal{V}, \mathcal{X}')$ *is the transition relation for the environment, which is an atomic proposition that relates the current proposition state and allowable environment variables at next step (i.e.,* $\mathcal{X}'$*).*
- $\rho_s(\mathcal{V}, \mathcal{X}', \mathcal{Y}')$ *is the transition relation for the system, which is an atomic proposition that relates the current proposition state and an input value at next step (i.e.,* $\mathcal{X}'$*) to allowable output variable values at next step (i.e.,* $\mathcal{Y}'$*)*[3]*.*
- $\phi_{\text{win}}$ *is the winning condition given by an LTL formula over* $\mathcal{V}$*.*

**Remark 5.4.** *Different from the pioneering work (Liu et al., 2013), where only switching mode p is chosen as the output variable, our contact planner switching strategy additionally designates the contact action s as a component of the synthesis consequence. Non-deterministic contact actions exist within a specific switching mode.*

---

[3]Here, it is assumed that environment takes action first and then system acts. Thus next output variable value $\mathcal{Y}'$ can depend on next input variable value $\mathcal{X}'$.

A winning strategy of the switched system for the pair $(\mathcal{TS}, \varphi)$ is defined as a partial function $(\gamma_0\gamma_1 \cdots \gamma_{i-1}, (q_i, e_i)) \mapsto (s_i, p_i)$, where a switching mode and a contact configuration are chosen according to the state sequence history and the current keyframe state and environment action in order to satisfy Eq. (5.6). All the specifications are satisfied whatever admissible uncontrollable environment actions are.

**Proposition 5.1 (Existence of a winning locomotion strategy).** *A winning WBL strategy exists for the game $\mathcal{G}$ in Definition 5.8 if and only if $(\mathcal{TS}, \varphi)$ is realizable.*

The realizability of $\varphi$ can be judged from two viewpoints: (i) desired system propositions are satisfied (i.e., $\varphi_s$ is true) or (ii) environment propositions are violated (i.e., $\varphi_e$ is false). The latter case is trivial since the synthesized automaton is no longer valid and no correct path exists. Fig. 5.4 shows a fragment of the WBL contact planner. For illustration convenience, we index both the environment action $\mathcal{E}$ in Eq. (5.8) and the system action $\mathcal{S}$ in Eq. (5.9) as $\{0, \ldots, 5\}$ in order, respectively. The system keyframe state $\mathcal{Q}$ is indexed as $\{0, \ldots, 10\}$ in order. For instance, when the automaton state is at number 25, we have environment state $e = 0$ and keyframe $q = 4$. The winning strategy assigns system action $s = 2$ and system switching mode $p = 4$. Self-transition exists in moderateUpward states (e.g., state 16) and moderateDownward states (e.g., state 12 and 25) while hugeDownward states (e.g., state 15) do not have a self-transition according to proposition (S1). There is no transition between states 12 and 16 due to infeasible keyframe state transition. States 18 and 19 in red nodes represent humanAppear and stairCrack scenarios, respectively.

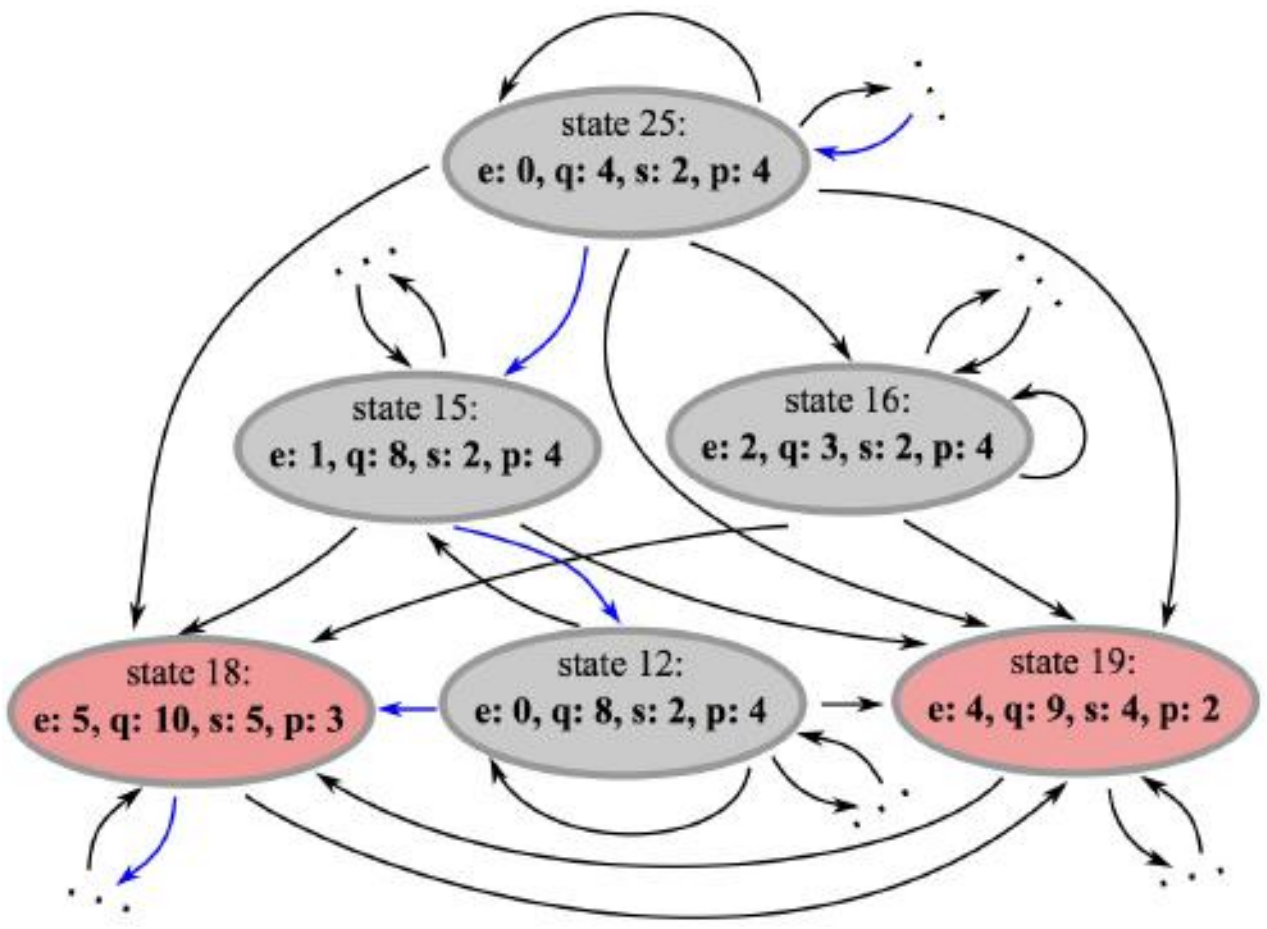

Figure 5.4: A fragment of the synthesized automaton for the WBL contact planner. Nondeterministic transitions are encoded in this automaton. The blue transitions represent a specific execution.

**Remark 5.5.** *The synthesized automaton incorporates nondeterministic transitions, caused by the following reasons: (i) Environment actions are nondeterministic. (ii) Given an environment action, several non-deterministic keyframe states can be chosen. (iii) Even when both an environment action and a keyframe state are given, non-deterministic system contact actions exist in certain transitions.*

In this section, we analyze the correctness of the low-level planner implementations in Section 5.1 for the discrete contact planner in Section 5.4. To begin with, we define the low-level trajectory as $\delta = (\boldsymbol{\xi}, \rho, \eta, \sigma)$. The high-level execution is defined as $\gamma = (q, e, s, p)$ in Definition 5.3.

**Definition 5.9 (Mapping from high-level execution to low-level trajectory).** *The low-level trajectory $\delta$ is a continuous implementation of the high-level execution $\gamma$, if there exists a sequence of non-overlapping phase intervals $\mathcal{H} = H_1 \cup H_2 \cup H_3 \ldots$ and $\cup_{i=1}^{\infty} H_i = \mathbb{R}^+$ such that $\forall \zeta \in H_k, \forall k \geq 1$, the following mappings hold*

$$\boldsymbol{\xi}(\zeta) \in T^{-1}(q_k), \rho(\zeta) = e_k, \eta(\zeta) = s_k, \sigma(\zeta) = p_k,$$

*where $T$ represents an abstraction mapping, which maps a certain continuous state $\boldsymbol{\xi}$ region into a discrete state $q$ (Liu et al., 2013).*

By this definition and stutter-equivalence (Baier et al., 2008), we can conclude $\delta \models \varphi$ if and only if $\gamma \models \varphi$. Additional assumptions need to be imposed in order to rule out Zeno behaviors, which are beyond the scope of this study. For detailed statements, reader can refer to (Liu et al., 2013) and the reference therein. Given these preliminaries, we can make the following statement:

**Theorem 5.1 (Correctness of the whole-body locomotion planner).** *Given an over-approximation model, a winning WBL strategy synthesized from the two-player game is guaranteed to be correctly implemented by the underlying low-level phase-space planner.*

*Proof.* From the previous proposition in Section 5.4, a winning WBL strategy synthesized from the WBL planner game solves the discrete locomotion planning problem. To study the correctness, we start by elaborating an over-approximated model of our locomotion planner. Initially, all the possible transitions in a given keyframe discretization (as shown in Fig. 5.3) are modeled in the finite transition system. According to environment actions and the LOD principle defined in proposition (S5) of Section 5.3.2, we remove unnecessary transitions and obtain a more accurate over-approximation model. Given this model, a winning switching strategy, represented as an automaton $\mathcal{A}_{\text{WBL}}$, is synthesized by solving the two-player game. According to $\mathcal{A}_{\text{WBL}}$, a sequence of system actions $s$ and switching modes $p$ are derived from a certain sequence of environment actions $e$ and discretized keyframe states $q$. To verify the correct implementation of a certain execution $\gamma$, we use the switching strategy semantics: given an initial state $\boldsymbol{\xi}(0)$ and an initial environment action $\rho(0) = e_0$, we assign $\eta(0) = s_0$ and $\sigma(0) = p_0$ according to $\mathcal{A}_{\text{WBL}}$. Then the continuous dynamics $\boldsymbol{\xi}(\zeta)$ evolve by following a specific mode $\dot{\boldsymbol{\xi}} = \boldsymbol{\mathcal{F}}_{p_0}(\boldsymbol{\xi}, \boldsymbol{u})$. Once a new environment action $e'$ is detected and this action satisfies all the environment assumptions, the switching mode $p$ is updated immediately based on $\mathcal{A}_{\text{WBL}}$. Given this new switching mode, the same procedure is repeated as above. Therefore, it is suggested that the low-level trajectory correctly implements one discrete execution of the synthesized automaton. □

**Remark 5.6.** *The preceding provable correctness is in view of the prerequisite of a realizable winning switching strategy. On the other hand, what if this synthesis problem is unrealizable? In this case, a refinement of the finite transition system can potentially result in a realizable switching strategy. This refinement is normally achieved by proposition preserving partition, such as splitting certain cells (Alur et al., 2015).*

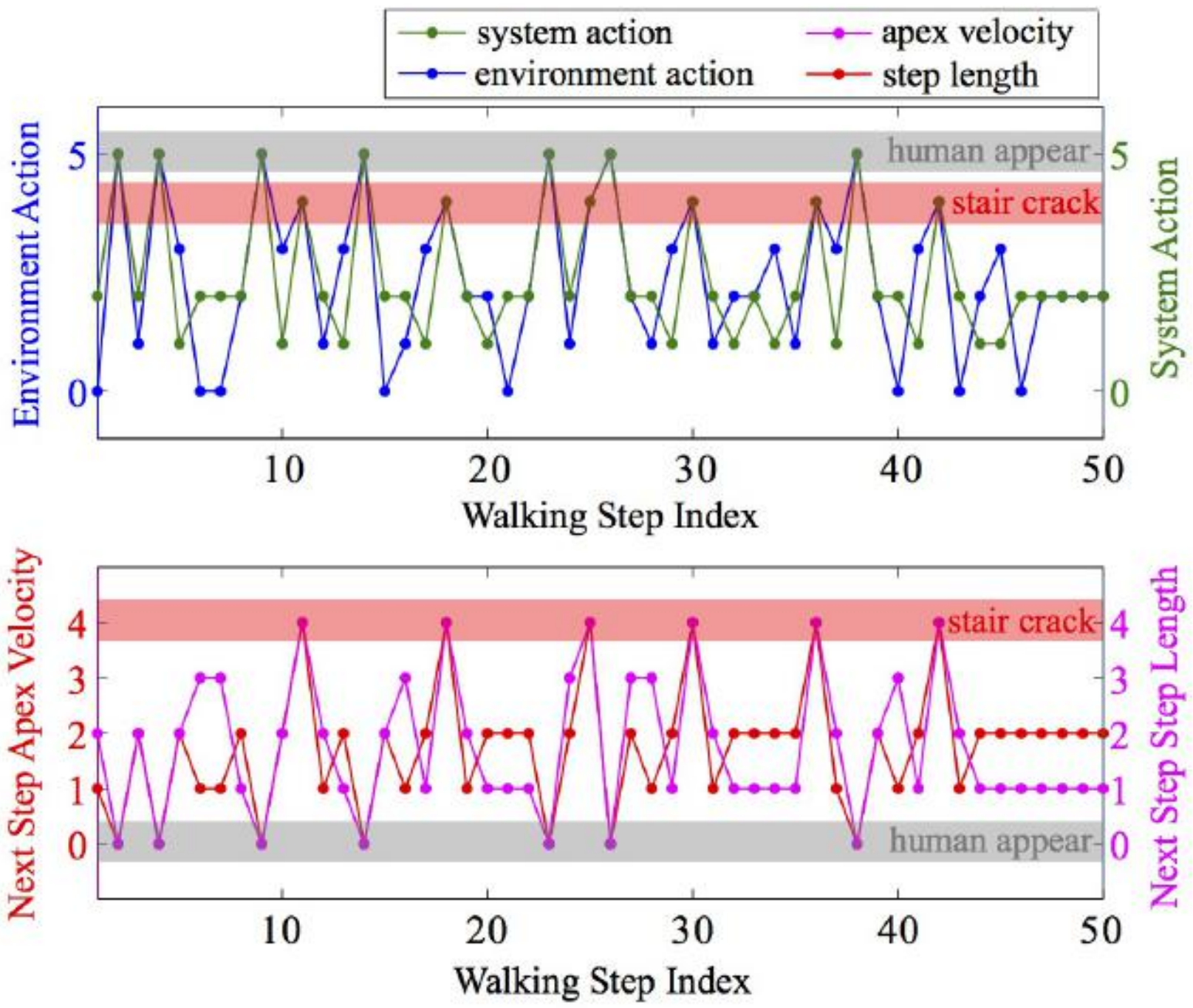

Figure 5.5: Environment actions, system actions and keyframe states. This figure illustrates the results of running the synthesized automaton for 50 walking steps. Actions and states are indexed and all the specifications are satisfied. Emergent behaviors, i.e, human appear and stair crack, are marked in the shaded regions.

## 5.5 Simulation Results

In this section, we demonstrate our locomotion results by using synthesized planner and low-level phase-space planner with switched modes. For illustration convenience, we index both environment action $\mathcal{E}$ in Eq. (5.8) and system action $\mathcal{S}$ in Eq. (5.9) as $\{0, \ldots, 5\}$ in order, respectively. System keyframe state $\mathcal{Q}$ is decomposed into apex velocity and step length. For both of them, Small, Normal and Large are assigned as $\{1, 2, 3\}$ in order while Stop and Swing are assigned as $\{0, 4\}$. Temporal Logic Planning (TuLiP) toolbox, a python-based correct-by-construction embedded control software (Wongpiromsarn et al., 2011), is used to synthesize the locomotion contact planner. The gr1c tool[4] tool involving CU Decision Diagram Package by Fabio Somenzi, is chosen as the underlying synthesis solver. *Realizability* of this planner is checked by an off-the-shelf function. If the formulae are *realizable*, the synthesized planner is guaranteed to satisfy all the proposed specifications.

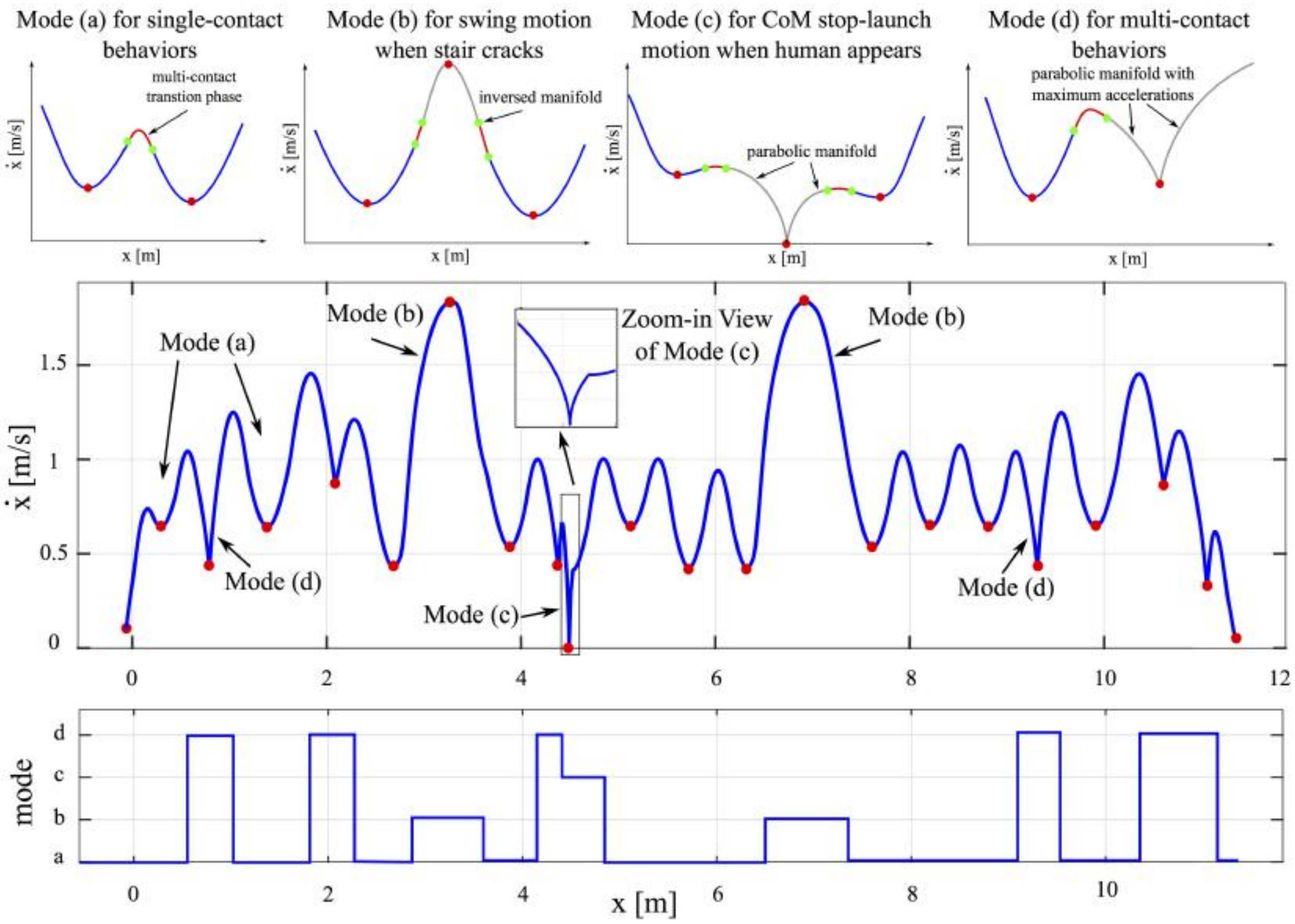

Figure 5.6: Phase-space trajectory and mode switchings for a 20-step WBL maneuver. The top four figures illustrate phase-space manifolds of the four locomotion modes. The mode switching is governed by the synthesized high-level contact planner. Among these steps, two stair cracks and one human appearance are taken into account.

[4]http://scottman.net/2012/gr1c

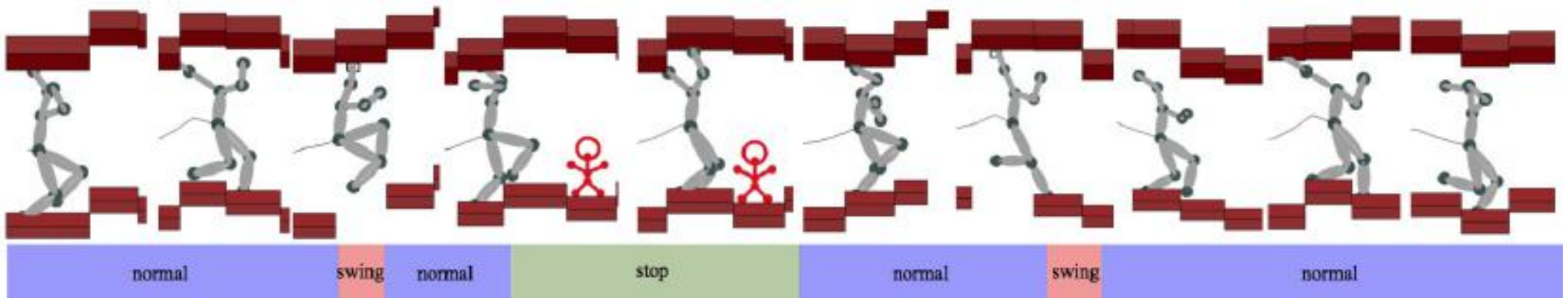

Figure 5.7: Snapshots of the whole-body locomotion behaviors. The snapshots shows a sequence of primary behaviors including swinging motion over the crashed stair and stopping motion when a human appears unexpectedly.

The resulting discrete planner is represented by a finite state automaton with 27 states and 148 transitions. Fig. 5.5 illustrates discrete environment and system actions, and their corresponding keyframe states. As it shows, based on non-deterministic environment actions and keyframe states, system action and system mode are designed as desired. In particular, two modeled emergent behaviors are highlighted in shaded regions. Non-determinism is observed in the system action and keyframe states.

As to the low-level planner, four different modes are invoked according to the high-level planner requests. Fig. 5.6 shows the synthesized CoM Sagittal phase-space trajectory of a 20-step walking. Among these steps, two stair cracks and one human appear are taken into account. Discretized keyframe states are designed based on the proposed two-player game. Fig. 5.7 illustrates dynamic motion snapshots. A simulation video is shown in (Uni, 2016).

## 5.6 Discussions and Conclusions

This chapter employs temporal-logic-based formal methods to synthesize high-level reactive planners for the generation of the complex locomotion be-haviors in cluttered environments. Contact decisions are determined according to the synthesized switching protocol. A particular focus has been given to the correctness of the overall implementation from the high-level to the low-level planners. We proposed new locomotion models at the low-level to deal with complex dynamic motions. The proposed hierarchical planning framework is validated through simulations of WBL maneuvers in a dynamic environment.

In the future, a couple of interesting problems awaits further explorations. For locomotion problems, the contact configurations and keyframe states may inherently obey a particular probabilistic model. Incorporating probabilistic models, such as Markov decision process (MDP) or POMDP, into the contact decisions will be our next focus. We plan to study probabilistic correctness guarantee and planner synthesis.

Our current planner assumes arm and leg contacts to switch synchronously. A future extension is to relax this assumption and study more sophisticated *yet* realistic contact dynamics involving switching asynchrony.

Reasoning robustness at the temporal logic level is another topic of our interest. Practical uncertainties tend to be caused by stair slippery (friction constraint violation) and large stair tilting angles. A few interesting references using signal temporal logic to quantify robustness are presented in (Farahani et al., 2015; Deshmukh et al., 2015; Sadraddini and Belta, 2015).

Our study assumes that the low-level motion planner can accurately implement the high-level command synthesized from the task planner. However, it may not be the case in practice. For instance, the terrain height can be uncertain, or the terrain becomes slippery, or the designed keyframe conflicts with robot kinematic constraints. In this case, the task planner should be aware of this execution failure, and re-plan the command timely. It is imperative to propose an interfacing layer (He et al., 2015; Dantam et al., 2016) in order to establish a feedback mechanism between task and motion planners. This area of research is quite active and provides a promising solution for achieving more realistic robotic behaviors interacting with dynamic environments.

Till now, we primarily focus on specifying the contact tasks, and it will be interesting to generalize the behaviors to more sophisticated tasks such as torso orientations, manipulation, grasping and iterations with humans.

# Chapter 6

# Stability and Impedance Performance of Distributed Feedback Controllers with Delays

Robotic systems are increasingly relying on distributed feedback controllers to tackle complex sensing and decision problems such as those found in highly articulated human-centered robots. These demands come at the cost of a growing computational burden and, as a result, larger controller latencies. To maximize robustness to mechanical disturbances by maximizing feedback control gains, this chapter emphasizes the necessity for compromise between high- and low-level feedback control effort in distributed controllers. Specifically, the effect of distributed impedance controllers is studied where damping feedback effort is executed in proximity to the control plant, and stiffness feedback effort is executed in a latency-prone centralized control process. Let us pose some critical questions regarding distributed stiffness-damping feedback controllers considered in this study: (A) Does controller stability have different sensitivity to stiffness and damping feedback delays? (B) If that is the case, what are the physical reasons for such a difference? To answer these

---

This chapter incorporates the results from the following publications: (Zhao, Paine, Kim, and Sentis, 2015b), (Zhao, Paine, and Sentis, 2014c). The results in Section 6.3 are joint works with Nicholas Paine and Kwan Suk Kim.

questions, this chapter studies the physical behavior of the proposed real-time distributed system using control analysis tools, such as the phase margin stability criterion, applied to the system's plant. Using these tools, our study reveals that system closed-loop stability and performance are much more sensitive to damping feedback delays than to stiffness feedback delays. We pursue a detailed analysis of this observation that leads to a physical understanding of the disparity. Then a practical controller breakdown gain rule is derived to aim at enabling control designers to consider the benefits of implementing their control applications in a distributed fashion.

To demonstrate the effectiveness of the proposed methods, this study implements tests on a high-performance actuator followed by experiments on a mobile base. First, a position step response is tested on an actuator under various combinations of stiffness and damping feedback delays. The experimental results show a high correlation to their corresponding simulation results. Second, the proposed distributed controller are applied to an implementation into an omnidirectional base. The results show a substantial increase in closed-loop impedance capabilities, which results in higher tracking accuracy with respect to the monolithic centralized controller counterpart approach.

The organization of this chapter is as follows: Section 6.1 proposes a distributed control architecture, which is simulated with observations of phase margin sensitivity to feedback delays; Then the fundamental reasons for this sensitivity discrepancy are analyzed in detail, and a servo breakdown gain rule is proposed correspondingly in Section 6.2; To validate this discrepancy, ex-perimental evaluations are shown in Section 6.3. Finally, Section 6.4 concludes this study and discusses future directions.

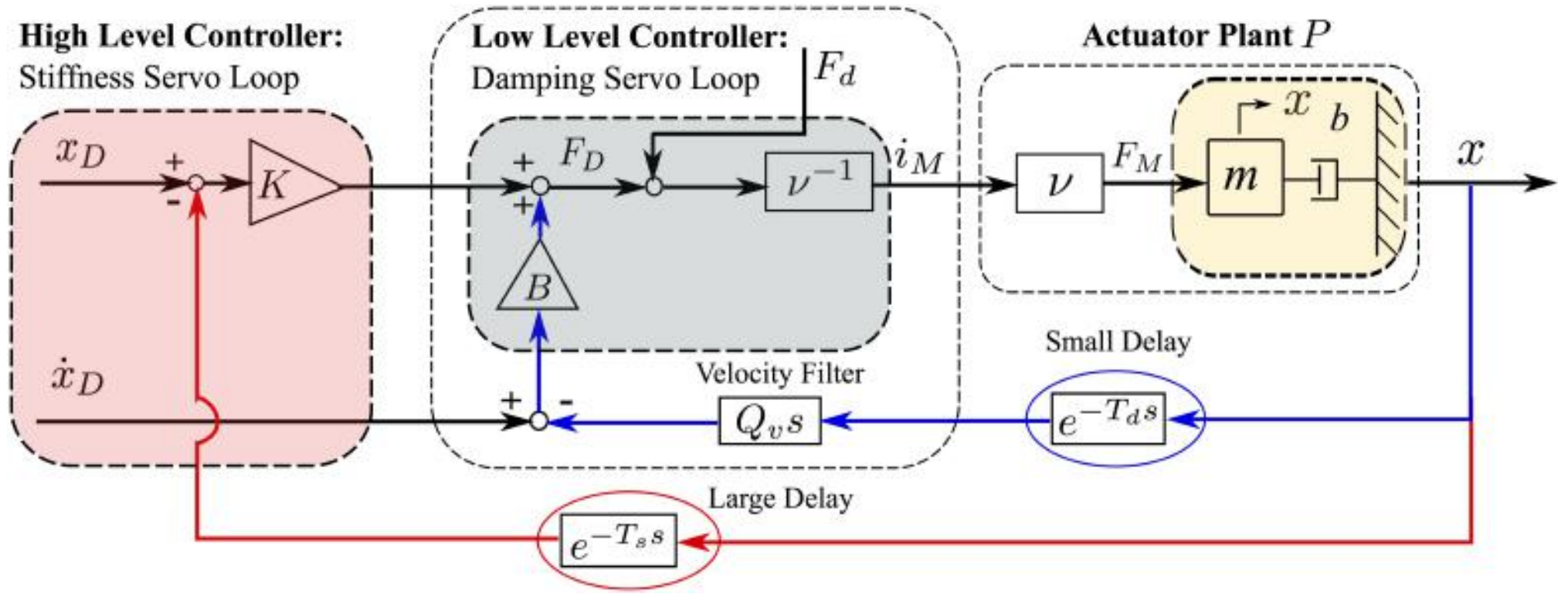

Figure 6.1: Single input / single output controller with distributed structure. A simple proportional-derivative control law is used to control an actuator. $P$ denotes the actuator plant with motor current input, $i_M$, and position output, $x$. $\nu^{-1}$ represents a scaling constant mapping the desired force, $F_D$, to the motor current, $i_M$. $K$ is the stiffness feedback gain while $B$ is the damping feedback gain.

## 6.1 Distributed Control Structure

This section describes the actuator model used to analyze closed-loop system stability, propose a basic distributed control architecture that delocalizes stiffness and damping servo loops, and analyzes the sensitivity of these control processes to loop delays. Many rigid electrical actuators like the ones used in modern robots can be approximately modeled as a second-order plant with a force acting on an inertia-damper pair (as shown in Fig. 6.1).

Considering a current-controlled motor, the control plant from current,

$i_M$, to position, $x$, is

$$P(s) = \frac{x(s)}{i_M(s)} = \frac{x(s)}{F_M(s)} \frac{F_M(s)}{i_M(s)} = \frac{\nu}{ms^2 + bs}, \tag{6.1}$$

where $F_M$ is the applied motor force, $\nu \triangleq F_M/i_M = \eta N k_\tau$, $\eta$ is the drivetrain efficiency, $N$ is the gear speed reduction and $k_\tau$ is the motor torque constant.

### 6.1.1 Closed-Loop Distributed Controller

Fig. 6.1 shows our proposed distributed controller built using a proportional-derivative feedback mechanism. It includes velocity feedback filtering ($Q_v s$), stiffness feedback delay ($T_s$), damping feedback delay ($T_d$), with $T_s \neq T_d$, stiffness feedback gain ($K$) and damping feedback gain ($B$). Excluding the unknown load ($F_d$), the desired motor force ($F_D$) in the Laplace domain associated with the proposed distributed controller is

$$F_D(s) = K(x_D - e^{-T_s s} x) + B(x_D s - e^{-T_d s} Q_v x s), \tag{6.2}$$

where $s$ is the Laplace variable, $x_D$ and $\dot{x}_D$ (i.e., $x_D s$ in the Laplace domain) are the desired output position and velocity, and $e^{-T_s s}$ and $e^{-T_d s}$ represent Laplace transforms of the time delays in the stiffness and damping feedback loops, respectively. Using Eqs. (6.1) and (6.2), one can derive the closed-loop transfer function from desired to output positions as

$$P_{CL}(s) = \frac{x}{x_D} = \frac{Bs + K}{ms^2 + (b + e^{-T_d s} B Q_v)s + e^{-T_s s} K}, \tag{6.3}$$

where $Q_v$ is chosen to be a first order low pass filter with a cutoff frequency $f_v$

$$Q_v(s) = \frac{2\pi f_v}{s + 2\pi f_v}. \tag{6.4}$$

To derive the open-loop transfer function (Ogata and Yang, 2010) of the distributed controller, one can re-write Eq. (6.3) as

$$P_{CL}(s) = \frac{\frac{Bs+K}{ms^2+bs}}{1 + P_{OL}(s)}, \tag{6.5}$$

where $P_{OL}(s) \triangleq P(s)H(s)$ is the open-loop transfer function,f

$$P_{OL}(s) = \frac{e^{-T_d s} B Q_v s + e^{-T_s s} K}{ms^2 + bs}, \tag{6.6}$$

$P(s)$ is the actuator's plant, and $H(s)$ is the so-called feedback transfer function.

The presence of delays and filtering causes the above closed loop plant to behave as a high order dynamic system for which typical gain selection methods do not apply. However, to make the problem tractable, one can define a dependency between the stiffness and damping gains using an idealized second order characteristic polynomial (Ogata and Yang, 2010)

$$s^2 + 2\zeta\omega_n s + \omega_n^2, \tag{6.7}$$

where $\omega_n$ is the so-called natural frequency and $\zeta$ is the so-called damping factor. In such case, the idealized characteristic polynomial (i.e. ignoring delays, $T_s = T_d = 0$, and filtering, $Q_v = 1$) associated with our closed loop plant of Eq. (6.3) would be

$$s^2 + (B + b)/m \cdot s + K/m. \tag{6.8}$$

Choosing the second order critically damped rule, $\zeta = 1$ and comparing Eqs. (6.7) and (6.8) one can get the gain dependency

$$B = 2\sqrt{mK} - b, \tag{6.9}$$

and the natural frequency,

$$f_n \triangleq \frac{\omega_n}{2\pi} = \frac{1}{2\pi}\sqrt{\frac{K}{m}}. \tag{6.10}$$

The second order dependency of Eq. (6.9) will be used for the rest of this dissertation for deriving new gain selection methods through the thorough analysis of the oscillatory behavior of the closed loop plant of Eq. (6.3). In particular our study will use the phase margin criterion and other visualizations tools to study how the complete system reacts to feedback delays and signal filtering. Phase margin is the additional phase value above $-180°$ when the magnitude plot crosses the 0 dB line (i.e., the gain crossover frequency). It is common to quantify system stability by its phase margin.

For the proposed distributed controller in Fig. 6.1, the damping feedback loop is labeled as low level (e.g. embedded) to emphasize that it is meant to be locally implemented to take advantage of high servo rates. On the other hand the stiffness loop is implemented in a high-level computational process close to external sensors and centralized models, for Operational Space Control (OSC) purposes. The OSC is normally used in human-centered robotic applications where controllers use task coordinates and global models for their operation. The simplified controller in Fig. 6.1 is used to illustrate the discrepancies in sensitivity to latencies between the servo loops. It does not correspond to a practical robot controller as it contains only a single degree of freedom. After analyzing this structure, we implement a similar distributed controller into a multi-axis robotic base shown in Fig. 6.10, which results in the simultaneous improvement of system stability while achieving Operational Space Control.

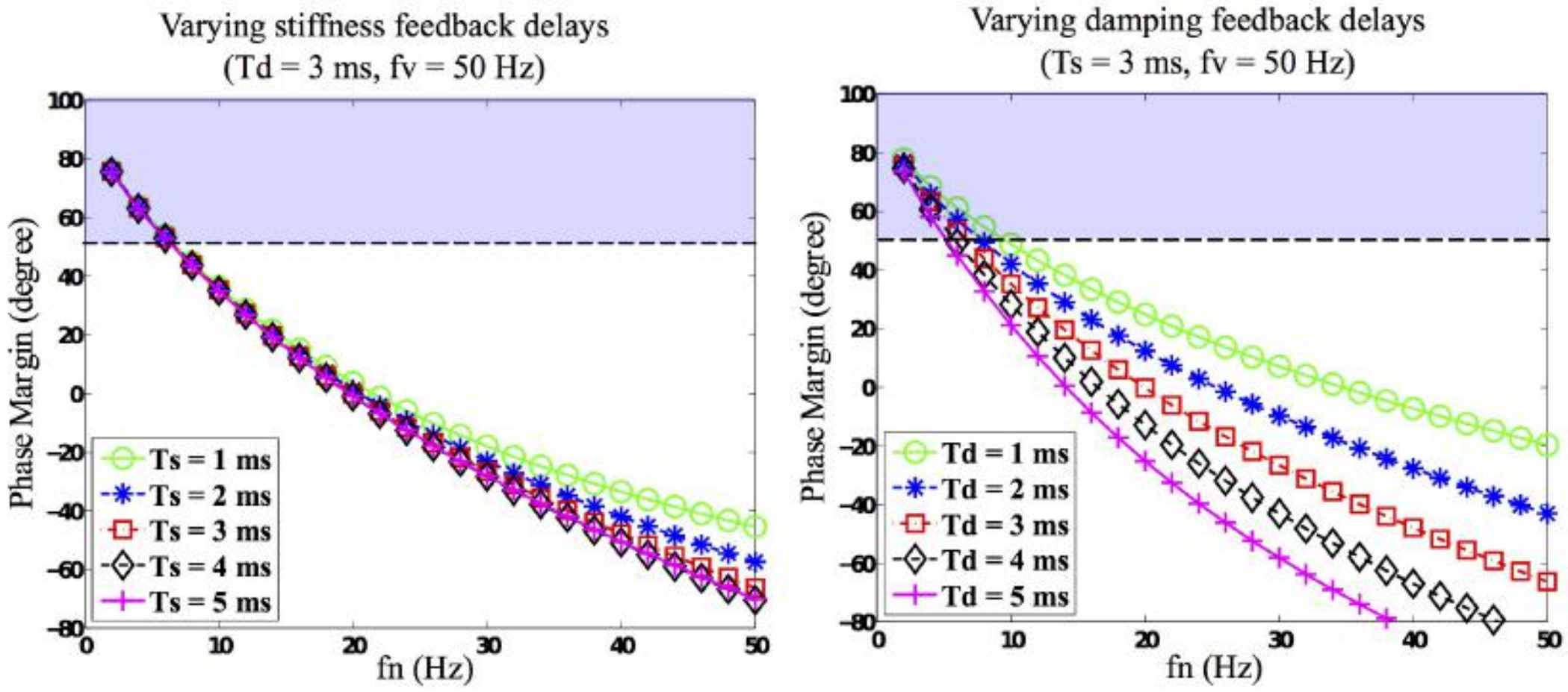

Figure 6.2: Phase margin sensitivity to time delays. This figure shows phase margin simulations of the open loop transfer function shown in (6.6) as a function of the natural frequency defined in (6.10) and the servo delays shown in Fig. 6.1. A phase margin of 0 degrees is considered marginally stable. Simulations indicate that phase margins less than 50 degrees exhibit oscillatory behavior (Paine and Sentis, 2015). The dashed line in this figure represents this threshold.

### 6.1.2 Phase Margin Sensitivity Comparison

This subsection focuses on utilizing frequency domain control methods to analyze the phase margin sensitivity to time delays on the distributed control architecture shown on Fig. 6.1. Different delay range scales are considered: (i) a small range scale ($1-5$ ms) to show detailed variations, and (ii) a larger range scale ($5-25$ ms) to cover practical delay ranges. These scales roughly correspond to embedded and centralized computational and communication processes found in highly articulated robots such as (Paine et al., 2015). Phase margin plots, are subsequently obtained for the controller of Eq. (6.3) and shown on Figs. 6.2 and 6.3 as a function of the natural frequency given in Eq. (6.10) and using the gain relationship of Eq. (6.9). All the simulations are carried out using Matlab@ software. Feedback delays are represented by the exponential term $e^{-Ts}$ in frequency domain. Phase margin is computed by using Matlab's margin() command based on the open loop transfer function.

In Fig. 6.2, delays ranging between 1 ms and 5 ms are simulated for both the stiffness and damping servos. The simulations are performed based on identical actuator parameters than those used in the experimental section, Section 6.3, i.e. passive output inertia $m = 256$ kg and passive damping $b = 1250$ Ns/m. Eqs. (6.9) and (6.10) can subsequently be used to derived the stiffness and damping feedback gains. It is noticeable that reducing either stiffness or damping feedback delays will increase the stability of the controller. But more importantly, it is clearly visible that phase margin behavior is much more sensitive to damping servo delays ($T_d$) than to stiffness servo delays ($T_s$). Not only there is a disparity on the behavior with respect to the delays, but phase margin is fairly insensitive to stiffness servo delays in the observed time scales. Such disparity and behavior is the central observation that motivates this study and the proposed distributed control architecture.

Fig. 6.3 simulates step position response of the controller for a range of relatively large stiffness delays and for two choices of damping delays, a short and a long one. The simulations are performed on the same actuator used in the experimental section, Section 6.3. The first point to notice here is that the phase margin values for subfigure (a) are significantly lower than for (d) due to the larger damping delay. Secondly, both (a) and (d) show small variations between the curves, corroborating the small sensitivity to stiffness delays that will be studied in Section 6.2. Corresponding step responses are shown along for various natural frequencies. It becomes clear that reducing damping delay significantly boosts stability even in the presence of fairly large stiffness delays. These results emphasize the significance of implementing damping terms at the fastest possible level (e.g. at the embedded level) while proportional (i.e. stiffness) servos can run in latency prone centralized processes. This conclusion is also validated by the Nichols diagram (Zhao et al., 2014c). We use frequency domain techniques to validate the stability discrepancies and robustness to stiffness delays shown in the phase margin diagrams above.

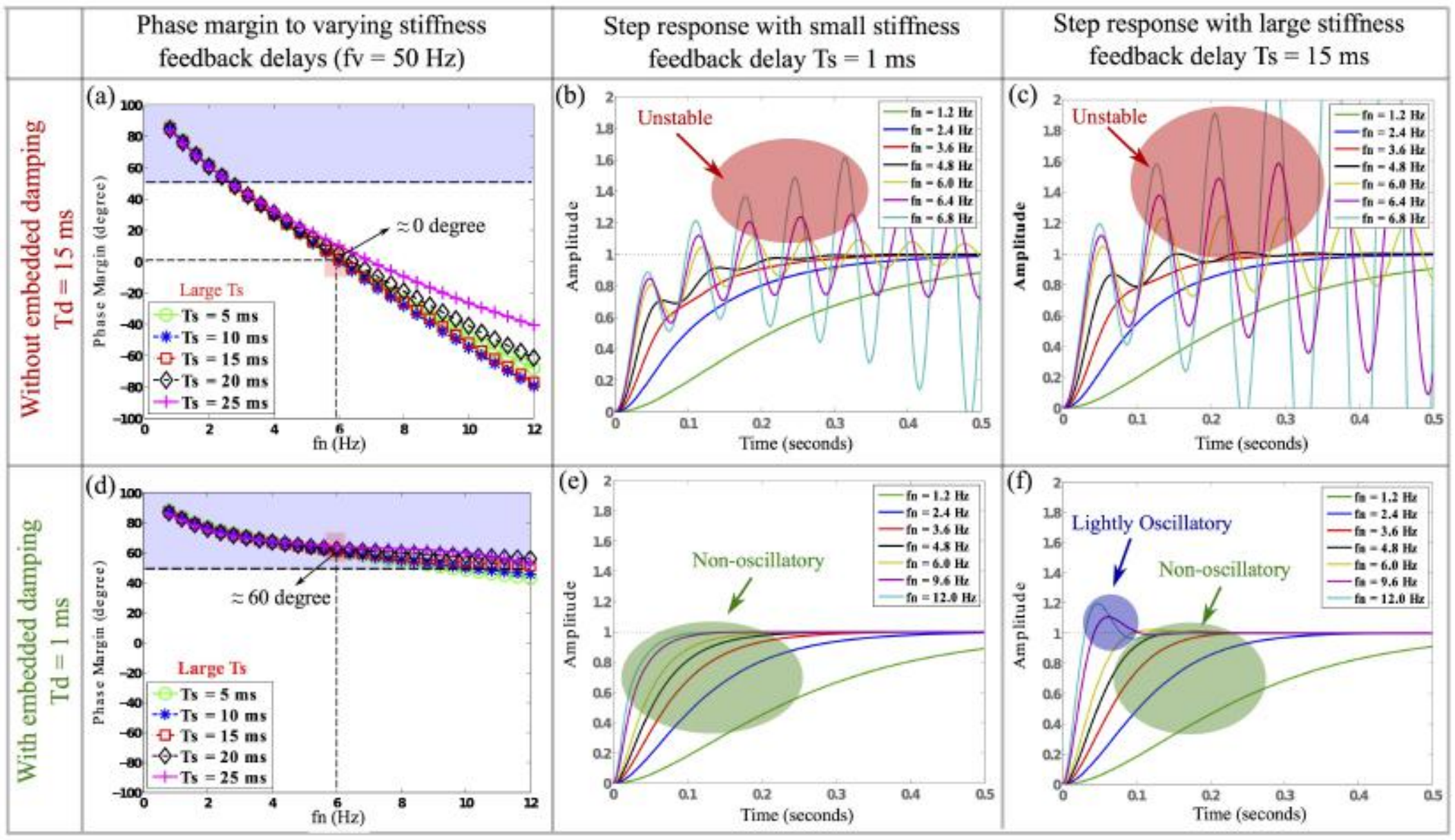

Figure 6.3: Comparison between step responses with slow and fast damping servos. The subfigures above compare the effects of damping feedback on slow or fast servo processes. The top row depicts damping feedback implemented with delays of 15 ms while the bottom row depicts a faster damping servo with delay of only 1 ms. For both rows, various stiffness delays are identically used ranging from 5 ms to 25 ms. Subfigures (a) and (d) show simulations of the phase margin once more as a function of the natural frequency, which in turn is a function of the feedback stiffness gain.

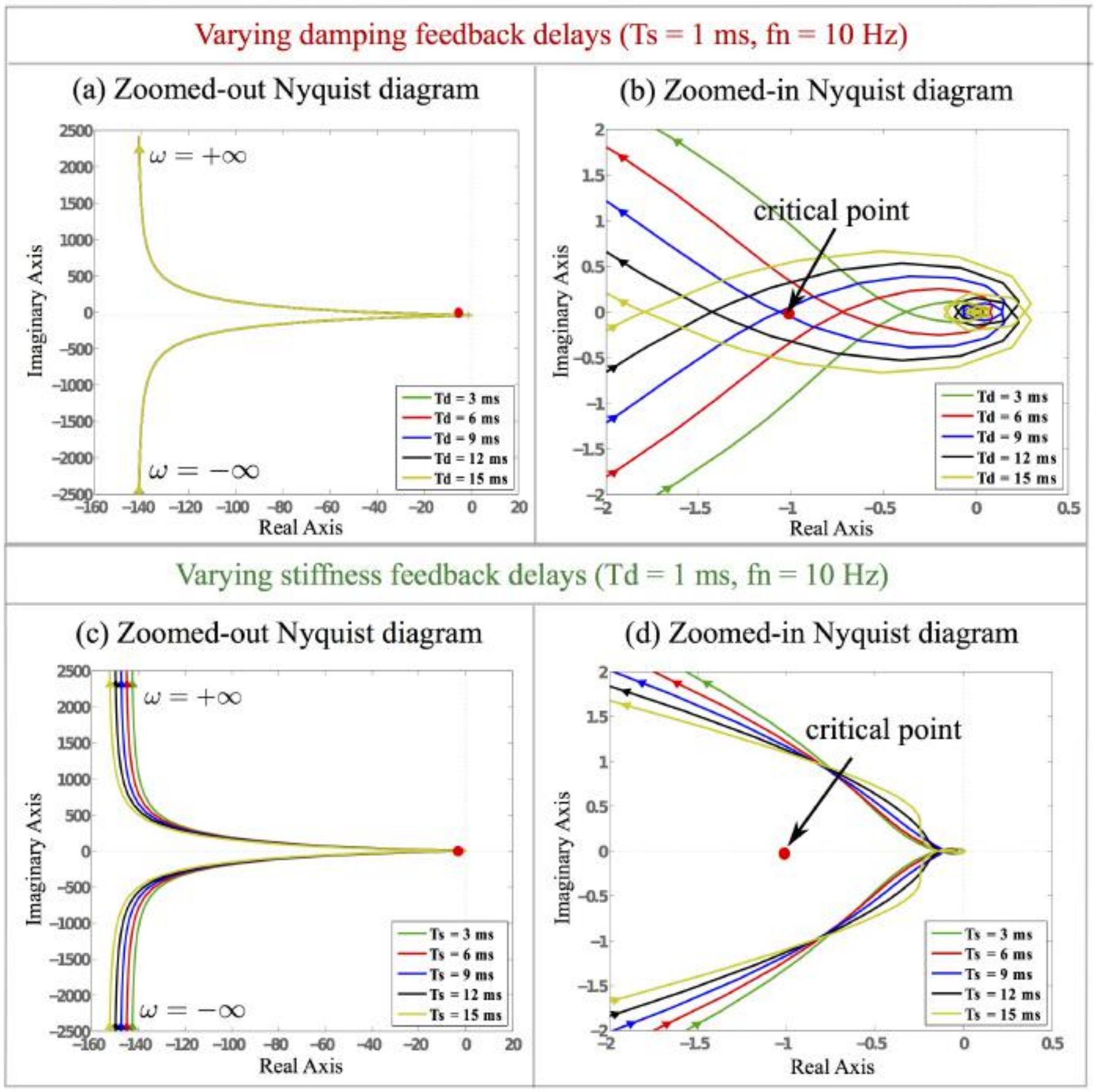

Figure 6.4: Nyquist diagrams with varying stiffness and damping feedback delays. These figures show Nyquist diagrams associated with the open loop transfer function in Eq. (6.6). On the two leftmost subfigures we show Nyquist diagram for large damping feedback delays while on the right side we show plots for small delays.

Notice that the open-loop transfer function $P_{OL}(s)$ in Eq. (6.6) has two non-positive poles, $s_1 = 0, s_2 = -m/b$, and therefore has no poles strictly on the right hand side. The resulting Nyquist criterion (Ogata and Yang, 2010) applied to the above transfer function is shown on the right side of Fig. 6.4 for varying stiffness and damping servo delays. We are looking for clockwise encirclements of the critical point $-1 + 0j$ by $P_{OL}(s)$, which would indicate instability of the closed loop plant $P_{CL}(s)$ according to the Nyquist criterion.

Comparing subfigures (b) and (d) in Fig. 6.4, we can see that when $T_s$ increases, the Nyquist plot does not encircle the critical point, indicating stability. On the other hand, when $T_d$ increases passed 9 ms, the Nyquist plot starts encircling the critical point, indicating instability. Once more, we have re-confirmed both the disparity on behavior due to delays of the stiffness and damping servos as well as the robustness of the stiffness servo to those delays.

## 6.2 Basis of Phase Margin Sensitivity Discrepancy

In the previous section, it was observed different behavior of the controller's phase margin depending on the nature of delay. Damping delay seems to affect much more the system's phase margin than stiffness delay. This section will analyze this physical phenomenon in more detail and reveal the conditions under which this disparity occurs. Detailed mathematical analysis is developed to find further physical structure for the causes of stability discrepancies between damping and stiffness delays. Let us re-visit the open loop transfer function of Eq. (6.6). The resulting open loop transfer function, including the low pass velocity filter of Eq. (6.4), in the frequency domain ($s = j\omega$) is

$$P_{OL}(\omega) = \frac{jA_1(\omega) + A_2(\omega)}{j\omega(jm\omega + b)(j\tau_v\omega + 1)}, \tag{6.11}$$

with

$$A_1(\omega) \triangleq B\,\omega\cos(T_d\,\omega) - K\sin(T_s\,\omega) + K\tau_v\omega\cos(T_s\omega),$$

$$A_2(\omega) \triangleq B\,\omega\sin(T_d\,\omega) + K\cos(T_s\,\omega) + K\tau_v\omega\sin(T_s\omega). \tag{6.12}$$

Note that Euler's Formula ($e^{-jx} = \cos x - j\sin x$) has been used to obtain the above results.

The phase margin, $PM \triangleq 180^\circ + \angle P_{OL}(\omega_g)$, of the plant (6.11), where $\angle.$ is the angle of the argument, is

$$PM = \mathrm{atan}\left[\frac{\mathrm{A_{1g}}}{\mathrm{A_{2g}}}\right] + 90^\circ - \mathrm{atan}\left[\frac{\mathrm{m}\,\omega_\mathrm{g}}{\mathrm{b}}\right] - \mathrm{atan}\left[\tau_\mathrm{v}\omega_\mathrm{g}\right], \tag{6.13}$$

with $\omega_g$ being the gain crossover frequency (Ogata and Yang, 2010) and $A_{ig} \triangleq A_i(\omega_g), i = \{1,2\}$. Following the derivations of Appendix F, we obtain the sensitivity equations below expressing variations of the phase margin with respect to stiffness and damping delays,

$$\frac{\partial PM}{\partial T_s} = \frac{\left[-K^2(\tau_v^2\omega_g^2+1) + K\,B\,\omega_g\,M\right]\,\omega_g}{A_{1g}^2 + A_{2g}^2}, \tag{6.14}$$

$$\frac{\partial PM}{\partial T_d} = \frac{\left[-B^2\omega_g^2 + K\,B\,\omega_g\,M\right]\,\omega_g}{A_{1g}^2 + A_{2g}^2}. \tag{6.15}$$

where

$$M = \sqrt{(\tau_v\omega_g)^2+1}\cdot\sin\Big((T_s - T_d)\,\omega_g + \phi\Big) \tag{6.16}$$

where the phase shift $\phi \triangleq \mathrm{atan}(-\tau_v\omega_g)$.

### 6.2.1 Gain Crossover Sensitivity Condition

From the control analysis of the distributed plant performed in previous sections, increasing damping delays decreases the phase margin. This observation means that the sensitivity of the phase margin to damping delays must be negative, i.e.

$$\frac{\partial PM}{\partial T_d} < 0. \tag{6.17}$$

Also from those analysis, it is observed that the phase margin is more sensitive to damping than to stiffness delays. This observation can be formulated as

$$\frac{\partial PM}{\partial T_d} < \frac{\partial PM}{\partial T_s}. \tag{6.18}$$

Let us re-organize the numerator of Eq. (6.15) to be written in the alternate form

$$\frac{\partial PM}{\partial T_d} = \frac{\left[-B\omega_g + KM\right] B\,\omega_g^2}{A_{1g}^2 + A_{2g}^2}. \tag{6.19}$$

An upper bound of the above equation occurs when the maximal condition $\sin\Big((T_s - T_d)\,\omega_g + \phi\Big) = 1$ is met, i.e.

$$\frac{\partial PM}{\partial T_d} \leq \frac{\left[-B\omega_g + K\sqrt{(\tau_v\omega_g)^2 + 1}\right] B\,\omega_g^2}{A_{1g}^2 + A_{2g}^2}. \tag{6.20}$$

Based on the above inequality, (6.17) is met if the following condition is met.

**Proposition 6.1** (**Gain crossover sensitivity condition**). *If the following gain crossover sensitivity condition holds,*

$$\omega_g > \frac{K}{\sqrt{B^2 - K^2\tau_v^2}}, \tag{6.21}$$

*then the Condition (6.17) is fulfilled.*

Obtaining a closed form solution for that condition would be very complex due to the presence of trigonometric terms. Therefore, the remainder of this section is to study under what circumstances Condition (6.21) holds.

At the same time, Inequality (6.18) can be re-written in the form

$$\frac{\partial PM}{\partial T_d} - \frac{\partial PM}{\partial T_s} = \frac{\left[-B^2\omega_g^2 + K^2(\tau_v^2\omega_g^2 + 1)\right]\omega_g}{A_{1g}^2 + A_{2g}^2} < 0, \tag{6.22}$$

where it has been subtracted the right hand sides of Eqs. (6.14) and (6.15) for the derivation. Notice that in that subtraction the sine functions cancel out. Coincidentally, the above inequality is also met if the gain crossover sensitivity condition (6.21) is fulfilled. In other words, that condition is sufficient to meet both Inequalities (6.17) and (6.18).

### 6.2.2 Servo Breakdown Gain Rule

To validate the gain crossover condition (6.21), this study solves for the gain crossover frequency, which consists of the frequency at which the magnitude of the open loop transfer function is equal to unity, i.e.

$$|P_{OL}(\omega_g)| = 1. \tag{6.23}$$

Using the plant (6.11), it can be shown that the above equation results in the equality (see the Appendix F for the derivations)

$$(B\omega_g)^2 + K^2(\tau_v^2\omega_g^2 + 1) - 2KB\omega_g M = \omega_g^2\Big((\omega_g m)^2 + b^2\Big)\Big(\tau_v^2\omega_g^2 + 1\Big). \tag{6.24}$$

The above equation is intractable in terms of deriving a closed loop expression of the gain crossover frequency. To tackle a solution this study introduces transformations of the parameters and numerically derives parameter ranges for which Condition (6.21) holds. Let us start by creating a new variable that allows to write (6.21) as an equality,

$$\delta \in [-1, \infty) \quad s.t. \quad \omega_g = (1+\delta)\,\frac{K}{\sqrt{B^2 - K^2\tau_v^2}}. \tag{6.25}$$

Thus, demonstrating the gain crossover sensitivity condition (6.21) is equivalent to demonstrating that $\delta > 0$. Rewriting Eq. (6.9) as $K = (B+b)^2/4m$ and substituting $K$ in the above equation, (6.25) can be further expressed as

$$\omega_g = (1+\delta)\,\frac{(B+b)^2}{\sqrt{16B^2m^2 - (B+b)^4\tau_v^2}}. \tag{6.26}$$

Dividing Eq. (6.24) by a new term $K^2\,U\,V$, with $U \triangleq \tau_v^2\omega_g^2 + 1$, and $V \triangleq B^2\,\omega_g^2/K^2$, while substituting $\omega_g$ on the right hand side of Eq. (6.24) by Eq. (6.26), and using $M$ as shown in Eq. (6.16), Eq. (6.24) becomes

$$\frac{1}{U} + \frac{1}{V} - \frac{2\sin\Big((T_s - T_d)\,\omega_g + \phi\Big)}{\sqrt{U \cdot V}} = \frac{\big(1+\delta\big)^2\,\big(B+b\big)^4}{16\,B^4 - B^2(B+b)^4\tau_v^2/m^2} + \left(\frac{b}{B}\right)^2. \tag{6.27}$$

Using Eq. (6.25) it can be further demonstrated that $V = (\tau_v\omega_g)^2 + (1+\delta)^2$. Thus $U$ and $V$ are only expressed in terms of $(\tau_v\omega_g)^2$. To further facilitate the analysis, let us introduce three more variables

$$\alpha \triangleq \sin\Big((T_s - T_d)\,\omega_g + \phi\Big) \in [-1, 1], \tag{6.28}$$

$$\beta \in (0, \infty) \quad s.t. \quad B = \beta\, m, \tag{6.29}$$

$$\gamma \in (0, \infty) \quad s.t. \quad B = \gamma\, b. \tag{6.30}$$

Notice that $\alpha$ can be interpreted as an uncertainty, $\beta$ is the ratio between damping gain and motor drive inertia and $\gamma$ is the ratio between damping gain and motor drive friction. Using these variables, (6.27) simplifies to

$$\frac{U + V - 2\alpha\sqrt{U \cdot V}}{U \cdot V} = \frac{\left(1+\delta\right)^2 (1+\gamma)^4}{16\,\gamma^4 - (1+\gamma)^4\beta^2\tau_v^2} + \frac{1}{\gamma^2}. \tag{6.31}$$

Using Eqs. (6.26), (6.29) and (6.30), the term $(\tau_v\omega_g)^2$ appearing in the variables $U$ and $V$ on Eq. (6.31) can be expressed as

$$(\tau_v\omega_g)^2 = \beta^2\tau_v^2 \frac{\left(1+\delta\right)^2 (1+\gamma)^4}{16\,\gamma^4 - (1+\gamma)^4\beta^2\tau_v^2} \tag{6.32}$$

Thus, Eq. (6.31) does not contain direct dependencies with $\omega_g$ and therefore can be represented as the nonlinear function

$$f(\alpha, \beta, \gamma, \delta, \tau_v) = 0 \tag{6.33}$$

Let us demonstrate under which conditions $\delta > 0$, which will imply that Eq. (6.21) holds. In this study, velocity filters with $\tau_v = 0.0032\,s$ are commonly used for achieving high performance control (Paine and Sentis, 2015), and therefore Eq. (6.31) will be solved for only that filter. Notice that it is not difficult to try new values of $\tau_v$ when needed. Additionally, when sampling Eq. (6.31) for the values of $\alpha$ shown in Eq. (6.28), it is observed that not only $\delta$ is fairly invariant to $\alpha$ but the lowest value of $\delta$ occurs for $\alpha = 1$. These behaviors are omitted here for space purposes. Therefore, as a particular solution, Eq. (6.31) is solved for the values

$$f(\alpha = 1, \beta, \gamma, \delta, \tau_v = 0.0032) = 0. \tag{6.34}$$

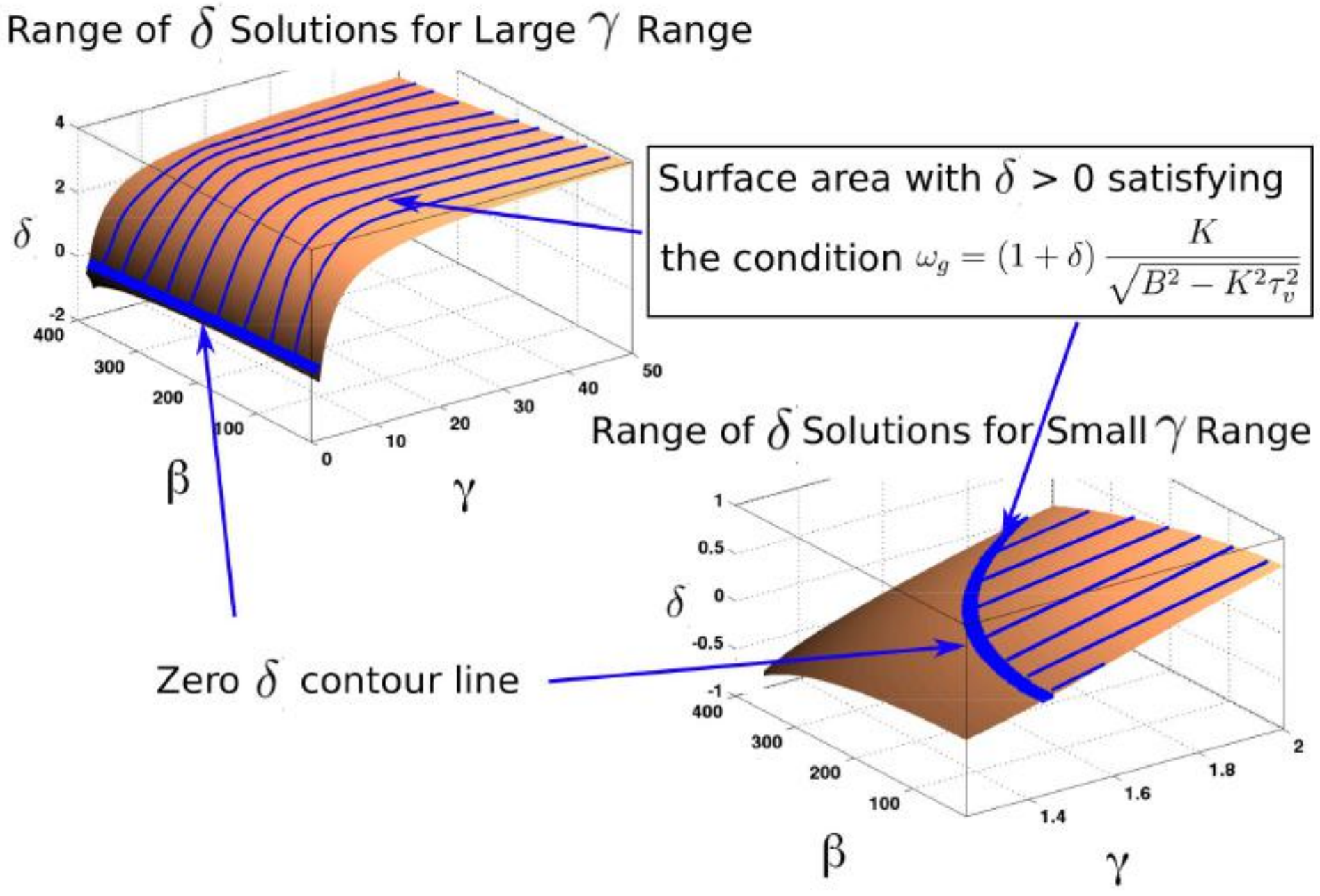

Figure 6.5: Controller values meeting the gain crossover sensitivity condition. The surfaces above show the range of feedback parameters that meet the gain crossover sensitivity condition of Eq. (6.21). $\delta > 0$ represents the excess gain ratio by which the condition is met. $\gamma > 0$ represents the ratio between damping feedback gain and passive damping. $\beta \in [10, 400]$ is chosen to cover a wide range of actuator parameters.

The above function is solved numerically and the solution surface is plotted in Fig. 6.5. The surfaces in this figure demonstrate that a wide range of practical gains, $\gamma$, meet the aforementioned gain crossover sensitivity condition. The values of the above surfaces are solved by numerically identifying the smallest real root of Eq. (6.31). In the bottom right surface, it can be seen that $\delta > 0$ for $\gamma > 2$, allowing us to state that using a distributed PD feedback control law like the one in Fig. 6.1 with the particular choice of the filter $\tau_v = 0.0032\,s$ and with damping gains greater than

$$B > 2\,b, \tag{6.35}$$

Table 6.1: UT-SEA/Valkyrie actuator parameters

| Actuator type | Output inertia $m$ | Passive damping $b$ | Damping gain B | Ratio $\gamma$ |
|---|---|---|---|---|
| UT-SEA | 360 kg | 2200 N·s/m | 50434 N·s/m | 22.92 |
| Valkyrie 1 | 270 kg | 10000 N·s/m | 46632 N·s/m | 4.66 |
| Valkyrie 2 | 0.4 kg·m$^2$ | 15 Nm·s/rad | 68 Nm·s/rad | 4.55 |
| Valkyrie 3 | 1.2 kg·m$^2$ | 35 Nm·s/rad | 196 Nm·s/rad | 5.60 |
| Valkyrie 4 | 0.8 kg·m$^2$ | 40 Nm·s/rad | 145 Nm·s/rad | 3.61 |
| Valkyrie 5 | 2.3 kg·m$^2$ | 50 Nm·s/rad | 360 Nm·s/rad | 7.20 |
| Valkyrie 6 | 1.5 kg·m$^2$ | 60 Nm·s/rad | 259 Nm·s/rad | 4.32 |

causes the phase margin to be more sensitive to damping delays than to stiffness delays. The threshold above can therefore be interpreted as a breakdown gain rule which is sufficient to meet the gain crossover sensitivity condition (6.21), and from which the aforementioned phase margin sensitivity discrepancy follows.

This threshold hints towards a general rule for breaking controllers down into distributed servos, as was illustrated in Fig. 6.1. Namely, if the maximum allowable feedback damping gain for a given servo rate is significantly larger than twice the passive actuator damping, then the controller's stiffness servo can be decoupled from the damping servo to a slower computational process without hurting the controller's stability.

### 6.2.3 Example: Real-World Actuators Analysis

As a means of demonstrating the utility of the breakdown gain rule of Eq. (6.35), here we analyze several real-world actuation systems. Our goal is to determine whether the properties of each system make them good candidates for distributed control schemes with decoupled stiffness and damping feedback loops.

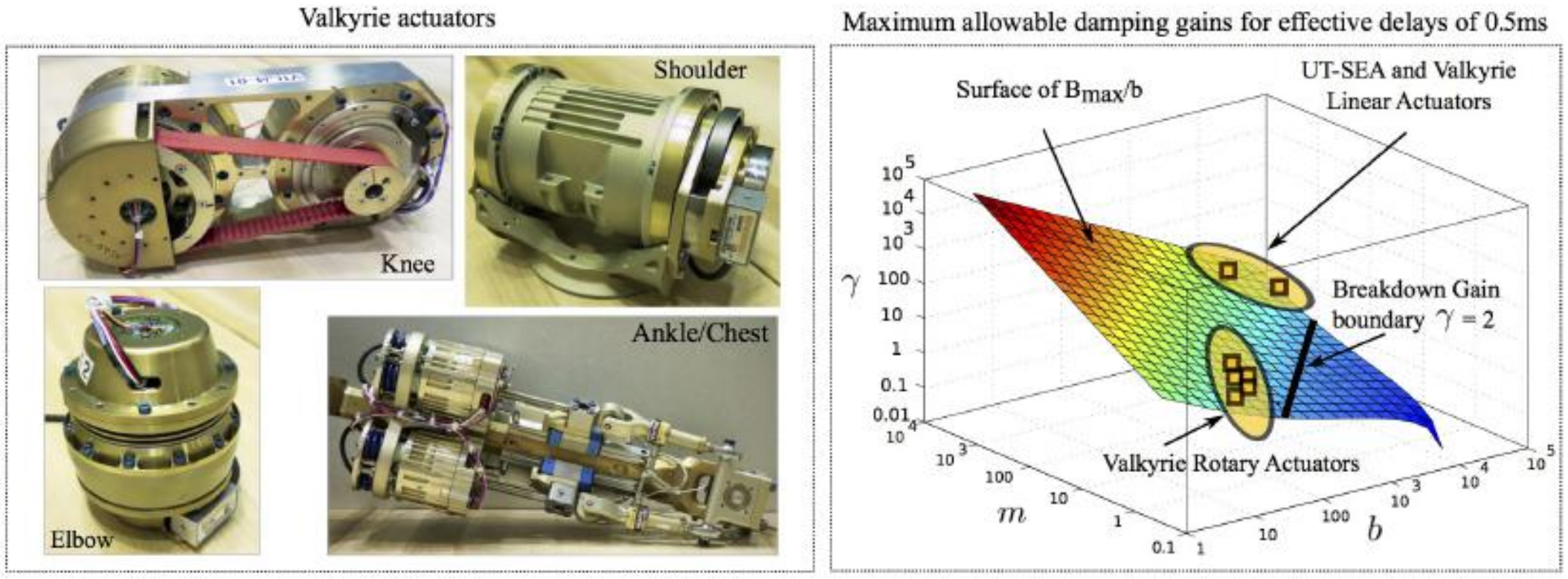

Figure 6.6: Various actuators meeting the servo breakdown gain rule. The left image shows various actuators from NASA that our group helped to build. The right image shows the surface of maximum allowable damping gains as a function of actuator parameters. $\gamma$ is the feedback damping gain ratio described in Eq. (6.30), and $m$ and $b$ are the output inertia and passive damping of the various actuators.

Table 6.1 shows actuator parameters for the Valkyrie humanoid and the UT-SEA actuator (Paine et al., 2014), as well as the maximum feedback damping gains that are implemented in those actuators to achieve maximum impedance control. Our lab has been involved in developing these two sets of actuators. In all instances, the embedded servos had effective delays of 0.5 ms. In order to compute the maximum feedback damping gains as a function of the previous servo rate, our recent work (Paine and Sentis, 2015) is used. In that work, a new rule is provided to compute maximum feedback gains for a phase margin of 50° given the actuator parameters and the servo rate. Fig. 6.6 shows, (i) pictures of various Valkyrie actuators, (ii) a surface depicting the maximum allowable damping gains as a function of actuator parameters, and (iii) Valkyrie's actuators mapped into the surface. The surface is computed for effective delays of 0.5 ms. Within the surface, it shows the line corresponding to the breakdown gain rule of Eq. (6.35). As can be seen, all actuators implement feedback damping gains that were above the breakdown gain boundary. It follows that those gains would be highly sensitive to damping servo delays.

To maintain these maximum actuator gains, servo latency for the damping process must not be increased. However, according to the servo breakdown rule, the stiffness servo processes shall be fairly insensitive to delays and therefore could be decoupled and implemented in a slower centralized process. Such decoupling is advantageous in multi-axis robots where centralized processes contain sensor and model information needed for Operational Space Control to coordinate the robot's movement. In Subsection 6.3.3 we discuss such an application for an omnidirectional mobile robotic base.

In the next section we study in detail the implementation of the proposed distributed control strategy in a new high-performance linear rigid actuator and an omnidirectional mobile base.

## 6.3 Experimental Evaluations

The proposed controller of Fig. 6.1 is implemented in our linear rigid actuator shown in Fig. 6.7. This actuator is equipped with a PC-104 form factor computer running Ubuntu Linux with an RTAI patched kernel (Paine et al., 2014). The PC communicates with the actuator using analog and quadrature signals through a custom signal conditioning board. Continuous signal time derivatives are converted to discrete form using a bilinear Tustin transform written in C. A load arm is connected to the output of the ball screw pushrod. Small displacements enable the actuator to operate in an approximately linear region of its load inertia. At the same time, the controller is simulated by using the closed loop plant given in Eq. (6.5). Identical parameters to the real actuator are used for the simulation, thus allowing us to compare both side by side. All the actuator tests are performed with a 1 kHz servo rate. Additional feedback delays are manually added by using a data buffer. Thus the total time delay is

$$T_{\text{total}} = \frac{T_{\text{sample}}}{2} + T_{\text{extra}}, \tag{6.36}$$

where the sampling period $T_{\text{sample}}$ is divided by two since the effective delay is the half of the sampling period (Hulin et al., 2008). The extra delay $T_{\text{extra}}$ represents a large communication delay in the centralized control structure.

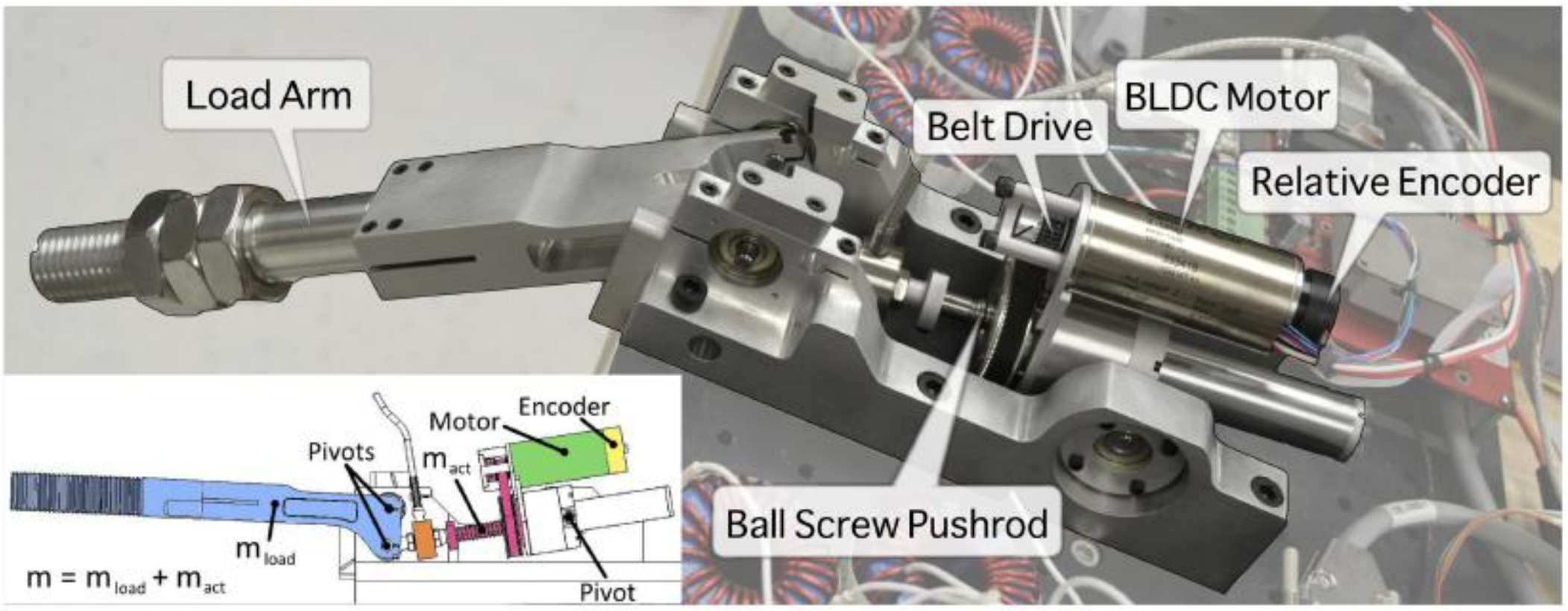

Figure 6.7: Linear UT actuator. This linear pushrod actuator has an effective output inertia of $m = 256$ kg and an approximate passive damping of $b = 1250$ Ns/m.

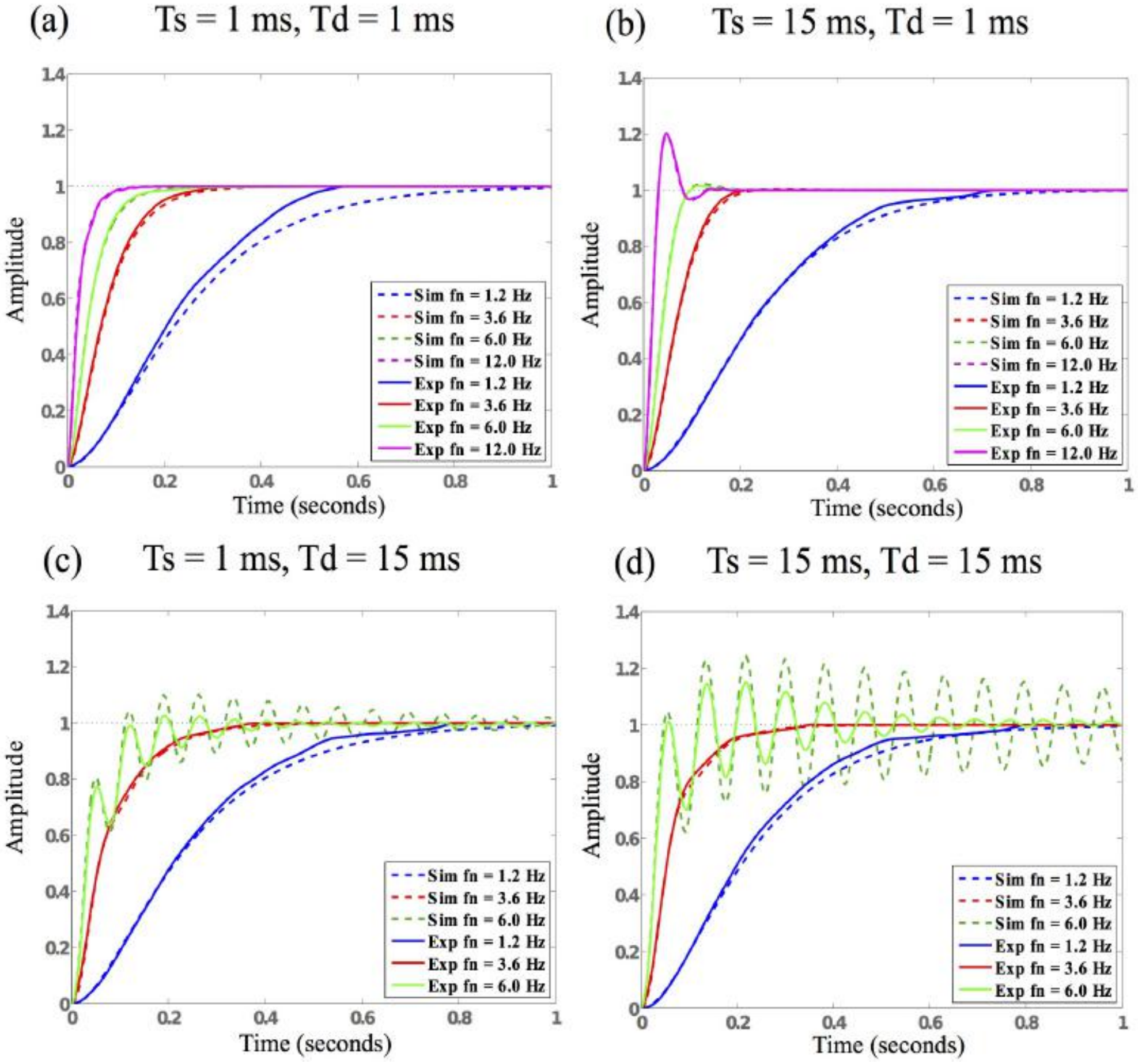

Figure 6.8: Step response experiment with the distributed controller. Subfigures (a) through (d) show various implementations on our linear rigid actuator corresponding to the simulations depicted on Fig. 6.3. Overlapped with the data plots, simulated replicas of the experiments are also shown to validate the proposed models. The experiments confirm the higher sensitivity of the actuator to damping than to stiffness delays.

### 6.3.1 Step Response

First, a test is performed on the actuator evaluating the response to a step input on its position. The results are shown in the bottom part of Fig. 6.8 which shows and compares the performance of the real actuator versus the sim-ulated closed loop controller. A step input comprising desired displacements between 0.131 m and 0.135 m of physical pushrod length is sent to the actuator. The main reason for constraining the experiment to a small displacement is to prevent current saturation of the motor driver. With a very high stiffness, it is easy to reach the 30 A limit for step responses. If current is saturated, then the experiment will deviate from the simulation. The step response is normalized between 0 and 1 for clarity. Various tests are performed for the same reference input with varying time delays in either or both the stiffness and damping loops. The four combinations of results are shown in the figure with delay values of 1 ms or 15 ms.

The first thing to notice is that there is a good correlation between the real and the simulated results both for smooth and oscillatory behaviors. Small discrepancies are attributed to unmodelled static friction and the effect of unmodelled dynamics. More importantly, the experiment confirms the anticipated discrepancy in delay sensitivity between the stiffness and damping loops. Large servo delays on the stiffness servo, corresponding to subfigures (a) and (b) have small effects on the step response. On the other hand, large servo delays on the damping servo, corresponding to subfigures (c) and (d), strongly affect the stability of the controller. In fact, for (c) and (d) the results corresponding to $f_n = 12$ Hz are omitted due to the actuator quickly becoming out of control. In contrast, the experiment in (b) can tolerate such high gains despite the large stiffness delay.

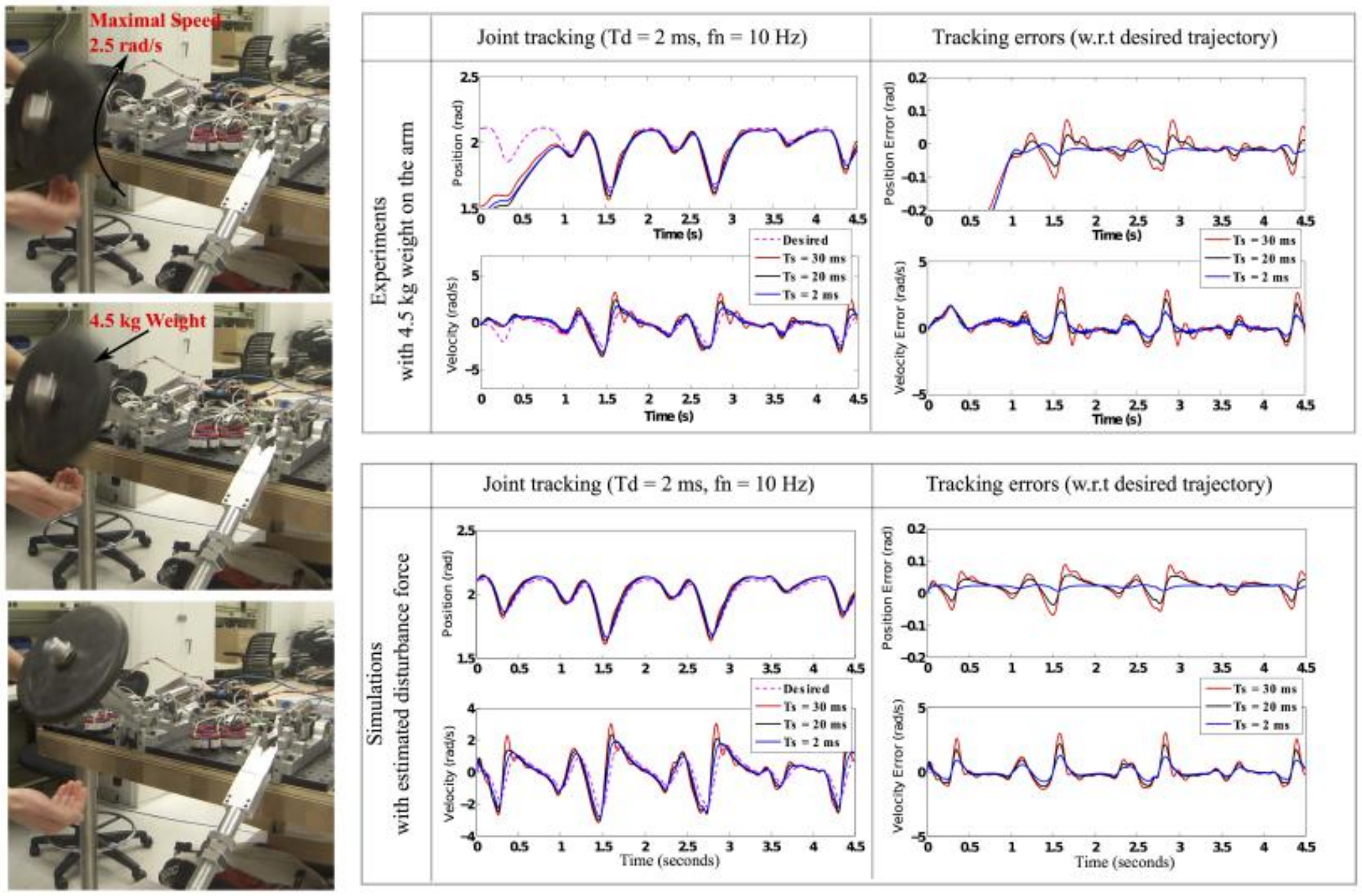

Figure 6.9: Trajectory tracking experiment under load disturbances with the distributed controller. The figures on the top row show snapshots of the testbed with an unmodelled load of 4.5 kg. The plots show trajectory tracking performance for a small damping delay, $T_d = 2$ ms and various stiffness delays ranging from $T_s = 2$ ms to $T_s = 30$ ms. Trajectory tracking errors on the bottom left remain relatively small despite the large stiffness delays, confirming the advantages of implementing damping feedback in a fast computational processes. A simulated experiment is also shown on the bottom right confirming good correlation between the real and simulated performance.

### 6.3.2 Trajectory Tracking with An Uncertain Load

Performance limits are explored at their fullest in the test shown in Fig. 6.9. Here, an unmodelled weight of 4.5kg is attached to the load arm which is also connected to the pushrod actuator through a pivot. The weight is unmodelled and therefore constitutes a disturbance. By estimation, the total disturbance torque that the controller has to deal with is $F_d = 16.84$ Nm. A trajectory with output angle variations within $[86°, 126°]$ is designed to test the controller's performance under the load disturbance. This trajectory is inspired by that of a fast bipedal locomotion knee joint motion in Chapter 3, with angular velocities varying between $\pm 2.5$ rad/s.

Table 6.2: Root mean square tracking errors

| Stiffness Delay $T_s$ (ms) | Experiment | | Simulation | |
|---|---|---|---|---|
| | Position Err (rad) | Velocity Err (rad/s) | Position Err (rad) | Velocity Err (rad/s) |
| 2 | 0.0182 | 0.3866 | 0.0204 | 0.3970 |
| 20 | 0.0247 | 0.6366 | 0.0289 | 0.6332 |
| 30 | 0.0360 | 0.8753 | 0.0386 | 0.8178 |

This experiment tests the tracking performance under the load disturbance on both the real actuator and also on a numerical simulation of the controller model depicted in Fig. 6.1. Disturbance forces for the numerical simulation are applied based on the position of the arm and considering only gravitational effects. Judging from the visualization of the errors in that figure and the root mean square of the errors depicted in Table 6.2, there is a good correlation between the real experiment and the simulation values for both the joint positions and angular velocities.

Once more, the test confirms the predicted robustness to stiffness delays that was studied in previous sections. Increasing 10 times $T_s$ from 2 ms to 20 ms only increases the root mean square (RMS) joint position and angular velocity errors by less than two fold as shown in Table 6.2. By using high gains, the controller ensures that joint tracking is accurate despite the large load disturbances. As shown in the previous table, the maximum RMS error for the tracked joint position is 0.0182 rad $\approx 1^\circ$ for stiffness delays of $T_s = 2$ ms and 0.036 rad $\approx 2^\circ$ for delays $T_s = 30$ ms.

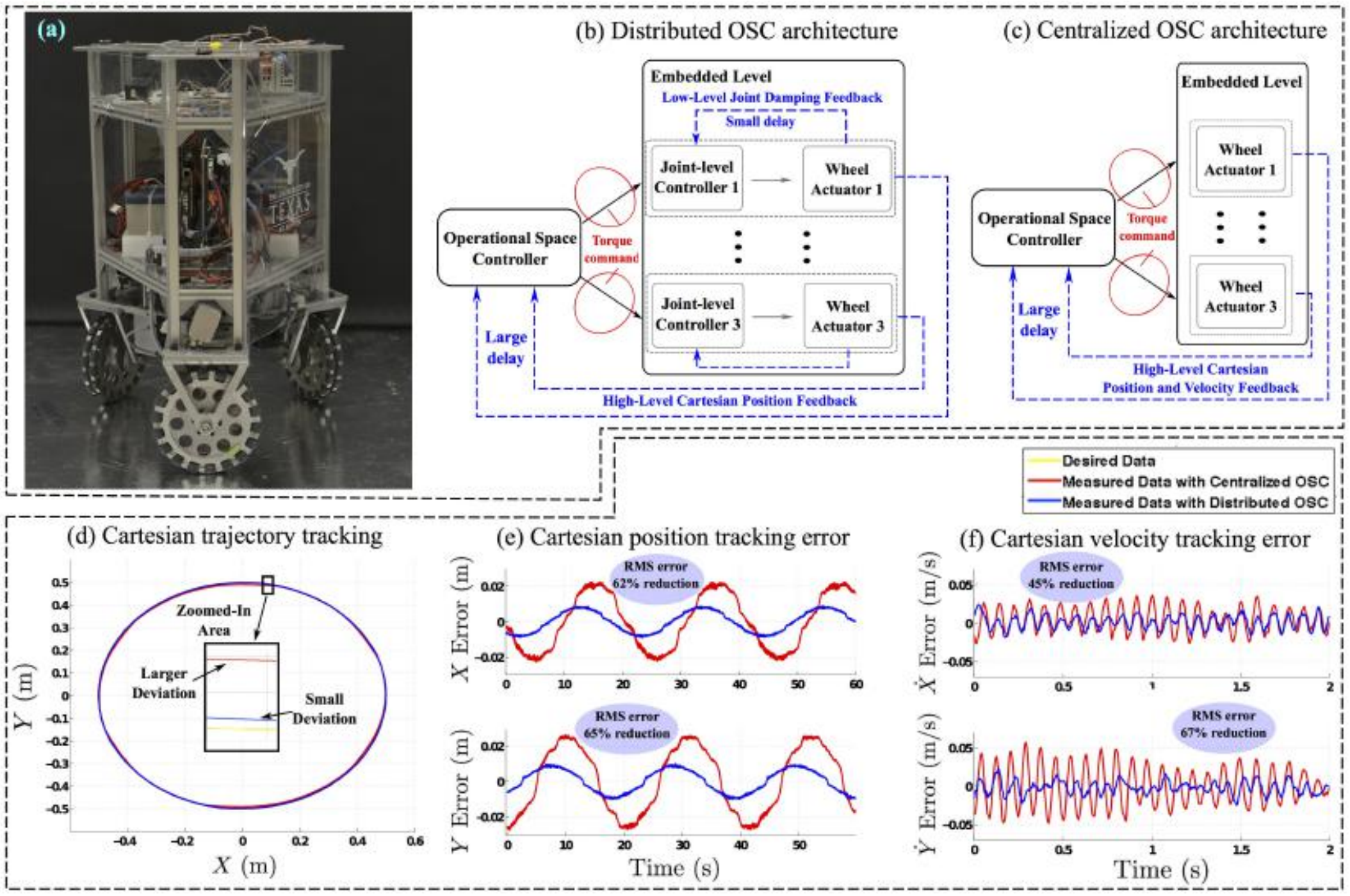

Figure 6.10: Omnidirectional mobile base with distributed and centralized OSC controllers. As a proof of concept we leverage the proposed distributed architectures to our robotic mobile base demonstrating significant improvements on tracking and stability.

### 6.3.3 Distributed Operational Space Control of A Mobile Robot

As a concept proof of the proposed distributed architecture on a multi-axis mobile platform, a Cartesian space feedback Operational Space Controller (Khatib, 1987b) is implemented on an omnidirectional mobile base. The original feedback controller was implemented as a centralized process (Kim et al., 2013) with no distributed topology at that time. The mobile base is equipped with a centralized PC computer running Linux with the RTAI real-time kernel. The PC connects with three actuator processors embedded next to the wheel drivetrains via EtherCat serial communications. The embedded processors do not talk to each other. The high level centralized PC on our robot, has a roundtrip latency to the actuators of 7ms due to process and bus communications, while the low level embedded processors have a servo rate of 0.5ms. Notice that 7ms is considered too slow for stiff feedback control. To accentuate even further the effect of feedback delay on the centralized PC, an additional 15ms delay is artificially introduced by using a data buffer. Thus, the high level controller has a total of 22 ms feedback delay.

An Operational Space Controller (OSC) is implemented in the mobile base using two different architectures. First, the controller is implemented as a centralized process, which will be called COSC, with all feedback processes taking place in the slow centralized processor and none in the embedded processors. In this case, the maximum stiffness gains should be severely limited due to the effect of the large latencies. Second a distributed controller architecture is implemented inspired by the one proposed in Fig. 6.1 but adapted to a distributed Operational Space Controller, which will be called DOSC. In this version, the Cartesian stiffness feedback servo is implemented in the centralized PC in the same way than in COSC, but the Cartesian damping feedback servo is removed from the centralized process. Instead, our study implements damping feedback in joint space (i.e. proportional to the wheel velocities) on the embedded processors. A conceptual drawing of these architectures is shown in Fig. 6.10. The metric used for performance comparison is based on the maximum achievable Cartesian stiffness feedback gains, and the Cartesian position and velocity tracking errors.

To implement the Cartesian stiffness feedback processes in both architectures, the Cartesian positions and orientations of the mobile base on the ground are computed using wheel odometry (more details are discussed in (Kim et al., 2013)). To achieve the highest stable stiffness gains, the following procedure is followed: (i) first, Cartesian stiffness gains are adjusted to zero while the damping gains in either Cartesian space (COSC) or joint space (DOSC) – depending on the controller architecture – are increased until the base starts vibrating; (ii) the Cartesian stiffness gains, on either architecture, are increased until the base starts vibrating or oscillating; (iii) a desired Cartesian circular trajectory is commanded to the base and the position and velocity tracking performance are recorded.

Based on these experiments, DOSC was able to attain a maximum Cartesian stiffness gain of 140 N/(m kg) compared to 30 N/(m kg) for COSC. This result means that the proposed distributed control architecture allowed the Cartesian feedback process to increase the Cartesian stiffness gain ($K_x$ in Fig. 6.11) by 4.7 times with respect to the centralized controller implementation. In terms of tracking performance, the results are shown in Fig. 6.10. Both Cartesian position and velocity tracking in DOSC are significantly more accurate. The proposed distributed architecture reduces Cartesian position root mean error between 62% and 65% while the Cartesian velocity root mean

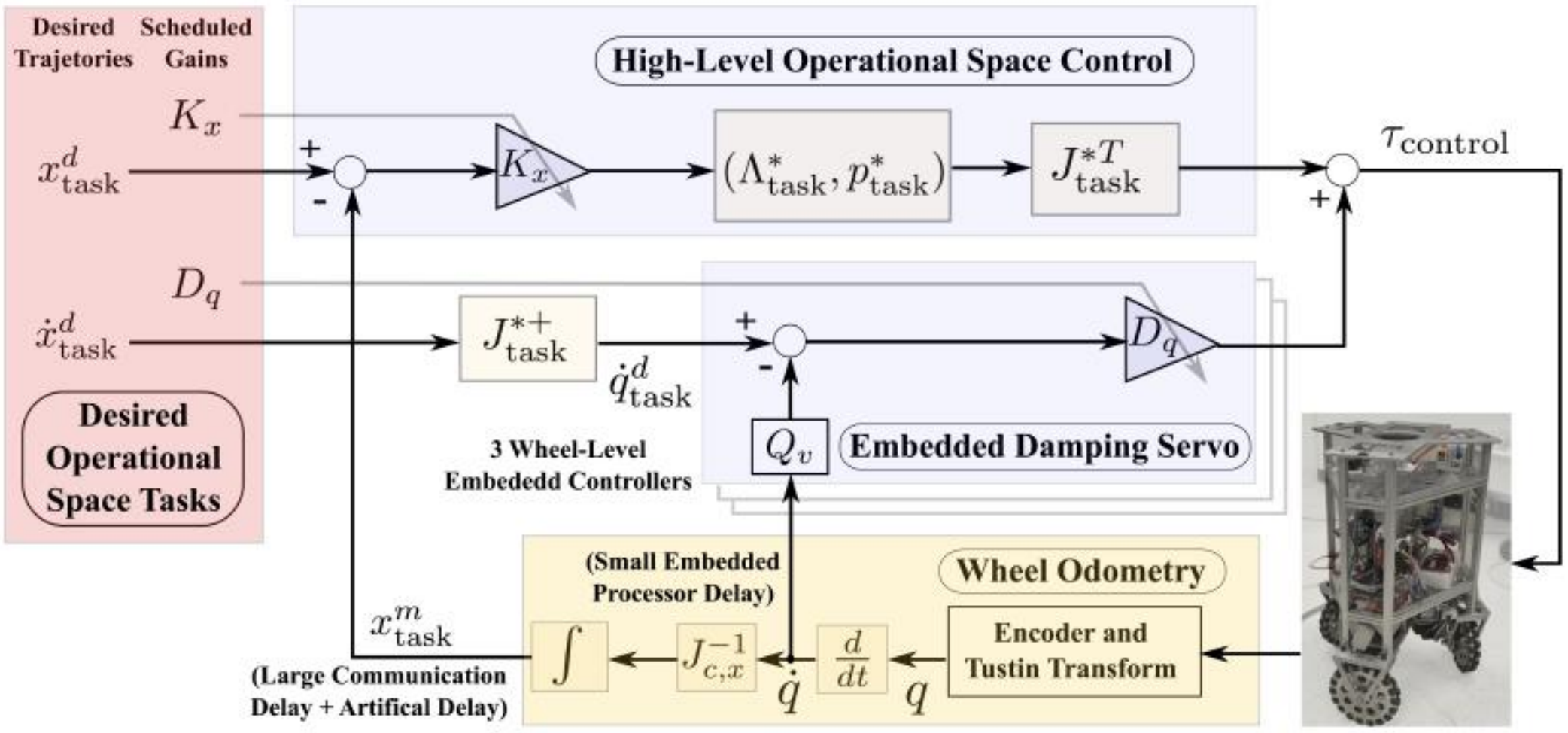

Figure 6.11: Detailed distributed Operational Space Control structure. The figure above illustrates details of the distributed Operational Space Controller used for the mobile base tracking experiment. $\Lambda^*_{\text{task}}$ and $p^*_{\text{task}}$ are the operational space inertia matrix and gravity based forces, respectively. $J^*_{\text{task}}$ is a contact consistent task Jacobin. More details about these matrices and vectors can be found in (Khatib, 1987b; Sentis, 2007).

error decreases between 45% and 67%.

The main contribution for this experiment lies in implementing operational space control in a distributed fashion and based on the observations performed on the previously simplified distributed controller. While the high-level operational space stiffness feedback loop suffers from large delays due to communication latencies and artificial delays (added by a data buffer), the embedded-level damping loop increases system stability. As a result, the proposed distributed architecture enables to achieve higher Cartesian stiffness gains $K_x$ for better tracking accuracy.

## 6.4 Discussions and Conclusions

The motivation for this work has been to study the stability and performance of distributed controllers where stiffness and damping servos are implemented in distinct processors. These types of controllers will become important as computation and communications become increasingly more complex in human-centered robotic systems. The focus has been first on studying the physical performance of a simple distributed controller. Simplifying the controller allows us to explore the physical effects of time delays in greater detail. Then the proposed architecture has been leveraged to a mobile base system as a proof of concept. Our focus on this work has been on high impedance behaviors. This focus contrasts with our previous work on low impedance control (Paine et al., 2014). However, both high and low impedance behaviors are important in human-centered robotics. For instance, high impedance behaviors are necessary to attain good position tracking in the presence of unmodelled actuator dynamics or external disturbances.

Using the phase margin frequency technique allowed us to reveal the severe effects of delays on the damping loop and appreciate the discrepancy with respect to the stiffness servo behavior. However, to reveal the physical reasons for this discrepancy, an in-depth mathematical analysis is performed based on phase margin sensitivity to time delays. This analysis allowed us to derive the physical condition for the discrepancy between delays. Further analysis revealed that the previous condition is met for high impedance controllers, in which the damping feedback gain is significantly larger than the passive damping actuator value. To confirm the observations and analytical derivations, we perform two experiments by using an actuator and a mobile base. In particular, the results have shown that decoupling stiffness servos to slower centralized processes does not significantly decrease system stability. As such, stiffness servo can be used to implement Operational Space Controllers which require centralized information such as robot models and external sensors.

Our next step is to develop a similar study for controllers incorporating an inner torque loop, such as those used for series elastic actuators (Paine et al., 2014). For this type of actuators our interest is to explore both high and low impedance capabilities (i.e., impedance range) under latencies and filtering and using distributed control concepts similar to those explored in this chapter. The challenge is that system dynamics become high order instead of second order and more advanced gain selection rules need to be designed for both impedance and torque gains. This gain design criterion will be the focus of next chapter.

# Chapter 7

# Impedance Control and Performance Measure of Series Elastic Actuators

The paradigm shift from rigid actuators to the series elastic actuators (see Fig. 7.1) opens new potentials to achieve a broad range of impedance performance. At the same time, new challenges inherently appear due to the high complexity of SEA dynamics. Advanced methods have been persistently called for the SEA control and performance evaluation. This chapter focuses on the impedance controller design and performance characterization of series elastic actuators.

## 7.1 SEA Modeling

In this section, we model a series elastic actuator with two nested control loops: an inner torque loop and an outer impedance loop. First, let us consider the SEA dynamics. As shown in Fig. 7.2, the spring force $\tau_k$ is

$$\tau_k = k(q_m - q_j). \tag{7.1}$$

---

This chapter incorporates the results from the following publications: (Zhao, Paine, and Sentis, 2014b; Zhao, Paine, Jorgensen, and Sentis, 2016e).

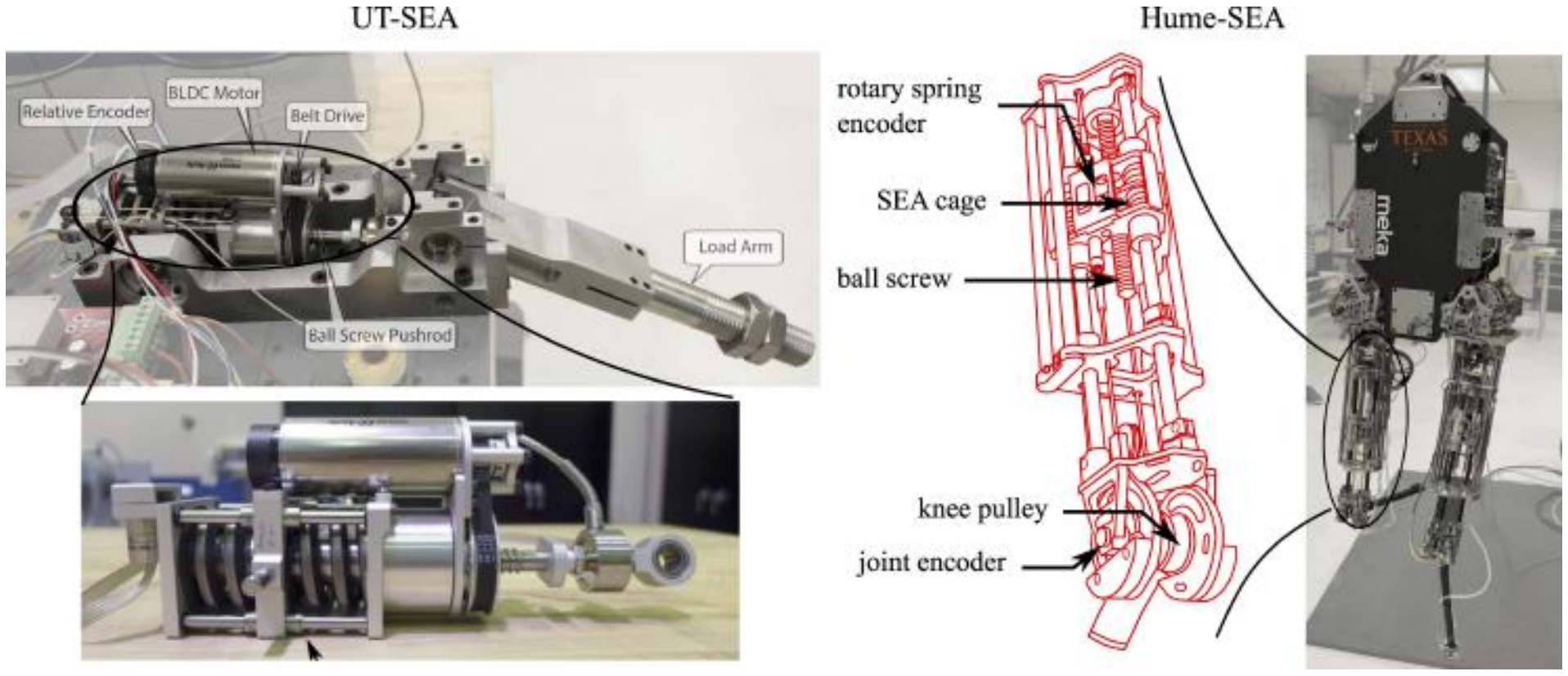

Figure 7.1: UT and Hume series elastic actuators. The left subfigures show a high-performance linear UT-SEA while the right ones show a SEA assembled in our biped robot.

For the joint side, we assume external force $\tau_d = 0$ (i.e., spring torque is equal to load torque) and have

$$\tau_k = I_j \ddot{q}_j + b_j \dot{q}_j. \tag{7.2}$$

Here we only model the viscous friction and more accurate Coulomb friction modeling is for future work. Thus, the load plant $P_L(s)$ is

$$P_L(s) = \frac{q_j(s)}{\tau_k(s)} = \frac{1}{I_j s^2 + b_j s}. \tag{7.3}$$

By Eqs. (7.1) and (7.2), we have the transfer function from motor position $q_m$ to joint position $q_j$.

$$\frac{q_j(s)}{q_m(s)} = \frac{k}{I_j s^2 + b_j s + k}. \tag{7.4}$$

Motor torque $\tau_m$ is represented as

$$\tau_m = I_m \ddot{q}_m + b_m \dot{q}_m + k(q_m - q_j) \tag{7.5}$$

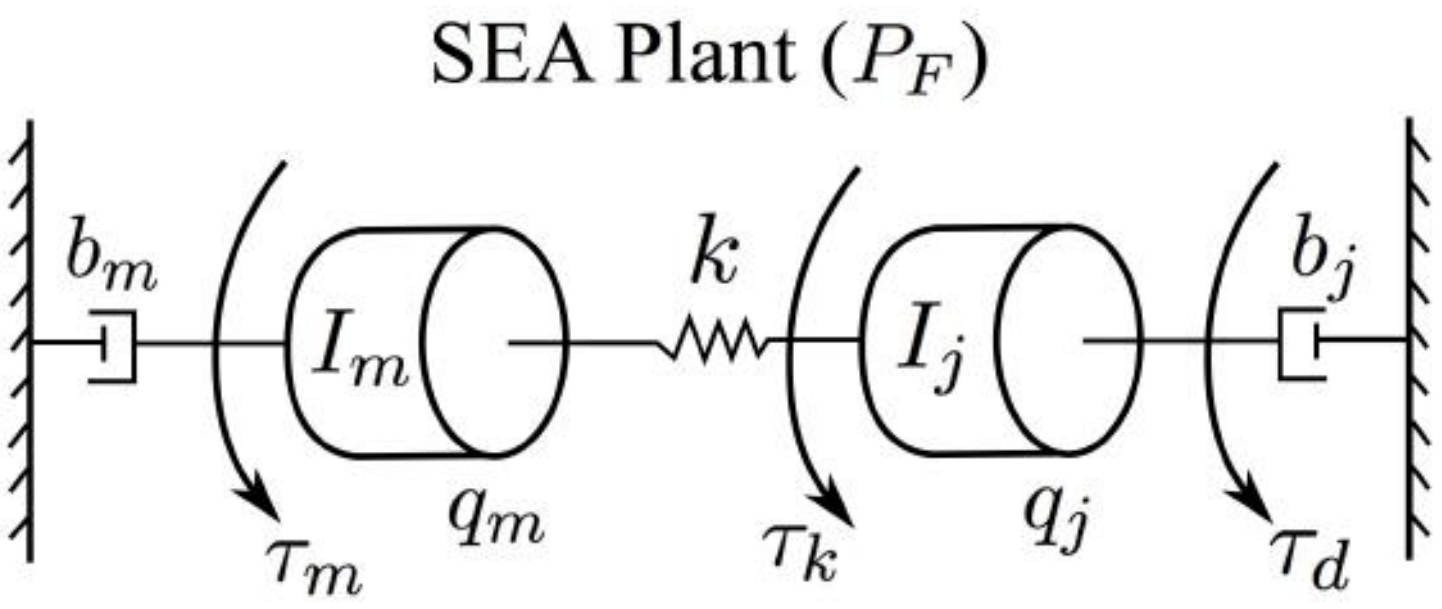

Figure 7.2: SEA model scheme. $q_m, q_j$ represents motor and joint positions, respectively. $k$ is the spring stiffness. $I_j, I_m$ is joint and motor inertias, respectively. $b_j, b_m$ are joint and motor damping coefficients, respectively. $I_m$ and $b_m$ are mapped to the joint coordinates by multiplying by the square of the gear reduction.

Combining this equation with Eq. (7.4) and defining $\Delta q = q_m - q_j$, we have the relationship between spring deflection and motor angle

$$r(s) = \frac{\Delta q(s)}{q_m(s)} = \frac{I_j s^2 + b_j s}{I_j s^2 + b_j s + k}. \tag{7.6}$$

Combining Eq. (7.1), spring force can be calculated

$$\tau_k(s) = k\Delta q(s) = kr(s)q_m(s). \tag{7.7}$$

Since the motor current $i_m$ and motor torque $\tau_m$ is related by $\tau_m(s)/i_m(s) = \beta = \eta N k_\tau$, with drivetrain efficiency $\eta$ (constant for simplicity, ignore dynamic model of drivetrain losses), gear speed reduction $N$ and motor torque constant $k_\tau$, the control plant $P_F(s)$ is

$$P_F(s) = \frac{\tau_k(s)}{i_m(s)} = \frac{\beta r(s) k}{I_m s^2 + b_m s + r(s)k}. \tag{7.8}$$

Then, from Fig. 7.3, the closed-loop torque control plant $P_C$ is

$$P_C(s) = \frac{\tau_k(s)}{\tau_{\text{des}}(s)} = \frac{P_F(\beta^{-1} + C)}{1 + P_F C e^{-T_\tau s}}. \tag{7.9}$$

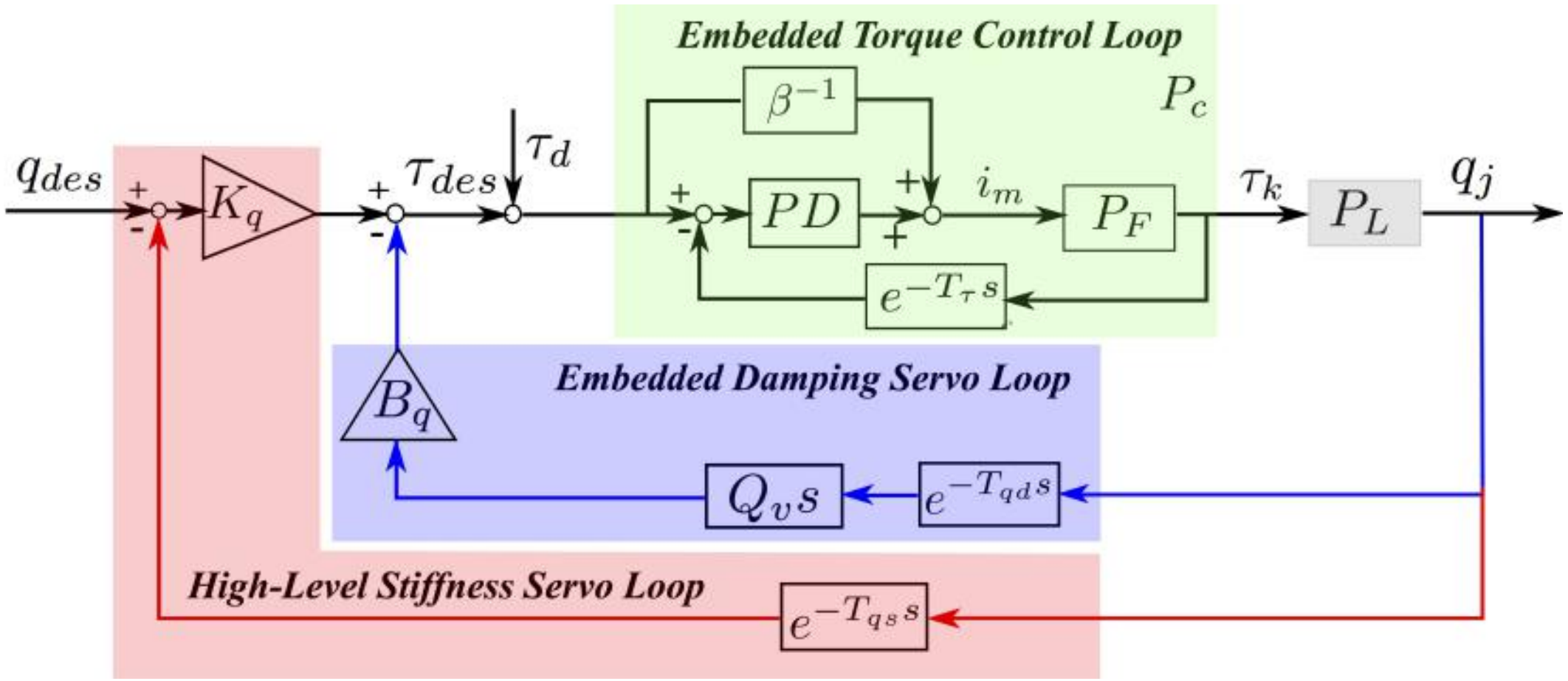

Figure 7.3: Joint-level SEA control block diagram. The inner torque loops have a proportional and derivative feedback and a feedforward loop with a mapping scaler $\beta^{-1}$. The outer impedance loops have stiffness and damping feedback. Delays in each loop are labeled as $e^{-Ts}$. First order low-pass filter is applied to both velocity and torque derivative. The motor has a current input $i_m$. $\tau_k$ is the spring torque. $P_C$ represents the embedded torque control module.

The torque feedback loop has a feedback delay $e^{-T_\tau s}$ and a PD compensator $C = K_\tau + B_\tau Q_{\tau d} s$, where $Q_{\tau d}$ is the filtering for torque derivative.

$$Q_{\tau d} = \frac{2\pi f_{\tau d}}{s + 2\pi f_{\tau d}}, \tag{7.10}$$

where $f_{\tau d}$ is the filter cut-off frequency. A feedforward term is designed to convert from desired torque $\tau_{\text{des}}$ to motor current $i_m$ as shown in Fig. 7.3. By Eqs. (7.3) and (7.9), we have the following transfer function

$$\frac{q_j(s)}{\tau_{\text{des}}(s)} = P_L P_C = \frac{P_F(\beta^{-1} + C)}{(1 + P_F C e^{-T_\tau s})(I_j s^2 + b_j s)}. \tag{7.11}$$

For the impedance feedback, we have the form as below

$$\tau_{\text{des}}(s) = K_q(q_{\text{des}} - e^{-T_{qs} s} q_j) - B_q e^{-T_{qd} s} Q_{qd} s q_j, \tag{7.12}$$

where $e^{-T_{qs}s}$ and $e^{-T_{qd}s}$ represent the feedback delays in stiffness and damping loops, respectively. The first order low-pass filter $Q_{qd}$ for joint velocity has the same form as Eq. (7.10) with a cut-off frequency $f_{qd}$. Alternatively, we can also send desired joint velocity to the embedded damping loop. In that case, another zero will show up in the numerator of Eq. (7.13). Since a zero only changes transient dynamics, it does not affect system stability. Using $P_L$ and $P_C$ in Eqs. (7.3) and (7.9), we can derive the system closed-loop transfer function $P_{CL}$ from $q_{\text{des}}$ to $q_j$,

$$\begin{aligned} P_{CL}(s) = \frac{q_j(s)}{q_{\text{des}}(s)} &= \frac{K_q P_C P_L}{1 + P_C P_L (e^{-T_{qd}s} B_q Q_{qd} s + e^{-T_{qs}s} K_q)} \\ &= \frac{K_q(1 + \beta K_\tau + \beta B_\tau Q_{\tau d} s)}{\sum_{i=0}^{4} D_i s^i}, \end{aligned} \tag{7.13}$$

with the coefficients

$$D_4 = I_m I_j / k,$$

$$D_3 = (I_j b_m + I_m b_j)/k + I_j \beta B_\tau Q_{\tau d} e^{-T_\tau s},$$

$$D_2 = I_j(1 + e^{-T_\tau s}\beta K_\tau) + I_m + b_j \beta B_\tau Q_{\tau d} e^{-T_\tau s} + \beta B_\tau B_q e^{-T_{qd}s} Q_{qd} Q_{\tau d} + b_j b_m / k,$$

$$D_1 = b_j(1 + e^{-T_\tau s}\beta K_\tau) + b_m + \beta B_\tau Q_{\tau d} K_q e^{-T_{qs}s} + e^{-T_{qd}s}(1 + \beta K_\tau) B_q Q_{qd},$$

$$D_0 = e^{-T_{qs}s}(1 + \beta K_\tau) K_q.$$

This transfer function is a sixth order system since the low pass filters $Q_{qd}$ and $Q_{\tau d}$ increase the order by two in total. Here we formulate it in fourth order form for convenience. An important issue to notice is that there is a zero in the numerator of Eq. (7.13). This zero is caused by the torque derivative term. This induced zero will shorten the rise time but also cause an overshoot in step response. However, it does not influence system stability, which is determined by the characteristic polynomial in the denominator. More advanced design of zero placement is for future work.

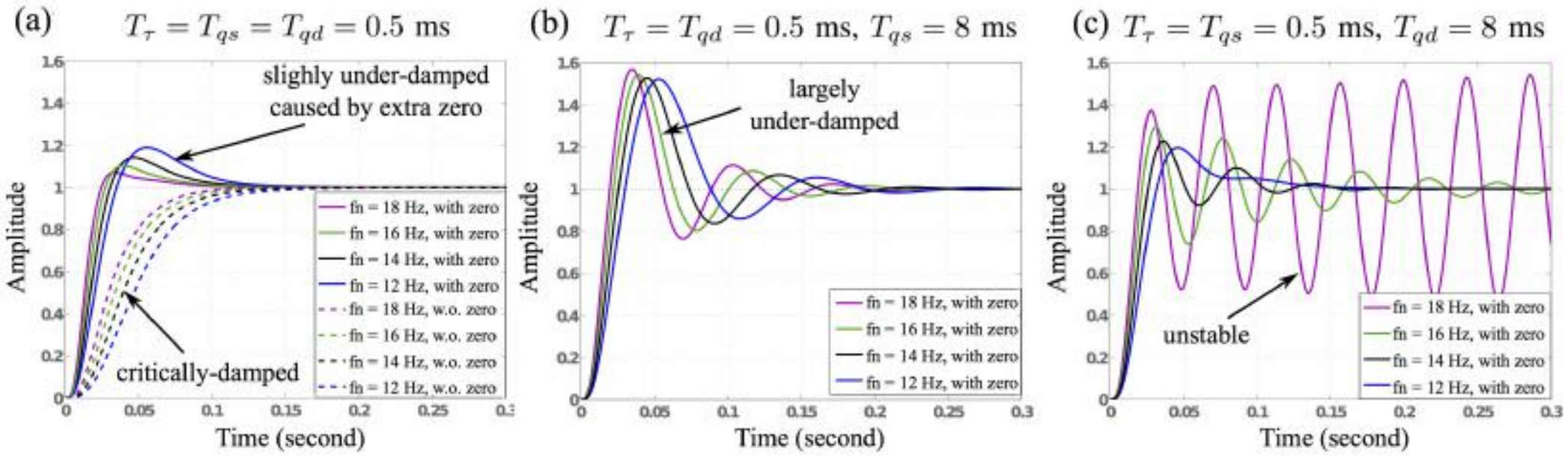

Figure 7.4: SEA step response with feedback delays. These subfigures demonstrate that large impedance feedback delays deteriorate the step response performance. Comparing subfigures (b) and (c), we observe that system stability is more sensitive to damping feedback delays than its stiffness counterpart.

## 7.2 SEA Gain Selection

The closed-loop transfer function derived in Eq. (7.13) is complex due to the cascaded impedance and torque feedback loops. This complexity makes the SEA controller design challenging. In this section, we propose a critically-damped criterion to design optimal SEA controller gains.

### 7.2.1 Critically-damped Gain Design Criterion

Impedance control gains of rigid actuators have been designed based on the critically-damped criterion of the second-order system in Chapter 6. As for high-order systems such as SEAs, such a critically-damped criterion is still missing. However, high-order systems usually can be approximated by the product of first- and second-order systems. To this end, we represent the fourth-order system in Eq. (7.13) (the feedback delays and filtering are ignored for problem tractability) by two second-order systems in multiplification, as presented in (Petit and Albu-Schaffer, 2011)

$$(s^2 + 2\zeta_1\omega_1 s + \omega_1^2)(s^2 + 2\zeta_2\omega_2 s + \omega_2^2), \tag{7.14}$$

which has four design parameters $\omega_1, \omega_2, \zeta_1, \zeta_2$. They are used to design four gains $K_q, B_q, K_\tau$, and $B_\tau$. First, we set $\zeta_1 = \zeta_2 = 1$ in Eq. (7.14) to obtain critically-damped performance. Second, we assume $\omega_2 = \omega_1$ for simplicity. An optimal pole placement design is left for future work. Let us define a natural frequency $f_n$ of Eq. (7.14) as

$$\omega_1 = \omega_2 \triangleq \omega_n = 2\pi f_n. \tag{7.15}$$

By comparing the denominators of Eqs. (7.13) and (7.14), we obtain the following nonlinear gain design criterion equations.

$$\begin{aligned}
&\frac{I_j b_m + I_m b_j + I_j \beta B_\tau k}{I_m I_j} = 4\omega_n, \\
&\frac{k(I_j(1 + \beta K_\tau) + I_m + \beta B_\tau(b_j + B_q)) + b_j b_m}{I_m I_j} = 6\omega_n^2, \\
&\frac{k(b_j + B_q)(1 + \beta K_\tau) + k(b_m + \beta B_\tau K_q)}{I_m I_j} = 4\omega_n^3, \\
&\frac{(1 + \beta K_\tau)k K_q}{I_m I_j} = \omega_n^4.
\end{aligned} \tag{7.16}$$

These four highly-coupled equations can be solved by Matlab's fsolve() function. Let us show an example as follows.

Table 7.1: Critically-damped SEA controller gains

| Frequency (Hz) | Impedance gains (Nm/rad, Nms/rad) | Torque gains (A/Nm, As/Nm) | Phase margin |
|---|---|---|---|
| $f_n = 12$ | $K_q = 65$<br>$B_q = 0.46$ | $K_\tau = 1.18$<br>$B_\tau = 0.057$ | 49.1° |
| $f_n = 14$ | $K_q = 83$<br>$B_q = 0.76$ | $K_\tau = 1.80$<br>$B_\tau = 0.067$ | 47.0° |
| $f_n = 16$ | $K_q = 103$<br>$B_q = 1.02$ | $K_\tau = 2.56$<br>$B_\tau = 0.077$ | 43.6° |
| $f_n = 18$ | $K_q = 124$<br>$B_q = 1.26$ | $K_\tau = 3.45$<br>$B_\tau = 0.087$ | 39.9° |
| $f_n = 20$ | $K_q = 148$<br>$B_q = 1.49$ | $K_\tau = 4.48$<br>$B_\tau = 0.097$ | 36.4° |

**Example 7.1.** *To validate this criterion, we test five natural frequency cases. Filter cut-off frequencies* $f_{vd} = 50$ *Hz,* $f_{\tau d} = 100$ *Hz and feedback delays* $T_\tau = T_{qs} = T_{qd} = 0.5$ *ms. The results are shown in Table 7.1. Increasing* $f_n$ *will lead to a uniform increase of all four gains. This property meets our expectation that increasing torque (or impedance) gains has the resulting effect of torque (or impedance) bandwidth increase. Noteworthily, we use the open-loop transfer function for the phase margin calculation.*

Given this gain design criterion, we further study the effect of feedback delays on the SEA stability. Since torque feedback are the inner loops, it always suffers a smaller delay, for instance $T_\tau = 0.5$ ms, than the outer impedance loops. As the step responses shown in Fig. 7.4, we obtain the same conclusion as that in Chapter 6. Namely, SEA stability is more sensitive to the damping feedback delay than its stiffness counterpart. This motivates us to implement the impedance feedback loops in a distributed pattern as shown in Fig. 7.3.

Another thing to note in Fig. 7.4 (a) is that, the larger $f_n$ is, the larger overshoot shows up, which seems counterintuitive. However, if observed closely, the solid magenta line with the largest $f_n$ already shows distortion and its phase margin value is 36.4°, smaller than other three cases. When the natural frequency increases to 30 Hz, obvious distortion shows up. Also, to analyze the effect of zero in Eq. (7.13), we simulate step responses without this zero, shown in dashed lines of subfigure (a). By comparison, we can observe this extra zero induces an overshoot.

### 7.2.2 Trade-Off between Impedance Control and Torque Control

During gain tunings of our SEA-equipped bipedal robot Hume (Kim et al., 2016a), which has a similar SEA control architecture in Fig. 7.3, we observed the following phenomenon: if we increase torque controller gains or decrease impedance controller gains, the robot became vulnerable to instability. To reason about this observation, we propose a gain scale definition as follows

**Definition 7.1 (SEA gain scale).** *The gain scale of a SEA cascaded controller is a scaling parameter $GS$ between adjusted gains $(K_{i_a}, B_{i_a})$ and nominal gains $(K_{i_n}, B_{i_n})$, $i \in \{\tau, q\}$,*

$$GS = \frac{K_{\tau_a}}{K_{\tau_n}} = \frac{K_{q_n}}{K_{q_a}}, \quad GS = \frac{B_{\tau_a}}{B_{\tau_n}} = \frac{B_{q_n}}{B_{q_a}}, \tag{7.17}$$

*where the adjusted gains denote actual gains in use while the nominal gains denote reference ones designed by the critically-damped gain design criterion.*

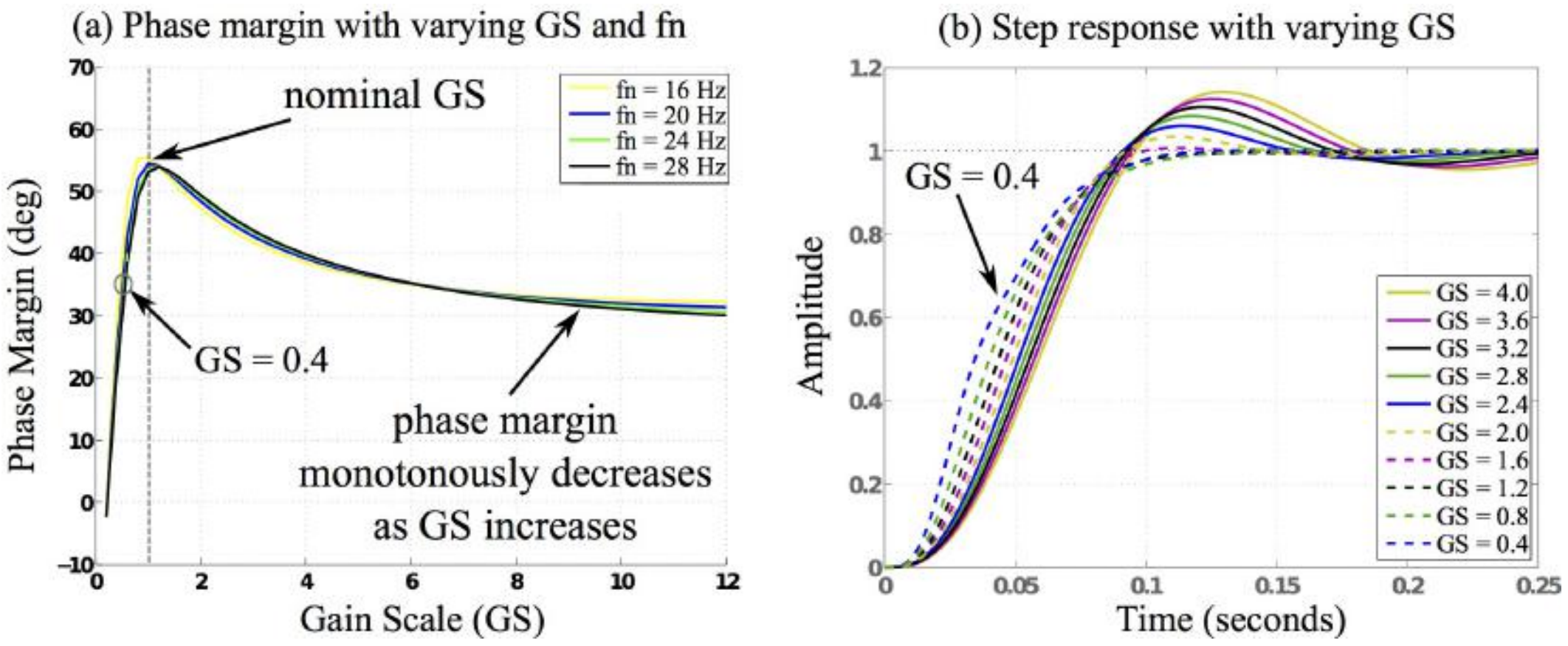

Figure 7.5: Optimality of the critically-damped gain design criterion. Subfigure (a) samples a variety of gain scales and natural frequencies. The optimal performance is achieved by the proposed critically-damped gain design criterion. Subfigure (b) shows: (i) larger overshoot but slow rise time when $GS > 1$; (ii) over-damped response with distortions when $GS < 1$.

For example, the controller gains corresponding to $f_n = 14$ Hz in Table 7.1 are a set of nominal gains. By Eq. (7.17), we have the following equation

$$K_{\tau_a} \cdot K_{q_a} = K_{\tau_n} \cdot K_{q_n}, \quad B_{\tau_a} \cdot B_{q_a} = B_{\tau_n} \cdot B_{q_n},$$

which maintains the same multiplication of nested proportional (or derivative) torque and impedance gains for either normal or adjusted conditions.

The work in (Focchi et al., 2016) obtained a similar observation that enlarging the inner loop controller bandwidth reduces the range of stable impedance control gains. There is a trade-off between a large torque bandwidth for accurate torque tracking and a low torque bandwidth for larger achievable impedance gain range. Nevertheless, their experimental validations are not persuasive since they do not decrease impedance gains when raising torque gains. Namely, the multiplication of two cascaded gains goes up. Ob-viously, the system stability will decrease. Our method instead maintains the same multiplication. Fig. 7.5 (a) shows the sampling results of different gain scales $GS$. As observed, $GS > 1$ means increasing torque gains while decreasing impedance gains. This deteriorates system stability and causes larger oscillatory step response as shown in Fig. 7.5(b). On the other hand, when $GS < 1$, decreasing torque gains while increasing impedance gains also decreases phase margin. For instance, $GS = 0.4$ corresponds to a $34°$ phase margin as shown in subfigure (a), and a distortion shows up in its step response of subfigure (b). We ignore delays and filtering to focus on the gain scale analysis. These tests validate the optimal performance (i.e., maximized phase-margin) by using our proposed critically-damped gain design criterion (i.e., $GS = 1$). In Section 7.4, we will use $GS$ to quantify the upper and lower impedance boundaries which are used to define a new impedance measure "Z-region". Impedance control is widely used for dynamic interaction between a robot and its physically contact environment (Hogan, 1985). In the next section, we will study the SEA impedance performance within a variety of frequency ranges, and the effects induced by feedback delays and filtering.

## 7.3 SEA Impedance Frequency Analysis

Given the SEA controller diagram in Fig. 7.3, we first derive the SEA impedance transfer function in this section, and analyze frequency characteristics over a wide range by considering delays and filtering. Note that, a higher natural frequency $f_n$ corresponds to a higher impedance.

### 7.3.1 Impedance Transfer Function

We define the SEA impedance transfer function via a joint velocity $\dot{q}_j$ input and a joint torque $\tau_j$ output. Based on zero desired joint position $q_{\text{des}}$, the SEA impedance $Z(s) = \tau_j(s)/(-sq_j(s))$ is formulated as follows

$$Z(s) = \frac{\tau_j(s)}{-sq_j(s)} = \frac{\sum_{i=0}^{4} N_{zi}s^i}{\sum_{i=0}^{5} D_{zi}s^i}, \tag{7.18}$$

with the numerator coefficients,

$$\begin{aligned}
N_{z4} &= I_m T_{f\tau} T_{fv} \beta k, \\
N_{z3} &= \beta k(I_m(T_{f\tau} + T_{fv}) + T_{f\tau} T_{fv} b_m), \\
N_{z2} &= I_m \beta k + \beta k b_m (T_{f\tau} + T_{fv}) + k k_\tau (T_{f\tau} + \beta(B_\tau + K_\tau T_{f\tau})) \\
&\quad (B_q e^{-T_{qd}s} + K_q T_{fv} e^{-T_{qs}s}), \\
N_{z1} &= b_m \beta k + B_q k k_\tau (1 + K_\tau \beta) e^{-T_{qd}s} + K_q k k_\tau \\
&\quad (T_{fv} + T_{f\tau} + \beta(B_\tau + K_\tau(T_{f\tau} + T_{fv})))e^{-T_{qs}s}, \\
N_{z0} &= K_q k k_\tau e^{-T_{qs}s}(1 + K_\tau \beta),
\end{aligned}$$

and the denominator coefficients,

$$\begin{aligned}
D_{z5} &= I_m T_{f\tau} T_{fv} \beta, D_{z4} = I_m \beta (T_{fv} + T_{f\tau}) + T_{fv} T_{f\tau} \beta b_m, \\
D_{z3} &= \beta I_m + \beta b_m (T_{f\tau} + T_{fv}) + T_{fv} k \beta (T_{f\tau} + k_\tau (B_\tau + K_\tau T_{f\tau}) e^{-T_\tau s}), \\
D_{z2} &= \beta(b_m + T_{f\tau} k + k k_\tau (B_\tau + K_\tau T_{f\tau}) e^{-T_\tau s}) + T_{fv} \beta k (1 + K_\tau k_\tau e^{-T_\tau s}), \\
D_{z1} &= \beta k (1 + K_\tau k_\tau e^{-T_\tau s}), D_{z0} = 0.
\end{aligned}$$

Note that, $Z(s)$ does not incorporate the joint inertia $I_j$ and damping $b_j$ at load side since they belong to the interacted environment. We model feedback delays and filtering, which are still missing in the literature of SEA controller architectures with PD outer-impedance control and inner-torque control. The result in (Vallery et al., 2008) studies one type of the cascaded control structure with a PI torque feedback controller and an inner-most motor velocity feedback controller. Nevertheless, this study merely models small sampling rate delays instead of large communication bus delays. Also, damping gains are neglected in their study. Instead, the SEA transfer function in Eq. (7.18) is accurate with no approximations.

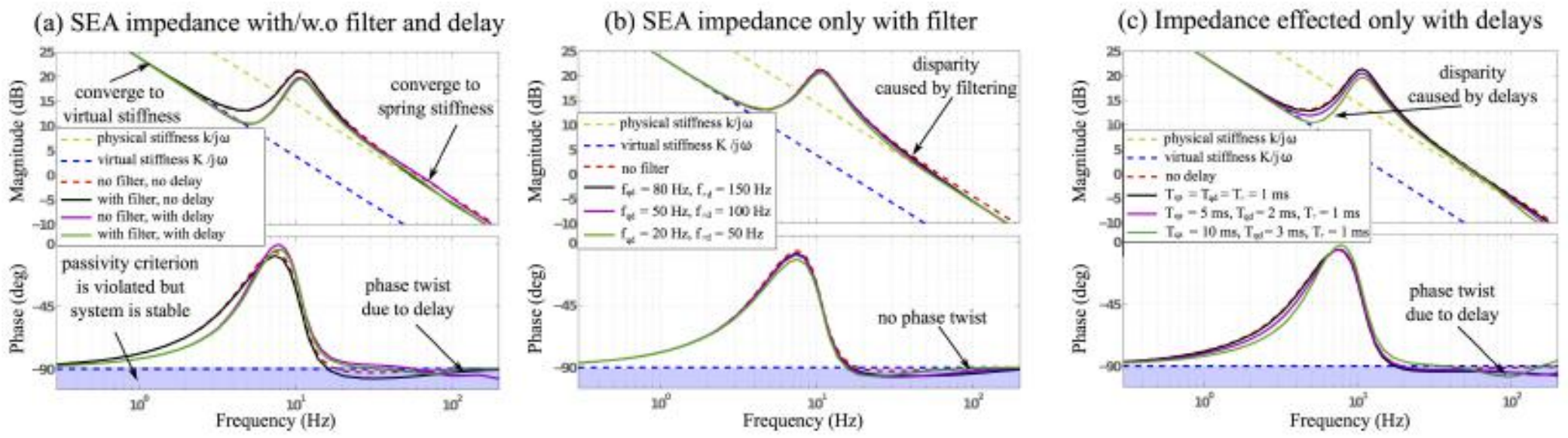

Figure 7.6: SEA impedance with feedback delays and filtering. At the low frequency range, SEA impedance converges to the virtual stiffness. A similar observation is shown in (Vallery et al., 2008). At the medium and high frequency ranges, it approaches to another impedance asymptote. (b) studies the filtering effect while (c) studies the feedback delay effect.

### 7.3.2 Effects of Feedback Delays and Filtering

The SEA impedance frequency characteristics are demonstrated in Fig. 7.6. The impedance of a physical spring $k$ and a virtual stiffness gain controller are shown by yellow and blue dashed lines, respectively. The ideal SEA impedance without delay and filtering is represented by a red dashed line. This simulations uses a natural frequency $f_n = 30$ Hz, corresponding to $K_q = 293.6$

Nm/rad, $B_q = 2.49$ Nms/rad, $K_\tau = 11.71$ A/Nm, $B_\tau = 0.146$ As/Nm. We analyze various scenarios either with or without feedback delays and filtering: (i) $Z_i(j\omega)$ is the ideal impedance without delays and filtering; (ii) $Z_f(j\omega)$ is the impedance only with filtering; (iii) $Z_d(j\omega)$ is the impedance only with delays; (iv) $Z_{fd}(j\omega)$ is the impedance with both delays and filtering. At the low frequency range, the SEA impedance converges to a virtual stiffness asymptote in all scenarios (when feedback delays are considered, we have $e^{-T_{qs}j\omega} \to 1, e^{-T_\tau j\omega} \to 1$ as $\omega \to 0$)

$$\lim_{\omega \to 0} Z_c(j\omega) = \lim_{\omega \to 0} \frac{N_{z0}}{j\omega \cdot D_{z1}} = \frac{K_q k_\tau (\beta^{-1} + K_\tau)}{j\omega \cdot (1 + K_\tau k_\tau)},$$

where $c \in \{i, f, d, fd\}$. The final expression keeps $j\omega$ to indicate a $-20$ dB/dec asymptote. The low frequency impedance $Z_c(j\omega)$ performs as a constant stiffness impedance $K_q/j\omega$ scaled by a constant $k_\tau(\beta^{-1} + K_\tau)/(1 + K_\tau k_\tau)$. This scaling applies to any PD-type cascaded impedance controller. Note that, $k_\tau \beta^{-1}$ is normally quite small. When $k_\tau K_\tau$ is large, $Z_c(j\omega)$ will approach to $K_q/j\omega$, i.e., a pure virtual spring.

As to the high frequency range, the impedance approaches to and sometimes twists around an asymptote, depending on delay and filtering conditions. First, let us start with the ideal case (i), i.e., without delays and filtering. This leads to $D_{z5} = D_{z4} = 0$.

$$\lim_{\omega \to +\infty} Z_i(j\omega) = \lim_{\omega \to +\infty} \frac{N_{z2}}{j\omega \cdot D_{z3}} = \frac{k(I_m + k_\tau B_\tau B_q)}{j\omega \cdot I_m},$$

which represents a constant stiffness-type impedance scaled from the passive spring stiffness $k/(j\omega)$. The red dashed lines in Fig. 7.6 illustrate this ideal

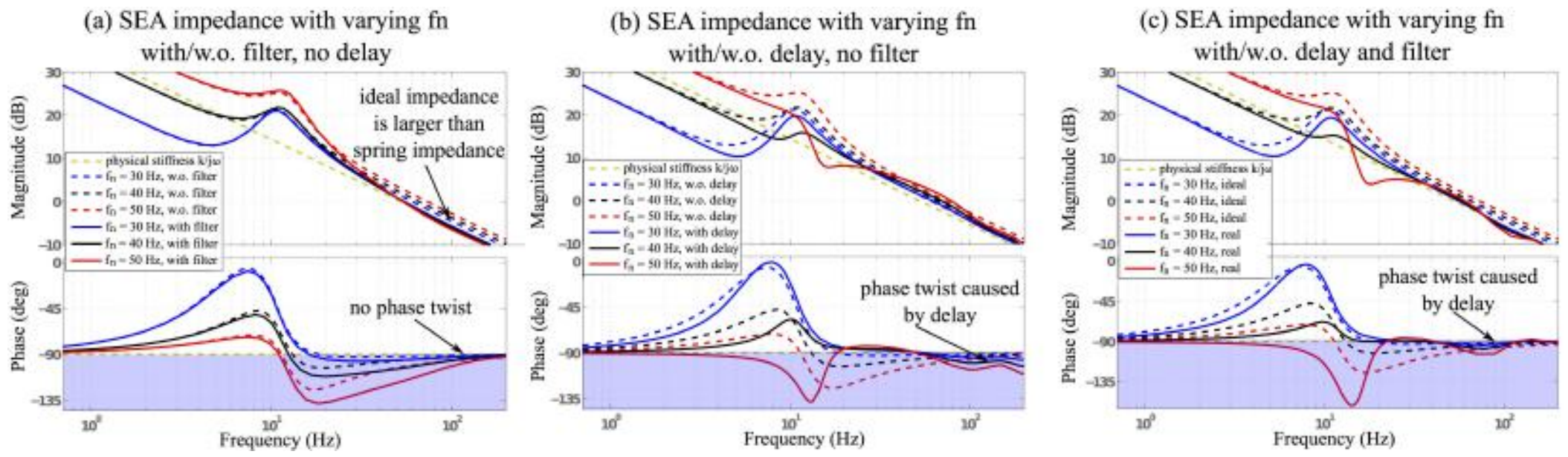

Figure 7.7: SEA impedance with varying natural frequencies $f_n$. Subfigures (a) and (b) show how feedback delay and filtering affect SEA impedance, respectively. In subfigure (c), we test the "real" cases with both filters and delays, and compare them with "ideal" cases with neither filter nor delays.

SEA impedance feature. Second, we derive the case (iii) only with delay, that is, $T_{fv} = T_{f\tau} = 0$. Then $D_{z5} = D_{z4} = 0$, and we obtain

$$\lim_{\omega \to +\infty} Z_d(j\omega) = \lim_{\omega \to +\infty} \frac{N_{z2}}{j\omega \cdot D_{z3}} = \frac{k(I_m + k_\tau B_\tau B_q e^{-T_{qd}s})}{j\omega \cdot I_m}$$

Since the complex number $e^{-T_{qd}s}$ rotates along the unit circle, the SEA impedance will periodically twist around the passive spring stiffness at the high frequency range. This is visualizable in Fig. 7.6 (c).

Third, in the case (ii) only with filtering, we have $T_{qs} = T_{qd} = T_\tau = 0$, and then obtain

$$\lim_{\omega \to +\infty} Z_f(j\omega) = \frac{N_{z4}}{j\omega D_{z5}} = \frac{k}{j\omega},$$

which represents a passive spring stiffness as shown in Fig. 7.6 (b). The curve does not twist thanks to the constant limit value. To verify that the behaviors aforementioned applicable to various natural frequencies, we analyze the SEA impedance performance with varying natural frequencies in Fig. 7.7. We use filters with $f_{qd} = 50$ Hz and $f_{\tau d} = 100$ Hz while feedback delays are chosen as $T_{qd} = T_\tau = 1$ ms, $T_{qs} = 10$ ms. First, these subfigures validate that a higher natural frequency $f_n$ results in a higher SEA impedance. Second, by comparing Fig. 7.7 (a) and (b) (or Fig. 7.6 (b) and (c)), we conclude that the feedback delay has a larger effect on the SEA impedance than the filtering.

Feedback and filtering indeed influence the SEA impedance. According to the passivity stability criterion (Colgate and Brown, 1994; Vallery et al., 2008; Ott et al., 2008), the phase value should stay within the range of $[-90°, 90°]$ (i.e., the blue region in Fig. 7.6. As the frequency responses show, the SEA impedance can penetrate the $-90°$ boundary value, which violates the passivity criterion. However, phase margins of the two cases without filtering in Fig. 7.6 (a) are larger than $54°$ while those of the other two cases with filtering are larger than $16°$. All four cases are stable but the passivity condition is violated. Similar behaviors can be observed in Fig. 7.7. To conclude, a non-passive system is not necessarily unstable and passivity-based stability criterion is conservative.

### 7.3.3 Effect of Load Inertia

This subsection studies the effect of load inertia on SEA impedance performance. A load inertia $I_j s$ is inserted into Eq. (7.18), i.e., $Z_l(j\omega) = Z(j\omega) + I_j s$. Then at the high frequency range, SEA impedance performs as a spring-mass impedance instead of a pure spring one. In particular, this impedance is dominated by the load inertia as shown in Fig. 7.8. We simulate three scenarios with different loads. The larger load inertia is, the smaller corner frequency it

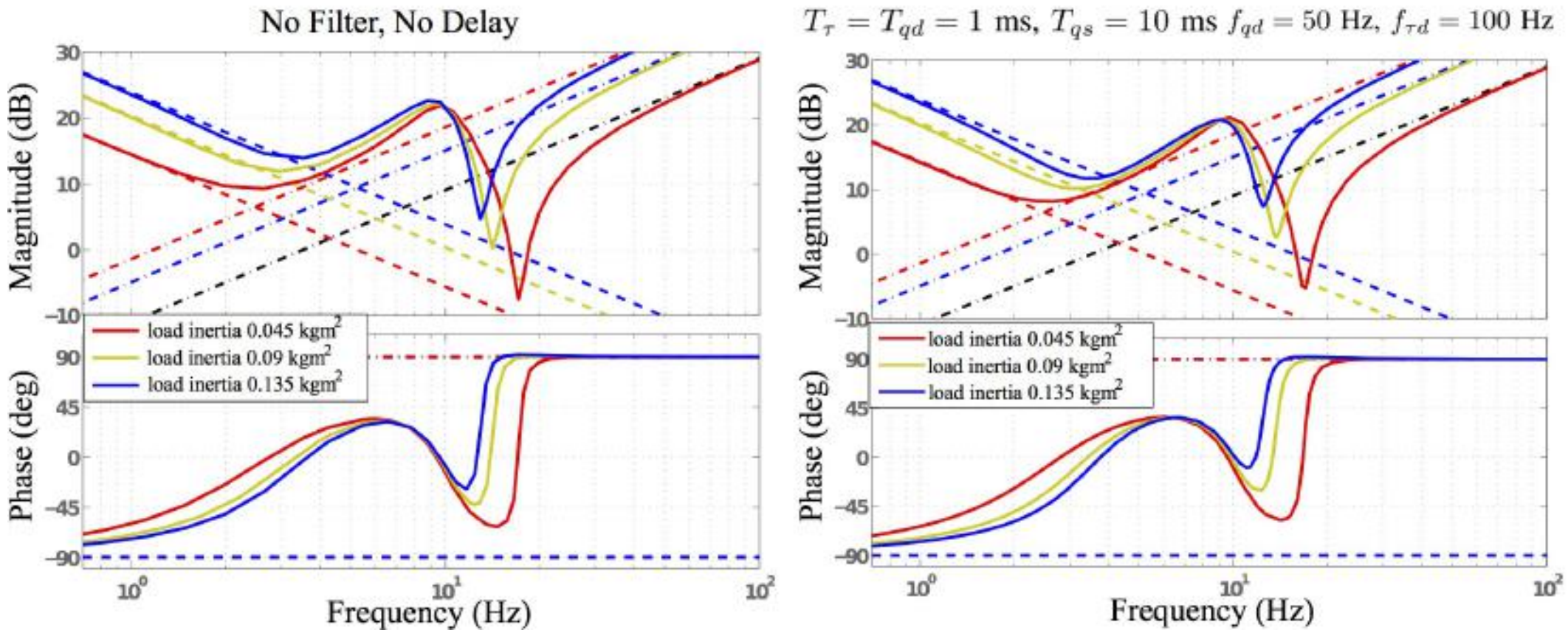

Figure 7.8: SEA impedance with varying load inertia. Three different load scenarios are illustrated in this figure. All of them use natural frequencies $f_n = 20$ Hz. When load inertia is considered, the SEA impedance approaches to the load inertia impedance curve $I_j \cdot j\omega$ at high frequencies.

has. Eq. (7.18) becomes $Z(j\omega) \to 0$ as $\omega \to +\infty$, and we have

$$\lim_{\omega \to +\infty} Z_l(j\omega) = \lim_{\omega \to +\infty} (Z(j\omega) + I_j \cdot j\omega) = I_j \cdot j\omega$$

where $I_j \cdot j\omega$ represents a 20 dB/dec asymptote at high frequencies (see Figure 7.8). Thus, the load inertia lead to high SEA impedance at the high frequency range.

## 7.4 SEA Impedance Characterization

In this section, we propose a metric to characterize SEA impedance performance in terms of both magnitude and frequency ranges[2]. The Z-width proposed in the haptics community (Colgate and Brown, 1994; Mehling et al.,

[2]The controller gains are designed by using the critically-damped gain design criterion and the gain scale defined in Section 7.2.

2005) only characterizes the achievable impedance magnitude. To quantify the achievable frequency range, we define a new concept of "Z-depth" – a frequency range within which achievable impedance magnitudes are non-zero (i.e., non-zero Z-width). Given the Z-width and Z-depth concepts, we can propose the following definition of Z-region.

**Definition 7.2 (SEA Z-region).** *The SEA impedance can be characterized by a Z-region, which is defined as a frequency-domain region composed of the achievable impedance magnitude range (Z-width) over a particular frequency range (Z-depth). Z-region can be estimated by the following metric*

$$Z_{\text{region}} = \int_{\omega_l}^{\omega_u} W(\omega)\Big|\log|Z_u(j\omega)| - \log|Z_l(j\omega)|\Big|d\omega, \tag{7.19}$$

*where $\omega_l$ and $\omega_u$ are lower and upper boundaries of the achievable frequency range, repsectively. $Z_u$ and $Z_l$ are the upper and lower boundries of the achievable impedances, respectively as shown in Fig. 7.9. $W(\omega)$ is a weighting value function depending on the frequency $\omega$.*

Eq. (7.19) adopts a log scale which is commonly used in frequency domain analysis. $\omega_u$ is determined by the intersection of $Z_u$ and $Z_l$ impedance lines while $\omega_l$ is a sufficiently small frequency and normally selected by the designer. The weighting function $W(\omega)$ is chosen by the designer, and can be used to emphasize a specific frequency range of interest. Another intuitive way to interpret the Z-region is that Z-region denotes the achievable frequency domain area.[3] To compute the Z-region, we present the following example.

[3]In theory, any infinitely small positive frequency should be included into the Z-depth.

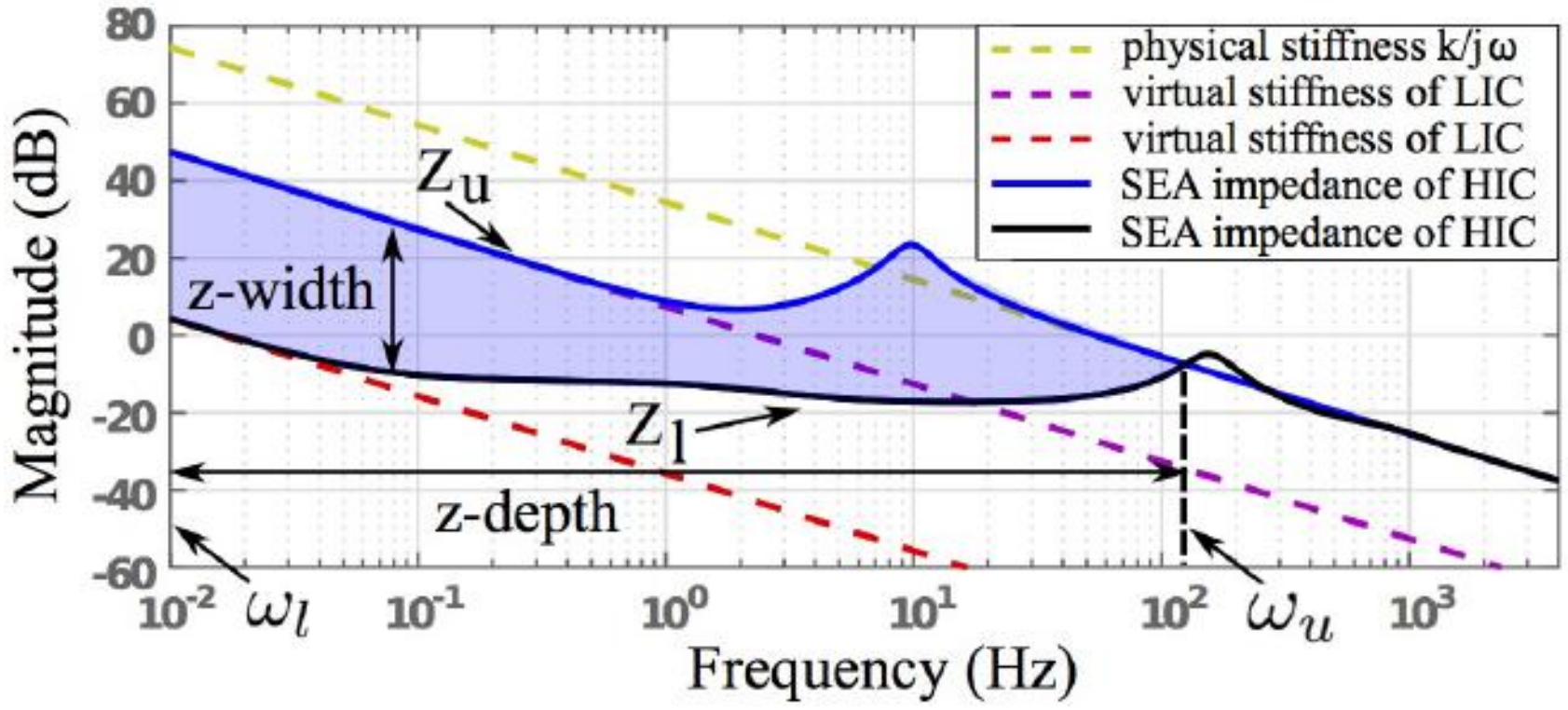

Figure 7.9: Z-region surrounded by the highest impedance controller (HIC) and the lowest impedance controller (LIC). This figure shows achievable SEA impedance ranges by modulating the gain scale as defined in Def. 7.1. The nominal gains are solved using the critically-damped gain design criterion.

**Example 7.2.** *To begin with, we choose* $W(\omega) = 1$ *since the impedance in different frequency ranges is equally treated. Natural frequency is chosen as* $f_n = 15$ *Hz. Filters have* $f_{vd} = 50$ *Hz,* $f_{\tau d} = 100$ *Hz, and feedback delays are chosen as* $T_\tau = T_{qd} = 1$ *ms and* $T_{qs} = 10$ *ms. Given these parameters, we sample for the highest impedance controller (HIC) (corresponds to the minimum gain scale* $GS_{\min}$*) without unstability. The result shows when* $GS_{\min} = 0.1$*, the phase margin becomes zero (marginally stable). As shown in Fig. 7.5, when* $GS > 1$ *and increasing* $GS$*, the phase margin approaches to a positive constant value. Thus, we manually set the maximum gain scale* $GS_{\max}$ *as* $10^3$*. As shown in Fig. 7.9, the Z-region is represented by the blue shaded region, surrounded by* $GS_{\max}$ *and* $GS_{\min}$ *impedance lines. The Z-depth is* $(\omega_l, \omega_u) = (10^{-2}, 120)$

---

However, our empirical experience reveals that the practically meaningful value of $\omega_l$ is usually within $[10^{-3}, 10^{-1}]$ Hz, depending on specific task requirements.

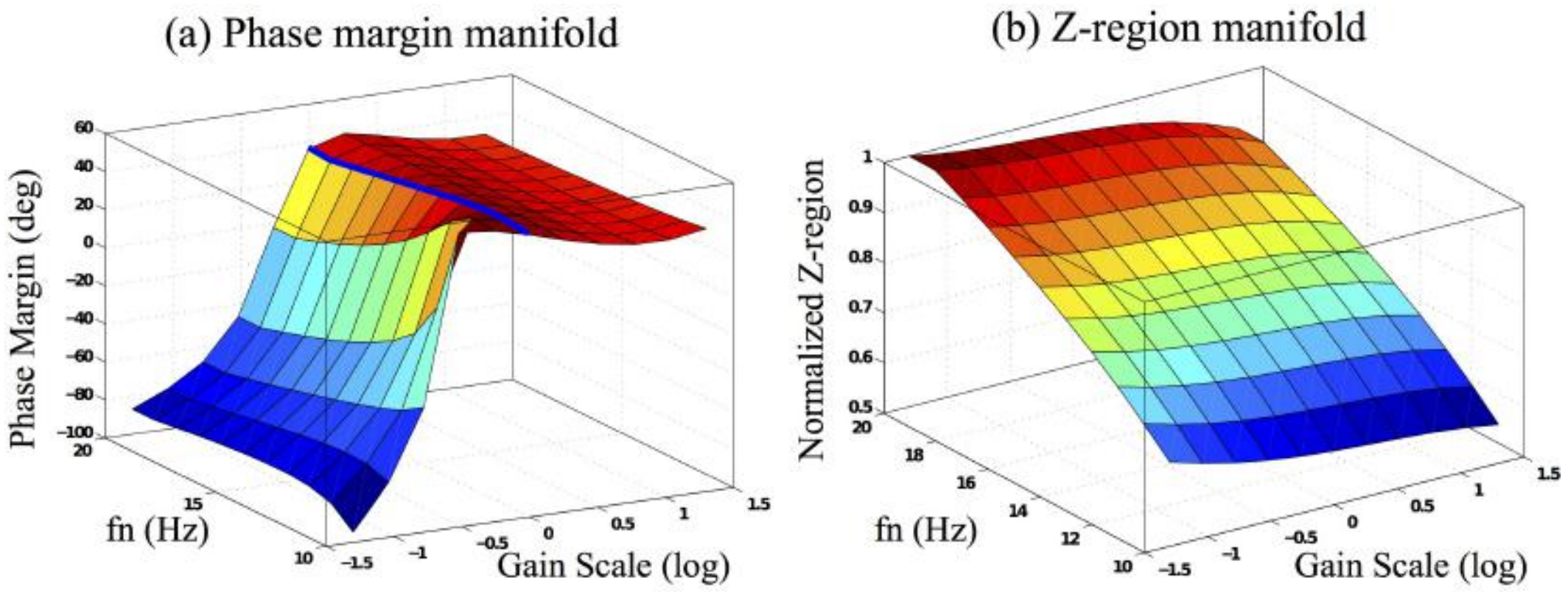

Figure 7.10: Stability and impedance performance maps. The phase margin manifold and Z-region manifold represent the system stability map and the impedance performance map, respectively. The gain scale $GS$ is expressed in the logarithmic scale. The searching range is $GS \in [0.05, 20]$. The natural frequency range is $f_n \in [10, 20]$ Hz.

*Hz (note that $\omega_l$ is empirically chosen)*[4].

Given the aforementioned phase margin analysis and the proposed $Z_{\text{region}}$ metric, let us generate SEA stability and impedance performance maps. The phase margin manifold and the Z-region manifold, as shown in Fig. 7.10, are sampled with respect to a wide range of natural frequencies $f_n$ and gain scales $GS$. $Z_{\text{region}}$ is normalized to the values within $[0, 1]$. The gain scale $GS$ uses a logarithmic scale for visualization convenience. As the peak blue line in subfigure (a) shows, the maximum phase margin is always achieved at around $GS = 1$ (i.e., 0 in the log scale) for all sampled natural frequencies $f_n$. This further validates the optimality of our critically-damped gain design criterion. Subfigure (b) shows the effects of natural frequency and gain scale on Z-region.

[4]We approximate $\omega_u$ by ignoring complex intersections induced by resonant peaks. In fact, within 100-200 Hz, the LIC impedance line has a small resonant peak which is induced by the feedback delays and filtering.

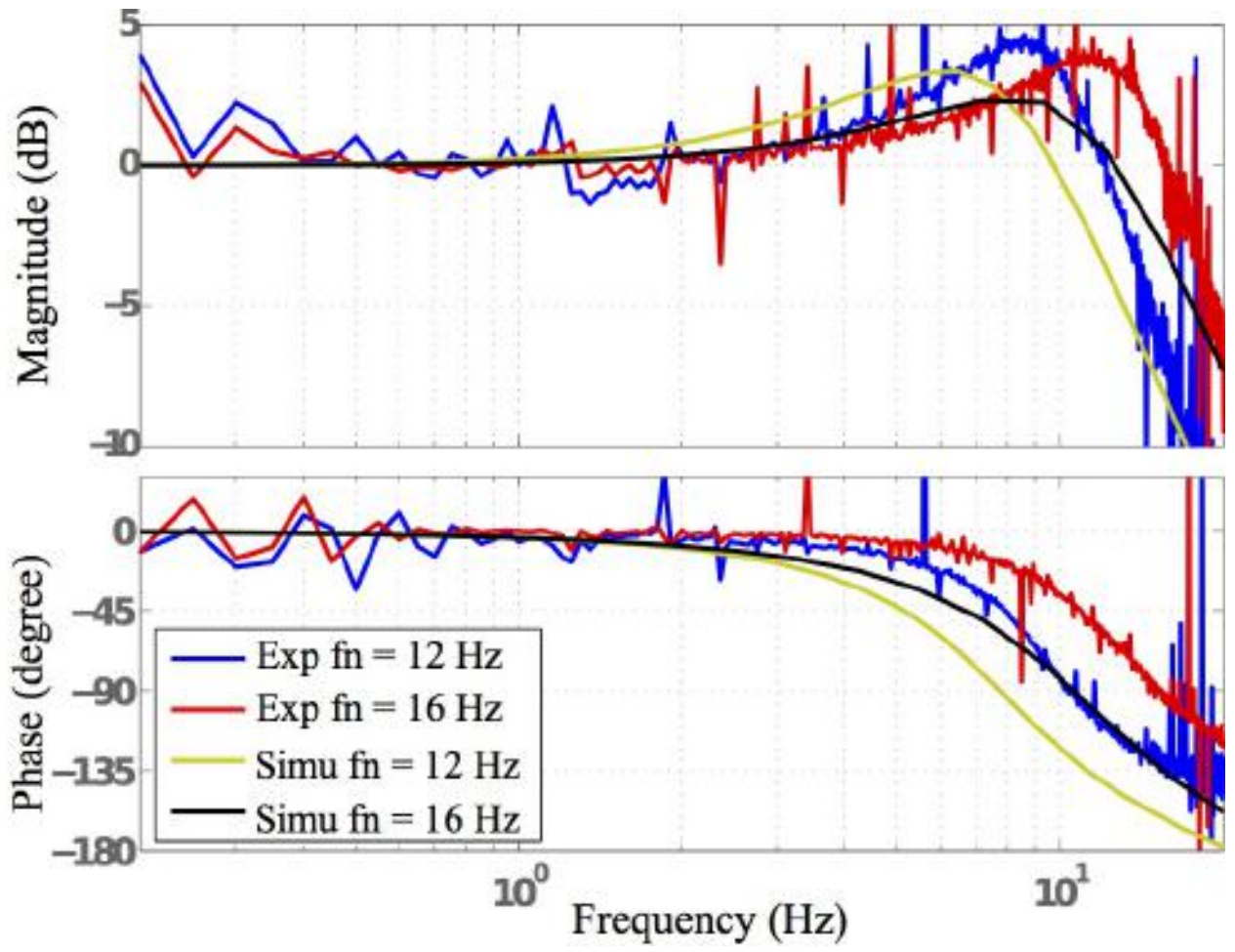

Figure 7.11: Bode diagrams with different natural frequencies $f_n$. This figure shows that increasing $f_n$ enlarges closed-loop bandwidth. At low frequencies, experiments match simulations quite well. At frequencies around resonant frequency, experiments show a larger resonant peak and a slightly larger bandwidth than simulations.

**Remark 7.1.** *This study takes one step towards quantifying the SEA impedance performance regarding simultaneously achievable magnitude and frequency ranges. Although Z-width was proposed to characterize the dynamic range of achievable stiffness-damping pairs, the pivotal frequency feature characterization is still unexploited in the literature.*

## 7.5 Experimental Evaluations

In this section, we use our UT-SEA test-bench to validate the proposed methods and criterion. The controller gains are designed based on the gain selection criterion in Section 7.2.1 and are shown in Table 7.1. In Fig. 7.11, we observe that a larger natural frequency results in a higher bandwidth of the closed-loop system. Simulations and experiments match each other except a slight discrepancy at the high-frequency range. As to the trade-off between torque gains and impedance gains, Fig. 7.12 demonstrates that when $GS > 1$, the system step response reveals a larger overshoot and a slower rise time. This result confirms our conjecture that increasing torque gains while decreasing impedance gains will deteriorate the closed-loop stability.

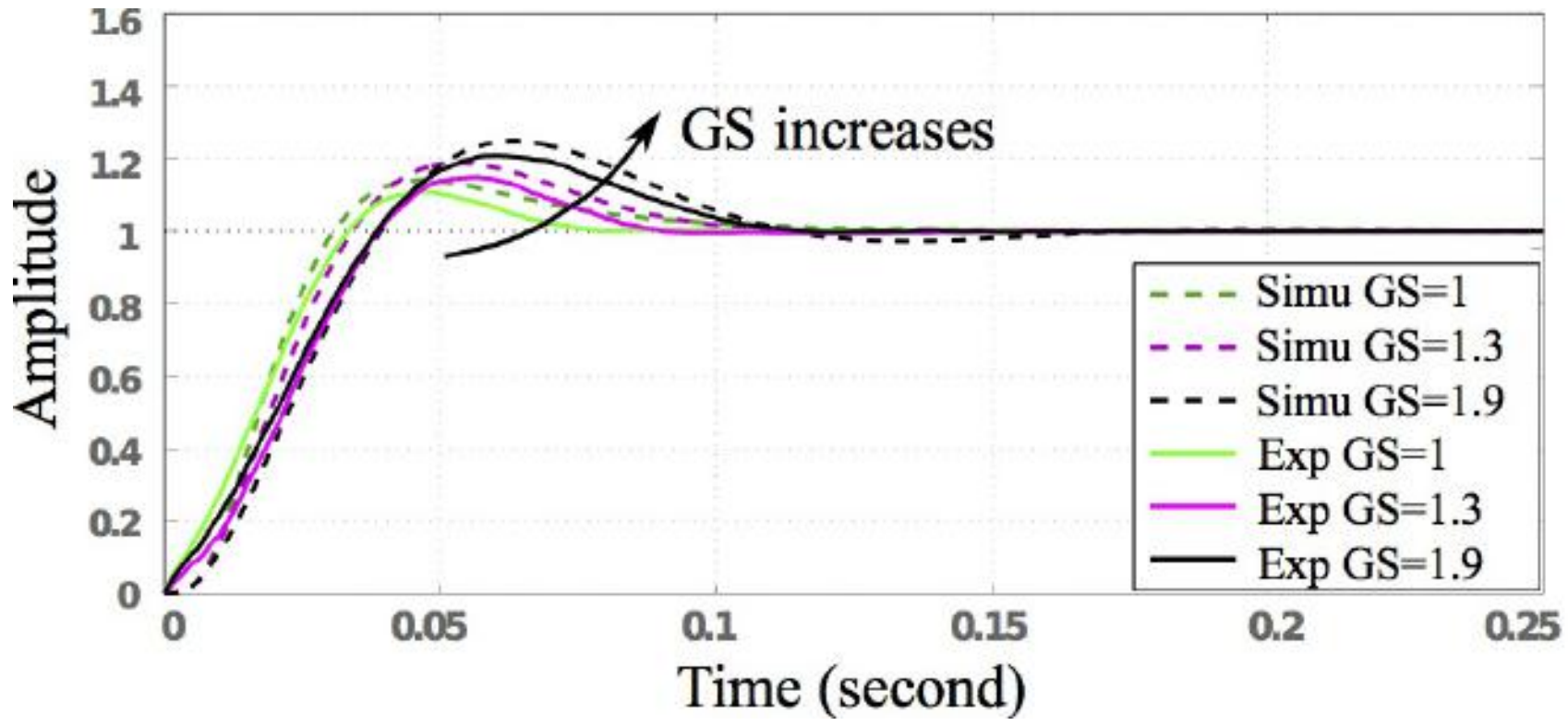

Figure 7.12: Comparison of simulations and experiments with different gain scales. Experimental data matches the simulations with a small discrepancy on the magnitude. We validate that increasing torque gains (decrease the impedance gains at the same time) deteriorate system stability.

As to the simulation and experiment discrepancies in the two tests above, one primary reason is that our SEA uses a timing belt inducing inevitable elasticity, which is is unrealistic to be modeled. Another probable reason is that simulation results use a constant joint pivot radius, which should be angle-dependent in practice. However, it is intractable to derive a transfer function with time-varying parameters. Thus, experiments are carried out around the horizontal arm position to minimize this radius error. Torque tracking performs almost identical at the low frequencies but reveals some discrepancies at the high frequencies.

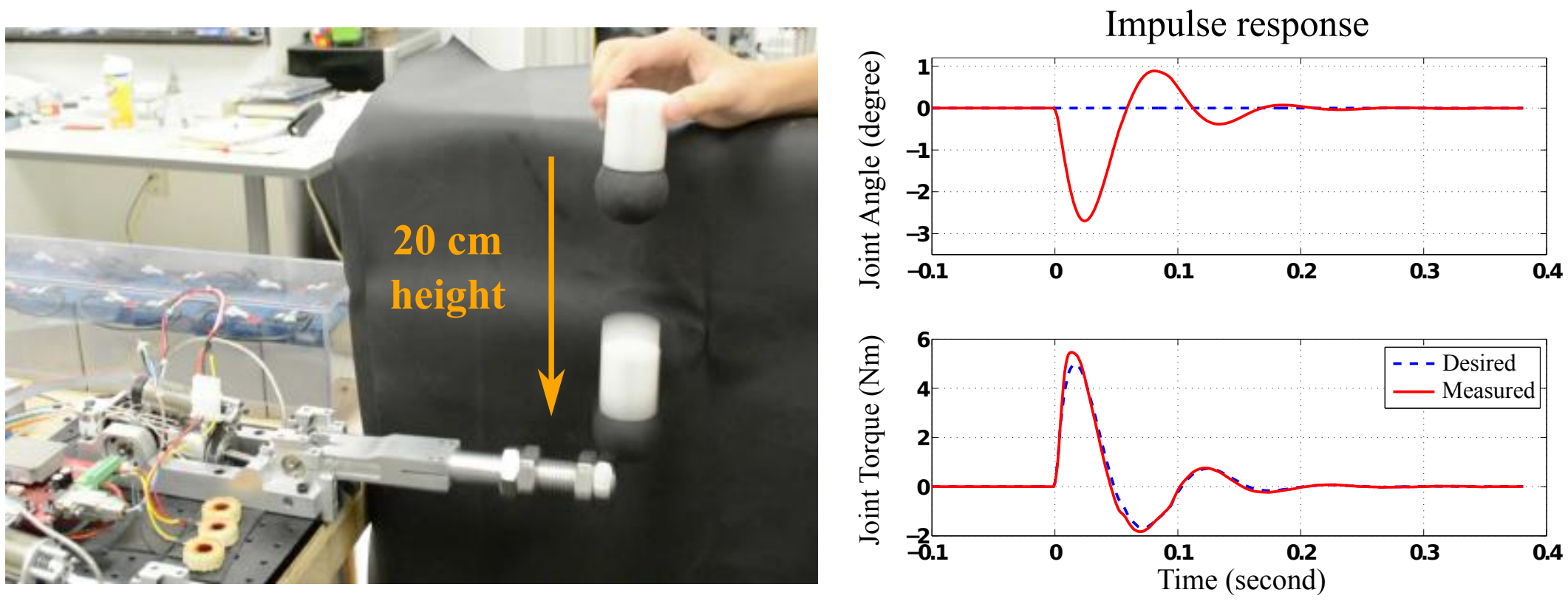

Figure 7.13: Impulse response. A ball is dropped from a constant height (20 cm) and exerts an impulse force on the arm end-effector. The maximum angle deviation is around 2.5 degrees. The arm recovers its initial position within 0.3 seconds. Joint torque has accurate force tracking.

Torque tracking under impact dynamics is essential for agile legged locomotion. As a proof of concept, we implement an impulse test and show how accurately our controller reacts to external impulse disturbances. As shown in Fig. 7.13, when a ball free falls from 20 cm height and hits the arm with an impulse force, the SEA actuator settles down promptly and recovers after approximately 0.3 seconds. Joint torque tracking is fairly accurate with only minor distortion around peak values.

## 7.6 Discussions and Conclusions

This chapter analyzed SEA impedance characteristics incorporating filtering and feedback delays at different frequency ranges. Controller gains in the cascaded control structure are designed based on a critically-damped fourth order gain selection method. By considering the trade-off between torque and impedance gains, we demonstrate our critically-damped gain selection method provides optimal gains in terms of phase margin based stability. Different from well-established Z-width work in the haptics field, we quantify the impedance range from the viewpoint of frequency characteristics. This feature provides another standpoint to understand SEA impedance performance and a new concept of Z-width is proposed correspondingly. Based on this concept, an optimization is proposed to compute the defined Z-region. Note that, our result is not limited by the conservative passivity criterion.

In the future, more comprehensive experimental implementations are supposed to evaluate the SEA impedance performance and Z-region metric. We will also explore the SEA impedance with different load inertias. Variable impedance controllers need to be designed for the legged robot applications.

Full-state feedback control, rather than the cascaded control structure, is also extensively invested for designing optimal controllers. It will be beneficial to compare the performance of these two control structures and benchmark the performance.

# Chapter 8

# Passivity of Time-Delayed Whole-Body Operational Space Control with SEA Dynamics

A class of distributed control architectures for latency-prone robotic systems was proposed in Chapter 6 and extended to the SEA cascaded control structure with inner-torque and outer-impedance feedback loops in Chapter 7. This paradigm motivates us to design a WBOSC with Cartesian position feedback at the centralized level and motor damping feedback at the embedded level. Note that, previous chapters mainly focus on SISO systems instead of the MIMO ones, which will be the focus of this chapter.

This chapter is outlined as follows. We first propose a theoretical formalism of the WBOSC architecture with embedded-level SEA dynamics in Section 8.1. Section 8.2 formulates a centralized-level WBOSC control with time delays. A passivity criterion is proposed by using Lyapunov-Krasovskii functionals in Section 8.3. Simulation results are shown in Section 8.4. The final section discusses future works. The proposed passivity has two inherent merits: (i) increase the robustness of the whole system, and (ii) provide

This chapter incorporates the results from the following publication: (Zhao and Sentis, 2016).

advantages in the compositional analysis of large-scale and complex control systems, such as the WBOSC coupled with SEA controllers examined in this Chapter. As far as the authors' knowledge, this is the first attempt to design time-delayed WBOSC with SEA dynamics while guaranteeing the system passivity.

## 8.1 Whole-Body Operational Space Control Formalism

Lagrange rigid multi-body dynamics are ubiquitously used in the robotics community (Featherstone, 2014; Chang, 2000). However, low-level actuator dynamics and time delays are largely overlooked but indeed pose a considerable threat to the closed-loop system stability and performance. In this Chapter, we propose a Whole-Body Operational Space Control (WBOSC) formalism incorporating embedded-level SEA dynamics in Section 8.1 and distributed delays in Section 8.2. An overall control diagram is shown in Fig. 8.1.

### 8.1.1 WBOSC with SEA Dynamics

Actuator dynamics are frequently ignored in the multi-body dynamics analysis due to its intrinsical complexity. The authors in (Ott et al., 2008; Braun et al., 2013; Spong, 1987) emphasized the importance of actuator dynamics in conventional rigid multi-body control and therefore, incorporated them into the whole-body controller formalism. Even so, extending the actuator-aware whole-body controller to free floating base dynamics with supporting contact constraints (Sentis et al., 2010) has not been explored yet. This mo-tivates our study in this Chapter. First, let us impose a few assumptions.

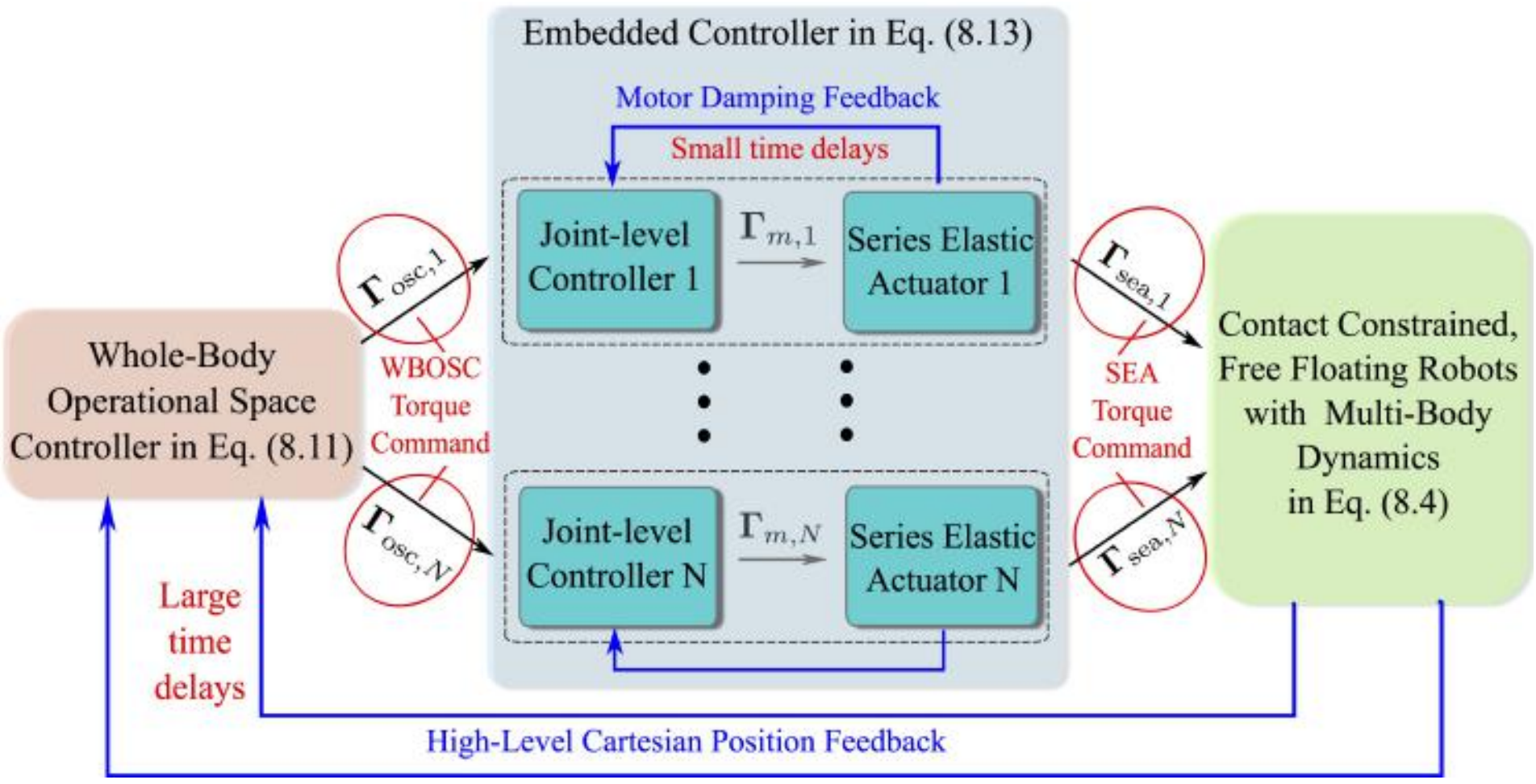

Figure 8.1: Time-delayed Whole-Body Operational Space Control architecture.

**Assumption 8.1.** *The elastic joint performs a relative small deformative motion such that the joint elasticity can be represented by a linear spring.*

**Assumption 8.2.** *The motor rotation axis coincides with the rotational symmetry, and thus rotor inertia has a uniform distribution around its rotational axis (Tomei, 1991).*

**Assumption 8.3.** *The rotor kinetic energy is dominated by pure rotation with respect to an inertial frame (Spong, 1987).*

To derive the equations of motion, we define kinetic and potential energies based on rigid multi-body dynamics and elastic actuator dynamics. The kinetic energy is represented as

$$\mathcal{T} = \frac{1}{2}\dot{\boldsymbol{q}}^T \boldsymbol{A}(\boldsymbol{q})\dot{\boldsymbol{q}} + \dot{\boldsymbol{q}}^T \boldsymbol{C}(\boldsymbol{q})\dot{\boldsymbol{\theta}} + \frac{1}{2}\dot{\boldsymbol{\theta}}^T \boldsymbol{B}\dot{\boldsymbol{\theta}}, \tag{8.1}$$

where $\boldsymbol{q} = (\boldsymbol{q}_b, \boldsymbol{q}_j)^T \in \mathbb{R}^{6+n}$ includes a free floating base state vector $\boldsymbol{q}_b \in \mathbb{R}^6$ composed of three prismatic joints and three rotational joints, and an actuated joint angle state vector $\boldsymbol{q}_j \in \mathbb{R}^n$; $\boldsymbol{\theta} \in \mathbb{R}^n$ corresponds to the motor angle state vector; $\boldsymbol{A}(\boldsymbol{q}) \succ \mathbf{0}$ is the inertia matrix of the multi-body dynamics; $\boldsymbol{C}(\boldsymbol{q})$ represents the coupled inertia between motor-side and joint-side dynamics; $\boldsymbol{B}$ is a constant diagonal matrix that denotes the rotor inertia after the gear ratio square scaling. By Assumption 8.3, the rotor inertia is the one along its principal axis of rotation, $z$-axis, represented by $I_{zz_i}$.

The potential energy has two components: gravitational energy $\mathcal{P}_g(\boldsymbol{q})$ and elastic potential energy $\mathcal{P}_e(\boldsymbol{q}, \boldsymbol{\theta})$.

$$\mathcal{P} = \mathcal{P}_g(\boldsymbol{q}) + \mathcal{P}_e(\boldsymbol{q}, \boldsymbol{\theta}). \tag{8.2}$$

By the Lagrangian $\mathcal{L} = \mathcal{T} - \mathcal{P}$ and Euler-Lagrangian derivations (Spong et al., 2006), we have

$$\begin{aligned} &\begin{pmatrix} \boldsymbol{A}(\boldsymbol{q}) & \boldsymbol{C}(\boldsymbol{q}) \\ \boldsymbol{C}^T(\boldsymbol{q}) & \boldsymbol{B} \end{pmatrix} \begin{pmatrix} \ddot{\boldsymbol{q}} \\ \ddot{\boldsymbol{\theta}} \end{pmatrix} + \begin{pmatrix} \boldsymbol{b} & \boldsymbol{b}_q \\ \boldsymbol{b}_\theta^T & \mathbf{0} \end{pmatrix} \begin{pmatrix} \dot{\boldsymbol{q}} \\ \dot{\boldsymbol{\theta}} \end{pmatrix} \\ &+ \begin{pmatrix} \boldsymbol{g}(\boldsymbol{q}) \\ \mathbf{0} \end{pmatrix} + \begin{pmatrix} \boldsymbol{J}_s^T \boldsymbol{F}_r \\ \mathbf{0} \end{pmatrix} = \begin{pmatrix} \boldsymbol{U}^T \boldsymbol{\Gamma}_{\text{sea}} \\ -\boldsymbol{\Gamma}_{\text{sea}} \end{pmatrix} + \begin{pmatrix} \mathbf{0} \\ \boldsymbol{\Gamma}_m \end{pmatrix}, \end{aligned} \tag{8.3}$$

where $\boldsymbol{b} = \boldsymbol{b}(\boldsymbol{q}, \dot{\boldsymbol{q}}, \dot{\boldsymbol{\theta}})$ denotes the Coriolis and centrifugal forces; $\boldsymbol{b}_q = \boldsymbol{b}_q(\boldsymbol{q}, \dot{\boldsymbol{q}})$ and $\boldsymbol{b}_\theta = \boldsymbol{b}_\theta(\boldsymbol{q}, \dot{\boldsymbol{q}})$ are derived from the Lagrange formalism in (Braun et al., 2013); $\boldsymbol{g(q)}$ denotes the gravitational forces; $\boldsymbol{F}_r$ denotes the ground reaction forces; $\boldsymbol{J}_s$ represents the Jacobian associated with all the supporting links; $\boldsymbol{U}$ is a selection matrix that chooses actuated joint states; $\boldsymbol{\Gamma}_{\text{sea}} = \boldsymbol{K}(\boldsymbol{\theta} - \boldsymbol{q}_j)$ denotes the sensed SEA torque from spring deflection, where the diagonal matrix $\boldsymbol{K}$ represents the joint spring stiffness; $\boldsymbol{\Gamma}_m$ represents the motor torque. The first block row in Eq. (8.3) represents the joint-side dynamics while the second block row represents the motor-side dynamics. A distinct feature of this formalism is its incorporation of the contact Jacobian force and free floating dynamics (Sentis, 2007), which characterize the features of under-actuated humanoid locomotion dynamics.

**Assumption 8.4.** *To make the multi-body dynamics tractable, we ignore the inertia coupling between joint-side and motor-side dynamics (Ott et al., 2008; Braun et al., 2013). That is,* $\boldsymbol{C}(\boldsymbol{q}) \approx \boldsymbol{0}, \boldsymbol{b} \approx \boldsymbol{b}(\boldsymbol{q}, \dot{\boldsymbol{q}}), \boldsymbol{b}_q \approx \boldsymbol{0}, \boldsymbol{b}_\theta \approx \boldsymbol{0}$.

**Proposition 8.1 (SEA-aware whole-body dynamics).** *Given Assumption 8.4, Eq. (8.3) results in the following SEA-aware whole-body dynamics*

$$\boldsymbol{A}(\boldsymbol{q})\ddot{\boldsymbol{q}} + \boldsymbol{N}(\boldsymbol{q}, \dot{\boldsymbol{q}}) + \boldsymbol{J}_s^T \boldsymbol{F}_r = \boldsymbol{U}^T \boldsymbol{\Gamma}_{\text{sea}}, \tag{8.4}$$

$$\boldsymbol{B}\ddot{\boldsymbol{\theta}} + \boldsymbol{\Gamma}_{\text{sea}} = \boldsymbol{\Gamma}_m, \tag{8.5}$$

$$\boldsymbol{\Gamma}_{\text{sea}} = \boldsymbol{K}(\boldsymbol{\theta} - \boldsymbol{q}_j), \tag{8.6}$$

*where* $\boldsymbol{N}(\boldsymbol{q}, \dot{\boldsymbol{q}}) = \boldsymbol{b}(\boldsymbol{q}, \dot{\boldsymbol{q}})\dot{\boldsymbol{q}} + \boldsymbol{g}(\boldsymbol{q})$.

Later on, we will design the WBOSC controller computing the desired joint torque command $\boldsymbol{\Gamma}_{\text{osc}}$, which is sent as control inputs to the embedded-level SEAs.

### 8.1.2 Embedded-Level SEA Controller

The embedded-level SEA controller comprises both torque and motor damping feedback loops (for instance, the control architecture shown in (Vallery et al., 2008)). By using torque feedback control, the control command $\boldsymbol{\Gamma}_{\text{osc}}$ sent to the embedded-level controller and the SEA torque $\boldsymbol{\Gamma}_{\text{sea}}$ actuating the robot joint are related as follows

$$\begin{aligned}\boldsymbol{\Gamma}_m =& \boldsymbol{B}\boldsymbol{B}_s^{-1}\boldsymbol{\Gamma}_{\text{osc}} + (\boldsymbol{I} - \boldsymbol{B}\boldsymbol{B}_s^{-1})\boldsymbol{\Gamma}_{\text{sea}} - \boldsymbol{B}\boldsymbol{B}_s^{-1} \cdot \boldsymbol{D}_\theta \dot{\boldsymbol{\theta}} \\ =& \boldsymbol{B}\boldsymbol{B}_s^{-1}\boldsymbol{\Gamma}_{\text{osc}} + \boldsymbol{\Gamma}_{\text{sea}} - \boldsymbol{B}\boldsymbol{B}_s^{-1}(\boldsymbol{\Gamma}_{\text{sea}} + \boldsymbol{D}_\theta \dot{\boldsymbol{\theta}}),\end{aligned} \tag{8.7}$$

where $\boldsymbol{\Gamma}_{\text{osc}}$ is the torque command computed from the centralized-level WBOSC controller, which will be designed in the next section; the torque feedback has a gain matrix $\boldsymbol{I} - \boldsymbol{B}\boldsymbol{B}_s^{-1}$, where the positive definite matrix $\boldsymbol{B}_s$ is the desired motor inertia matrix (Ott et al., 2008). From a physical interpretation, the torque feedback control is designed in the form of motor inertia shaping, which makes the passivity analysis tractable. Since we aim to reduce the effect of motor inertia on the joint-side dynamics, we choose $\boldsymbol{B}_s \prec \boldsymbol{B}$. As $\boldsymbol{B}_s$ approaches zero, $\boldsymbol{I} - \boldsymbol{B}\boldsymbol{B}_s^{-1}$ becomes more negative, which implies larger torque feedback gains.

**Remark 8.1.** *If the torque derivative term $\boldsymbol{D}\boldsymbol{K}^{-1}\dot{\boldsymbol{\Gamma}}_{\text{sea}}$ is modeled on the left-hand side of Eq. (8.5), a corresponding torque derivative feedback term can be added in Eq. (8.7).*

Besides this torque feedback loop, Eq. (8.7) contains a motor damping feedback $\boldsymbol{D}_\theta \dot{\boldsymbol{\theta}}$ replacing the centralized-level Cartesian damping feedback. Combining Eqs. (8.5) and (8.7), we have

$$\boldsymbol{\Gamma}_{\text{osc}} = \boldsymbol{B}_s \ddot{\boldsymbol{\theta}} + \boldsymbol{\Gamma}_{\text{sea}} + \boldsymbol{D}_\theta \dot{\boldsymbol{\theta}}, \tag{8.8}$$

As a result, the overall SEA-aware multi-body dynamics are represented by the Lagrangian dynamics in Section 8.1.1 and the embedded-level SEA controller in Section 8.1.2. Plugging Eq. (8.8) into Eq. (8.4), we have

$$\boldsymbol{A}(\boldsymbol{q})\ddot{\boldsymbol{q}} + \boldsymbol{U}^T\boldsymbol{B}_s\ddot{\boldsymbol{\theta}} + \boldsymbol{U}^T\boldsymbol{D}_\theta\dot{\boldsymbol{\theta}} + \boldsymbol{N}(\boldsymbol{q}, \dot{\boldsymbol{q}}) + \boldsymbol{J}_s^T\boldsymbol{F}_r = \boldsymbol{U}^T\boldsymbol{\Gamma}_{\text{osc}}. \tag{8.9}$$

Compared with conventional rigid multi-body dynamics, the following new terms emerge: shaped motor inertia $\boldsymbol{U}^T\boldsymbol{B}_s\ddot{\boldsymbol{\theta}}$ and embedded motor damping feedback $\boldsymbol{U}^T\boldsymbol{D}_\theta\dot{\boldsymbol{\theta}}$. The ignored friction compensation and torque derivative can also be modeled as necessary (Albu-Schäffer et al., 2007).

## 8.2 Centralized-level WBOSC with Time Delays

In this section, we design the centralized-level Whole-Body Operational Space controller (WBOSC) on the left side of the control diagram in Fig. 8.1. The WBOSC with dynamically consistent contact constraints (Khatib, 1987a; Sentis et al., 2010) is expressed as

$$\boldsymbol{\Lambda}_{t|s}\ddot{\boldsymbol{x}} + \boldsymbol{\mu}_{t|s} + \boldsymbol{p}_{t|s} + \boldsymbol{F}_c = \bar{\boldsymbol{J}}_{t|s}^T(\boldsymbol{U}\boldsymbol{N}_s)^T\boldsymbol{\Gamma}_{\text{osc}}, \tag{8.10}$$

where $\boldsymbol{\Lambda}_{t|s}$ is the task space inertia matrix under contact constraints; $\boldsymbol{\mu}_{t|s}$ represents the centrifugal and Coriolis force; $\boldsymbol{p}_{t|s}$ denotes the gravitational force; $\boldsymbol{F}_c \in \mathbb{R}^6$ is a reaction force acting on the task point; $\boldsymbol{J}^* = \boldsymbol{J}_{t|s}\overline{\boldsymbol{U}\boldsymbol{N}}_s$ is the support consistent reduced Jacobian. The subscript $t|s$ represents that the *task* is projected in the space consistent with *supporting constraints*. For more details, please refer to Def. 2.2.4 in (Sentis, 2007).

Time delays intensely degrade the real-time control performance of humanoid robots (Fok et al., 2015). Chapter 6 reveals that system stability and tracking performance is more sensitive to damping time delays rather than its stiffness counterpart. Thus Cartesian damping feedback is allocated to the embedded-level and represented by the motor damping feedback. In the WBOSC formalism, we denote the centralized-level round-trip time delay as $T_H$ and the embedded-level round-trip time delay as $T_L$. These delays are time-varying and can be induced by communication channels, filtering, and computation. In general, $T_H \gg T_L$ since $T_H$ is usually dominated by large communication delays. Now we propose a theorem of the time-delayed Operational Space Controller with SEA dynamics as follows.

**Theorem 8.1 (SEA-aware Operational Space Control with time delays).** *For contact-free motion control (i.e., $\boldsymbol{F}_c = 0$), the SEA torque command $\boldsymbol{\Gamma}_{\text{sea}}$ at the embedded level is*

$$\begin{aligned}\boldsymbol{\Gamma}_{\text{sea}}(t) = \boldsymbol{J}^{*T}_{(-d_0)}\Big(&\boldsymbol{\Lambda}_{t|s,(-d_0)}\boldsymbol{K}_x\big(\boldsymbol{x}_d(t - \frac{T_H + T_L}{2}) - \boldsymbol{x}(t - T_H - \frac{T_L}{2})\big)\Big)\\ &- \boldsymbol{B}_s\ddot{\boldsymbol{\theta}}(t) - \boldsymbol{D}_\theta\dot{\boldsymbol{\theta}}(t) + \bar{\boldsymbol{g}}(\boldsymbol{\theta})_{(-d_0)},\end{aligned} \tag{8.11}$$

*where the matrices with subscript $-d_0$ are evaluated at the instant $t - d_0 = t - T_H - T_L/2$.*

*Proof.* Via the Theorem 2.2.1 in (Sentis, 2007), the time-delayed Operational

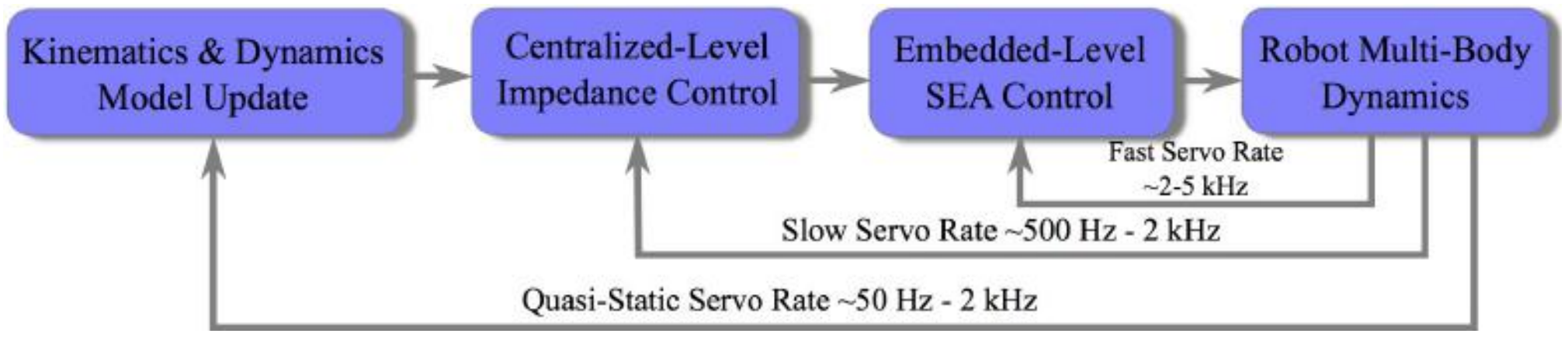

Figure 8.2: A conceptual diagram for hierarchical control/dynamics layers with different servo rates.

Space torque command at the centralized level is defined as

$$\begin{aligned}\boldsymbol{\Gamma}_{\text{osc}}(t) &= \boldsymbol{J}^{*T}(t-\frac{T_H}{2})\cdot\boldsymbol{\Lambda}_{t|s}(t-\frac{T_H}{2})\ddot{\boldsymbol{x}}(t)+\bar{\boldsymbol{g}}(\boldsymbol{\theta}(t-\frac{T_H}{2}))\\ &= \boldsymbol{J}^{*T}_{(-T_H/2)}\cdot\boldsymbol{\Lambda}_{t|s,(-T_H/2)}\boldsymbol{K}_x\big(\boldsymbol{x}_d(t)-\boldsymbol{x}(t-\frac{T_H}{2})\big)+\bar{\boldsymbol{g}}(\boldsymbol{\theta})_{(-T_H/2)},\quad (8.12)\end{aligned}$$

where $\boldsymbol{\mu}_{t|s}$ is ignored for simplicity. The subscript $-T_H/2$ stands for the instant $t-T_H/2$. The linear acceleration $\ddot{\boldsymbol{x}}(t)$ is defined as

$$\ddot{\boldsymbol{x}}(t)=\boldsymbol{K}_x\big(\boldsymbol{x}_d(t)-\boldsymbol{x}(t-\frac{T_H}{2})\big). \quad (8.13)$$

By Eq. (8.8), $\boldsymbol{\Gamma}_{\text{sea}}(t)$ can be reformulated as

$$\boldsymbol{\Gamma}_{\text{sea}}(t)=\ \boldsymbol{\Gamma}_{\text{osc}}(t-\frac{T_H+T_L}{2})-\boldsymbol{B}_s\ddot{\boldsymbol{\theta}}(t)-\boldsymbol{D}_\theta\dot{\boldsymbol{\theta}}(t) \quad (8.14)$$

Combining the equation above with Eqs. (8.12), we have the result in Eq. (8.11).

□

As a brief summary, this WBSOC formulation is distinct from conventional Operational Space Control in terms of the following aspects: (i) Embedded-level SEA dynamics in Eq. (8.14) are modeled. (ii) The fast embedded damping feedback $\boldsymbol{D}_\theta\dot{\boldsymbol{\theta}}$ takes the place of the slow centralized-level Cartesian velocity feedback. (iii) Distributed time delays are incorporated. Fig. 8.2 shows a diagram of nested control and dynamics layers with different servo rates. In certain cases, the kinematics and dynamics model uses the same update rate as that of the centralized-level controller (Kim et al., 2016a).

## 8.3 Passivity of Time-Delayed WBOSC

Given the time-delayed WBOSC formalism above, this section proposes a Lyapunov-Krasovskii functional to derive a passivity criterion for the overall closed-loop system. We subsequently generalize this passivity criterion to prioritized multi-task control.

### 8.3.1 Preliminaries

To guarantee the passivity condition, a majority of the existing literature derives the controller by only using motor angles $\boldsymbol{\theta}$ and their derivative (Tomei, 1991; Zollo et al., 2005). In turn, the passivity condition merely holds with respect to motor states. However, it is more physically meaningful to target joint-state-based passivity. In this Chapter, we aim at this type of passivity by a one-to-one mapping $\boldsymbol{\theta}_0 = \boldsymbol{h}(\boldsymbol{q}_0)$ between equilibrium points $\boldsymbol{\theta}_0$ and $\boldsymbol{q}_0$ under certain relaxed assumptions (Albu-Schäffer et al., 2007). The mapping is defined as

$$\boldsymbol{\theta}_0 = \boldsymbol{h}(\boldsymbol{q}_0) = \boldsymbol{q}_0 + \boldsymbol{K}^{-1}\boldsymbol{l}(\boldsymbol{q}_0), \tag{8.15}$$

with a function $\boldsymbol{l}(\boldsymbol{q})$ relating to feedback control and gravity compensation

$$\boldsymbol{l}(\boldsymbol{q}) = \boldsymbol{J}^*(\boldsymbol{q})^T \boldsymbol{\Lambda}(\boldsymbol{q}) \boldsymbol{K}_x \tilde{\boldsymbol{x}}(\boldsymbol{q}) + \boldsymbol{g}(\boldsymbol{q}), \tag{8.16}$$

where the Cartesian position error is $\tilde{\boldsymbol{x}}(\boldsymbol{q}_0) = \boldsymbol{x}_d(\boldsymbol{q}_0) - \boldsymbol{x}(\boldsymbol{q}_0)$; $\boldsymbol{l}(\boldsymbol{q}_0) = \boldsymbol{l}(\boldsymbol{q})\big|_{\boldsymbol{q}=\boldsymbol{q}_0} = \boldsymbol{K}(\boldsymbol{\theta}_0 - \boldsymbol{q}_0)$. Let us define a new state variable $\overline{\boldsymbol{q}}$, as a function of the motor angle $\boldsymbol{\theta}$ only, that is equal to the joint angle $\boldsymbol{q}$ at static state. Indeed, $\overline{\boldsymbol{q}}$ can not be solved from $\boldsymbol{\theta}$ analytically by Eq. (8.15). Thus, we adopt an iterative computational method (Albu-Schäffer et al., 2007) to compute $\overline{\boldsymbol{q}}$. More details of this method are provided in the contraction mapping principle of Appendix G. As a result, the static joint state is expressed as $\overline{\boldsymbol{q}}(\boldsymbol{\theta}) = \boldsymbol{h}^{-1}(\boldsymbol{\theta})$. The merit of using variable $\overline{\boldsymbol{q}}$ rather than $\boldsymbol{q}$ or $\boldsymbol{\theta}$ is to make the passivity analysis tractable since a coupling between motor and joint positions is canceled. Meanwhile, the joint-side Cartesian stiffness is maintained.

Given the state $\overline{\boldsymbol{q}}$ defined above, the centralized-level WBOSC in Eq. (8.12) is reformulated as

$$\begin{aligned}\boldsymbol{\Gamma}_{\text{osc}}(t) = {} & \boldsymbol{J}^*(\overline{\boldsymbol{q}})^T_{(-T_H/2)} \boldsymbol{\Lambda}(\overline{\boldsymbol{q}})_{(-T_H/2)} \boldsymbol{K}_x \Big( \boldsymbol{x}_d(\overline{\boldsymbol{q}}(t)) - \boldsymbol{x}(\overline{\boldsymbol{q}}(t - \frac{T_H}{2})) \Big) \\ & + \overline{\boldsymbol{g}}(\boldsymbol{\theta})_{(-T_H/2)},\end{aligned} \tag{8.17}$$

where $\overline{\boldsymbol{q}}(t) = \overline{\boldsymbol{q}}(\boldsymbol{\theta}(t))$, $\overline{\boldsymbol{q}}_0 = \overline{\boldsymbol{q}}(\boldsymbol{\theta}_0)$ at the static equilibrium. $\dot{\boldsymbol{x}}(\overline{\boldsymbol{q}}) = \boldsymbol{J}^*(\overline{\boldsymbol{q}}) \dot{\overline{\boldsymbol{q}}}(t)$. The subscript $t|s$ is omitted for clarity.

**Remark 8.2.** *The objective of gravity compensation is to find a* $\overline{\boldsymbol{q}}$ *such that* $\boldsymbol{g}(\overline{\boldsymbol{q}}) = \boldsymbol{g}(\boldsymbol{q}) = \overline{\boldsymbol{g}}(\boldsymbol{\theta})$ *in the quasi-static condition.* $\boldsymbol{g}(\overline{\boldsymbol{q}})$ *corresponds to the gravitational force* $\boldsymbol{p}_{t|s}$ *in the WBOSC of Eq. (8.10).*

The overall time-delayed WBOSC is composed of Eqs. (8.4), (8.14) and (8.17), which will be used in the passivity analysis.

Another issue to take into account is the joint-angle-based Jacobian, where we have

$$\dot{\boldsymbol{x}}(\overline{\boldsymbol{q}}(\boldsymbol{\theta})) = \boldsymbol{J}^*(\overline{\boldsymbol{q}})\dot{\overline{\boldsymbol{q}}}(\boldsymbol{\theta}) = \underbrace{\boldsymbol{J}^*(\overline{\boldsymbol{q}})\big(\dot{\overline{\boldsymbol{q}}}(\boldsymbol{\theta}) - \dot{\boldsymbol{\theta}}(t)\big)}_{\boldsymbol{v}_1} + \underbrace{\boldsymbol{J}^*(\overline{\boldsymbol{q}})\dot{\boldsymbol{\theta}}(t)}_{\boldsymbol{v}_2}.$$

As will be shown in the proof, the passivity criterion requires a motor-angle-based Jacobian mapping, however. To this end, we adopt the energy tank method in (Dietrich et al., 2016) and introduce a new coordinate $\hat{\boldsymbol{x}}(\overline{\boldsymbol{q}}(\boldsymbol{\theta}))$ defined as

$$\dot{\hat{\boldsymbol{x}}}(\overline{\boldsymbol{q}}(\boldsymbol{\theta})) = \boldsymbol{u} + \boldsymbol{v}_2, \tag{8.18}$$

where $\dot{\hat{\boldsymbol{x}}}$ intends to track $\dot{\boldsymbol{x}}$ while satisfying the passivity condition. $\boldsymbol{u}$ is a velocity state deviating from $\boldsymbol{v}_1$ to maintain the passivity. To compensate for the effect of $\boldsymbol{u}$, we define an energy storage function $E_s$ as

$$E_s = \frac{1}{2}s^2, \tag{8.19}$$

which has the flow rate of change

$$\dot{s} = -\frac{1}{2s}\bar{d}_1\boldsymbol{u}^T\boldsymbol{Q}\boldsymbol{u} - \frac{1}{s}\bar{d}_1\boldsymbol{u}^T\boldsymbol{Q}\boldsymbol{v}_2. \tag{8.20}$$

We define the deviating term $\boldsymbol{u}$ as

$$\boldsymbol{u} = \begin{cases} \boldsymbol{v}^{\mathrm{ref}} & s > \epsilon \qquad (8.21\mathrm{a}) \\ \boldsymbol{0} & \text{else} \qquad (8.21\mathrm{b}) \end{cases}$$

with

$$\boldsymbol{v}^{\mathrm{ref}} = \boldsymbol{v}_1 + \boldsymbol{K}_b\big(\boldsymbol{x}(\overline{\boldsymbol{q}}(\boldsymbol{\theta})) - \hat{\boldsymbol{x}}(\overline{\boldsymbol{q}}(\boldsymbol{\theta}))\big). \tag{8.22}$$

If $s > \epsilon > 0$ (i.e., a non-empty energy tank), $\boldsymbol{u}$ converges to $\boldsymbol{v}_1$ by using the propositional feedback control $\boldsymbol{K}_b\big(\boldsymbol{x}(\overline{\boldsymbol{q}}(\boldsymbol{\theta})) - \hat{\boldsymbol{x}}(\overline{\boldsymbol{q}}(\boldsymbol{\theta}))\big)$. Passivity and null space performance are simultaneously guaranteed in this case. If $s \leq \epsilon$ (i.e., an empty energy tank), $\boldsymbol{u} = \boldsymbol{0}$, which indicates the deviation of $\dot{\hat{\boldsymbol{x}}}(\overline{\boldsymbol{q}}(\boldsymbol{\theta}))$ from $\dot{\boldsymbol{x}}(\overline{\boldsymbol{q}}(\boldsymbol{\theta}))$. In this case, the null space control performance is compromised to guarantee the system passivity. Note that Eq. (8.21) is designed to avoid a zero division of $1/s$ in Eq. (8.20). The parameter $\epsilon$ should be chosen close enough to zero such that the energy tank will never be empty (i.e., without the sacrifice of null space control performance). More details of the energy tank design are provided in (Dietrich et al., 2016).

Necessary propositions for the passivity criterion are shown in the Supplementary Material. From now on, without explicit notations, we assume the matrices $\boldsymbol{J}^*$, $\boldsymbol{\Lambda}$ and $\bar{\boldsymbol{g}}(\boldsymbol{\theta})$ represent the time-delayed ones evaluated at time $t - d_0$ in Eq. (8.11) for the sake of clarity. Additionally, we propose an assumption for upper bounds of time delays.

**Assumption 8.5.** (Upper bounds of time delays) *Since the round trip delays $T_H$ and $T_L$ are time-varying, we define the upper bounds of the following delays: $d_1 = T_H + T_L/2 \leq \bar{d}_1$, $d_2 = (T_H + T_L)/2 \leq \bar{d}_2$, which will be used afterwards.*

### 8.3.2 Passivity of WBOSC

This subsection derives a delay-dependent passivity criterion for the WBOSC with SEA dynamics. Let us first construct a Lyapunov-Krasovskii functional of the concerned time-delayed system (Gu et al., 2003) with $V = V_1 + V_2 + V_3 + V_4$

$$\begin{aligned}
V_1 &= \frac{1}{2}\dot{\boldsymbol{q}}^T \boldsymbol{A}(\boldsymbol{q})\dot{\boldsymbol{q}} + \mathcal{P}_g(\boldsymbol{q}) + \int_0^t \big(\dot{\boldsymbol{q}}^T(\delta)\boldsymbol{J}_s^T \boldsymbol{F}_r(\delta)\big)d\delta,\\
V_2 &= \frac{1}{2}\dot{\boldsymbol{\theta}}^T \boldsymbol{B}_s\dot{\boldsymbol{\theta}} + \frac{1}{2}(\boldsymbol{\theta} - \boldsymbol{q}_j)^T \boldsymbol{K}(\boldsymbol{\theta} - \boldsymbol{q}_j),\\
V_3 &= -V_{\bar{l}}(\boldsymbol{\theta}) = \frac{1}{2}\tilde{\boldsymbol{x}}(\overline{\boldsymbol{q}}(\boldsymbol{\theta}))^T \boldsymbol{\Lambda}\boldsymbol{K}_x\tilde{\boldsymbol{x}}(\overline{\boldsymbol{q}}(\boldsymbol{\theta})) - \frac{1}{2}\boldsymbol{l}^T(\overline{\boldsymbol{q}}(\boldsymbol{\theta}))\boldsymbol{K}^{-1}\boldsymbol{l}(\overline{\boldsymbol{q}}(\boldsymbol{\theta})) - \mathcal{P}_g(\overline{\boldsymbol{q}}),\\
V_4 &= \frac{1}{2}\int_{-\bar{d}_1}^0 \int_{t+r}^t \dot{\tilde{\boldsymbol{x}}}^T(\xi)\boldsymbol{Q}\dot{\tilde{\boldsymbol{x}}}(\xi)d\xi dr + E_s,
\end{aligned}$$

where $\boldsymbol{Q} \succ 0$. $V_1$ is an energy function corresponding to the multi-body dynamics in the joint-side subsystem of Fig. 8.3 while $V_2$ corresponds to the SEA actuator dynamics in the controller-side subsystem. $V_3 = -V_{\bar{l}}(\boldsymbol{\theta})$ represents a potential function for $\bar{\boldsymbol{l}}(\boldsymbol{\theta}) = \boldsymbol{l}(\overline{\boldsymbol{q}}(\boldsymbol{\theta}))$ of Eq. (8.16) and satisfies

$$\frac{\partial V_{\bar{l}}(\boldsymbol{\theta})}{\partial \boldsymbol{\theta}} = \bar{\boldsymbol{l}}(\boldsymbol{\theta})^T = \boldsymbol{l}(\overline{\boldsymbol{q}}(\boldsymbol{\theta}))^T. \tag{8.23}$$

For more details about this potential function, please refer to (Appendix, (Ott et al., 2008)). The purpose of defining this potential function is to cancel certain terms in $\dot{V}_2$ and $\dot{V}_3$, as will be illustrated later. $V_4$ is a delay compensation term to be used together with Proposition G.3 in Appendix G.

Let us first prove the passivity of the joint-side subsystem via $V_1$. By the following properties

$$\frac{\partial \mathcal{P}_g(\boldsymbol{q})}{\partial \boldsymbol{q}} = \boldsymbol{g}(\boldsymbol{q}),\ \frac{\partial \mathcal{P}_g(\overline{\boldsymbol{q}})}{\partial \overline{\boldsymbol{q}}} = \boldsymbol{g}(\overline{\boldsymbol{q}}), \tag{8.24}$$

and Eq. (8.4) and Property 2, the derivative of $V_1$ is

$$\dot{V}_1 = \dot{\boldsymbol{q}}^T \boldsymbol{U}^T \boldsymbol{\Gamma}_{\mathbf{sea}}. \tag{8.25}$$

Choosing the storage function of the joint-side subsystem in Fig. 8.3 as $S_l(\boldsymbol{q}, \dot{\boldsymbol{q}}) = V_1$, we have

$$\dot{S}_l(\boldsymbol{q}, \dot{\boldsymbol{q}}) = \dot{V}_1 = \dot{\boldsymbol{q}}^T \boldsymbol{U}^T \boldsymbol{\Gamma}_{\mathbf{sea}} = \dot{\boldsymbol{q}}_j^T \boldsymbol{\Gamma}_{\mathbf{sea}}. \tag{8.26}$$

Thus, the passivity of the mapping $\boldsymbol{\Gamma}_{\mathbf{sea}} \to \dot{\boldsymbol{q}}$ of Fig. 8.3 is guaranteed due to the inherent passive property of the physical system. Next, we prove the passivity of the controller-side subsystem in Fig. 8.3. Given a few mathematical manipulations in Appendix G, the derivative of $V_2$ is

$$\begin{aligned} \dot{V}_2 = & -\dot{\boldsymbol{\theta}}^T \boldsymbol{D}_\theta \dot{\boldsymbol{\theta}} - \dot{\boldsymbol{q}}_j^T \boldsymbol{\Gamma}_{\mathbf{sea}} + \dot{\boldsymbol{\theta}}^T \boldsymbol{J}^{*T} \boldsymbol{\Lambda} \boldsymbol{K}_x \int_{t-d_1}^{t} \dot{\hat{\boldsymbol{x}}}(\xi) d\xi \\ & + \dot{\boldsymbol{\theta}}^T \boldsymbol{J}^{*T} \boldsymbol{\Lambda} \boldsymbol{K}_x \big(\hat{\boldsymbol{x}}_d - \hat{\boldsymbol{x}}\big) + \dot{\boldsymbol{\theta}}^T \bar{\boldsymbol{g}}(\boldsymbol{\theta}). \end{aligned} \tag{8.27}$$

As for $\dot{V}_3$ and $\dot{V}_4$, we have

$$\begin{aligned} \dot{V}_3 = & -\frac{\partial \dot{V}_{\bar{l}}(\boldsymbol{\theta})}{\partial \boldsymbol{\theta}} \dot{\boldsymbol{\theta}} = -\boldsymbol{l}^T(\overline{\boldsymbol{q}}(\boldsymbol{\theta}))\dot{\boldsymbol{\theta}} = -\dot{\boldsymbol{\theta}}^T \big(\boldsymbol{J}^{*T} \boldsymbol{\Lambda} \boldsymbol{K}_x \tilde{\boldsymbol{x}}(\overline{\boldsymbol{q}}(\boldsymbol{\theta})) + \boldsymbol{g}(\overline{\boldsymbol{q}})\big), \\ \dot{V}_4 = & \frac{1}{2}\bar{d}_1 \dot{\hat{\boldsymbol{x}}}^T \boldsymbol{Q} \dot{\hat{\boldsymbol{x}}} - \frac{1}{2}\int_{t-d_1}^{t} \dot{\hat{\boldsymbol{x}}}(\xi)^T \boldsymbol{Q} \dot{\hat{\boldsymbol{x}}}(\xi) d\xi + s\dot{s}. \end{aligned}$$

Without explicit notations, we define $\hat{\boldsymbol{x}}(\overline{\boldsymbol{q}}(\boldsymbol{\theta})) = \hat{\boldsymbol{x}}(t) = \hat{\boldsymbol{x}}$, $\hat{\boldsymbol{x}}_d = \hat{\boldsymbol{x}}_d(t)$. For the components in $\dot{V}_4$, we have

$$\begin{aligned} \frac{1}{2}\bar{d}_1 \dot{\hat{\boldsymbol{x}}}^T \boldsymbol{Q} \dot{\hat{\boldsymbol{x}}} + s\dot{s} = & \frac{1}{2}\bar{d}_1 (\boldsymbol{u} + \boldsymbol{v}_2)^T \boldsymbol{Q} (\boldsymbol{u} + \boldsymbol{v}_2) - \frac{1}{2}\bar{d}_1 \boldsymbol{u}^T \boldsymbol{Q} \boldsymbol{u} - \bar{d}_1 \boldsymbol{u}^T \boldsymbol{Q} \boldsymbol{v}_2 \\ = & \frac{1}{2}\bar{d}_1 \boldsymbol{v}_2^T \boldsymbol{Q} \boldsymbol{v}_2 = \frac{1}{2}\bar{d}_1 \dot{\boldsymbol{\theta}}^T(t) \boldsymbol{J}^{*T}(\overline{\boldsymbol{q}}) \boldsymbol{Q} \boldsymbol{J}^{*}(\overline{\boldsymbol{q}}) \dot{\boldsymbol{\theta}}(t), \end{aligned}$$

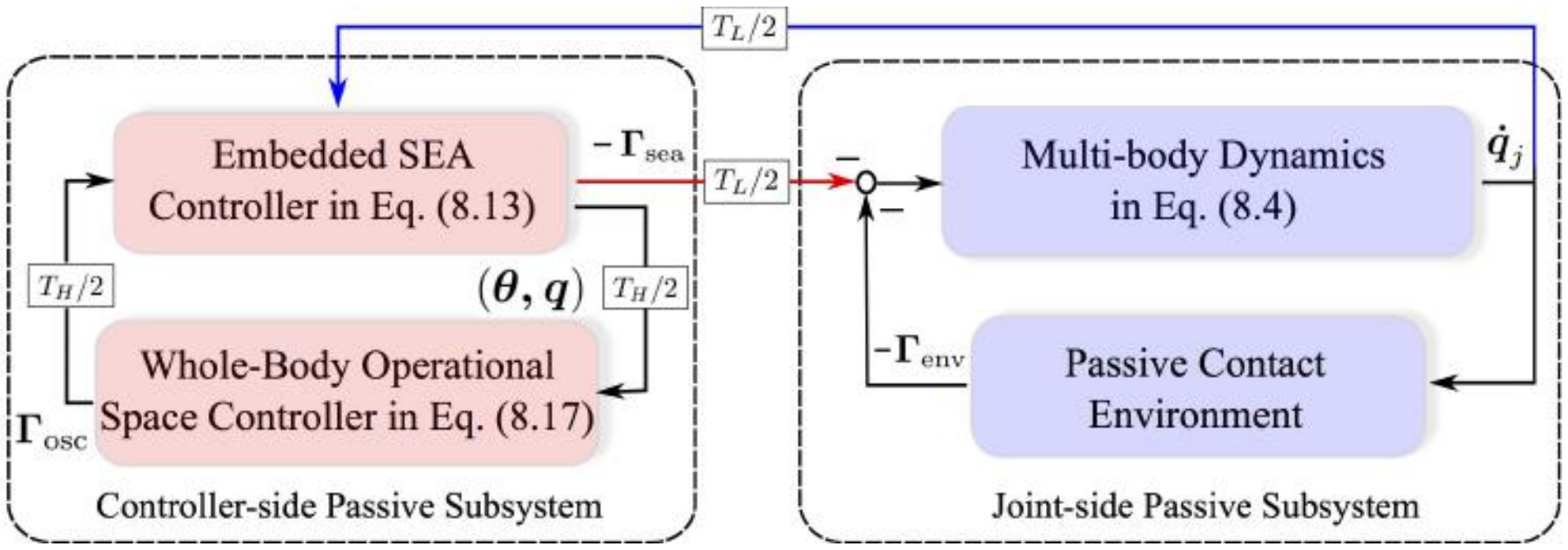

Figure 8.3: Two passive subsystems interconnected in a feedback configuration. High- and low-level time delays $T_H$ and $T_L$ are labeled among subsystem blocks.

which is a favorable quadratic term of $\dot{\boldsymbol{\theta}}(t)$. Additionally, by Property G.3 in Appendix G, we have

$$\dot{\boldsymbol{\theta}}^T \boldsymbol{J}^{*T} \boldsymbol{\Lambda} \boldsymbol{K}_x \int_{t-d_1}^{t} \dot{\boldsymbol{x}}(\xi) d\xi - \frac{1}{2} \int_{t-d_1}^{t} \dot{\boldsymbol{x}}(\xi)^T \boldsymbol{Q} \dot{\boldsymbol{x}}(\xi) d\xi \leq \frac{1}{2} \bar{d}_1 \dot{\boldsymbol{\theta}}^T(t) \boldsymbol{P} \dot{\boldsymbol{\theta}}(t), \quad (8.28)$$

where $\boldsymbol{P} = \boldsymbol{J}^{*T} \boldsymbol{\Lambda} \boldsymbol{K}_x \boldsymbol{Q}^{-1} \boldsymbol{K}_x \boldsymbol{\Lambda} \boldsymbol{J}^*$. Since $\boldsymbol{Q}^{-1} \succ 0$ and $(\boldsymbol{J}^{*T} \boldsymbol{\Lambda} \boldsymbol{K}_x)^T = \boldsymbol{K}_x \boldsymbol{\Lambda} \boldsymbol{J}^*$, $\boldsymbol{P}$ is positive definite. Let us define the storage function of the controller-side subsystem as $S_c(\boldsymbol{q}, \boldsymbol{\theta}, \dot{\boldsymbol{\theta}}) = V_2 + V_3 + V_4$ and take its derivative

$$\dot{S}_c(\boldsymbol{q}, \boldsymbol{\theta}, \dot{\boldsymbol{\theta}}) = \dot{V}_2 + \dot{V}_3 + \dot{V}_4 \leq -\dot{\boldsymbol{\theta}}^T \big( \boldsymbol{D}_\theta - \frac{1}{2} \bar{d}_1 \boldsymbol{P} - \frac{1}{2} \bar{d}_1 \boldsymbol{J}^{*T}(\overline{\boldsymbol{q}}) \boldsymbol{Q} \boldsymbol{J}^*(\overline{\boldsymbol{q}}) \big) \dot{\boldsymbol{\theta}} - \dot{\boldsymbol{q}}_j^T \boldsymbol{\Gamma}_{\mathbf{sea}}.$$

If the first quadratic term is negative definite, the passivity of $S_c(\boldsymbol{q}, \boldsymbol{\theta}, \dot{\boldsymbol{\theta}})$ is guaranteed. That is, the mapping $\dot{\boldsymbol{q}} \rightarrow -\boldsymbol{\Gamma}_{\mathbf{sea}}$ is passive in Fig. 8.3. Combining Eqs. (8.25)-(8.28), we have

$$\dot{V} = \dot{V}_1 + \dot{V}_2 + \dot{V}_3 + \dot{V}_4 \leq \dot{\boldsymbol{\theta}}^T \big( -\boldsymbol{D}_\theta + \frac{1}{2} \bar{d}_1 \boldsymbol{P} + \frac{1}{2} \bar{d}_1 \boldsymbol{J}^{*T}(\overline{\boldsymbol{q}}) \boldsymbol{Q} \boldsymbol{J}^*(\overline{\boldsymbol{q}}) \big) \dot{\boldsymbol{\theta}}. \quad (8.29)$$

To guarantee $\dot{V} \leq 0$, we propose the following theorem.

**Theorem 8.2** (**Passivity criterion of single task control**). *If there exists a positive-definite matrix* $\boldsymbol{Q}$ *and a positive time delay scalar* $\bar{d}_1$ *such that the following LMI holds:*

$$\begin{pmatrix} -\boldsymbol{D}_\theta + \frac{1}{2}\bar{d}_1 \boldsymbol{J}^{*T}(\overline{\boldsymbol{q}})\boldsymbol{Q}\boldsymbol{J}^*(\overline{\boldsymbol{q}}) & \frac{1}{2}\bar{d}_1 \boldsymbol{J}^{*T}(\overline{\boldsymbol{q}})\boldsymbol{\Lambda}\boldsymbol{K}_x \\ * & -\frac{1}{2}\bar{d}_1\boldsymbol{Q} \end{pmatrix} \preceq \boldsymbol{0}, \tag{8.30}$$

*with* $*$ *denoting the transpose of the corresponding matrix blocks, then the interconnected feedback system is passive and the motor velocity state* $\dot{\boldsymbol{\theta}}$ *is bounded.*

*Proof.* Based on the result in Eq. (8.29), the delay-dependent criterion in Eq. (8.30) is derived by the Schur complement (see Proposition G.5 in Appendix G). □

Eq. (8.30) shows that as the motor damping feedback gain matrix $\boldsymbol{D}_\theta$ increases, larger time delays are allowable. We will validate this property in the simulation.

Note that, the matrices $\boldsymbol{J}^*(\overline{\boldsymbol{q}})$ and $\boldsymbol{\Lambda}$ in Eq. (8.30) are evaluated at time instant $t - T_H - T_L/2$. These matrices are treated as quasi-static ones given they are updated at a relatively slow servo rate. Therefore, we solve this LMI numerically. Compared to time-invariant LMI solutions, the feasible solution range of $\boldsymbol{D}_\theta$ and $\bar{d}_1$ in Eq. (8.30) becomes constrained. Given a fixed $\boldsymbol{D}_\theta$, the allowable maximum delay $\bar{d}_1$ is solvable (Hua and Liu, 2010).

Based on the theorem above, we can generalize the passivity criterion to prioritized multi-task control as follows.

**Corollary 8.1 (Passivity criterion of prioritized multi-task control).** *Consider $N$ prioritized Whole-Body Operational Space tasks. If there exists a set of positive-definite matrices $\boldsymbol{Q}_i, i \in [1, N]$ and a positive time delay scalar $\bar{d}_1$ such that the following LMI holds,*

$$\begin{pmatrix} -\boldsymbol{D}_\theta + \frac{1}{2}\bar{d}_1 \sum_{i=1}^{N} \boldsymbol{J}^{*T}_{i|\mathrm{prec}(i)}(\overline{\boldsymbol{q}})\boldsymbol{Q}_i \boldsymbol{J}^{*}_{i|\mathrm{prec}(i)}(\overline{\boldsymbol{q}}) & \frac{1}{2}\bar{d}_1\boldsymbol{M}_1 & \cdots & \frac{1}{2}\bar{d}_1\boldsymbol{M}_N \\ * & -\frac{1}{2}\bar{d}_1\boldsymbol{Q}_1 & \mathbf{0} & \mathbf{0} \\ * & * & \ddots & \mathbf{0} \\ * & * & * & -\frac{1}{2}\bar{d}_1\boldsymbol{Q}_N \end{pmatrix} \preceq \mathbf{0}, \tag{8.31}$$

*where $\boldsymbol{M}_i = \boldsymbol{J}^{*T}_{i|\mathrm{prec}(i)}(\overline{\boldsymbol{q}})\boldsymbol{\Lambda}_{i|\mathrm{prec}(i)}\boldsymbol{K}_{x,i}$, then the interconnected feedback system is passive for this prioritized multi-task control and the motor velocity $\dot{\boldsymbol{\theta}}$ is bounded.*

*Proof.* Given the prioritized multi-task control structure in (Corollary 3.2.2, (Sentis, 2007)), this proof is derived by following a procedure similar to that of Theorem 8.2. A few mathematical terms are augmented as below.

$$\hat{V}_4 = \frac{1}{2}\sum_{i}^{N}\int_{-\bar{d}_1}^{0}\int_{t+r}^{t} \dot{\tilde{\boldsymbol{x}}}_i^T(\xi)\boldsymbol{Q}_i\dot{\tilde{\boldsymbol{x}}}_i(\xi)d\xi dr + E_s, \tag{8.32}$$

and a potential function $\hat{\boldsymbol{l}}(\boldsymbol{q})$ is

$$\hat{\boldsymbol{l}}(\boldsymbol{q}) = \sum_{i}^{N} \boldsymbol{J}^{*T}_{i|\mathrm{prec}(i)}(\boldsymbol{q})\boldsymbol{\Lambda}_{i|\mathrm{prec}(i)}(\boldsymbol{q})\boldsymbol{K}_{x,i}\tilde{\boldsymbol{x}}_i(\boldsymbol{q}) + \boldsymbol{g}(\boldsymbol{q}),$$

then by following the same procedure, we have

$$\dot{\hat{V}} \leq \dot{\boldsymbol{\theta}}^T\big(-\boldsymbol{D}_\theta + \frac{1}{2}\bar{d}_1\sum_{i}^{N}\boldsymbol{P}_i + \frac{1}{2}\bar{d}_1\sum_{i}^{N}\boldsymbol{J}^{*T}_{i|\mathrm{prec}(i)}(\overline{\boldsymbol{q}})\boldsymbol{Q}_i\boldsymbol{J}^{*}_{i|\mathrm{prec}(i)}(\overline{\boldsymbol{q}})\big)\dot{\boldsymbol{\theta}},$$

with $\boldsymbol{P}_i = \boldsymbol{M}_i\boldsymbol{Q}_i^{-1}\boldsymbol{M}_i^T$. Then the result in Eq. (8.31) follows. □

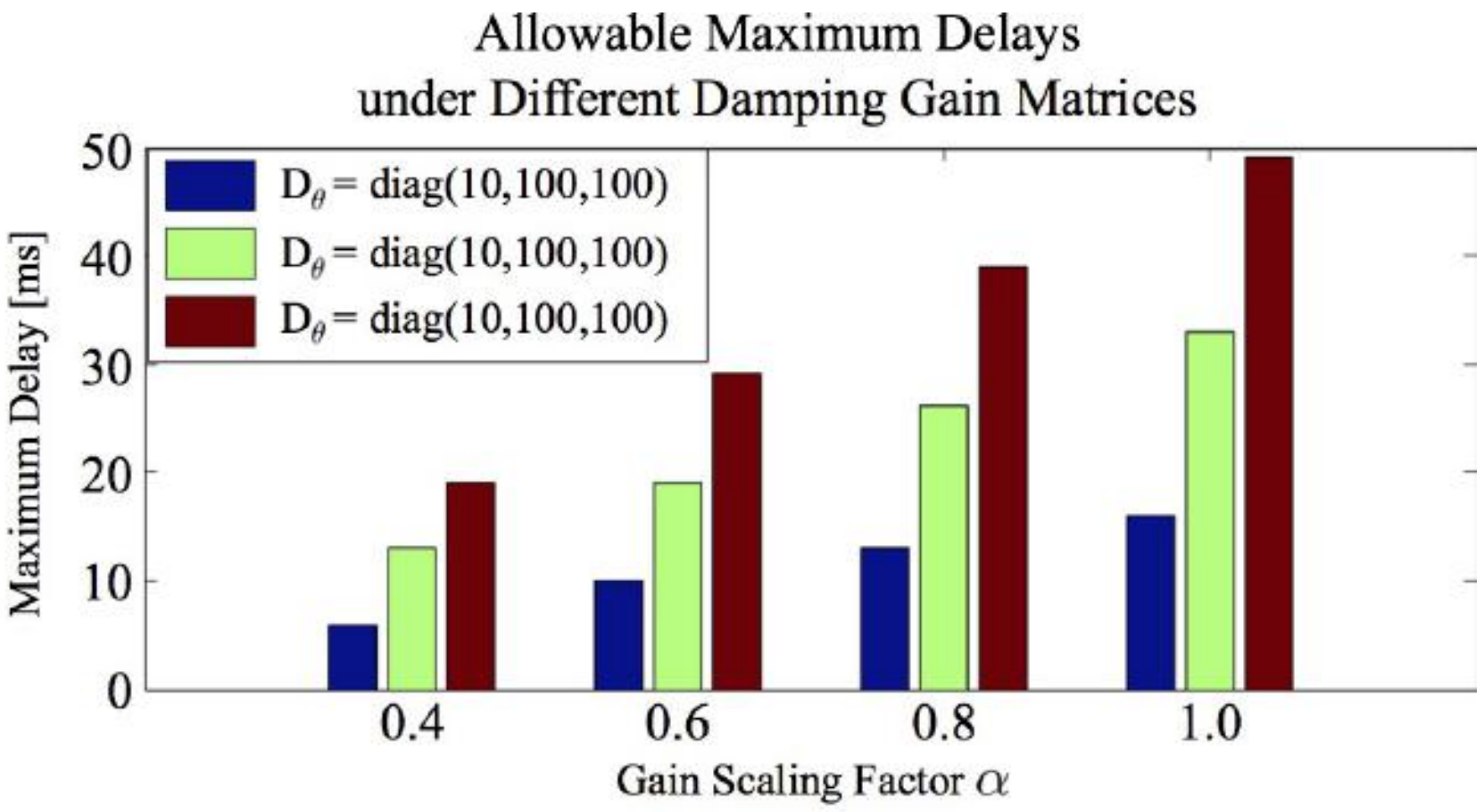

Figure 8.4: Allowable maximum delays. The motor damping gain matrix we use is $\alpha \cdot \boldsymbol{D}_\theta$, where $\alpha$ is a scaling factor. Increasing $\alpha$ enables a larger maximum delay $\bar{d}_1$. For each scaling factor $\alpha$, three tests are conducted by adjusting the first element of $\boldsymbol{D}_\theta$, i.e., the hip abduction/adduction motor damping gain. As this damping gain increases, we can achieve a larger allowable maximum delay.

## 8.4 Simulation Results

As a proof of concept, this section uses two simulations to (i) validate the proposed passivity criterion and (ii) test the torque control performance of WBOSC. The dynamic simulation adopts the recursive dynamics algorithm in (Chang, 2000) for the free floating multi-body dynamics of a point-feet bipedal robot with 3 DOF per leg. Model parameters are consistent with our Hume bipedal robot (Kim et al., 2016a). The locomotion scenario is a 7-step bipedal walking process over rough terrain in Chapter 3. For the sake of simplicity, we only concern the dynamics of one leg. The operational space task is assigned as the 6-DOF center of mass positions and orientations.

In the first simulation, we solve the allowable maximum time delay via the LMI-based passivity criterion in Eq. (8.30). We specify *a priori* the nom-inal motor damping gain matrix $\boldsymbol{D}_\theta = \text{diag}\{10, 100, 100\}$ Nms/rad for one leg (i.e., hip abduction/adduction motor, hip flexion/extension motor, and knee flexion/extension motor) and the CoM Cartesian stiffness gain matrix $\boldsymbol{K}_x = \text{diag}\{100, 100, 100, 50, 50, 50\}$ N/m for CoM positions and orientations. We numerically evaluate the system matrices $\boldsymbol{J}^*(\overline{\boldsymbol{q}})$ and $\boldsymbol{\Lambda}$ at each time instant of the entire locomotion process, and compute the maximum time delay which guarantees the feasibility of all LMIs associated with all these system matrices. A bisection algorithm is used together with the off-the-shelf Matlab LMI-optimization solver (Boyd et al., 1994) to search the maximum delay solution. Fig. 8.4 shows the maximum delays under different damping feedback gain matrices. The result indicates that as the gains in the motor damping matrix $\boldsymbol{D}_\theta$ increase, the WBOSC can tolerate larger maximum delays without becoming unstable. This result is consistent with the passivity criterion of Eq. (8.30).

In the second simulation, we test the rough terrain locomotion under time-varying delays. We choose the delays as $T_H(t) = 20 + 10\sin(t)$ ms and $T_L = 3 + 2\sin(t)$ ms. Fig. 8.5 reveals that the SEA torque command $\boldsymbol{\Gamma}_{\text{sea}}$ has a phase lag to the WBOSC command $\boldsymbol{\Gamma}_{\text{osc}}$, which is caused by the feedforward channel delay. For $\boldsymbol{\Gamma}_{\text{sea}}$ with different $\beta$, it is observed that $\boldsymbol{\Gamma}_{\text{sea}}$ with the bigger $\beta$ (black line) experiences an inferior torque transparency (i.e., a larger deviation from $\boldsymbol{\Gamma}_{\text{osc}}$). This phenomenon is explainable below: a larger $\boldsymbol{B}_s$ leads to a smaller torque feedback gain $\boldsymbol{I} - \boldsymbol{B}\boldsymbol{B}_s^{-1}$. Correspondingly, the actuator dynamics make a larger deviation of $\boldsymbol{\Gamma}_{\text{sea}}$ from $\boldsymbol{\Gamma}_{\text{osc}}$. This conclusion can also be drawn from the torque relationship in Eq. (8.14). For more details about simulated robot parameters, please refer to the result in (Kim et al., 2016a).

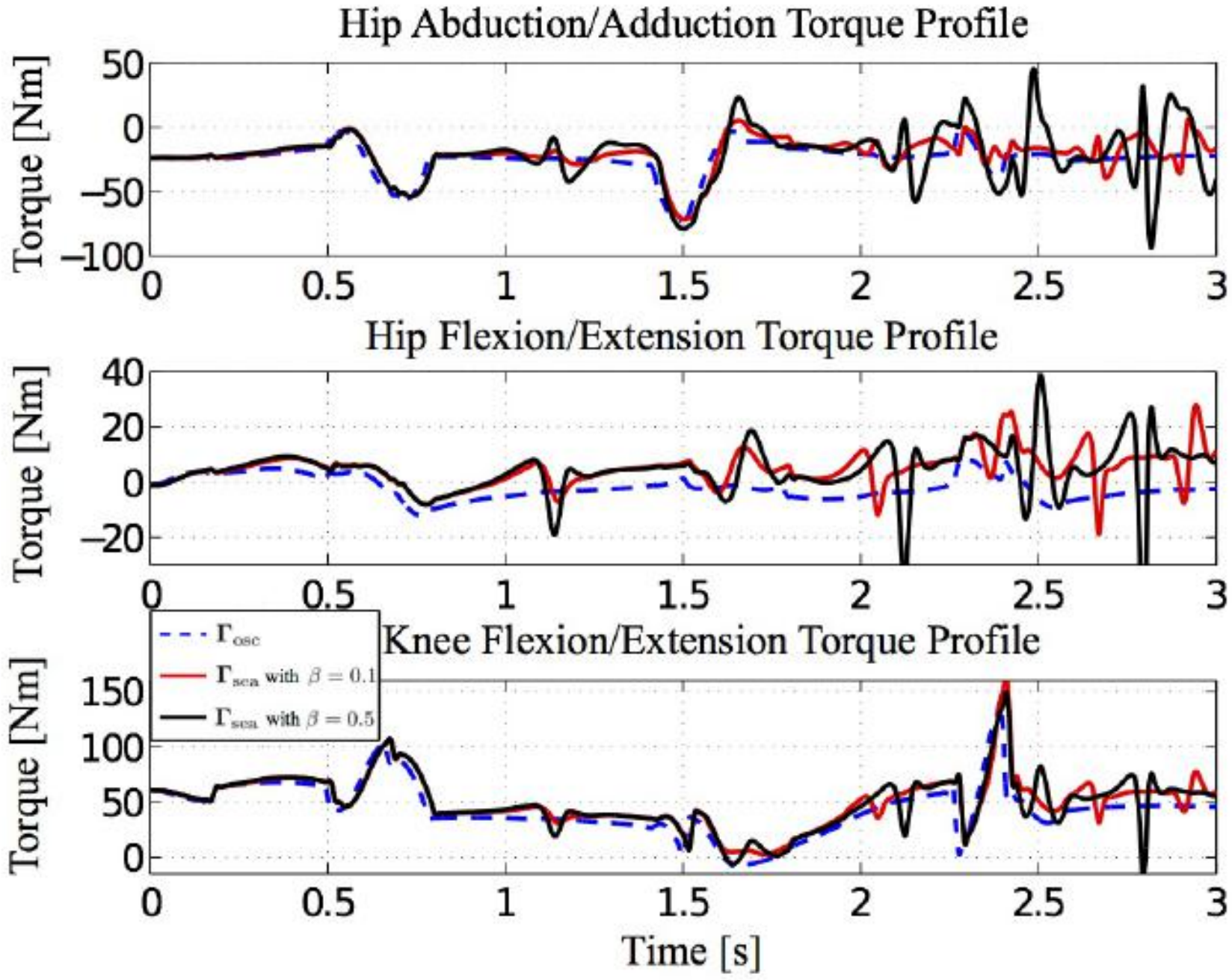

Figure 8.5: WBOSC and SEA torque profiles under different motor inertias. The shaped motor inertia matrix $\boldsymbol{B}_s = \beta \cdot \boldsymbol{B}$, where $\boldsymbol{B} = \text{diag}\{0.1357, 0.093, 0.093\}$ kg·m$^2$ is the diagonal physical motor inertia matrix of one 3-DOF Hume leg. $\beta$ denotes a scaling factor.

## 8.5 Discussions and Conclusions

In this Chapter, we propose a class of time-delayed Whole-Body Operational Space Control (WBOSC) with the series elastic actuator (SEA) dynamics. A novel Lyapunov-Krasovskii functional is designed to derive a delay-dependent LMI-based passivity criterion. This criterion is evaluated in a dynamic locomotion simulation. This line of work lays the groundwork for achieving high-performance SEA-aware WBOSC with time delays while guaranteeing the passivity.

Given the results of this Chapter, the following conclusions for the passivity-based WBOSC are reached:

- The passivity criterion in Eq. (8.30) indicates that a larger stiffness gain matrix $\boldsymbol{K}_x$ is achievable given a larger motor damping gain matrix $\boldsymbol{D}_\theta$ or a smaller time delay $\bar{d}_1$. These parameter relationships are consistent with those in (Gil et al., 2009) established for virtual stiffness, virtual damping, and time delay.

- Increasing $\boldsymbol{D}_\theta$ enhances the WBOSC passivity as shown in Eq. (8.30), but deteriorates the SEA torque transparency as shown Eq. (8.14) (i.e., $\boldsymbol{\Gamma}_{\text{sea}}$ has a larger deviation from $\boldsymbol{\Gamma}_{\text{osc}}$). This trade-off is analogous to the conflict of stability and transparency widely studied in the teleoperation community (Lawrence, 1993).

- The passivity criterion of the prioritized multi-task in Eq. (8.31) has a smaller solution range since finding multiple feasible $Q_i, i \in [1, N]$ simultaneously poses more constraints. This conclusion meets our expectation.

A few promising problems that still need to be addressed in the future are summarized below:

- Solve the time-varying system matrices of the passivity criteria in a more elegant way, such as the methods of solving time-varying LMIs online in (Guo and Zhang, 2014) and convexifying matrices as a polytope of parametric matrices (Gahinet et al., 1996);

- Model more realistic round trip delays, such as asymmetric delays in feedforward and feedback channels. Literature from teleoperation and network communication fields (Lee and Spong, 2006; Hua and Liu, 2010) will be instructive references;

- Validate the criterion in Eq. (8.31) for the prioritized multi-task control;

- Design an optimal controller for the centralized-level WBOSC, and establish the mapping between the centralized-level and embedded-level damping gain matrices. (Braun et al., 2013) is a good reference;

- Conduct experimental validations on our legged robot (Kim et al., 2016a) and mobile manipulator (Sentis et al., 2013).

- Extend this work to more sophisticated task scenarios, such as balancing, mobility, interaction with humans, collision avoidance and so on.

# Chapter 9

# Discussions, Conclusions and Future Works

### Dissertation Summary

This dissertation presents a planning, control, and decision-making framework for humanoid robots with agile mobility behaviors and high performance. A particular research concentration is on the robust, optimal and real-time performance. We present these works in the following three parts.

First, we present a phase-space planning framework for dynamic humanoid robot locomotion over rough terrain in Chapters 3-4. Based on a centroidal momentum model, Chapter 3 formulates a hybrid planning framework including a series of pivotal components: a step transition solver, a robust hybrid automaton, and a steering direction model. In Chapter 4, we define a new class of locomotion phase-space manifolds, used as a Riemannian distance metric to measure deviations resulting from external disturbances or model uncertainties. To achieve robust locomotion, we propose a hybrid control method composed of a dynamic programming control based on the robust metric and a closed-form solution of discrete foot placement re-planning. The robust capabilities of our proposed framework are illustrated in (i) locomotion simulations of dynamic maneuvers over diverse challenging terrains and under external dis-turbances; (ii) experimental implementations of the dynamic balancing on our point-feet bipedal robot. Our objective is to provide a general motion planning framework for dynamic rough terrain locomotion, which can be leveraged to other legged robot platforms.

Second, Chapter 5 takes a step toward formally synthesizing high-level reactive planners for the whole-body locomotion in constrained environments epitomized by rough terrains, crashed terrain and sudden human appearance in the scene. We formulate a two-player temporal logic game between the contact planner and its possibly-adversarial environment. The resulting discrete planner satisfies the given task specifications expressed as a fragment of temporal logic and is executed by a low-level phase-space planner. Given centroidal momentum dynamics, we devise various low-level locomotion modes associated with specific contact configurations. The provable correctness of the low-level execution of the synthesized discrete planner is guaranteed through the so-called simulation relations. Simulations of dynamic locomotion in constrained environments support the effectiveness of the outlined hierarchical planner protocol. We conjecture that this theoretical advance acts as an entry point for the humanoid community to employ formal methods to verify and synthesize high-level task planners.

Third, we propose a distributed control architecture for the latency-prone humanoid robotic systems in Chapter 6. The distributed impedance controllers are analyzed where damping feedback effort is executed in proximity to the control plant, and stiffness feedback effort is implemented in a latency-prone centralized control process. We pursue a detailed analysis of a central observation that the stability of high impedance distributed controllers is highly sensitive to damping feedback delay but much less to stiffness feedback delay. As to series elastic actuators (SEAs) in Chapter 7, we propose a critically-damped gain selection criterion for a SEA cascaded control structure with inner torque and outer impedance feedback loops. A trade-off between inner torque gains and outer impedance gains is observed. Our study indicates that the proposed gain design criterion solves optimal controller gains by maximizing system stability. Based on this controller design, we analyze the effects of feedback delays and filtering on SEA impedance performance in the frequency domain. We propose a novel impedance performance metric, defined as "Z-region", which simultaneously quantifies achievable impedance magnitude range (Z-width) and frequency range (Z-depth). Maximizing Z-region enables SEA-equipped robots to achieve a broad range of Cartesian impedance tasks without alternating the control structure. Then Chapter 8 generalize this distributed control strategy to time-delayed Whole-Body Operational Space with SEA dynamics. A delay-dependent passivity criterion is proposed by treating the overall system as two interconnected subsystems in a feedback configuration. The distributed control strategy is validated through experimental validations on a UT rigid actuator, a series elastic actuator, an omnidirectional mobile base and a bipedal robot. This line of work opens the way to design distributed feedback controllers incorporating practical factors.

**Discussions and Future Work**

Although a series of simulations and experiments have been presented throughout this dissertation, there is huge room for improvements as the subject of future work. Aside from the future work mentioned in the Conclusion and Discussion Sections of previous chapters, we also would like to put our effort on the following aspects:

- (i) Extensive experimental validations of our rough terrain planning strategy. Although a modified phase-space planner has been implemented on our Hume biped robot (Kim et al., 2016a), more work undoubtedly needs to be implemented to achieve a general framework for legged locomotion in 3D rough terrains. When reference trajectories are applied to a real robot, modeling errors, sensor noise and external perturbations will likely cause the robot to deviate from planned trajectories. All of these errors require robust planners based on real-time sensory feedback.

- (ii) Study robot adaptation to human collisions. A good reference is shown (Kim et al., 2016b). Guiding and assisting humans traversing through confined environments becomes more important topics, especially in rescue environments. How to modulate the impedance between humans and robots properly are interesting topics to be explored.

- (iii) During multi-contact phases, we use the polynomial fitting to smoother finite time phase transition maneuvers. In the future we plan to introduce dynamics based multi-contact transition maneuvers. Specifically when using point contacts, multi-contact dynamics have passive modes that have been ignored so far. Also, till now the contacts are assumed only to occur at foot contact points. However, they could occur at any points on the robot body which is an interesting problem for both interaction control and re-planning.

- (iv) For the wedge jumping scenario, we would like to explore how prioritized control of internal forces, linear and angular momentum, affects extreme locomotion capabilities and performance.

- (v) In-depth phase-space reachability analysis should be analyzed by incorporating robot physical and environmental constraints, as well as computational speed and delays restrictions. We target increasing the flexibility of our planning strategy to more general constrained and unconstructed environments.

- (vi) An interesting scientific topic is to evaluate the physical limits of extreme locomotion given limited control capabilities. It will be attractive to quantify walking, running or climbing efficiency, and maximum jumping capabilities.

- (vii) An open question worthy to think about in the long-term is what type of motion planning framework is capable of generating behaviors of a team of humanoid and mobile robots, such as legged robots, manipulators and quadrotors. It is highly demanding to propose a generalized planning framework, "Systems of Systems". The planner should reason at not only system-level but also individual-robot-level. Entropy-based optimal control in (Atanasov, 2015) provides promising insights.

Robustness is an essential property of humanoid control and planning in the presence of model uncertainties and external disturbances. It can be reasoned from various perspectives. In this dissertation, we primarily quantify the robustness within three different layers, from the actuator-level control to the locomotion phase-space planner to the high-level task planner.

- As for the actuator-level control, the proposed high-impedance control reveals robust tracking performance in the face of unmodeled load dynamics. Despite no explicit explorations, the passivity property in Chapter 8 intrinsically enhances the system robustness with respect to unmodeled robot dynamics and unknown passive environments (Albu-Schäffer et al., 2007; Ott et al., 2008). Other prior works on passivity-based robust control are referred to (Spong et al., 2006; Ortega et al., 2013). In the future, we are interested in exploring the parametric robustness (Zhou et al., 1996).

- The phase-space motion planner defines invariant and recoverability bundles to quantify control-dependant robust phase-space regions. Given these robust bundles, we propose a Riemannian metric to quantify the phase-space deviation and explicitly design a robust optimal control strategy for the nominal manifold recovery when external disturbances occur. The resulting phase-space manifolds are sequentially composed to robustly track a set of keyframe states. Relevant works on robust motion planning are in (Lozano-Perez et al., 1984; Latombe et al., 1991; Majumdar, 2016). A key distinction of our work is that the under-actuated nonlinear robot dynamics are dominant, and thus we focus on reasoning the robustness in phase-space instead of configuration space. In the fine motion planning literature, sequential composition of funnels (Burridge et al., 1999) or preimage backchaining (Lozano-Perez et al., 1984) were introduced to produce dynamic robot behaviors. The notion of a funnel is introduced as a metaphor for the robustness of robot behaviors.

- As to the high-level task planner, the contact planner does not know the dynamical environment *a priori*, although a set of admissible environmental events is predefined. The synthesized reactive planner guarantees proper contact actions according to the sensed upcoming environment events at runtime. This reactive mechanism is treated as a robustness property to unforeseen environment events, i.e. the planner always reacts in a correct way whatever the adversarial environment events are. In the future, we will introduce an interfacing layer (He et al., 2015; Dantam et al., 2016) to allow communications between the task and motion planners. In that case, the task planner can detect the motion planner failures and re-plan a new command at runtime.

# Appendices

# Appendix A

# Phase-Space Manifold

The desired behavior of the outputs lie in manifolds $\mathcal{M}_i$ as shown in Eqs. (3.7). Here we present a brief review of space curves and surfaces that relate to the phase-space manifold (PSM) and present a Riemannian geometry metric that can be generalized to this family of problems.

The trajectory of the center of mass (CoM) of the robot is a space curve in 3D, $\boldsymbol{p}_{\mathrm{CoM}} = \{x, y, z\} \in \mathbb{R}^3$. Also, for a particular output $y_i$ (i.e., one element of Eq. (3.1b)), if we consider the case of an output-task with relative order $r_i = 3$, the manifold $\mathcal{M}_i \in \mathbb{R}^{r_i}$ is a space curve $\mathcal{C}_i$ in Euclidean three-dimensional space (see Fig. A.1). We assume that the curve is parametrized by an arc-length parameter $\zeta$ that we refer to as the progression variable. Hence the position vector $\boldsymbol{\rho}_i$ of any point on the curve can be defined by specifying the value of $\zeta$,

$$\boldsymbol{\rho}_i(\zeta) = \sum_{k=1}^{r_i} \xi_k(\zeta)\mathbf{E}_k = \xi_1(\zeta)\mathbf{E}_1 + \xi_2(\zeta)\mathbf{E}_2 + \xi_3(\zeta)\mathbf{E}_3. \tag{A.1}$$

where, $\mathbf{E}_k$ is the unit vector in the $k$-axis of the Euclidean space and $\xi_k(\zeta)$ is the projection coordinate of $\boldsymbol{\rho}$ on $\mathbf{E}_k$. A unit tangent vector $\mathbf{e}_t$ to the curve can also be defined,

$$\mathbf{e}_t = \frac{\partial \boldsymbol{\rho}_i}{\partial \zeta}. \tag{A.2}$$

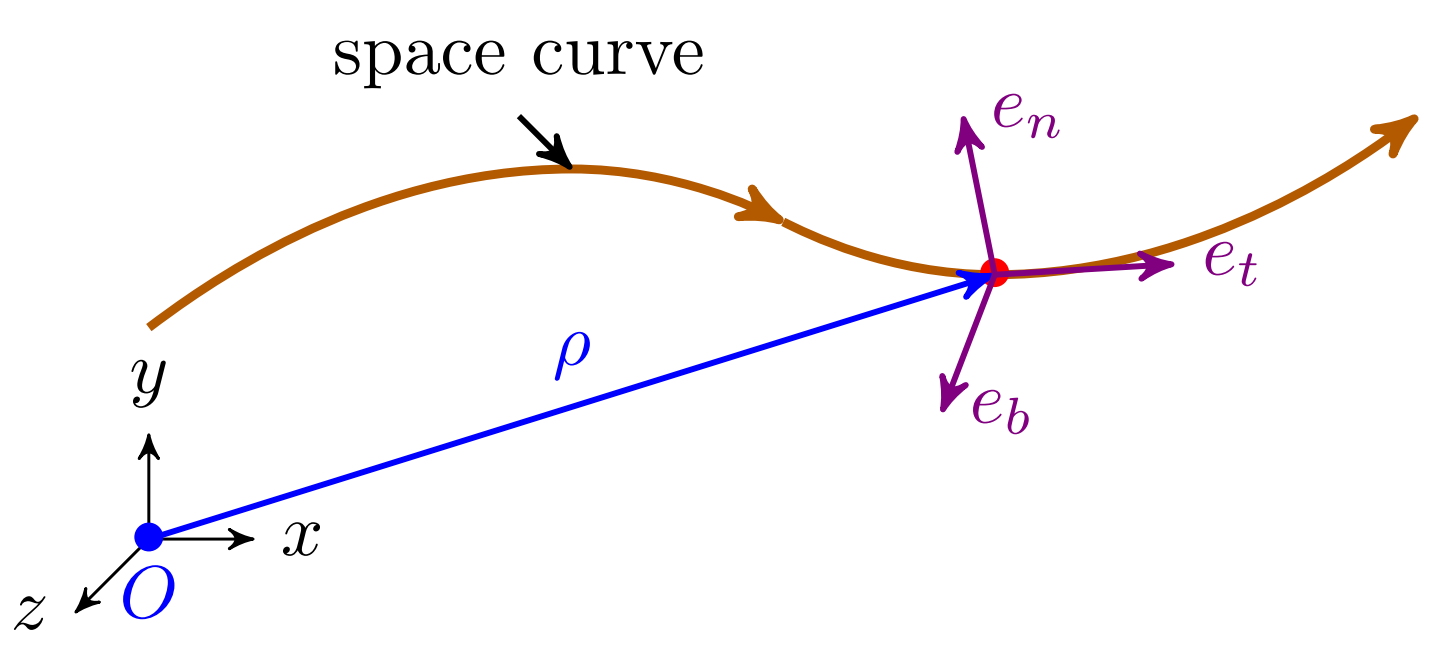

Figure A.1: A space curve showing the evolution of the Frenet triad.

The derivative of this vector defines the curvature $\kappa$ and the unit normal vector $\mathbf{e}_n$,

$$\frac{\partial \mathbf{e}_t}{\partial \zeta} = \kappa \mathbf{e}_n, \quad \text{where} \quad \kappa = \left\| \frac{\partial \mathbf{e}_t}{\partial \zeta} \right\|. \tag{A.3}$$

In the case of a space surface (where, $\boldsymbol{\rho}_i$ belongs to a manifold $\mathcal{M}_i$ in the output phase-space), instead of a vector $\mathbf{e}_t$, we have a tangent manifold, denoted by $T_{\mathcal{M}_i}$. The tangent space at any point can be mapped to the vector $\boldsymbol{\chi}_i \in \mathbb{R}^{r_i - 1}$ that spans $T_{\mathcal{M}_i}$. Without loss of generality, the actual motion is a specific line in space curve $\mathcal{M}_i$. The tangent vector in the manifold $\mathcal{M}_i$ is $\boldsymbol{e}_\zeta = (\mathbf{e}_i)_t$ while the cotangent vector in the manifold is $\mathbf{e}_\sigma = (\mathbf{e}_i)_n$. $\mathbf{e}_\sigma$ denotes the normal deviation distance from the surface $\sigma_i$. For $r_i = 3$, the binormal vector, $\mathbf{e}_b$ is orthogonal to $\mathbf{e}_t$ and $\mathbf{e}_n$. These three vectors are called the Frenet space. These three Frenet frame vectors are proportional to the first three derivatives of the curve $\boldsymbol{\rho}$, as a benefit of taking the arc length $\zeta$ as the parameter.

In disturbance-free cases, the system will remain in the manifold if it starts on it. It can be considered as the zero dynamics of the surface deviation, $\sigma_i$.

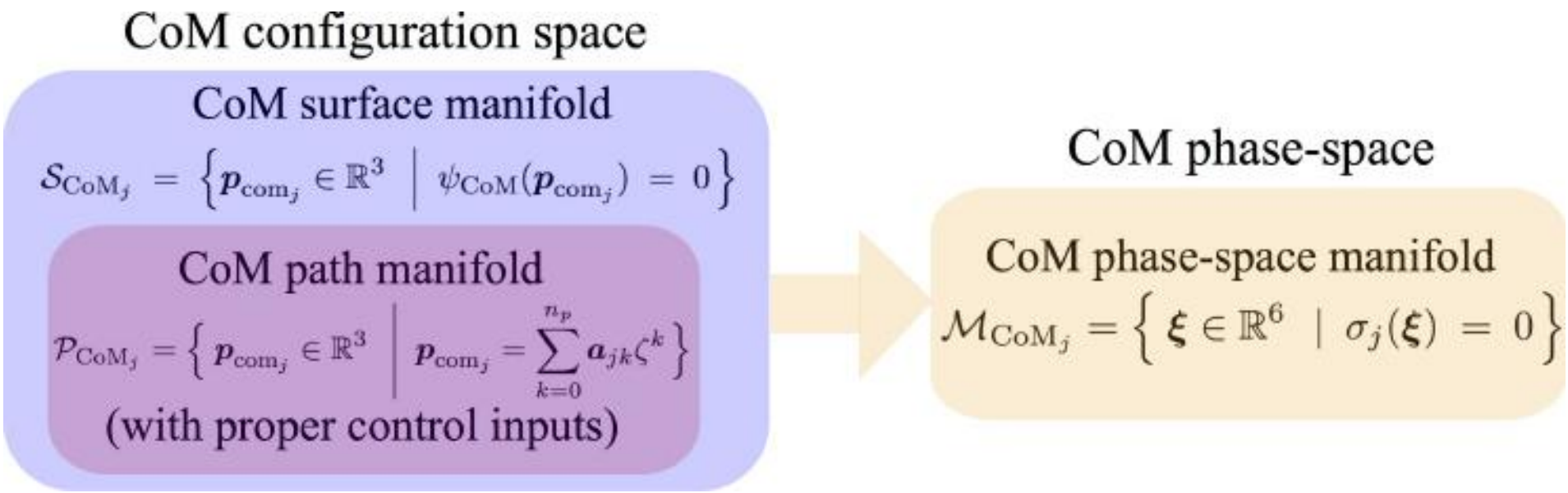

Figure A.2: Relationship between the CoM configuration space manifolds and the CoM phase-space manifold.

When disturbance occurs, the state may escape the manifold and the controller should bring it back for recovery. To define a metric on the manifold itself and normal to it, we use Riemannian Geometry. In general, we treat each manifold $\mathcal{M}_i$ in Eq. (3.7) of the task-space $i^{\mathrm{th}}$-coordinate as independent from each other. The actual task manifold, $\mathcal{M}$ is the intersection of all $\mathcal{M}_i$ manifolds,

$$\mathcal{M} = \bigcap_i \mathcal{M}_i. \tag{A.4}$$

This manifold also has a tangent manifold $T_{\mathcal{M}} \in \mathbb{R}^r$, where $r = \sum_i r_i$. Each manifold $\mathcal{M}_i$, separately have a null-space (cotangent manifold, $T^*_{\mathcal{M}_i}$) and their intersection is the task cotangent manifold, $T^*_{\mathcal{M}}$.

In relation to the manifolds presented in Section 3.2, the relationship between CoM configuration manifolds and CoM phase-space manifolds can be visualized via the set diagram of Fig. A.2.

## A.1 Proof of Tangent Phase-Space Manifold

In the desired case, angular momentum $\tau_y$ in the Eq. (3.14) is zero. Sagittal inverted pendulum dynamics are simplified to $\ddot{x} = \omega^2(x - p_x)$, where $\omega$ is constant within one step. Since foot placement $p_x$ is also a constant in one step, thus $\dot{p}_x = 0, \ddot{p}_x = 0$. The equation above is equivalent to $\ddot{x} - \ddot{p}_x = \omega^2(x - p_x)$. Now let us define a transformation $\tilde{x} = x - p_x$, then we have $\ddot{x}' = \omega^2\tilde{x}$. By Laplace transformation, we have $s^2x(s) - \tilde{x}_0 - s\dot{\tilde{x}}_0 = \omega^2 x(s)$, where $\tilde{x}_0 = x_0 - p_x$, $(x_0, \dot{x}_0)$ is an initial condition. Then we have

$$x(s) = \mathscr{L}^{-1}\{\frac{\tilde{x}_0 + s\dot{\tilde{x}}_0}{s^2 - \omega^2}\} \tag{A.5}$$

Solving this equation, we have the analytical solution

$$\tilde{x}(t) = \frac{\tilde{x}_0(e^{\omega t} + e^{-\omega t})}{2} + \frac{\dot{\tilde{x}}_0(e^{\omega t} - e^{-\omega t})}{2\omega} = \tilde{x}_0\text{cosh}(\omega t) + \frac{1}{\omega}\dot{\tilde{x}}_0\text{sinh}(\omega t) \tag{A.6}$$

and further

$$\dot{\tilde{x}}(t) = \frac{\omega\tilde{x}_0(e^{\omega t} - e^{-\omega t})}{2} + \frac{\dot{\tilde{x}}_0(e^{\omega t} + e^{-\omega t})}{2} = \omega\tilde{x}_0\text{sinh}(\omega t) + \dot{\tilde{x}}_0\text{cosh}(\omega t) \tag{A.7}$$

since $\tilde{x}_0 = x_0 - p_x, \dot{\tilde{x}}_0 = \dot{x}_0 - \dot{p}_x = \dot{x}_0, \dot{\tilde{x}} = \dot{x} - \dot{p}_x = \dot{x}$, we have

$$x(t) = (x_0 - p_x)\text{cosh}(\omega t) + \frac{1}{\omega}\dot{x}_0\text{sinh}(\omega t) + p_x \tag{A.8}$$

$$\dot{x}(t) = \omega(x_0 - p_x)\text{sinh}(\omega t) + \dot{x}_0\text{cosh}(\omega t) \tag{A.9}$$

Now we have the following state space formulation

$$\begin{pmatrix} x(t) - p_x \\ \dot{x}(t) \end{pmatrix} = \begin{pmatrix} x_0 - p_x & \dot{x}_0/\omega \\ \dot{x}_0 & \omega(x_0 - p_x) \end{pmatrix} \begin{pmatrix} \text{cosh}(\omega t) \\ \text{sinh}(\omega t) \end{pmatrix} \quad \Rightarrow$$

$$\begin{pmatrix} \text{cosh}(\omega t) \\ \text{sinh}(\omega t) \end{pmatrix} = \frac{1}{\omega(x_0 - p_x)^2 - \dot{x}_0^2/\omega} \begin{pmatrix} \omega(x_0 - p_x) & -\dot{x}_0/\omega \\ -\dot{x}_0 & x_0 - p_x \end{pmatrix} \begin{pmatrix} x - p_x \\ \dot{x} \end{pmatrix}$$

since $\cosh^2(x) - \sinh^2(x) = 1$, one has

$$(\omega(x_0 - p_x)(x - p_x) - \dot{x}_0\dot{x}/\omega)^2 - (-\dot{x}_0(x - p_x) + \dot{x}(x_0 - p_x))^2 = (\omega(x_0 - p_x)^2 - \dot{x}_0^2/\omega)^2 \tag{A.10}$$

After expanding the square terms, manipulating and moving all terms to one side, we obtain

$$(x_0 - p_x)^2\Big(2\dot{x}_0^2 - \dot{x}^2 + \omega^2(x - x_0)(x + x_0 - 2p_x)\Big) - \dot{x}_0^2(x - p_x)^2 + \dot{x}_0^2(\dot{x}^2 - \dot{x}_0^2)/\omega^2 = 0 \tag{A.11}$$

which is defined as the phase-space manifold $\sigma(x, \dot{x}, x_0, \dot{x}_0, p_x)$ in Proposition 4.1.

## A.2 Proof of Cotangent Phase-Space Manifold

Taking the derivative of Eq. (4.2), we have

$$d\sigma = \frac{\partial\sigma}{\partial x}dx + \frac{\partial\sigma}{\partial\dot{x}}d\dot{x}, \quad \text{where,} \quad \frac{\partial\sigma}{\partial x} = -2\dot{x}_{\text{apex}}^2(x - x_{\text{foot}}), \quad \frac{\partial\sigma}{\partial\dot{x}} = 2\dot{x}_{\text{apex}}^2\dot{x}/\omega^2.$$

The $\sigma$-normal vector is given by it's gradient, $\mathbf{e}_n = \big(-2\dot{x}_{\text{apex}}^2(x - x_{\text{foot}})\ ,\ 2\dot{x}_{\text{apex}}^2\dot{x}/\omega^2\big)^T$, and the $\sigma$-tangent vector is orthogonal, $\mathbf{e}_t = \big(2\dot{x}_{\text{apex}}^2\dot{x}/\omega^2\ ,\ 2\dot{x}_{\text{apex}}^2(x - x_{\text{foot}})\big)^T$. Since $\zeta$ is orthogonal to $\sigma$, the tangent vector of $\zeta$ is the normal vector of $\sigma$, i.e.,

$$d\zeta = \frac{\partial\zeta}{\partial x}dx + \frac{\partial\zeta}{\partial\dot{x}}d\dot{x}, \quad \text{with} \quad \frac{\partial\zeta}{\partial x} = 2\dot{x}_{\text{apex}}^2\dot{x}/\omega^2, \quad \frac{\partial\zeta}{\partial\dot{x}} = -2\dot{x}_{\text{apex}}^2(x - x_{\text{foot}})$$

A $\zeta$-isoline (constant $\zeta$) requires $d\zeta = 0$, which further implies

$$\frac{d\dot{x}}{dx} = -\frac{\dot{x}}{\omega^2(x - x_{\text{foot}})} \quad \Rightarrow \quad \omega^2\int_{\dot{x}_0}^{\dot{x}}\frac{d\dot{x}}{\dot{x}} = -\int_{x_0}^{x}\frac{dx}{x - x_{\text{foot}}} \tag{A.12}$$

then we have

$$\ln(\frac{\dot{x}}{\dot{x}_0})^{\omega^2} + \ln\frac{x - x_{\text{foot}}}{x_0 - x_{\text{foot}}} = 0 \quad \Rightarrow \quad (\frac{\dot{x}}{\dot{x}_0})^{\omega^2}\frac{x - x_{\text{foot}}}{x_0 - x_{\text{foot}}} = 1 \tag{A.13}$$

Thus, we can define the cotangent manifold as

$$\zeta = \zeta_0(\frac{\dot{x}}{\dot{x}_0})^{\omega^2}\frac{x - x_{\text{foot}}}{x_0 - x_{\text{foot}}} \tag{A.14}$$

where the constant $\zeta_0$ is a nonnegative scaling factor. In this case, $(x_0, \dot{x}_0)$ is an initial condition where $\zeta = \zeta_0$. $(x_0, \dot{x}_0)$ cannot be the apex condition, we could use $(x_0, \dot{x}_0) = (\dot{x}_{\text{apex}}/\omega, \dot{x}_{\text{apex}})$, which is a point in the asymptote. All of these three terms, i.e., $\zeta_0, x_0$ and $\dot{x}_0$, do not affect the orthogonality of the tangent and cotangent manifolds. They are chosen such that $\zeta$ values are scaled to reasonable values within one walking step (e.g., $\zeta \in [0, 1]$).

# Appendix B

# PIPM Locomotion Dynamics

## B.1 Derivation of PIPM Dynamics

We first expand Eq. (3.13) as follows

$$(z - z_{\text{foot}_q}) \cdot m\ddot{x} = (x - x_{\text{foot}_q}) \cdot m(\ddot{z} + g) - \tau_y, \tag{B.1}$$

$$-(z - z_{\text{foot}_q}) \cdot m\ddot{y} = (y - y_{\text{foot}_q}) \cdot m(\ddot{z} + g) - \tau_x, \tag{B.2}$$

$$(x - x_{\text{foot}_q}) \cdot m\ddot{y} = (y - y_{\text{foot}_q}) \cdot m\ddot{x} - \tau_z. \tag{B.3}$$

Given the 3D surface in Eq. (3.16), we differentiate it twice and obtain

$$\ddot{z} = a_q\ddot{x} + b_q\ddot{y}. \tag{B.4}$$

Substituting Eq. (B.4) into Eqs. (B.1) and (B.2), we have

$$(a_q x + b_q y + c_q - z_{\text{foot}_q})\ddot{x} - (x - x_{\text{foot}_q})(a_q\ddot{x} + b_q\ddot{y} + g) + \tau_y/m = 0, \tag{B.5}$$

$$(a_q x + b_q y + c_q - z_{\text{foot}_q})\ddot{y} - (y - y_{\text{foot}_q})(a_q\ddot{x} + b_q\ddot{y} + g) + \tau_x/m = 0. \tag{B.6}$$

Combining Eqs. (B.3) and (B.5), we obtain

$$\ddot{x} = \frac{(x - x_{\text{foot}_q})g}{a_q x_{\text{foot}_q} + b_q y_{\text{foot}_q} + c_q - z_{\text{foot}_q}} - \frac{\tau_y + b_q\tau_z}{m(a_q x_{\text{foot}_q} + b_q y_{\text{foot}_q} + c_q - z_{\text{foot}_q})}. \tag{B.7}$$

Combining Eqs. (B.3) and (B.6), we obtain

$$\ddot{y} = \frac{(y - y_{\text{foot}_q})g}{a_q x_{\text{foot}_q} + b_q y_{\text{foot}_q} + c_q - z_{\text{foot}_q}} - \frac{\tau_x + a_q\tau_z}{m(a_q x_{\text{foot}_q} + b_q y_{\text{foot}_q} + c_q - z_{\text{foot}_q})}. \tag{B.8}$$

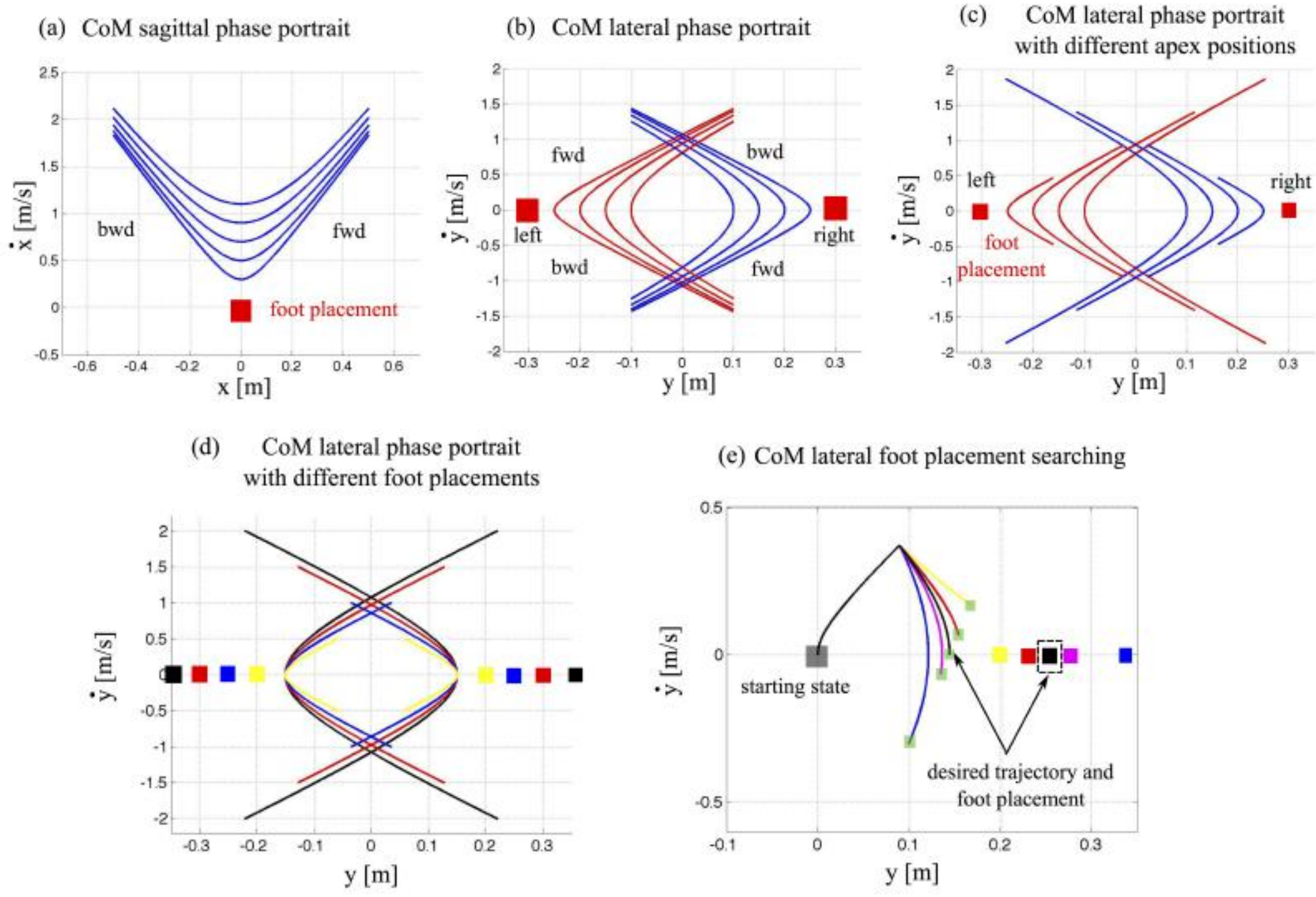

Figure B.1: Numerical integration of sagittal and lateral dynamics. Numerical Integration is used for phase-space prediction of sagittal and lateral prismatic inverted pendulum dynamics in Eq. (3.14).

By combining Eqs. (B.4), (B.7) and (B.8), $\ddot{z}$ can be derived accordingly. By defining a phase-space asymptotic slope $\omega_q$ as shown in Eq. (3.15), we can obtain Eq. (3.14) for the prismatic inverted pendulum based control system.

## B.2 Numerical Integration

For the nonlinear Eq. (3.14), we assume that $\ddot{x}$ is approximately constant for small increments of time. Since $f$ is normally highly nonlinear, numerical integration algorithms are normally used. More details are explained in (Sentis and Slovich, 2011). The pipeline for finding state-space trajectories goes as follows: (1) choose a very small time perturbations $\epsilon$, (2) given known velocities $\dot{x}_k$ and accelerations $\ddot{x}_k$, we get the next velocity $\dot{x}_{k+1}$, (3) get the next position $x_{k+1}$, (4) plot the points $(x_{k+1}, \dot{x}_{k+1})$ in the phase-plane. We also notice, that we can iterate this recursion both forward and backward. If we iterate backward we then need to choose a negative perturbation $\epsilon$. Based on this pipeline, one-step sagittal and lateral manifold with different initial conditions are shown in Fig. B.1. The phase diagrams (a) and (b) correspond to the sagittal and lateral CoM phase behaviors given desired foot contact locations (red boxes), a desired CoM surface of motion, and initial position and velocity conditions. If we consider timing issues on the lateral plane as discussed in Chapter 4, we can derive two different trajectories shown in (c) and (d). (c) shows lateral CoM behaviors given a fixed lateral foot placement and varying starting conditions. (d) corresponds to CoM trajectories derived given varying lateral foot placements and a fixed starting conditions. In (e), we analyze lateral CoM trajectories of two consecutive steps with varying lateral foot placements. As the foot placement moves further apart, the acceleration becomes larger and the CoM position transverses (at the apex point) less in the y direction.

## B.3 Multi-Contact Maneuvers

Our objective here is to incorporate multi-contact transitions into our gait planner to make it look more natural. To this end, we cut out a portion of the phase curves around step transitions and fit a $5^{\text{th}}$-order polynomial with the desired smooth behavior. In this fitting, desired boundary values of position, velocity and acceleration are endowed by the gait designer. More importantly, it is needed to also take into account time constraints in such a way that the sagittal and lateral behaviors are exactly synchronized. Boundary and timing conditions allow us to calculate the coefficients of the polynomials. For more mathematical derivations, please refer to (Zhao and Sentis, 2012).

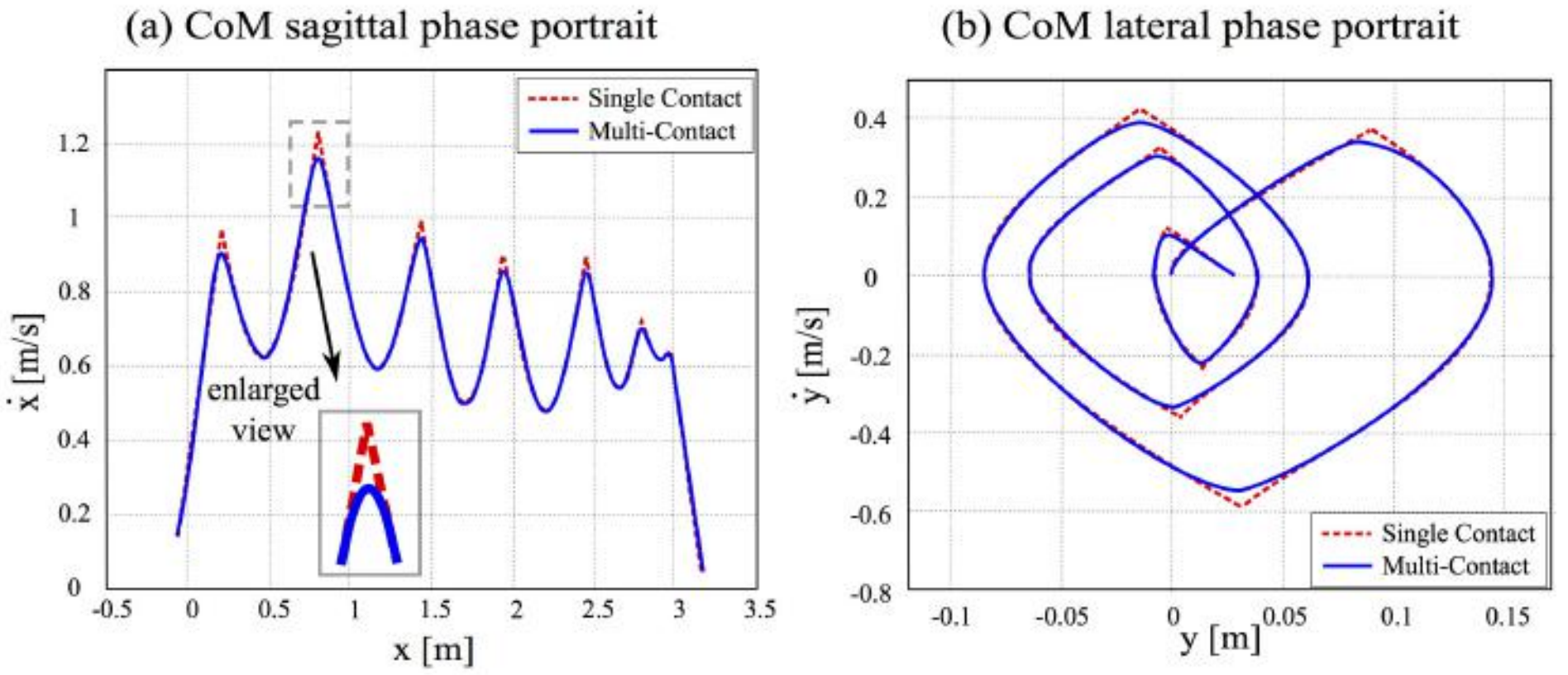

Figure B.2: Integration of multi-contact transition phases. The plots (a) and (b) are similar to their counterparts of Fig. 3.2 but with an addition of a multi-contact phase. A user decides the duration of the multi-contact phase with respect to the overall step and then chooses the velocity and acceleration profile during multi-contact.

In our case, we design a multi-contact phase that takes place during 25% of the time of any given step. We show the result in Fig. B.2. The dotted rectangle in plot (a) of the previous figure depicts the time window for multi-contact. This percentage is adjustable to other values based on the desired walking profile. To determine the feasibility of the curves we extract internal forces using the multi-contact-grasp matrix in Section 3.3.2 and then determine if they are feasible given surface friction constraints.

# Appendix C

# Robust Hybrid Automaton

Mathematical notations of the robust hybrid automaton in Section 3.5.1 are described in detail as

- $\zeta$ is the phase-space progression variable;
- $\mathcal{Q}$ is the set of discrete states;
- $\mathcal{X}$ is the set of continuous states. The system state is augmented to $\boldsymbol{s} := \zeta \times \mathcal{Q} \times \mathcal{X}$ in a hybrid state space;
- $\mathcal{U} := \{\boldsymbol{u}_q, q \in \mathcal{Q}\}$, is the set of control inputs. $\mathcal{U} = \{\boldsymbol{u}_c\} \cup \{\boldsymbol{u}_d\}$ where $\boldsymbol{u}_c, \boldsymbol{u}_d$ are continuous and discrete control inputs, respectively;
- $\mathcal{W}$ is the set of disturbances;
- $\mathcal{F}$ is the vector field, with $\mathcal{F} : \zeta \times \mathcal{Q} \times \mathcal{X} \times \mathcal{U} \times \mathcal{W} \rightarrow T_{\mathcal{X}}$, where $T_{\mathcal{X}}$ is tangent bundle of $\mathcal{X}$;
- $\mathcal{I} := \zeta \times \mathcal{Q} \times \mathcal{X}$, is the initial condition;
- $\mathcal{D}(q) : \mathcal{Q} \rightarrow 2^{\mathcal{X}}, q \in \mathcal{Q}$, is the domain[1];
- $\mathcal{R} := \{\mathcal{R}_q, q \in \mathcal{Q}\}$, is the collection of recoverability sets;

[1] $2^{\mathcal{X}}$ represents the power set (all the subsets) of $\mathcal{X}$.

- $\mathcal{B} := \{\mathcal{B}_q, q \in \mathcal{Q}\}$, is the collection of invariant bundles;
- $\mathcal{E}(q, q+1) := \mathcal{Q} \times \mathcal{Q}$, is the edge;
- $\mathcal{G}(q, q+1) : \mathcal{Q} \times \mathcal{Q} \to 2^{\mathcal{X}_q}$ is the guard, which is abbreviated as $\mathcal{G}_{q \to q+1}$; $\mathcal{G}(q, q+1) = \cup_{\tau,\mu} \mathcal{G}_{\mu}^{[\tau]}$, is the transition set[2] (notations are shown in Table 3.1);
- $\mathcal{T}(q, q+1) : \mathcal{Q} \times \mathcal{Q} \to 2^{\mathcal{X}_{q+1}}, q, q+1 \in \mathcal{Q}$, is the transition termination set;
- $\Delta^{[\tau]}_{\mu(q \to q+1)}(\boldsymbol{s}_q^-, \boldsymbol{u}_q^-, w_d^-)$, is the transition map (see Eq. C.1).

Based on the automaton above, the hybrid system can be represented by

$$\Sigma_q : \begin{cases} \{\boldsymbol{\mathcal{F}}^+, \boldsymbol{u}^+_{q+1}, \boldsymbol{x}^+_{q+1}\} & \leftarrow \quad \Delta^{[\tau]}_{\mu(q \to q+1)}(\boldsymbol{s}_q^-, \boldsymbol{u}_q^-, w_d^-), \\ & \quad \text{if } \ (\boldsymbol{s}_q^-, \boldsymbol{u}_q^-, w_d^-) \in \mathcal{G}_{\mu}^{[\tau]}(q, q+1) \\ \dot{\boldsymbol{x}}_q = \boldsymbol{\mathcal{F}}(q, \boldsymbol{x}_q, \boldsymbol{u}_q, w_d), & \quad \text{otherwise} \end{cases} \tag{C.1}$$

where, $\boldsymbol{s}_q(\zeta_q) = (\zeta_q, q, \boldsymbol{x}_q)$ is the hybrid automaton state. This automaton has non-periodic orbits, since our planning focuses on irregular terrain locomotion. A directed diagram of this non-periodic automaton is shown in Fig. 3.4.

To the best of authors' knowledge, no result till now has been achieved for robust hybrid locomotion automaton and ours turns out to be the first. In Section 4.2, more details will be provided for how this automaton governs the hierarchical optimization sequence. To demonstrate the usefulness of this hybrid automaton, an example of the planning sequence is provided as follows.

[2] More details are provided in (Branicky et al., 1998).

For example, consider a phase-space trajectory that contains a step (two states) $\mathcal{Q} = \{q, q+1\}$ that are sequentially connected (e.g., left and right feet). Given an initial condition $(\zeta_0, q, \boldsymbol{x}_q(\zeta_0)) \in \mathcal{I}$, the system will evolve following the system $\Sigma_q$ as long as the continuous state $\boldsymbol{x}_q$ remains in $\mathcal{D}(q)$ (e.g., left foot in the ground, right foot swinging). If at some point $\boldsymbol{x}_q$ reaches the guard $\mathcal{G}(q, q+1)$ (e.g., right foot touches the ground) of some edge $\mathcal{E}(q, q+1)$, the discrete state switches to $q+1$. At the same time the continuous state gets reset to some value by $\Delta(q, q+1, \boldsymbol{x}^+_{q+1})$ (e.g., left foot is swinging and right foot is in the ground). After this discrete transition, continuous evolution resumes and the whole process is repeated.

**Algorithm 3** Overall Hybrid Robust Locomotion Planning Structure

1: Initialize walking step index $k \leftarrow 1$, discrete state $q$, initial condition $\mathcal{I}_q$, $\epsilon$ for invariant bundle $\mathcal{B}_q(\epsilon)$, stage update indicator $b_{\text{update}} \leftarrow$ **false**.
2: **while** $\boldsymbol{x}_q \notin \mathcal{G}_a^{[\delta_j]}(q, q+1)$ **do**
3:     **if** $\boldsymbol{x}_q \in \mathcal{G}_d^{[\delta_j]}(q, q_{\text{dist}})$ **then**
4:         Execute $\Delta_d^{[\delta_j]}(q, q_{\text{dist}}, \boldsymbol{x}_{q_{\text{dist}}})$ and quantize disturbed state $(\boldsymbol{x}_q)_{\text{dist}}$.
5:         Generate optimal policies $(\zeta, \dot{x}, \ddot{x}, \tau, \omega, \mathcal{L}, \mathcal{V})_{\text{opt}}$ by dynamic programming.
6:         $b_{\text{update}} \leftarrow$ **true**.
7:         Compute the phase-space manifold $\sigma_{\text{trans}}$ by Eq. (4.1) at transition phase.
8:         **if** $(\boldsymbol{x}_q)_{\text{trans}} \notin \mathcal{R}_q$ **then**
9:             Re-plan $x_{\text{foot}_{q+1}}$ by Eq. (4.14) and search $y_{\text{foot}_{q+1}}$ by Algorithm 2.
10:         **end if**
11:     **end if**
12:     Compute $\sigma_{i+1}$ over domain $\mathcal{D}_q$ by Eq. (4.1).
13:     **if** $b_{\text{update}}$ is **true then**
14:         Update stage index $i_{\text{stage}}$ of recovery optimal control inputs.
15:     **end if**
16:     **if** $\boldsymbol{x}_q \notin \mathcal{B}_q$ **then**
17:         Compute $\boldsymbol{u}_{c_{i+1}} =: (\tau_y, \omega)_{i_{\text{stage}}}$ by Eq. (4.7a) and assign $\ddot{x}_{i+1} \leftarrow \ddot{x}_{\text{opt}}(i_{\text{stage}})$.
18:     **else**
19:         Compute $\boldsymbol{u}_{c_{i+1}} =: (\tau_y, \omega)_{i+1}$ by Eq. (4.7b) and assign $\ddot{x}_{i+1}$ by Eq. (3.14).
20:     **end if**
21:     Evolve $(x_{i+1}, \dot{x}_{i+1})$ over domain $\mathcal{D}_q$ numerically.
22:     $i \leftarrow i + 1$.
23: **end while**
24: $q \leftarrow q + 1$, re-assign $\mathcal{I}_{q+1}$, $b_{\text{update}} \leftarrow$ **false** and jump to line 2 for next walking step.

# Appendix D

# Dynamic Programming

Dynamic programming divides a multi-period planning problem into simpler subproblems at different stages. In contrast to using a traditional time discretization, our study discretizes the CoM sagittal position. Our objective is to generate recovery control policies offline for all admissible disturbance and store them as a policy table. Therefore, recovery can be achieved by looking up the table when disturbances are detected at runtime. We implement a grid-search backward DP. The cost function in Eq. (4.5) can be defined as the value function $\mathcal{V}(q, \boldsymbol{x}_N)$. According to Bellman's equation, one step optimization at $n^{\text{th}}$ stage is

$$\mathcal{V}_n(q, \boldsymbol{x}_n) = \min_{\boldsymbol{u}_{\boldsymbol{x}}^c} \; \mathcal{L}_n(q, \boldsymbol{x}_n, \boldsymbol{u}_{\boldsymbol{x}}^c) + \mathcal{V}_{n+1}(q, \boldsymbol{x}_{n+1}), \tag{D.1}$$

which is known as principle of optimality for discrete systems[1]. Note that, the control input $\boldsymbol{u}_{\boldsymbol{x}}^c$ as well as CoM acceleration are assumed to be constant within one stage. For the integral of $\mathcal{L}_n(q, \boldsymbol{x}_n, \boldsymbol{u}_{\boldsymbol{x}}^c)$, linear interpolation is used to estimate CoM velocities. The velocities of two consecutive stages satisfy $\dot{x}_{n+1} = \dot{x}_n + \ddot{x}_n T_n$, where $T_n$ is the duration of one stage. Given the constant acceleration within one stage, we have $\delta x_n = (\dot{x}_{n+1} + \dot{x}_n) T_n / 2$. Combining the two equations above, we can derive the constant acceleration

$$\ddot{x}_n = \frac{\dot{x}_{n+1}^2 - \dot{x}_n^2}{2\delta x_n}, \tag{D.2}$$

which is used to seed values to the equality constraint in the optimization problem of Eq. (4.5), allowing us to solve for the continuous control input.

[1]Since the cost is computed iteratively in a backward way, the stage index $n$ decreases. In this value iteration algorithm, $\mathcal{V}_{n+1}(q, \boldsymbol{x}_{n+1})$ represents the total optimal cost from $(n+1)^{\text{th}}$ stage to the terminal $N^{\text{th}}$ stage for all feasible states $\dot{x}_{n+1}$. Accordingly, we solve the optimal control sequence from $(n+1)^{\text{th}}$ to $N^{\text{th}}$ stage for all $\dot{x}_{n+1}$. Then for $n^{\text{th}}$ stage, we only need to solve the optimal cost from $n^{\text{th}}$ to $(n+1)^{\text{th}}$ stage.

# Appendix E

# Linear Temporal Logic

Temporal logic is composed of propositions, standard Boolean operators and temporal operators. It has been widely used in motion planning communities to represent properties, requirements and specifications of robotic systems. For instance, the following desired properties can be expressed: *response* (if $a$, then $b$), *safety* (always $a$), *liveness* (infinitely often $a$) and *stability* (eventually always $a$). In this study, linear temporal logic (LTL) is used to specify linear time properties and the reasons are in twofold: i), the formulae can be synthesized into low-level controllers in a convenient way. and ii), it can capture many complex robot behaviors. Given an environment $\mathcal{E}$ and the controllable robot system state $\mathcal{S}$, we define an augmented variable $\boldsymbol{r} \in \mathcal{R} = \mathcal{E} \cup \mathcal{S}$.

**Definition E.1** (**Linear-time property**). *A linear-time (LT) property $P$ over atomic propositions in $AP$ is a set of infinite sequences over $2^{AP}$.*

**LTL Syntax:** An LTL formula $\varphi$ is composed of atomic propositions $\pi \in AP$, which are statements on a Boolean system variable that can either be true or false. The generic form of a LTL formula has the following grammar,

$$\varphi ::= \pi \Big| \neg\varphi \Big| \varphi_1 \wedge \varphi_2 \Big| \varphi_1 \vee \varphi_2 \Big| \bigcirc \varphi \Big| \varphi_1 \mathcal{U} \varphi_2$$

where the Boolean constants true and false are expressed by false = ¬true and true = $\varphi \vee \neg\varphi$, and we have the temporal operators $\bigcirc$ (”next”), $\mathcal{U}$ (”until”), ¬ (”negation”) and ∧ (”conjunction”). We can also define ∨ (”disjunction”), ⇒ (”implication”), ⇔ (”equivalence”). Another two key operators in LTL are ◇ (”Eventually”) and □ (”Always”). We can interpret them $\Diamond\varphi :=$ true $\mathcal{U}\varphi$ for “Eventually” and $\Box\varphi := \neg\Diamond\neg\varphi$ for “Always”. Combinations of these operators can describe a variety of specifications for system behaviors.

**LTL Semantics:** An LTL formula is interpreted over an infinite sequence of states $(\boldsymbol{q}, \boldsymbol{p}) = (q_1, p_1)(q_2, p_2)(q_3, p_3)\ldots$. We define $(\boldsymbol{q}_i, \boldsymbol{p}_i) = (q_i, p_i)(q_{i+1}, p_{i+1})$ $(q_{i+2}, p_{i+2})\ldots$ as a run from $i^{\text{th}}$ position. It is said that a LTL formula $\varphi$ holds at $i^{\text{th}}$ position of $(\boldsymbol{q}, \boldsymbol{p})$, represented as $(q_i, p_i) \models \varphi$, if and only if (iff) $\varphi$ holds for the remaining sequence of $(\boldsymbol{q}, \boldsymbol{p})$ starting at $i^{\text{th}}$ position. The semantics of LTL are defined inductively as

$$(q_i, p_i) \models \neg\varphi \text{ iff } (q_i, p_i) \not\models \varphi$$
$$(q_i, p_i) \models \varphi_1 \wedge \varphi_2 \text{ iff } (q_i, p_i) \models \varphi_1 \wedge (q_i, p_i) \models \varphi_2$$
$$(q_i, p_i) \models \varphi_1 \vee \varphi_2 \text{ iff } (q_i, p_i) \models \varphi_1 \vee (q_i, p_i) \models \varphi_2$$
$$(q_i, p_i) \models \bigcirc\varphi \text{ iff } (q_{i+1}, p_{i+1}) \models \varphi$$
$$(q_i, p_i) \models \varphi_1 \mathcal{U} \varphi_2 \text{ iff } \exists j \geq i, \text{s.t.} (q_j, p_j) \models \varphi_2 \text{ and } (q_k, p_k) \models \varphi_1, \forall i \leq k \leq j$$

By these definitions, the notation $\bigcirc\varphi$ represents that $\varphi$ is true at the next “step” (i.e., next position in the sequence), $\Box\varphi$ represents $\varphi$ is always true (i.e., true at every position of the sequence), $\Diamond\varphi$ represents that $\varphi$ is eventually true at some position of the sequence, $\Box\Diamond\varphi$ represents that $\varphi$ is true infinitely often (i.e., eventually become true starting from any position), and $\Diamond\Box\varphi$ represents that $\varphi$ is eventually always true (i.e., always become true after some position of the sequence) (Baier et al., 2008).

# Appendix F

# Sensitivity Discrepancy Equation Derivations

### Derivation of Eqs. (6.14) and (6.15)

Using the product differentiation rule on Eq. (6.14), one obtains,

$$\frac{\partial \text{atan}\left(A_{1g}/A_{2g}\right)}{\partial T_s} = \frac{\partial \text{atan}\left(A_{1g}/A_{2g}\right)}{\partial\left(A_{1g}/A_{2g}\right)} \cdot \frac{\partial\left(A_{1g}/A_{2g}\right)}{\partial T_s}. \tag{F.1}$$

Using differentiation rules, the following applies,

$$\frac{\partial \text{atan}\left(A_{1g}/A_{2g}\right)}{\partial\left(A_{1g}/A_{2g}\right)} = \frac{1}{1+\left(A_{1g}/A_{2g}\right)^2}. \tag{F.2}$$

Also, it is straightforward to derive, once again, using the product differentiation rule, the following equalities,

$$\begin{aligned}
\frac{\partial\left(A_{1g}/A_{2g}\right)}{\partial T_s} &= \frac{\partial A_{1g}}{\partial T_s}\frac{1}{A_{2g}} + \frac{\partial\left(1/A_{2g}\right)}{\partial T_s}A_{1g} \\
&= \frac{\left[\begin{array}{l} -A_{2g}\Big(K\omega_g\cos(T_s\,\omega_g) + K\,\tau_v\,\omega_g^2\sin(T_s\,\omega_g)\Big) \\ +A_{1g}\Big(K\omega_g\sin(T_s\,\omega_g) - K\,\tau_v\,\omega_g^2\cos(T_s\,\omega_g)\Big)\end{array}\right]}{A_{2g}^2} \\
&= \frac{\left[\begin{array}{l} -K^2(\tau_v^2\omega_g^2+1) + KB\omega_g\Big(\sin\left(\left(T_s-T_d\right)\omega_g\right) \\ -\tau_v\omega_g\cos\left(\left(T_s-T_d\right)\omega_g\right)\Big)\end{array}\right]\omega_g}{A_{2g}^2},
\end{aligned}$$

which make use of the trigonometric rules $\cos^2(x)+\sin^2(x) = 1$ and $\sin(x-y) = \sin(x)\cos(y)-\cos(x)\sin(y)$. Gathering the above derivations and putting them together according to Eq. (F.1) yields,

$$\frac{\partial \mathrm{atan}\,(A_{1g}/A_{2g})}{\partial T_s} = \frac{\Big[ -K^2(\tau_v^2\omega_g^2+1) + KB\omega_g\, M \quad \Big]\,\omega_g}{A_{1g}^2 + A_{2g}^2}$$

with $M \triangleq \sin\Big((T_s - T_d)\,\omega_g\Big) - \tau_v\omega_g\cos\Big((T_s - T_d)\,\omega_g\Big)$. Using basic trigonometry, one can further simplify $M$ to

$$M = \sqrt{\tau_v^2\omega_g^2+1}\cdot\sin\Big((T_s - T_d)\,\omega_g + \phi\Big) \tag{F.3}$$

where the phase shift $\phi \triangleq \mathrm{atan}(-\tau_v\omega_g)$. The derivation of Eq. (6.15) is very similar to the above and is therefore omitted for space purpose. □

**Derivation of Eq. (6.24)**

Using the gain crossover frequency of Eq. (6.23) on the numerator of Eq. (6.11) yields,

$$|jA_{1g} + A_{2g}|^2 = A_{1g}^2 + A_{2g}^2. \tag{F.4}$$

Using the expressions of $A_{1g}$ and $A_{2g}$ given in Eq. (6.12) yields,

$$A_{1g}^2 + A_{2g}^2 = (B\omega_g)^2 + K^2(\tau_v^2\omega_g^2+1) - 2KB\omega_g M,$$

which also makes use of the trigonometric rules $\cos^2(x) + \sin^2(x) = 1$ and $\sin(x-y) = \sin(x)\cos(y) - \cos(x)\sin(y)$. Combining the above norm of the numerator of Eq. (6.11) with the norm of its denominator one gets

$$|P_{OL}(\omega_g)|^2 = \frac{(B\omega_g)^2 + K^2(\tau_v^2\omega_g^2+1) - 2KB\omega_g\, M}{\omega_g^2\Big((m\omega_g)^2 + b^2\Big)\Big((\tau_v\omega_g)^2+1\Big)} = 1, \tag{F.5}$$

which is equivalent to Eq. (6.24). □

# Appendix G

# Passivity Preliminary

### Contraction Mapping Principle

By Eq. (15), the iteration equation becomes $\hat{\boldsymbol{q}}_{n+1} = \boldsymbol{\theta} - \boldsymbol{K}^{-1}\boldsymbol{l}(\hat{\boldsymbol{q}}_n)$. By contraction mapping principle Ott et al. (2008); Vidyasagar (2002), we have

$$\lim_{n\to\infty} \hat{\boldsymbol{q}}_n = \overline{\boldsymbol{q}}, \tag{G.1}$$

as long as the equation below is satisfied

$$||\frac{\partial \boldsymbol{l}(\boldsymbol{q})}{\partial \boldsymbol{q}}|| \leq \alpha \leq \frac{1}{||\boldsymbol{K}^{-1}||},\ \forall \boldsymbol{q} \in \mathbb{R}^n. \tag{G.2}$$

In practice, one or two iterations will result in the convergence to $\overline{\boldsymbol{q}}$ given a common accuracy requirement Albu-Schäffer et al. (2007). A physical interpretation of Eq. (G.2) is: considering Eq. (8.16) and ignoring its gravity, we have

$$\frac{\partial \boldsymbol{l}(\boldsymbol{q})}{\partial \boldsymbol{q}} = \frac{\partial}{\partial \boldsymbol{q}}[\boldsymbol{J}^*(\boldsymbol{q}_0)^T \boldsymbol{\Lambda}(\boldsymbol{q}_0)\boldsymbol{K}_x \boldsymbol{J}^*(\boldsymbol{q}_0)\tilde{\boldsymbol{q}}_0] = \boldsymbol{J}^*(\boldsymbol{q}_0)^T \boldsymbol{\Lambda}(\boldsymbol{q}_0)\boldsymbol{K}_x \boldsymbol{J}^*(\boldsymbol{q}_0) = \boldsymbol{K}_q.$$

Thus, Eq. (G.2) is equivalent to

$$||\boldsymbol{K}_q|| \leq \boldsymbol{K},$$

which means that the desired joint stiffness $\boldsymbol{K}_q$ mapped from the Cartesian space stiffness can not exceed the joint spring stiffness $\boldsymbol{K}$. This is consistent with our impedance analysis of the series elastic actuators in Chapter 7.

**Propositions of Passivity Analysis**

**Proposition G.1.** (Inertia Matrix) *The inertia matrix $\boldsymbol{A}(\boldsymbol{q})$ is positive definite*

$$\boldsymbol{A}(\boldsymbol{q}) = \boldsymbol{A}^T(\boldsymbol{q}) > \boldsymbol{0},\ \forall \boldsymbol{q} \in \mathbb{R}^n$$

**Proposition G.2.** (Skew symmetry) *For free-floating multi-body dynamics, the matrix $\dot{\boldsymbol{A}}(\boldsymbol{q}) - 2\boldsymbol{b}(\boldsymbol{q}, \dot{\boldsymbol{q}})$ is skew symmetric, i.e.,*

$$\dot{\boldsymbol{A}}(\boldsymbol{q}) - 2\boldsymbol{b}(\boldsymbol{q}, \dot{\boldsymbol{q}}) = 0$$

*where $\boldsymbol{b}(\boldsymbol{q}, \dot{\boldsymbol{q}})$ is the matrix of centrifugal and Coriolis forces.*

**Proposition G.3.** (Inequality Property) *For a positive-definite matrix $\Phi$, the following inequality holds:*

$$\pm 2m^T(t) \int_{t-d(t)}^{t} n(\xi) d\xi - \int_{t-d(t)}^{t} n^T(\xi) \Phi n(\xi) d\xi \le \bar{d}\, m^T(t) \Phi^{-1} m^T(t)$$

*where $m(\cdot)$ and $n(\cdot)$ are vectors, $d(t)$ is a time-varying scalar with $0 \le d(t) \le \bar{d}$. Note that, the first term in the left-hand side of inequality above is sign independent since $\bar{d}\, m^T(t)\Phi^{-1}m^T(t) = \bar{d}\,(-m^T(t))\Phi^{-1}(-m^T(t))$.*

**Proposition G.4.** (Passivity) *A system, interacting with the environment or sustaining external disturbance, is passive*

$$\int_0^T \dot{\boldsymbol{q}}^T(\delta) \tau(\delta) d\delta \ge -\gamma,\ \forall\, T > 0$$

*where the constant* $\gamma \geq 0, \dot{\boldsymbol{q}}^T(\delta)\tau(\delta)$ *represents the instantaneous power and* $\int_0^T \dot{\boldsymbol{q}}^T(\delta)\tau(\delta)d\delta$ *denotes the energy generated by the system over time interval* $[0, T]$. *This property means the system can only produce a finite amount of energy which is bounded by* $-\gamma$. *We use this proposition to prove the passivity of the time-delayed WBOSC.*

**Proposition G.5.** (Schur Complement) *The linear matrix inequality (LMI) Boyd et al. (1994)*

$$\begin{pmatrix} \boldsymbol{A}(\boldsymbol{x}) & \boldsymbol{B}(\boldsymbol{x}) \\ \boldsymbol{B}^T(\boldsymbol{x}) & \boldsymbol{C}(\boldsymbol{x}) \end{pmatrix} > 0$$

*where* $\boldsymbol{A}(\boldsymbol{x}) = \boldsymbol{A}^T(\boldsymbol{x}), \boldsymbol{B}(\boldsymbol{x}) = \boldsymbol{B}^T(\boldsymbol{x})$ *is equivalent to the following inequalities*

$$\boldsymbol{C}(\boldsymbol{x}) > 0, \quad \boldsymbol{A}(\boldsymbol{x}) - \boldsymbol{B}(\boldsymbol{x})\boldsymbol{C}^{-1}(\boldsymbol{x})\boldsymbol{B}^T(\boldsymbol{x}) > 0$$

### Derivation of $\dot{V}_2$

Here we assume the matrices $\boldsymbol{J}^{*T}$, $\boldsymbol{\Lambda}$ and $\bar{\boldsymbol{g}}(\boldsymbol{\theta})$ are evaluated at time $t - d_0$. Namely, $\boldsymbol{J}^{*T} = \boldsymbol{J}^{*T}_{(-d_0)}, \boldsymbol{\Lambda} = \boldsymbol{\Lambda}_{(-d_0)}, \bar{\boldsymbol{g}}(\boldsymbol{\theta}) = \bar{\boldsymbol{g}}(\boldsymbol{\theta})_{(-d_0)}$. Since $\boldsymbol{B}_s$ and $\boldsymbol{K}$ are constant matrices, $\boldsymbol{x}_d$ is a constant input (i.e., a regulation problem) and considering Eq. (8.11), the derivative of $V_2$ is

$$
\begin{aligned}
\dot{V}_2 &= \dot{\boldsymbol{\theta}}^T \boldsymbol{B}_s \ddot{\boldsymbol{\theta}} + (\boldsymbol{\theta} - \boldsymbol{q}_j)^T \boldsymbol{K} (\dot{\boldsymbol{\theta}} - \dot{\boldsymbol{q}}_j) \\
&= \dot{\boldsymbol{\theta}}^T \Big( - \boldsymbol{D}_\theta \dot{\boldsymbol{\theta}} - \boldsymbol{J}^{*T} \boldsymbol{\Lambda} \boldsymbol{K}_x \big( \boldsymbol{x}(t - T_H - \frac{T_L}{2}) - \boldsymbol{x}_d(t - \frac{T_H + T_L}{2}) \big) + \bar{\boldsymbol{g}}(\boldsymbol{\theta}) \Big) \\
&\quad - \dot{\boldsymbol{q}}_j^T \boldsymbol{\Gamma}_{\boldsymbol{sea}} \\
&= - \dot{\boldsymbol{\theta}}^T \boldsymbol{D}_\theta \dot{\boldsymbol{\theta}} - \dot{\boldsymbol{q}}_j^T \boldsymbol{\Gamma}_{\boldsymbol{sea}} + \dot{\boldsymbol{\theta}}^T \boldsymbol{J}^{*T} \boldsymbol{\Lambda} \boldsymbol{K}_x \big( \boldsymbol{x}_d(t - \frac{T_H + T_L}{2}) - \boldsymbol{x}(t - T_H - \frac{T_L}{2}) \big) \\
&\quad + \dot{\boldsymbol{\theta}}^T \bar{\boldsymbol{g}}(\boldsymbol{\theta}) \\
&= - \dot{\boldsymbol{\theta}}^T \boldsymbol{D}_\theta \dot{\boldsymbol{\theta}}(t) - \dot{\boldsymbol{q}}_j^T \boldsymbol{\Gamma}_{\boldsymbol{sea}} - \dot{\boldsymbol{\theta}}^T \boldsymbol{J}^{*T} \boldsymbol{\Lambda} \boldsymbol{K}_x \big( \boldsymbol{x}_d(t) - \boldsymbol{x}_d(t - d_2) \big) \\
&\quad + \dot{\boldsymbol{\theta}}^T \boldsymbol{J}^{*T} \boldsymbol{\Lambda} \boldsymbol{K}_x \big( \boldsymbol{x}(t) - \boldsymbol{x}(t - d_1) \big) + \dot{\boldsymbol{\theta}}^T \boldsymbol{J}^{*T} \boldsymbol{\Lambda} \boldsymbol{K}_x \big( \boldsymbol{x}_d(t) - \boldsymbol{x}(t) \big) + \dot{\boldsymbol{\theta}}^T \bar{\boldsymbol{g}}(\boldsymbol{\theta}) \\
&= - \dot{\boldsymbol{\theta}}^T \boldsymbol{D}_\theta \dot{\boldsymbol{\theta}} - \dot{\boldsymbol{q}}_j^T \boldsymbol{\Gamma}_{\boldsymbol{sea}} + \dot{\boldsymbol{\theta}}^T \boldsymbol{J}^{*T} \boldsymbol{\Lambda} \boldsymbol{K}_x \int_{t-d_1}^{t} \dot{\boldsymbol{x}}(\xi) d\xi + \dot{\boldsymbol{\theta}}^T \boldsymbol{J}^{*T} \boldsymbol{\Lambda} \boldsymbol{K}_x \big( \boldsymbol{x}_d(t) \\
&\quad - \boldsymbol{x}(t) \big) + \dot{\boldsymbol{\theta}}^T \bar{\boldsymbol{g}}(\boldsymbol{\theta}).
\end{aligned}
$$

Note that, since $\boldsymbol{x}_d$ is constant, thus $\boldsymbol{x}_d(t) - \boldsymbol{x}_d(t - d_2) = 0$ and the term $\dot{\boldsymbol{\theta}}^T \boldsymbol{J}^{*T} \boldsymbol{\Lambda} \boldsymbol{K}_x \big( \boldsymbol{x}_d(t) - \boldsymbol{x}_d(t - d_2) \big)$ is cancelled out. Then the result in Eq. (8.27) follows.

# Bibliography

Rough terrain locomotion video, youtube channel. https://youtu.be/F8uTHsqn1dc, 2016.

Whole-body locomotion video, youtube channel. https://youtu.be/urp7xu8vx3s, 2016.

Muhammad Abdallah and Ambarish Goswami. A biomechanically motivated two-phase strategy for biped upright balance control. In *IEEE-RAS International Conference on Robotics and Automation*, pages 1996–2001, 2005.

Amir Ali Ahmadi and Anirudha Majumdar. Dsos and sdsos optimization: Lp and socp-based alternatives to sum of squares optimization. In *Annual Conference on Information Sciences and Systems*, pages 1–5, 2014.

Mostafa Ajallooeian, Soha Pouya, Alexander Sproewitz, and Auke Jan Ijspeert. Central pattern generators augmented with virtual model control for quadruped rough terrain locomotion. In *IEEE-RAS International Conference on Robotics and Automation*, 2013.

Alin Albu-Schäffer, Christian Ott, and Gerd Hirzinger. A unified passivity-based control framework for position, torque and impedance control of flexible joint robots. *The International Journal of Robotics Research*, 26(1): 23–39, 2007.

Rajeev Alur, Salar Moarref, and Ufuk Topcu. Pattern-based refinement of assume-guarantee specifications in reactive synthesis. In *Tools and Algorithms for the Construction and Analysis of Systems*, pages 501–516. 2015.

Aaron D Ames, Paulo Tabuada, Bastian Schürmann, Wen-Loong Ma, Shishir Kolathaya, Matthias Rungger, and Jessy W Grizzle. First steps toward formal controller synthesis for bipedal robots. In *International Conference on Hybrid Systems: Computation and Control*, pages 209–218, 2015.

Robert J Anderson and Mark W Spong. Bilateral control of teleoperators with time delay. *IEEE Transactions on Automatic control*, 34(5):494–501, 1989.

Marco Antoniotti and Bud Mishra. Discrete event models+ temporal logic= supervisory controller: Automatic synthesis of locomotion controllers. In *IEEE-RAS International Conference on Robotics and Automation*, volume 2, pages 1441–1446, 1995.

Omür Arslan and Uluc Saranli. Reactive planning and control of planar spring–mass running on rough terrain. *IEEE Transactions on Robotics*, 28(3):567–579, 2012.

Nikolay Asenov Atanasov. *Active information acquisition with mobile robots.* PhD thesis, University of Pennsylvania, Philadelphia, USA, 2015.

Hervé Audren, Joris Vaillant, Abderrahmane Kheddar, Adrien Escande, Kunihiko Kaneko, and Erika Yoshida. Model preview control in multi-contact motion-application to a humanoid robot. In *IEEE/RSJ International Conference on Intelligent Robots and Systems*, pages 4030–4035, 2014.

Android Authority. Alphabets schaft shows off its bipedal robot, and its kind of freaky. http://www.androidauthority.com/alphabet-schaft-bipedal-robot-685340/, 2016.

Christel Baier, Joost-Pieter Katoen, et al. *Principles of model checking*, volume 26202649. MIT press Cambridge, 2008.

Grégory Batt, Calin Belta, and Ron Weiss. Temporal logic analysis of gene networks under parameter uncertainty. *IEEE Transactions on Automatic Control*, 53(Special Issue):215–229, 2008.

Calin Belta, Antonio Bicchi, Magnus Egerstedt, Emilio Frazzoli, Eric Klavins, and George J Pappas. Symbolic planning and control of robot motion [grand challenges of robotics]. *IEEE Robotics & Automation Magazine*, 14(1):61–70, 2007.

Amit Bhatia, Lydia E Kavraki, and Moshe Y Vardi. Motion planning with hybrid dynamics and temporal goals. In *IEEE Conference on Decision and Control*, pages 1108–1115, 2010.

Roderick Bloem, Barbara Jobstmann, Nir Piterman, Amir Pnueli, and Yaniv Saar. Synthesis of reactive (1) designs. *Journal of Computer and System Sciences*, 78(3):911–938, 2012.

Thiago Boaventura, Gustavo A Medrano-Cerda, Claudio Semini, Jonas Buchli, and Darwin G Caldwell. Stability and performance of the compliance controller of the quadruped robot hyq. In *IEEE/RSJ International Conference on Intelligent Robots and Systems*, pages 1458–1464, 2013.

Karim Bouyarmane and Abderrahmane Kheddar. Multi-contact stances planning for multiple agents. In *IEEE-RAS International Conference on Robotics and Automation*, pages 5246–5253, 2011.

Karim Bouyarmane, Joris Vaillant, François Keith, and Abderrahmane Kheddar. Exploring humanoid robots locomotion capabilities in virtual disaster response scenarios. In *IEEE-RAS International Conference on Humanoid Robots*, pages 337–342, 2012.

Stephen P Boyd, Laurent El Ghaoui, Eric Feron, and Venkataramanan Balakrishnan. *Linear matrix inequalities in system and control theory*, volume 15. SIAM, 1994.

Michael S Branicky, Vivek S Borkar, and Sanjoy K Mitter. A unified framework for hybrid control: Model and optimal control theory. *IEEE Transactions on Automatic Control*, 43(1):31–45, 1998.

David J Braun, Florian Petit, Felix Huber, Sami Haddadin, Patrick Van Der Smagt, Alin Albu-Schaffer, and Sethu Vijayakumar. Robots driven by compliant actuators: Optimal control under actuation constraints. *IEEE Transactions on Robotics*, 29(5):1085–1101, 2013.

Stephen P Buerger and Neville Hogan. Complementary stability and loop shaping for improved human–robot interaction. *IEEE Transactions on Robotics*, 23(2):232–244, 2007.

Robert R Burridge, Alfred A Rizzi, and Daniel E Koditschek. Sequential composition of dynamically dexterous robot behaviors. *The International Journal of Robotics Research*, 18(6):534–555, 1999.

Maureen Byko. Personification: The materials science and engineering of humanoid robots. `http://www.tms.org/pubs/journals/JOM/0311/Byko-0311.html`, 2003.

Katie Byl and Russ Tedrake. Metastable walking machines. *The International Journal of Robotics Research*, 28(8):1040–1064, 2009. doi: 10.1177/0278364909340446.

Stéphane Caron, Quang-Cuong Pham, and Yoshihiko Nakamura. Zmp support areas for multi-contact mobility under frictional constraints. *arXiv preprint arXiv:1510.03232*, 2015.

G A Cavagna and R Margaria. Mechanics of walking. *Journal of Applied Physiology*, 21(1):271–278, 1966. ISSN 8750-7587. URL `http://jap.physiology.org/content/21/1/271`.

Kyong-Sok Chang. *Efficient algorithms for articulated branching mechanisms: dynamic modeling, control, and simulation.* PhD thesis, Stanford, 2000.

Kenneth Y Chao, Matthew J Powell, Aaron D Ames, and Pilwon Hur. Unification of locomotion pattern generation and control lyapunov function-based quadratic programs, 2016.

Gordon Cheng, Sang-Ho Hyon, Jun Morimoto, Aleš Ude, Joshua G Hale, Glenn Colvin, Wayco Scroggin, and Stephen C Jacobsen. Cb: A humanoid research platform for exploring neuroscience. *Advanced Robotics*, 21(10): 1097–1114, 2007.

Sandeep Chinchali, Scott C Livingston, Ufuk Topcu, Joel W Burdick, and Richard M Murray. Towards formal synthesis of reactive controllers for dexterous robotic manipulation. In *IEEE-RAS International Conference on Robotics and Automation*, pages 5183–5189, 2012.

Shu-Yun Chung and Oussama Khatib. Contact-consistent elastic strips for multi-contact locomotion planning of humanoid robots. In *IEEE-RAS International Conference on Robotics and Automation*, pages 6289–6294, 2015.

J Edward Colgate and J Michael Brown. Factors affecting the z-width of a haptic display. In *IEEE-RAS International Conference on Robotics and Automation*, pages 3205–3210, 1994.

J Edward Colgate and Gerd G Schenkel. Passivity of a class of sampled-data systems: Application to haptic interfaces. *Journal of robotic systems*, 14(1): 37–47, 1997.

Hongkai Dai and Russ Tedrake. Optimizing robust limit cycles for legged locomotion on unknown terrain. In *IEEE Conference on Control and Decision*, pages 1207–1213, 2012.

Hongkai Dai and Russ Tedrake. L2-gain optimization for robust bipedal walking on unknown terrain. In *IEEE-RAS International Conference on Robotics and Automation*, pages 3116–3123, 2013.

Hongkai Dai and Russ Tedrake. Planning robust walking motion on uneven terrain via convex optimization. 2016.

Hongkai Dai, Andrés Valenzuela, and Russ Tedrake. Whole-body motion planning with centroidal dynamics and full kinematics. In *IEEE-RAS International Conference on Humanoid Robots*, pages 295–302, 2014.

Neil T. Dantam, Zachary K. Kingston, Swarat Chaudhuri, and Lydia E. Kavraki. Incremental task and motion planning: A constraint-based approach. In *Proceedings of Robotics: Science and Systems*, AnnArbor, Michigan, June 2016. doi: 10.15607/RSS.2016.XII.002.

DARPA. The DARPA robotics challenge. http://www.theroboticschallenge.org, 2014. Accessed: 2014-02-27.

Alessandro De Luca, Bruno Siciliano, and Loredana Zollo. Pd control with on-line gravity compensation for robots with elastic joints: Theory and experiments. *Automatica*, 41(10):1809–1819, 2005.

Jonathan A DeCastro and Hadas Kress-Gazit. Synthesis of nonlinear continuous controllers for verifiably correct high-level, reactive behaviors. *The International Journal of Robotics Research*, 34(3):378–394, 2015.

Robin Deits and Russ Tedrake. Footstep planning on uneven terrain with mixed-integer convex optimization. In *IEEE-RAS International Conference on Humanoid Robots*, pages 279–286. IEEE, 2014.

Jyotirmoy V Deshmukh, Alexandre Donzé, Shromona Ghosh, Xiaoqing Jin, Garvit Juniwal, and Sanjit A Seshia. Robust online monitoring of signal temporal logic. In *Runtime Verification*, pages 55–70. Springer, 2015.

Alexander Dietrich, Christian Ott, and Alin Albu-Schäffer. An overview of null space projections for redundant, torque-controlled robots. *The International Journal of Robotics Research*, 2015.

Alexander Dietrich, Christian Ott, and Stefano Stramigioli. Passivation of projection-based null space compliance control via energy tanks. *IEEE Robotics and Automation Letters*, 1(1):184–191, 2016.

Myron A Diftler, JS Mehling, Muhammad E Abdallah, Nicolaus A Radford, Lyndon B Bridgwater, Adam M Sanders, Roger Scott Askew, D Marty Linn, John D Yamokoski, FA Permenter, et al. Robonaut 2-the first humanoid robot in space. In *IEEE-RAS International Conference on Robotics and Automation*, 2011.

Nicola Diolaiti, Günter Niemeyer, Federico Barbagli, and J Kenneth Salisbury. Stability of haptic rendering: Discretization, quantization, time delay, and coulomb effects. *Robotics, IEEE Transactions on*, 22(2):256–268, 2006.

Alexandre Donzé and Oded Maler. Robust satisfaction of temporal logic over real-valued signals. In *International Conference on Formal Modeling and Analysis of Timed Systems*, pages 92–106. Springer, 2010.

J. Englsberger, C. Ott, M.A. Roa, A. Albu-Schaffer, and G. Hirzinger. Bipedal walking control based on capture point dynamics. In *IEEE/RSJ International Conference on Intelligent Robots and Systems*, pages 4420–4427, 2011.

Johannes Englsberger, Twan Koolen, Sylvain Bertrand, Jerry Pratt, Christian Ott, and Alin Albu-Schaffer. Trajectory generation for continuous leg forces during double support and heel-to-toe shift based on divergent component of motion. In *IEEE/RSJ International Conference on Intelligent Robots and Systems*, pages 4022–4029, 2014.

Johannes Englsberger, Pawel Kozlowski, and Christian Ott. Biologically inspired deadbeat control for running on 3d stepping stones. In *IEEE-RAS International Conference on Humanoid Robots*, pages 1067–1074, 2015a.

Johannes Englsberger, Christian Ott, and Alin Albu-Schaffer. Three-dimensional bipedal walking control based on divergent component of motion. *IEEE Transactions on Robotics*, 31(2):355–368, 2015b.

Tom Erez and William D Smart. Bipedal walking on rough terrain using manifold control. In *IEEE/RSJ International Conference on Intelligent Robots and Systems*, pages 1539–1544, 2007.

Adrien Escande, Nicolas Mansard, and Pierre-Brice Wieber. Hierarchical quadratic programming: Fast online humanoid-robot motion generation. *The International Journal of Robotics Research*, 2014.

Georgios E Fainekos, Antoine Girard, Hadas Kress-Gazit, and George J Pappas. Temporal logic motion planning for dynamic robots. *Automatica*, 45 (2):343–352, 2009.

Samira S Farahani, Vasumathi Raman, and Richard M Murray. Robust model predictive control for signal temporal logic synthesis. *IFAC-PapersOnLine*, 48(27):323–328, 2015.

Roy Featherstone. *Rigid body dynamics algorithms*. Springer, 2014.

Siyuan Feng, X Xinjilefu, Weiwei Huang, and Christopher G Atkeson. 3d walking based on online optimization. In *IEEE-RAS International Conference on Humanoid Robots*, pages 21–27, 2013.

Siyuan Feng, Eric Whitman, X Xinjilefu, and Christopher G Atkeson. Optimization based full body control for the atlas robot. In *IEEE-RAS International Conference on Humanoid Robots*, pages 120–127, 2014.

Siyuan Feng, Eric Whitman, X Xinjilefu, and Christopher G Atkeson. Optimization-based full body control for the darpa robotics challenge. *Journal of Field Robotics*, 32(2):293–312, 2015.

B. Fernández-Rodríguez. *Control of Multivariable Nonlinear Systems by the Sliding Mode Method.* PhD thesis, Massachusetts Institute of Technology, Cambridge, Massachusetts, US., 1988. Ph.D. Thesis.

Michele Focchi, Gustavo A Medrano-Cerda, Thiago Boaventura, Marco Frigerio, Claudio Semini, Jonas Buchli, and Darwin G Caldwell. Robot impedance control and passivity analysis with inner torque and velocity feedback loops. *Control Theory and Technology*, pages 1–16, 2016.

Chien-Liang Fok, Gwendolyn Johnson, Luis Sentis, Aloysius Mok, and John D Yamokoski. Controlit!a software framework for whole-body operational space control. *International Journal of Humanoid Robotics*, 2015.

Emilio Frazzoli. *Robust hybrid control for autonomous vehicle motion planning.* PhD thesis, Massachusetts Institute of Technology, 2001.

Jie Fu and Ufuk Topcu. Synthesis of joint control and active sensing strategies under temporal logic constraints. *IEEE Transactions on Automatic Control*, 2016. doi: 10.1109/TAC.2016.2518639.

Pascal Gahinet, Pierre Apkarian, and Mahmoud Chilali. Affine parameter-dependent lyapunov functions and real parametric uncertainty. *IEEE Transactions on Automatic control*, 41(3):436–442, 1996.

Zhenyu Gan, Thomas Wiestner, Michael A Weishaupt, Nina M Waldern, and C David Remy. Passive dynamics explain quadrupedal walking, trotting, and tölting. *Journal of Computational and Nonlinear Dynamics*, 11(2), 2016.

Huijun Gao, Tongwen Chen, and Tianyou Chai. Passivity and passification for networked control systems. *SIAM Journal on Control and Optimization*, 46(4):1299–1322, 2007.

Jorge Juan Gil, Emilio Sánchez, Thomas Hulin, Carsten Preusche, and Gerd Hirzinger. Stability boundary for haptic rendering: Influence of damping and delay. *Journal of Computing and Information Science in Engineering*, 9(1):011005, 2009.

Jeremy H Gillula, Gabriel M Hoffmann, Haomiao Huang, Michael P Vitus, and Claire Tomlin. Applications of hybrid reachability analysis to robotic aerial vehicles. *The International Journal of Robotics Research*, 2011.

Jessy W Grizzle, Christine Chevallereau, Ryan W Sinnet, and Aaron D Ames. Models, feedback control, and open problems of 3d bipedal robotic walking. *Automatica*, 50(8):1955–1988, 2014.

GuoYing Gu, LiMin Zhu, ZhenHua Xiong, and Han Ding. Design of a distributed multiaxis motion control system using the ieee-1394 bus. *IEEE Transactions on Industrial Electronics*, 57(12):4209–4218, 2010.

Keqin Gu, Jie Chen, and Vladimir L Kharitonov. *Stability of time-delay systems*. Springer, 2003.

Dongsheng Guo and Yunong Zhang. Zhang neural network for online solution of time-varying linear matrix inequality aided with an equality conversion. *IEEE transactions on neural networks and learning systems*, 25(2):370–382, 2014.

Kaveh Akbari Hamed, Brian G Buss, and Jessy W Grizzle. Exponentially stabilizing continuous-time controllers for periodic orbits of hybrid systems: Application to bipedal locomotion with ground height variations. *The International Journal of Robotics Research*, 2015.

Blake Hannaford. Stability and performance tradeoffs in bi-lateral telemanipulation. In *Robotics and Automation, 1989. Proceedings., 1989 IEEE International Conference on*, pages 1764–1767. IEEE, 1989.

Blake Hannaford and Jee-Hwan Ryu. Time-domain passivity control of haptic interfaces. *IEEE Transactions on Robotics and Automation*, 18(1):1–10, 2002.

Kris Hauser. Fast interpolation and time-optimization with contact. *The International Journal of Robotics Research*, 33(9):1231–1250, 2014.

Keliang He, Morteza Lahijanian, Lydia E Kavraki, and Moshe Y Vardi. Towards manipulation planning with temporal logic specifications. In *IEEE-*

*RAS International Conference on Robotics and Automation*, pages 346–352, 2015.

Bernd Henze, Máximo A Roa, and Christian Ott. Passivity-based whole-body balancing for torque-controlled humanoid robots in multi-contact scenarios. *The International Journal of Robotics Research*, 2016.

Katharina Hertkorn, Thomas Hulin, Philipp Kremer, Carsten Preusche, and Gerd Hirzinger. Time domain passivity control for multi-degree of freedom haptic devices with time delay. In *IEEE-RAS International Conference on Robotics and Automation*, pages 1313–1319, 2010.

Alexander Herzog, Nicholas Rotella, Sean Mason, Felix Grimminger, Stefan Schaal, and Ludovic Righetti. Momentum control with hierarchical inverse dynamics on a torque-controlled humanoid. *Autonomous Robots*, 40(3):473–491, 2016.

Daan GE Hobbelen and Martijn Wisse. A disturbance rejection measure for limit cycle walkers: The gait sensitivity norm. *IEEE Transactions on Robotics*, 23(6):1213–1224, 2007.

At L Hof. The 'extrapolated center of mass' concept suggests a simple control of balance in walking. *Human movement science*, 27(1):112–125, 2008.

Andreas Hofmann. *Robust execution of bipedal walking tasks from biomechanical principles.* PhD thesis, Computer Science Department, Massachusetts Institute of Technology,, 2006.

Neville Hogan. Impedance control: An approach to manipulation: Part ii implementation. *Journal of dynamic systems, measurement, and control*, 107(1):8–16, 1985.

Michael A Hopkins, Dennis W Hong, and Alexander Leonessa. Compliant locomotion using whole-body control and divergent component of motion tracking. In *IEEE-RAS International Conference on Robotics and Automation*, 2013.

Michael A Hopkins, Alexander Leonessa, Brian Y Lattimer, and Dennis W Hong. Optimization-based whole-body control of a series elastic humanoid robot. *International Journal of Humanoid Robotics*, 13(01), 2016.

Chang-Chun Hua and Xiaoping P Liu. Delay-dependent stability criteria of teleoperation systems with asymmetric time-varying delays. *IEEE Transactions on Robotics*, 26(5):925–932, 2010.

Christian Hubicki, Jesse Grimes, Mikhail Jones, Daniel Renjewski, Alexander Spröwitz, Andy Abate, and Jonathan Hurst. Atrias: Design and validation of a tether-free 3d-capable spring-mass bipedal robot. *The International Journal of Robotics Research*, 2016.

Thomas Hulin, Carsten Preusche, and Gerd Hirzinger. Stability boundary for haptic rendering: Influence of human operator. In *IEEE/RSJ International Conference on Intelligent Robots and Systems*, pages 3483–3488, 2008.

Thomas Hulin, Alin Albu-Schäffer, and Gerd Hirzinger. Passivity and stability boundaries for haptic systems with time delay. *IEEE Transactions on Control Systems Technology*, 22(4):1297–1309, 2014.

Marco Hutter, C David Remy, Mark A Hoepflinger, and Roland Siegwart. Efficient and versatile locomotion with highly compliant legs. *IEEE/ASME Transactions on Mechatronics*, 18(2):449–458, 2013.

Marco Hutter, Hannes Sommer, Christian Gehring, Mark Hoepflinger, Michael Bloesch, and Roland Siegwart. Quadrupedal locomotion using hierarchical operational space control. *The International Journal of Robotics Research*, 2014.

S Hyon and Gordon Cheng. Disturbance rejection for biped humanoids. In *IEEE-RAS International Conference on Robotics and Automation*, pages 2668–2675, 2007.

Sang-Ho Hyon, Joshua G Hale, and Gordon Cheng. Full-body compliant human–humanoid interaction: balancing in the presence of unknown external forces. *IEEE Transactions on Robotics*, 23(5):884–898, 2007.

Auke Jan Ijspeert. Central pattern generators for locomotion control in animals and robots: a review. *Neural Networks*, 21(4):642–653, 2008.

A. Isidori. Nonlinear control systems: An introduction. In M. Thoma, editor, *Lecture Notes in Control and Information Sciences*. Springer-Verlag, Berlin, 1985.

M Jantsch, Steffen Wittmeier, and Alois Knoll. Distributed control for an anthropomimetic robot. In *IEEE/RSJ International Conference on Intelligent Robots and Systems*, pages 5466–5471, October 2010.

S. Kajita, F. Kanehiro, K. Kaneko, K. Fujiwara, K. Yokoi, and H. Hirukawa. A realtime pattern generator for biped walking. In *IEEE-RAS International Conference on Robotics and Automation*, volume 1, pages 31–37. IEEE, 2002.

S. Kajita, F. Kanehiro, K. Kaneko, K. Fujiwara, K. Harada, K. Yokoi, and H. Hirukawa. Biped walking pattern generation by using preview control of zero-moment point. In *IEEE-RAS International Conference on Robotics and Automation*, pages 1620–1626, 2003a.

Shuuji Kajita, Fumio Kanehiro, Kenji Kaneko, Kiyoshi Fujiwara, Kensuke Harada, Kazuhito Yokoi, and Hirohisa Hirukawa. Resolved momentum control: Humanoid motion planning based on the linear and angular momentum. In *IEEE/RSJ International Conference on Intelligent Robots and Systems*, pages 1644–1650, 2003b.

O. Khatib. A unified approach for motion and force control of robot manipulators: The operational space formulation. *The International Journal of Robotics Research*, 3(1):43–53, 1987a.

O. Khatib. A Unified Approach to Motion and Force Control of Robot Manipulators: The Operational Space Formulation. *International Journal of Robotics and Automation*, RA–3(1):43–53, February 1987b.

Donghyun Kim, Gray Thomas, and Luis Sentis. Continuous cyclic stepping on 3d point-foot biped robots via constant time to velocity reversal. In *International Conference on Control Automation Robotics & Vision*, pages 1637–1643, 2014.

Donghyun Kim, Ye Zhao, Gray Thomas, Benito Fernandez, and Luis Sentis. Stabilizing series-elastic point-foot bipeds using whole-body operational space control. *IEEE Transactions on Robotics, In Press*, 2016a.

Jong-Phil Kim, Sang-Yun Baek, and Jeha Ryu. A force bounding approach for multi-degree-of-freedom haptic interaction. *IEEE/ASME Transactions on Mechatronics*, 20(3):1193–1203, 2015.

Jung-Yup Kim, Ill-Woo Park, Jungho Lee, Min-Su Kim, Baek-Kyu Cho, and Jun-Ho Oh. System design and dynamic walking of humanoid robot khr-2. In *Robotics and Automation, Proceedings of the IEEE International Conference on*, 2005.

Kwan Suk Kim, Alan S Kwok, and Luis Sentis. Contact sensing and mobility in rough and cluttered environments. In *European Conference on Mobile Robots*, pages 274–281, 2013.

Kwan Suk Kim, Travis Llado, and Luis Sentis. Full-body collision detection and reaction with omnidirectional mobile platforms: a step towards safe human–robot interaction. *Autonomous Robots*, 40(2):325–341, 2016b.

Marius Kloetzer and Calin Belta. Automatic deployment of distributed teams of robots from temporal logic motion specifications. *IEEE Transactions on Robotics*, 26(1):48–61, 2010.

Taku Komura, Howard Leung, Shunsuke Kudoh, and James Kuffner. A feedback controller for biped humanoids that can counteract large perturbations during gait. In *IEEE-RAS International Conference on Robotics and Automation*, pages 1989–1995, 2005.

Twan Koolen, Tomas De Boer, John Rebula, Ambarish Goswami, and Jerry Pratt. Capturability-based analysis and control of legged locomotion, part 1: Theory and application to three simple gait models. *The International Journal of Robotics Research*, 31(9):1094–1113, 2012.

Twan Koolen, Sylvain Bertrand, Gray Thomas, Tomas de Boer, Tingfan Wu, Jesper Smith, Johannes Englsberger, and J Pratt. Design of a momentum-based control framework and application to the humanoid robot atlas. *International Journal of Humanoid Robotics*, 2015.

Hadas Kress-Gazit, Georgios E Fainekos, and George J Pappas. Temporal-logic-based reactive mission and motion planning. *IEEE Transactions on Robotics*, 25(6):1370–1381, 2009.

M Kudruss, Maximilien Naveau, Olivier Stasse, Nicolas Mansard, C Kirches, Philippe Souères, and K Mombaur. Optimal control for whole-body motion generation using center-of-mass dynamics for predefined multi-contact configurations. In *IEEE-RAS International Conference on Humanoid Robots*, pages 684–689, 2015.

Scott Kuindersma, Frank Permenter, and Russ Tedrake. An efficiently solvable quadratic program for stabilizing dynamic locomotion. In *IEEE-RAS International Conference on Robotics and Automation*, pages 2589–2594, 2014.

Scott Kuindersma, Robin Deits, Maurice Fallon, Andrés Valenzuela, Hongkai Dai, Frank Permenter, Twan Koolen, Pat Marion, and Russ Tedrake. Optimization-based locomotion planning, estimation, and control design for the atlas humanoid robot. *Autonomous Robots*, 40(3):429–455, 2016.

Arthur D Kuo. Energetics of actively powered locomotion using the simplest walking model. *Journal of Biomechanical Engineering*, 124(1):113–120, 2002.

Arthur D Kuo. The six determinants of gait and the inverted pendulum analogy: A dynamic walking perspective. *Human Movement Science*, 26(4): 617–656, 2007.

Arthur D Kuo and Felix E Zajac. Human standing posture: multi-joint movement strategies based on biomechanical constraints. *Progress in brain research*, 97:349–358, 1992.

Jean-Claude Latombe, Anthony Lazanas, and Shashank Shekhar. Robot motion planning with uncertainty in control and sensing. *Artificial Intelligence*, 52(1):1–47, 1991.

Dale A Lawrence. Stability and transparency in bilateral teleoperation. *IEEE Transactions on Robotics and Automation*, 9(5):624–637, 1993.

Dongjun Lee and Mark W Spong. Passive bilateral teleoperation with constant time delay. *IEEE Transactions on Robotics*, 22(2):269–281, 2006.

Sung-Hee Lee and Ambarish Goswami. Ground reaction force control at each foot: A momentum-based humanoid balance controller for non-level and non-stationary ground. In *IEEE/RSJ International Conference on Intelligent Robots and Systems*, pages 3157–3162, 2010.

Daniel Liberzon. *Switching in systems and control*. Springer Science & Business Media, 2012.

Jun Liu, Ufuk Topcu, Necmiye Ozay, and Richard M Murray. Synthesis of reactive control protocols for differentially flat systems. In *IEEE Conference on Decision and Control*, 2012.

Jun Liu, Necmiye Ozay, Ufuk Topcu, and Richard M Murray. Synthesis of reactive switching protocols from temporal logic specifications. *IEEE Transactions on Automatic Control*, 58(7):1771–1785, 2013.

Yiping Liu, Patrick M Wensing, David E Orin, and Yuan F Zheng. Trajectory generation for dynamic walking in a humanoid over uneven terrain using a 3d-actuated dual-slip model. In *IEEE/RSJ International Conference on Intelligent Robots and Systems*, pages 374–380, 2015.

Tomas Lozano-Perez, Matthew T Mason, and Russell H Taylor. Automatic synthesis of fine-motion strategies for robots. *The International Journal of Robotics Research*, 3(1):3–24, 1984.

Lu Lu and Bin Yao. A performance oriented multi-loop constrained adaptive robust tracking control of one-degree-of-freedom mechanical systems: Theory and experiments. *Automatica*, 50(4):1143–1150, 2014.

John Lygeros, Claire Tomlin, and Shankar Sastry. Hybrid systems: modeling, analysis and control. *preprint*, 1999.

Anirudha Majumdar. *Funnel Libraries for Real-Time Robust Feedback Motion Planning.* PhD thesis, Massachusetts Institute of Technology, 2016.

Ian R Manchester and Jack Umenberger. Real-time planning with primitives for dynamic walking over uneven terrain. In *IEEE-RAS International Conference on Robotics and Automation*, pages 4639–4646, 2014.

Ian R Manchester, Uwe Mettin, Fumiya Iida, and Russ Tedrake. Stable dynamic walking over uneven terrain. *The International Journal of Robotics Research*, 2011.

Spyros Maniatopoulos, Philipp Schillinger, Vitchyr Pong, David C Conner, and Hadas Kress-Gazit. Reactive high-level behavior synthesis for an atlas humanoid robot. In *IEEE-RAS International Conference on Robotics and Automation*, pages 4192–4199, 2016.

Nicolas Mansard, Oussama Khatib, and Abderrahmane Kheddar. A unified approach to integrate unilateral constraints in the stack of tasks. *IEEE Transactions on Robotics*, 25(3):670–685, 2009.

J Corless Martin and L George. Continuous state feedback guaranteeing uniform ultimate boundedness for uncertain dynamic systems. *IEEE Transactions on Automatic Control*, 26(5):1139, 1981.

Matt Mason. The mechanics of manipulation. In *IEEE-RAS International Conference on Robotics and Automation*, volume 2, pages 544–548. IEEE, 1985.

Joshua S Mehling, J Edward Colgate, and Michael A Peshkin. Increasing the impedance range of a haptic display by adding electrical damping. In *First Joint Eurohaptics Conference and Symposium on Haptic Interfaces for Virtual Environment and Teleoperator Systems. World Haptics Conference*, pages 257–262, 2005.

Ian M Mitchell, Alexandre M Bayen, and Claire J Tomlin. A time-dependent hamilton-jacobi formulation of reachable sets for continuous dynamic games. *IEEE Transactions on automatic control*, 50(7):947–957, 2005.

Igor Mordatch, Martin De Lasa, and Aaron Hertzmann. Robust physics-based

locomotion using low-dimensional planning. *ACM Transactions on Graphics (TOG)*, 29(4):71, 2010.

Igor Mordatch, Emanuel Todorov, and Zoran Popović. Discovery of complex behaviors through contact-invariant optimization. *ACM Transactions on Graphics (TOG)*, 31(4):43, 2012.

M. Morisawa, S. Kajita, K. Kaneko, K. Harada, F. Kanehiro, K. Fujiwara, and H. Hirukawa. Pattern generation of biped walking constrained on parametric surface. In *IEEE-RAS International Conference on Robotics and Automation*, pages 2405–2410, 2005.

Mitsuharu Morisawa, Shuuji Kajita, Fumio Kanehiro, Kenji Kaneko, Kanako Miura, and Kazuhito Yokoi. Balance control based on capture point error compensation for biped walking on uneven terrain. In *IEEE-RAS International Conference on Humanoid Robots*, pages 734–740, 2012.

Federico L Moro, Michael Gienger, Ambarish Goswami, Nikos G Tsagarakis, and Darwin G Caldwell. An attractor-based whole-body motion control (wbmc) system for humanoid robots. In *IEEE-RAS International Conference on Humanoid Robots*, pages 42–49, 2013.

Mohamad Mosadeghzad, Gustavo A Medrano-Cerda, Jody A Saglia, Nikos G Tsagarakis, and Darwin G Caldwell. Comparison of various active impedance control approaches, modeling, implementation, passivity, stability and trade-offs. In *IEEE/ASME International Conference on Advanced Intelligent Mechatronics*, pages 342–348, 2012.

Keiji Nagatani, Seiga Kiribayashi, Yoshito Okada, Kazuki Otake, Kazuya Yoshida, Satoshi Tadokoro, Takeshi Nishimura, Tomoaki Yoshida, Eiji Koyanagi, Mineo Fukushima, and Shinji Kawatsuma. Emergency response to the nuclear accident at the Fukushima Daiichi Nuclear Power Plants using mobile rescue robots. *Journal of Field Robotics*, 30(1):44–63, 2013. ISSN 1556-4967. doi: 10.1002/rob.21439. URL `http://dx.doi.org/10.1002/rob.21439`.

Jun Nakanishi, Rick Cory, Michael Mistry, Jan Peters, and Stefan Schaal. Operational space control: A theoretical and empirical comparison. *The International Journal of Robotics Research*, 27(6):737–757, 2008.

Quan Nguyen and Koushil Sreenath. Optimal robust control for bipedal robots through control lyapunov function based quadratic programs. In *Robotics: Science and Systems*, 2015.

Günter Niemeyer and Jean-Jacques E Slotine. Telemanipulation with time delays. *The International Journal of Robotics Research*, 23(9):873–890, 2004.

Katsuhiko Ogata and Yanjuan Yang. Modern control engineering. 2010.

Allison M Okamura. Methods for haptic feedback in teleoperated robot-assisted surgery. *Industrial Robot: An International Journal*, 31(6):499–508, 2004.

David E Orin and Ambarish Goswami. Centroidal momentum matrix of a humanoid robot: Structure and properties. In *IEEE/RSJ International Conference on Intelligent Robots and Systems.*, pages 653–659, 2008.

David E Orin, Ambarish Goswami, and Sung-Hee Lee. Centroidal dynamics of a humanoid robot. *Autonomous Robots*, 35(2-3):161–176, 2013.

Romeo Ortega, Julio Antonio Loría Perez, Per Johan Nicklasson, and Hebertt Sira-Ramirez. *Passivity-based control of Euler-Lagrange systems: mechanical, electrical and electromechanical applications.* Springer Science & Business Media, 2013.

Christian Ott, A Kugi, and G Hirzinger. On the passivity-based impedance control of flexible joint robots. *IEEE Transactions on Robotics*, 24(2):416–429, 2008.

Christian Ott, Jordi Artigas, and Carsten Preusche. Subspace-oriented energy distribution for the time domain passivity approach. In *IEEE/RSJ International Conference on Intelligent Robots and Systems*, pages 665–671, 2011.

Christian Ott, Alexander Dietrich, and Alin Albu-Schäffer. Prioritized multi-task compliance control of redundant manipulators. *Automatica*, 53:416–423, 2015.

Nicholas Paine, Sehoon Oh, and Luis Sentis. Design and control considerations for high-performance series elastic actuators. *IEEE/ASME Transactions on Mechatronics*, 19(3):1080–1091, 2014.

Nicholas Paine, Joshua S Mehling, James Holley, Nicolaus A Radford, Gwendolyn Johnson, Chien-Liang Fok, and Luis Sentis. Actuator control for the nasa-jsc valkyrie humanoid robot: A decoupled dynamics approach for torque control of series elastic robots. *Journal of Field Robotics*, 32(3): 378–396, 2015.

Nick Paine and Luis Sentis. A closed-form solution for selecting maximum critically damped actuator impedance parameters. *ASME Journal of Dynamic Systems, Measurement, and Control*, 137(4), 2015.

Hae-Won Park, Alireza Ramezani, and JW Grizzle. A finite-state machine for accommodating unexpected large ground-height variations in bipedal robot walking. *IEEE Transactions on Robotics*, 29(2):331–345, 2013.

Pablo A Parrilo. *Structured semidefinite programs and semialgebraic geometry methods in robustness and optimization*. PhD thesis, Citeseer, 2000.

Florian Petit and A Albu-Schaffer. State feedback damping control for a multi dof variable stiffness robot arm. In *IEEE-RAS International Conference on Robotics and Automation*, pages 5561–5567, 2011.

Quang-Cuong Pham, Stéphane Caron, and Yoshihiko Nakamura. Kinodynamic planning in the configuration space via admissible velocity propagation. In *Robotics: Science and Systems*, 2013.

Michael Posa, Scott Kuindersma, and Russ Tedrake. Optimization and sta-

bilization of trajectories for constrained dynamical systems. In *IEEE-RAS International Conference on Robotics and Automation*, 2016.

Gill A Pratt, Pace Willisson, Clive Bolton, and Andreas Hofman. Late motor processing in low-impedance robots: Impedance control of series-elastic actuators. In *American Control Conference*, volume 4, pages 3245–3251, 2004.

Jerry Pratt, Chee-Meng Chew, Ann Torres, Peter Dilworth, and Gill Pratt. Virtual model control: An intuitive approach for bipedal locomotion. *The International Journal of Robotics Research*, 20(2):129–143, 2001.

Jerry Pratt, John Carff, Sergey Drakunov, and Ambarish Goswami. Capture point: A step toward humanoid push recovery. In *IEEE-RAS International Conference on Humanoid Robots*, pages 200–207, 2006.

Carsten Preusche, Gerd Hirzinger, J-H Ryu, and Blake Hannaford. Time domain passivity control for 6 degrees of freedom haptic displays. In *IEEE/RSJ International Conference on Intelligent Robots and Systems*, volume 3, pages 2944–2949, 2003.

Marc H Raibert. Legged robots that balance. *MIT press*, 1986.

Subramanian Ramamoorthy and Benjamin J Kuipers. Trajectory generation for dynamic bipedal walking through qualitative model based manifold learning. In *Robotics and Automation, 2008. ICRA 2008. IEEE International Conference on*, pages 359–366. IEEE, 2008.

Vasumathi Raman, Alexandre Donzé, Dorsa Sadigh, Richard M Murray, and Sanjit A Seshia. Reactive synthesis from signal temporal logic specifications. In *Proceedings of the International Conference on Hybrid Systems: Computation and Control*, pages 239–248. ACM, 2015.

Alireza Ramezani, Jonathan W Hurst, Kaveh Akbari Hamed, and JW Grizzle. Performance analysis and feedback control of atrias, a three-dimensional bipedal robot. *Journal of Dynamic Systems, Measurement, and Control*, 136(2):021012, 2014.

Oscar E Ramos and Kris Hauser. Generalizations of the capture point to nonlinear center of mass paths and uneven terrain. In *IEEE-RAS International Conference on Humanoid Robots*, pages 851–858, 2015.

Jee-Hwan Ryu and Moon-Young Yoon. Memory-based passivation approach for stable haptic interaction. *IEEE/ASME Transactions on Mechatronics*, 19(4):1424–1435, 2014.

Layale Saab, Oscar E Ramos, François Keith, Nicolas Mansard, Philippe Soueres, and Jean-Yves Fourquet. Dynamic whole-body motion generation under rigid contacts and other unilateral constraints. *IEEE Transactions on Robotics*, 29(2):346–362, 2013.

Dorsa Sadigh and Ashish Kapoor. Safe control under uncertainty with probabilistic signal temporal logic. In *Proceedings of Robotics: Science and Systems*, AnnArbor, Michigan, June 2016. doi: 10.15607/RSS.2016.XII.017.

Sadra Sadraddini and Calin Belta. Robust temporal logic model predictive control. In *Annual Allerton Conference on Communication, Control, and Computing*, pages 772–779, 2015.

Cenk Oguz Saglam and Katie Byl. Robust policies via meshing for metastable rough terrain walking. *Proceedings of Robotics: Science and Systems, Berkeley, USA*, 2014.

Yoshiaki Sakagami, Ryujin Watanabe, Chiaki Aoyama, Shinichi Matsunaga, Nobuo Higaki, and Kikuo Fujimura. The intelligent asimo: System overview and integration. In *IEEE/RSJ International Conference on Intelligent Robots and Systems*, 2002.

Vítor MF Santos and Filipe MT Silva. Design and low-level control of a humanoid robot using a distributed architecture approach. *Journal of Vibration and Control*, 12(12):1431–1456, 2006.

Tom Schouwenaars, Bernard Mettler, Eric Feron, and Jonathan P How. Robust motion planning using a maneuver automation with built-in uncertainties. In *American Control Conference, 2003. Proceedings of the 2003*, volume 3, pages 2211–2216. IEEE, 2003.

L. Sentis, J. Park, and O. Khatib. Compliant control of multi-contact and center of mass behaviors in humanoid robots. *IEEE Transactions on Robotics*, 26(3):483–501, June 2010.

Luis Sentis. *Synthesis and control of whole-body behaviors in humanoid systems*. PhD thesis, Stanford University, 2007.

Luis Sentis and Mike Slovich. Motion planning of extreme locomotion maneuvers using multi-contact dynamics and numerical integration. In *IEEE-RAS International Conference on Humanoid Robots*, pages 760 –767, oct. 2011.

Luis Sentis, Josh Petersen, and Roland Philippsen. Implementation and stability analysis of prioritized whole-body compliant controllers on a wheeled humanoid robot in uneven terrains. *Autonomous Robots*, 35(4):301–319, 2013.

Rangoli Sharan. *Formal methods for control synthesis in partially observed environments: application to autonomous robotic manipulation*. PhD thesis, California Institute of Technology, 2014.

Thomas B Sheridan. Space teleoperation through time delay: review and prognosis. *IEEE Transactions on Robotics and Automation*, 9(5):592–606, 1993.

Sigurd Skogestad and Ian Postlethwaite. *Multivariable feedback control: analysis and design*, volume 2. Wiley New York, 2007.

IEEE Spectrum. Walk-man team built brand new, highly custom robot for drc finals. `http://spectrum.ieee.org/automaton/robotics/humanoids/walkman-humanoid-robot-iit`, 2015.

IEEE Spectrum. Boston dynamics' marc raibert on next-gen atlas: "a huge amount of work". `http://spectrum.ieee.org/automaton/robotics/humanoids/boston-dynamics-marc-raibert-on-nextgen-atlas`, 2016.

Mark W Spong. Modeling and control of elastic joint robots. *Journal of dynamic systems, measurement, and control*, 109(4):310–318, 1987.

Mark W Spong, Seth Hutchinson, and Mathukumalli Vidyasagar. *Robot modeling and control*, volume 3. Wiley New York, 2006.

Koushil Sreenath and Vijay Kumar. Dynamics, control and planning for cooperative manipulation of payloads suspended by cables from multiple quadrotor robots. *Robotics: Science and Systems*, 2013.

Koushil Sreenath, Connie R Hill Jr, and Vijay Kumar. A partially observable hybrid system model for bipedal locomotion for adapting to terrain variations. In *International conference on Hybrid systems: computation and control*, pages 137–142, 2013.

Manoj Srinivasan and Andy Ruina. Computer optimization of a minimal biped model discovers walking and running. *Nature*, 439(7072):72–75, 2006.

Benjamin Stephens. Humanoid push recovery. In *IEEE-RAS International Conference on Humanoid Robots*, pages 589–595, 2007.

Benjamin J Stephens and Christopher G Atkeson. Push recovery by stepping for humanoid robots with force controlled joints. In *IEEE-RAS International Conference on Humanoid Robots*, pages 52–59, 2010.

Stefano Stramigioli. *Modeling and IPC control of interactive mechanical systemsa coordinate-free approach.* 2001.

Paulo Tabuada. *Verification and control of hybrid systems: a symbolic approach.* Springer Science & Business Media, 2009.

Nevio Luigi Tagliamonte, Dino Accoto, and Eugenio Guglielmelli. Rendering viscoelasticity with series elastic actuators using cascade control. In *IEEE-RAS International Conference on Robotics and Automation*, pages 2424–2429, 2014.

Toru Takenaka, Takashi Matsumoto, and Takahide Yoshiike. Real time motion generation and control for biped robot-1 st report: Walking gait pattern generation. In *IEEE/RSJ International Conference on Intelligent Robots and Systems*, pages 1084–1091, 2009.

Yuval Tassa, Tom Erez, and Emanuel Todorov. Synthesis and stabilization of complex behaviors through online trajectory optimization. In *IEEE/RSJ International Conference on Intelligent Robots and Systems*, pages 4906–4913, 2012.

Russ Tedrake, Ian R Manchester, Mark Tobenkin, and John W Roberts. Lqr-trees: Feedback motion planning via sums-of-squares verification. *The International Journal of Robotics Research*, 2010.

Gray C Thomas and Luis Sentis. Towards computationally efficient planning of dynamic multi-contact locomotion. In *IEEE/RSJ International Conference on Intelligent Robots and Systems*, 2016.

International Business Times. Japanese robot teaches itself how to move heavy weights like a human. `http://www.ibtimes.co.uk/japanese-robot-teaches-itself-how-move-heavy-weights-like-human-1504729`, 2016.

Patrizio Tomei. A simple pd controller for robots with elastic joints. *IEEE Transactions on Automatic Control*, 36(10):1208–1213, 1991.

Vadim I Utkin. Sliding modes in control and optimization. *Springer Science & Business Media*, 2013.

Heike Vallery, Jan Veneman, Edwin Van Asseldonk, Ralf Ekkelenkamp, Martin Buss, and Herman Van Der Kooij. Compliant actuation of rehabilitation robots. *IEEE Robotics & Automation Magazine*, 15(3):60–69, 2008.

Arjan Van der Schaft. *L2-gain and passivity techniques in nonlinear control.* Springer Science & Business Media, 2012.

Mathukumalli Vidyasagar. *Nonlinear systems analysis*, volume 42. Siam, 2002.

Chuanfeng Wang and Donghai Li. Decentralized pid controllers based on probabilistic robustness. *Journal of Dynamic Systems, Measurement, and Control*, 133(6), 2011.

Patrick M Wensing and David E Orin. Generation of dynamic humanoid behaviors through task-space control with conic optimization. In *IEEE-RAS International Conference on Robotics and Automation*, pages 3103–3109, 2013.

Eric R Westervelt, Jessy W Grizzle, Christine Chevallereau, Jun Ho Choi, and Benjamin Morris. *Feedback control of dynamic bipedal robot locomotion*, volume 28. CRC press, 2007.

P-B Wieber. Trajectory free linear model predictive control for stable walking in the presence of strong perturbations. In *IEEE-RAS International Conference on Humanoid Robots*, pages 137–142, 2006.

David A Winter. Human balance and posture control during standing and walking. *Gait & posture*, 3(4):193–214, 1995.

Tichakorn Wongpiromsarn, Ufuk Topcu, Necmiye Ozay, Huan Xu, and Richard M Murray. Tulip: a software toolbox for receding horizon temporal logic planning. In *International conference on Hybrid systems: computation and control*, pages 313–314, 2011.

Tichakorn Wongpiromsarn, Ufuk Topcu, and Richard M Murray. Receding horizon temporal logic planning. *IEEE Transactions on Automatic Control*, 57(11):2817–2830, 2012.

Jia-chi Wu and Zoran Popović. Terrain-adaptive bipedal locomotion control. *ACM Transactions on Graphics (TOG)*, 29(4):72, 2010.

T Yang, ER Westervelt, A Serrani, and James P Schmiedeler. A framework for the control of stable aperiodic walking in underactuated planar bipeds. *Autonomous Robots*, 27(3):277–290, 2009.

Shen Yin, Hongyan Yang, and Okyay Kaynak. Coordination task triggered formation control algorithm for multiple marine vessels. *IEEE Trans. on Industrial Electronics*, 2016. doi: 10.1109/TIE.2016.2574301.

Seung-kook Yun and Ambarish Goswami. Momentum-based reactive stepping controller on level and non-level ground for humanoid robot push recovery. In *IEEE/RSJ International Conference on Intelligent Robots and Systems*, pages 3943–3950, 2011.

Wente Zeng and M Chow. A reputation-based secure distributed control methodology in d-ncs. *IEEE Transactions on Industrial Electronics*, 61(11): 6294–6303, 2014.

Hui-Hua Zhao, Wen-Loong Ma, Michael B Zeagler, and Aaron D Ames. Human-inspired multi-contact locomotion with amber2. In *ACM/IEEE International Conference on Cyber-Physical Systems*, pages 199–210, 2014a.

Jie Zhao, Steffen Schutz, and Karsten Berns. Biologically motivated push recovery strategies for a 3d bipedal robot walking in complex environments. In *IEEE International Conference on Robotics and Biomimetics*, pages 1258–1263, 2013a.

Ye Zhao and Luis Sentis. A three dimensional foot placement planner for locomotion in very rough terrains. In *IEEE-RAS International Conference on Humanoid Robots*, pages 726–733, 2012.

Ye Zhao and Luis Sentis. Passivity of time-delayed whole-body operational space controllers with series elastic actuation. In *IEEE-RAS International Conference on Humanoid Robots, Under Review*, 2016.

Ye Zhao, Donghyun Kim, Benito Fernandez, and Luis Sentis. Phase space planning and robust control for data-driven locomotion behaviors. In *IEEE-RAS International Conference on Humanoid Robots*, pages 80–87, 2013b.

Ye Zhao, Nicholas Paine, and Luis Sentis. Feedback parameter selection for impedance control of series elastic actuators. In *IEEE-RAS International Conference on Humanoid Robots*, pages 999–1006, 2014b.

Ye Zhao, Nicholas Paine, and Luis Sentis. Sensitivity comparison to loop latencies between damping versus stiffness feedback control action in distributed controllers. In *ASME Dynamic Systems and Control Conference*, 2014c.

Ye Zhao, Donghyun Kim, Gray Thomas, and Luis Sentis. Hybrid multi-contact dynamics for wedge jumping locomotion behaviors. In *Proceedings of the International Conference on Hybrid Systems: Computation and Control*. ACM, 2015a.

Ye Zhao, Nicholas Paine, Kwan Suk Kim, and Luis Sentis. Stability and performance limits of latency-prone distributed feedback controllers. *IEEE Transactions on Industrial Electronics*, 62:7151–7162, 2015b.

Ye Zhao, Benito Fernandez, and Luis Sentis. Robust phase-space planning for agile legged locomotion over various terrain topologies. In *Proceedings of Robotics: Science and Systems*, AnnArbor, Michigan, June 2016a. doi: 10.15607/RSS.2016.XII.010.

Ye Zhao, Benito Fernandez, and Luis Sentis. A framework for planning and controlling non-periodic bipedal locomotion. *arXiv preprint arXiv:1511.04628*, 2016b.

Ye Zhao, Benito Fernandez, and Luis Sentis. Robust phase-space planning for agile legged locomotion over various terrain topologies. *Robotics: Science and System*, 2016c.

Ye Zhao, Jonathan Samir Matthis, Sean L. Barton, Mary Hayhoe, and Luis Sentis. Exploring visually guided locomotion over complex terrain: A phase-space planning method. In *Proceedings of Dynamic Walking*, 2016d.

Ye Zhao, Nicholas Paine, Steven Jorgensen, and Luis Sentis. Impedance control and performance measure of series elastic actuators. *IEEE Transactions on Industrial Electronics, Under Review*, 2016e.

Ye Zhao, Ufuk Topcu, and Luis Sentis. High-level reactive planner synthesis for whole-body locomotion in constrained environments. In *IEEE Conference on Decision and Control*, 2016f.

Ye Zhao, Ufuk Topcu, and Luis Sentis. Towards formal planner synthesis for unified legged and armed locomotion in constrained environments. In *Proceedings of Dynamic Walking*, 2016g.

Kemin Zhou, John Comstock Doyle, Keith Glover, et al. Robust and optimal control. *Prentice hall New Jersey*, 40, 1996.

Loredana Zollo, Bruno Siciliano, Alessandro De Luca, Eugenio Guglielmelli, and Paolo Dario. Compliance control for an anthropomorphic robot with elastic joints: Theory and experiments. *Journal of Dynamic Systems, Measurement, and Control*, 127(3):321–328, 2005.

# Vita

Ye Zhao received the Bachelor of Engineering (B.E.) degree in Control Science and Engineering from Harbin Institute of Technology, China, in 2011, and the Mater of Science in Engineering (M.S.E.) degree in Mechanical Engineering from The University of Texas at Austin (UT Austin), Texas, USA, in 2013. Since then, he has been pursuing his Ph.D. degree at The Human Centered Robotics Laboratory, Mechanical Engineering, UT Austin. His long-term research goal is to devote himself to planning, control, optimization, and decision-making of collaborative humanoid and mobile robots operating in cluttered environments and interacting with and working alongside humans.

Contact email: yezhao@utexas.edu.

This dissertation was typeset with LaTeX[†] by the author.

[†]LaTeX is a document preparation system developed by Leslie Lamport as a special version of Donald Knuth's TeX Program.